\documentclass[a4paper,10pt]{article}
\advance\hoffset by -0.5truein
\usepackage[utf8]{inputenc}
\usepackage[T1]{fontenc}
\usepackage[english]{babel}
\usepackage{graphicx}
\usepackage{adjustbox}
\usepackage{subcaption}
\usepackage{epstopdf}
\usepackage{diagbox}
\usepackage{tikz}
\usepackage{booktabs} 
\usepackage{multirow}
\usepackage{float}
\usepackage{placeins}
\usepackage{amsmath,amsfonts,amssymb,bm}
\usepackage{changepage} 
\usepackage{placeins}

\usepackage[english]{babel} 
\usepackage{xcolor}
\usepackage{url}
\usepackage{pdflscape}

\usepackage{latexsym}
\usepackage{enumerate}
\usepackage{caption}
\addto\captionspolish{}
\usepackage{changepage}
\newfont{\mm}{eufm10 scaled 1200}
\begin{document}





\newpage
\centerline{\Large\bf   Robustness and Invariance of Hybrid Metaheuristics}
\vspace{0.2cm}
\centerline{\Large\bf under Objective Function Transformations}
\vspace{0.4cm}


\centerline{\large Grzegorz Sroka } \vspace{4mm}
\centerline{\footnotesize\it Department of Analysis Nonlinear, Rzesz\'ow University of Technology, Powsta\'nc\'ow Warszawy 12,}
\centerline{\footnotesize\it  35-959 Rzesz\'ow, Poland   }
\centerline{\footnotesize\it  e-mail: gsroka@prz.edu.pl}
\vspace{0.31cm}
\centerline{\large Sławomir T. Wierzchoń} \vspace{4mm}
\centerline{\footnotesize\it   Institute of Computer Science, Polish Academy of Sciences, }
\centerline{\footnotesize\it   ul. Jana Kazimierza 5, 01-248 Warsaw, Poland}

\vspace{0,25cm}

{\footnotesize  {\bf Abstract.}  \it  \  This paper evaluates the robustness and structural invariance of hybrid population-based metaheuristics under various objective space transformations. A lightweight plug-and-play hybridization operator is applied to nineteen state-of-the-art algorithms-including differential evolution (DE), particle swarm optimization (PSO), and recent bio-inspired methods-without modifying their internal logic.
Benchmarking on the CEC-2017 suite across four dimensions (10, 30, 50, 100) is performed under five transformation types: baseline, translation, scaling, rotation, and constant shift. Statistical comparisons based on Wilcoxon and Friedman tests, Bayesian dominance analysis, and convergence trajectory profiling consistently show that differential-based hybrids (e.g., hIMODE, hSHADE, hDMSSA) maintain high accuracy, stability, and invariance under all tested deformations.
In contrast, classical algorithms-especially PSO- and HHO-based variants-exhibit significant performance degradation under non-separable or distorted landscapes. The findings confirm the superiority of adaptive, structurally resilient hybrids for real-world optimization tasks subject to domain-specific transformations.}

\vskip0.01cm  
{\footnotesize  {\bf Keywords.}  \it  \ Global optimization, Population-based algorithms, Meta-heuristics, Hybridization, Invariance analysis, Statistical analysis}

\section{Introduction}

The originality and innovation of the present work lie in the introduction of a universal, plug-and-play hybridization module proposed by Oszust, Sroka, and Cymerys \cite{5}, which only requires access to the current objective function values for the population and uses basic inverse regression (calibration), without modifying the internal code of the base algorithm. Unlike earlier studies \cite{1}, \cite{6}, which primarily explored isolated algorithm pairs, we demonstrate the integration of this operator across a broad spectrum of contemporary metaheuristics. Furthermore, we contribute a novel evaluation of their robustness not only to translation, scaling, and rotation of the search space but also to additive shifts of the objective function-an often overlooked but practically relevant transformation in engineering applications. The operator generates anticipated solutions of a targeted quality and replaces the weakest individuals in the population, thereby mitigating premature convergence and loss of diversity, while preserving the original evolutionary mechanisms.

Unlike most prior studies-often limited to a single convergence metric-our experiments utilize extensive benchmark sets, robust statistical evaluation (including Friedman with post-hoc CD tests, Wilcoxon comparisons, and Bayesian analysis), and a thorough assessment of four types of invariance. This comprehensive methodology enhances insight into algorithmic stability and generality, while also informing the design of efficient hybrid models ready for industrial applications-from structural engineering to automation, robotics, data processing, and artificial intelligence. With minimal implementation overhead, the Oszust, Sroka, and Cymerys operator \cite{5} becomes a practical tool for researchers and practitioners alike. Simultaneously, our detailed analysis of function landscape transformations improves the interpretability of results and facilitates their transfer to real-world problem settings.

\section{Related work}

The hybridization of metaheuristics has emerged as a key direction in contemporary optimization research, enabling the synergistic combination of complementary algorithmic features to significantly enhance performance. In the landscape of modern optimization techniques-drawing inspiration from both natural processes and physical phenomena-it is increasingly difficult to identify a single universally effective method. This growing complexity has led to a surge in interest in hybrid frameworks, including approaches that incorporate predicted candidates into the population. Existing efforts have primarily focused on combining memetic algorithms with local refiners or on sequentially integrating global and local search strategies \cite{1}, \cite{5}, \cite{6}.

Our experimental verification encompasses ten representative modern population-based algorithms. Due to its tightly integrated covariance adaptation, CMA-ES \cite{2},\cite{3},\cite{4}, could not be hybridized with the proposed operator and was therefore
included only in its standard form. Its variants, such as IPOP-CMA-ES \cite{0j} and PSO-CMA-ES \cite{6a}, implement population restarts and adaptive switching between fast PSO-based search and CMA-ES exploitation. SPSO2011 \cite{0}, a standard version of Particle Swarm Optimization with guaranteed rotational and shift invariance via strict topology and velocity clamping, remains largely unexplored in hybrid contexts - unlike its modified counterparts, including CLPSO \cite{4d}, QPSO \cite{5c}, and FIPS \cite{5oa} - making it an ideal baseline for hybridization.
MPA \cite{00}, inspired by marine predator foraging strategies, has seen variants incorporating the Nelder-Mead method \cite{5ax}, chaotic initialization with Gaussian mutation \cite{4ax}, and opposition-based learning with compound mutation \cite{0B}; however, none have explicitly investigated the impact of decision space transformations. In this study, hMPA denotes the hybridized variant of MPA. SHADE \cite{5x} -- further developed in variants such as GL-SHADE \cite{5ab} and LSHADE-cnEpSi \cite{4aa} -- leverages historical success memories. IMODE \cite{5a}, the winner of CEC-2020, employs multi-strategy mutation for effective optimization, yet - except for its IMODE-II variant \cite{5b} -remains unexamined in hybrid settings, making it a promising candidate for further exploration.
HGS \cite{7}, a metaheuristic algorithm modeling hunger-driven hunting strategies with adaptive weight mechanisms, has already been successfully hybridized with differential evolution \cite{4yy}, opposition-based optimization \cite{4cc}, and chaotic approaches \cite{4fa}. Its dynamic balance between exploration and exploitation has proven effective for solving complex optimization problems. In contrast, HHO - a swarm-based algorithm inspired by Harris hawk hunting behavior \cite{4a}-has been combined with chaos theory \cite{0D}, differential evolution \cite{0a}, Nelder-Mead method \cite{8}, and Cuckoo Search \cite{4ab}. SMA \cite{4b}, inspired by slime mold oscillations, has been enhanced through hybridization \cite{9}, mutation strategies \cite{0A}, and local search \cite{4ac}; while the improved DMSSA \cite{4c} incorporates CMS and ADMS mutations, no further hybrid extensions have been reported. The recently introduced AROA \cite{0C}, based on competing attraction-repulsion forces, has not yet been integrated into hybrid models. 

The selection of the Marine Predators Algorithm (MPA), Slime Mould Algorithm (SMA), Dynamic Modified Salp Swarm Algorithm (DMSSA), Henry Gas Solubility Optimization (HGS), and Harris Hawks Optimization (HHO) in this study represents a deliberate response to a well-recognized gap in the metaheuristics literature, rather than an inclination toward exotic algorithmic metaphors. While most comparative studies concentrate on well-established and widely cited optimizers, our decision reflects the necessity to rigorously assess a group of algorithms that, although increasingly popular and frequently modified or hybridized in recent years, have seldom been systematically compared under unified experimental conditions. The aim here is not only to evaluate their performance under challenging scenarios -- such as transformations of the objective function -- but also to determine their actual potential when integrated with a lightweight, universal hybridization operator. The use of this operator enables a deeper investigation of their adaptive capabilities, robustness under structural perturbations, and ability to exploit additional information derived from population dynamics to enhance the exploration-exploitation balance.

This approach directly addresses the concerns raised in the influential paper by Aranha et al., \textit{Metaphor-based metaheuristics, a call for action: the elephant in the room}~\cite{Aran}, where the authors critically examine the rapidly expanding number of metaphor-inspired heuristics, many of which lack solid theoretical grounding or empirical justification. Rather than contributing yet another metaphor-driven procedure, this study focuses on evaluating the real-world utility and algorithmic behavior of selected methods that, despite their metaphorical origins, have been subject to increasing practical use and hybrid adaptations. 
By benchmarking them under uniform conditions and assessing their compatibility with an external operator, we contribute to a more objective understanding of their value beyond their initial narrative appeal. In doing so, the findings provide a coherent and evidence-based perspective that supports the informed development of hybrid optimization strategies across diverse application domains.

\section{CEC-2017 Benchmark}

To evaluate the proposed hybrids, the standard CEC-2017 benchmark suite (Competitions on Evolutionary Computation 2017) \cite{0ab} was employed. This suite consists of 30 single-objective optimization problems (with one, F2, commonly omitted due to a known implementation issue, resulting in 29 problems being actually used), grouped into categories: unimodal (1--3), basic multimodal (4--10), hybrid (11--20), and composite (21--30) functions.
A key characteristic of this benchmark is that each function is shifted and rotated in the decision space, making it a transformed variant of classical test functions-thus requiring the algorithm to handle complex search space geometries and ensuring robustness against coordinate transformations.
The benchmark was executed using MATLAB R2022a on Ubuntu 20.04 LTS. All benchmark functions were recompiled locally using the command \textit{mex benchmark\_func.c} with the \textit{-largeArrayDims} option enabled, which ensures compatibility with high-dimensional input vectors (up to $dim = 100$). This recompilation was required because the original \textit{.mex} files were compiled for Windows platforms \textit{.mexw64} and are not directly compatible with Linux environments.
The results obtained from all runs were used to compute the mean, median, and standard deviation, which were subsequently used for statistical testing using the Friedman test and post hoc Nemenyi comparisons.
The choice of CEC-2017 is well justified by its broad recognition in the metaheuristics literature \cite{0aaa},\cite{0C}. Leveraging this benchmark allows for reliable comparison of the obtained results with those from other state-of-the-art studies.

\section{Population-Based Operator}

In the presented hybrid algorithms, a pivotal component is the population-based operator introduced by Oszust, Sroka, and Cymerys as part of the concept proposed in \cite{5}. The core idea behind this mechanism is as follows: the operator analyzes the current population of solutions generated by the base algorithm and estimates a target objective function value for a new, predicted candidate solution. This target value is determined based on the fitness distribution of the population and the adopted search strategy. Subsequently, a calibration problem is solved to generate a feature vector that corresponds to the predicted function value.

Through this process, a subset of the weakest individuals in the population is replaced by forecasted candidates, which is intended to reinforce exploitation in promising regions of the search space. Importantly, all other components of the base algorithm-such as crossover, mutation, and population update mechanisms-remain unchanged. This design ensures high flexibility, allowing the operator to be seamlessly integrated with virtually any population-based metaheuristic. In essence, this operator strengthens selective pressure using global information about the population without interfering with the native solution-generation processes.

Originally verified within hybridizations of Particle Swarm Optimization (PSO) and nine other population-based metaheuristics -- Marine Predators Algorithm (MPA), Salp Swarm Algorithm (SSA), Butterfly Optimization Algorithm (BOA), Grasshopper Optimization Algorithm (GOA), Dragonfly Algorithm (DA), Moth-Flame Optimization (MFO), Grey Wolf Optimizer (GWO), Multi-Verse Optimizer (MVO), and Wind Driven Optimization (WDO) -- this study extends the operator's application to more recent algorithms such as HGS, SMA, and AROA. This broader experimentation enables a comprehensive evaluation of the operator’s universality and its actual impact across different search architectures.

The primary objective of incorporating this operator is to improve the exploration-exploitation balance: by introducing predicted candidates, the search process is guided toward more attractive regions, while the original mechanisms operating on the remaining population preserve the algorithm’s ability for effective global exploration.

\vspace{-0.11cm}
\section{Computational Environment and Experimental Methodology}

All experiments were conducted in MATLAB R2022a under a Linux-based operating system on a workstation equipped with an Intel Core i7-6700K processor (4 GHz) and 64 GB of RAM, using a unified codebase. Implementations of all algorithms, including their hybrid variants, were based on source codes published 
in the original articles or provided in the official repositories by the respective authors. Each algorithm was executed using the parameter settings recommended in the literature.

All algorithms were evaluated under identical conditions. Each run was terminated by a fixed budget of 100000 function evaluations (FES= total number of function evaluations), independent of the problem dimensionality. No explicit limit on the number of generations was imposed; runs stopped when the FES budget was exhausted. The population size was fixed at 20 individuals in all experiments to ensure uniform settings across algorithms and dimensions and to allow more generational updates within the same computational cost. Smaller populations have also been reported as competitive under constrained evaluation budgets \cite{0ctk}.

The search domain was the hypercube $[-100, 100]^{dim}$. We considered four dimensionalities, $dim = 10, 30, 50, 100$. This setup 
allowed for an in-depth assessment of both the scalability of the algorithms and their robustness to increasing dimensionality. For each algorithm--function pair, 30 independent runs were performed. The resulting distributions were analyzed with non-parametric statistical tests (Friedman with post-hoc Nemenyi and pairwise Wilcoxon signed-rank) to assess the significance of performance differences.

The implementation of the population operator \cite{5} in hybrid variants of the algorithms was carried out as follows: after each generation, the $g = \lfloor 0.1 \times \text{agents} \rfloor$ worst-performing individuals were identified and replaced with new candidate solutions generated by the operator. 
All other components of the algorithm remained unchanged, ensuring that any differences in behavior could be attributed solely to the integration of the operator. During the experiments, the objective function values were recorded at each iteration, allowing for the simultaneous assessment of both the final 
solution quality and the convergence speed. Following the original methodology, the replacement rate was fixed at $10\%$, which has been shown to provide a favorable balance between search intensification and diversity maintenance, and thus no further tuning of $g$ was performed in this study.

\section{Analysis of Tabular Results}

\subsection{Reporting Methodology and Experimental Result Structure}
To ensure a reliable and rigorous assessment of the performance of the analyzed metaheuristic algorithms on the CEC-2017 benchmark suite, a standardized reporting methodology was adopted, based on best practices established in leading publications in the fields of evolutionary computation and global optimization 
\cite{0ab},\cite{0kb},\cite{0BB},\cite{0CC},\cite{8a}.

Each of the 19 compared algorithms, including their hybrid variants, was independently executed 30 times for each of the 29 benchmark functions and across four different decision space dimensionalities ($dim = 10, 30, 50, 100$). Based on the results obtained, three primary descriptive statistics were computed:\\
\textbf{Mean} - the average final objective function value (calculated as the mean of the means obtained across 30 runs for each function),\\
\textbf{Med} - the median value of the results,\\
\textbf{Std} - the standard deviation, serving as an indicator of result dispersion and algorithmic stability.\\
The mean-of-means approach was selected due to its ability to smooth out stochastic fluctuations and reduce the influence of outliers, thereby enabling a more robust comparison of overall algorithmic efficiency. Additionally, a \textit{ranking-based performance summary} was included based on median values:\\
\textbf{Sum rank} - the cumulative rank of each algorithm compared to others across all test functions. For the main experiments (19 algorithms, 29 functions), the theoretical range is from $29$ (best possible) to $551$ (worst possible). For the invariance experiments, where CMA-ES is omitted (18 algorithms), the range is from $29$ to $522$.\\
\textbf{Mean rank} - the average rank per function.\\
To evaluate the statistical significance of the observed performance differences among algorithms, two classical non-parametric tests were applied:\\
\textbf{Friedman test} - used to perform a global assessment of differences between methods (Friedman p-values are available in the full dataset); a p-value < 0.05 was considered evidence of statistically significant differences within the group and justified further post-hoc analysis.\\
\textbf{Wilcoxon signed-rank test} - conducted for each pairwise comparison between algorithms. For each method, the number of statistically significant wins and losses ($+/-$) against other algorithms was recorded, along with the corresponding p-value. In all tables of this type, the last column for each dimensionality reports this p-value, summarizing the statistical significance of the performance differences between the given algorithm and all others in pairwise comparisons for that specific dimensionality. A p-value below $0.05$ indicates statistically significant superiority or inferiority, while a value equal to or above $0.05$ suggests no statistically significant difference. Pairs with $p < 0.05$ were considered statistically significant.\\

In addition to these tests, the Nemenyi post-hoc procedure and modern Bayesian analysis techniques, were also employed to complement the classical non-parametric framework. This combined approach provided a more comprehensive evaluation of performance differences, with detailed discussion of these tests and their outcomes presented in the subsequent sections dedicated to results interpretation and algorithm comparison.

This multi-faceted evaluation framework -- combining descriptive statistical measures, non-parametric tests, and rank-based assessment -- supports both quantitative performance comparison and qualitative analysis of stability and relative dominance across algorithms.

\subsection{Analysis of Tabular Results: Comparative Performance and Statistical Significance of Algorithms}

This subsection provides a comprehensive interpretation of the data reported in Tab.\ref{tab1}, with particular emphasis on the influence of problem dimensionality, variability of the results, relative algorithm rankings, and the statistical significance of performance differences as established by Wilcoxon signed-rank and Friedman tests.

\tiny
\setlength{\tabcolsep}{4pt}
\begin{table}[H]
\begin{adjustwidth}{-1.0cm}{-2.6cm}
\caption{\footnotesize{Statistical results for 10 algorithms and their 9 hybrids on original CEC-2017 functions across 4 dims.}}
\scalebox{0.67}{
\renewcommand{\arraystretch}{0.9}
\begin{tabular}{@{}l
ccccccc
@{\hspace{0.5cm}}
ccccccc
@{}}
\toprule
\multirow{2}{*}{Algorithm} & \multicolumn{7}{c}{dim = 10} & \multicolumn{7}{c}{dim = 30}\\
\cmidrule(lr){2-8} \cmidrule(lr){9-15}
 & Mean & Med & Std & Sum rank & Mean rank & +/- & p-value & Avg & Med & Std & Sum rank & Mean rank & +/- & p-value \\
\midrule
CMA-ES  & 1.5E+08 & 4.0E+03 & 2.9E+08 & 551 & 19.0 & 0/522 & 3.6E-10 & 1.1E+11 & 3.1E+04 & 2.6E+11 & 551 & 19.0 & 0/522 & 5.5E-10 \\
hSPSO2011  & 1.6E+04 & 1.9E+03 & 2.0E+04 & 374.5 & 12.9 & 146/293 & 7.0E-02 & 5.9E+04 & 2.8E+03 & 3.9E+04 & 381 & 13.1 & 128/311 & 6.4E-02 \\
SPSO2011  & 1.0E+04 & 2.0E+03 & 1.8E+04 & 381 & 13.1 & 133/301 & 7.8E-02 & 5.3E+04 & 2.7E+03 & 3.6E+04 & 367 & 12.7 & 138/302 & 5.9E-02 \\
hMPA  & 2.6E+03 & 1.6E+03 & 5.2E+03 & 252 & 8.7 & 270/214 & 2.7E-02 & 1.9E+04 & 2.3E+03 & 1.7E+04 & 292 & 10.1 & 223/239 & 3.7E-02 \\
MPA  & 1.7E+03 & 1.6E+03 & 1.8E+01 & 211 & 7.3 & 330/175 & 1.1E-02 & 4.5E+03 & 2.2E+03 & 2.1E+03 & 232 & 8.0 & 289/186 & 3.4E-02 \\
hSHADE  & -6.1E-01 & 1.4E+00 & 4.7E+01 & 91 & 3.1 & 411/3 & 1.5E-01 & -2.2E-01 & 4.5E-01 & 5.2E+01 & 88 & 3.0 & 382/5 & 1.4E-01 \\
SHADE  & -4.9E-01 & 1.0E+00 & 4.7E+01 & 87 & 3.0 & 410/5 & 1.3E-01 & 6.2E-01 & 7.9E-01 & 5.4E+01 & 103 & 3.6 & 382/13 & 1.2E-01 \\
hIMODE  & -1.5E+00 & 1.0E+00 & 5.1E+01 & 78 & 2.7 & 415/0 & 1.4E-01 & -2.3E-01 & 1.3E+00 & 5.5E+01 & 102 & 3.5 & 385/2 & 1.4E-01 \\
IMODE  & 1.4E+00 & 3.2E+00 & 5.3E+01 & 85 & 2.9 & 410/16 & 1.2E-01 & 4.5E-01 & 8.7E-01 & 5.7E+01 & 117 & 4.0 & 391/6 & 1.3E-01 \\
hHGS  & 1.1E+04 & 2.6E+03 & 1.4E+04 & 439 & 15.1 & 86/377 & 4.3E-02 & 1.4E+05 & 3.2E+03 & 3.3E+05 & 430 & 14.8 & 106/367 & 3.3E-02 \\
HGS  & 1.2E+04 & 2.0E+03 & 1.5E+04 & 365 & 12.6 & 158/284 & 6.4E-02 & 1.3E+04 & 3.0E+03 & 9.1E+03 & 372 & 12.8 & 161/306 & 3.8E-02 \\
hHHO  & 5.8E+04 & 2.3E+03 & 7.1E+04 & 504 & 17.4 & 38/450 & 2.4E-02 & 3.5E+05 & 3.3E+03 & 3.1E+05 & 502 & 17.2 & 42/457 & 1.6E-02 \\
HHO  & 1.1E+04 & 2.0E+03 & 1.6E+04 & 477 & 16.4 & 55/420 & 3.8E-02 & 3.6E+05 & 3.3E+03 & 1.5E+05 & 481 & 16.6 & 52/431 & 2.6E-02 \\
hSMA  & 3.6E+03 & 1.9E+03 & 1.1E+03 & 337 & 11.6 & 161/281 & 6.9E-02 & 2.1E+04 & 2.9E+03 & 9.6E+03 & 337 & 11.6 & 165/257 & 8.8E-02 \\
SMA  & 3.9E+03 & 1.9E+03 & 9.7E+02 & 339 & 11.7 & 161/281 & 6.5E-02 & 1.9E+04 & 2.9E+03 & 1.1E+04 & 322 & 11.1 & 175/249 & 8.2E-02 \\
hDMSSA  & 9.7E+01 & 9.2E+01 & 5.3E+00 & 174 & 6.0 & 377/145 & 3.6E-20 & 3.0E-01 & 2.2E-01 & 5.2E+01 & 97 & 3.3 & 382/8 & 1.4E-01 \\
DMSSA  & -4.2E-01 & 6.7E-01 & 4.7E+01 & 94 & 3.2 & 411/3 & 1.4E-01 & 4.6E-01 & 1.0E+00 & 5.3E+01 & 102 & 3.5 & 383/9 & 1.3E-01 \\
hAROA  & 5.4E+03 & 1.7E+03 & 9.2E+03 & 381.5 & 13.2 & 138/329 & 4.4E-02 & 6.7E+04 & 2.9E+03 & 3.5E+04 & 361 & 12.5 & 154/291 & 6.1E-02 \\
AROA  & 3.8E+03 & 1.7E+03 & 7.4E+03 & 290 & 10.0 & 229/240 & 4.6E-02 & 1.9E+04 & 2.4E+03 & 1.5E+04 & 272 & 9.4 & 237/214 & 4.9E-02 \\
\bottomrule
\end{tabular}
}
\vspace{1.1mm}
\scalebox{0.67}{
\renewcommand{\arraystretch}{0.9}
\begin{tabular}{@{}l
ccccccc  
@{\hspace{0.5cm}}
ccccccc
@{}}
\toprule
\multirow{2}{*}{Algorithm} & \multicolumn{7}{c}{dim=50} & \multicolumn{7}{c}{dim=100} \\
\cmidrule(lr){2-8} \cmidrule(lr){9-15}
 & Mean & Med & Std & Sum rank & Mean rank & +/- & p-value & Avg & Med & Std & Sum rank & Mean rank & +/- & p-value \\
\midrule
CMA-ES  & 3.4E+13 & 1.7E+05 & 1.1E+14 & 551 & 19.0 & 0/522 & 2.1E-11 & 8.4E+14 & 5.8E+05 & 1.9E+15 & 551 & 19.0 & 0/522 & 2.0E-11 \\
hSPSO2011  & 8.8E+05 & 3.3E+03 & 2.4E+05 & 401 & 13.8 & 116/331 & 6.2E-02 & 3.1E+06 & 6.0E+03 & 1.0E+06 & 407 & 14.0 & 123/355 & 3.6E-02 \\
SPSO2011  & 8.3E+05 & 3.3E+03 & 2.7E+05 & 369 & 12.7 & 135/316 & 6.0E-02 & 3.0E+06 & 5.7E+03 & 1.2E+06 & 383 & 13.2 & 138/337 & 4.0E-02 \\
hMPA  & 4.3E+05 & 3.2E+03 & 3.3E+05 & 321 & 11.1 & 200/253 & 4.8E-02 & 6.0E+06 & 5.3E+03 & 2.8E+06 & 344 & 11.9 & 168/294 & 4.9E-02 \\
MPA  & 1.2E+05 & 2.9E+03 & 7.8E+04 & 248 & 8.6 & 277/198 & 3.5E-02 & 7.1E+05 & 4.1E+03 & 3.6E+05 & 258 & 8.9 & 268/207 & 3.8E-02 \\
hSHADE  & 1.1E+00 & 1.9E+00 & 5.3E+01 & 103 & 3.6 & 382/12 & 1.3E-01 & -6.2E-01 & -7.9E-02 & 5.2E+01 & 99 & 3.4 & 382/10 & 1.3E-01 \\
SHADE  & 9.4E-01 & 2.3E+00 & 5.6E+01 & 97 & 3.3 & 384/11 & 1.3E-01 & -1.9E-01 & -5.7E-01 & 5.6E+01 & 119 & 4.1 & 383/20 & 1.1E-01 \\
hIMODE  & 8.0E-01 & 5.1E-01 & 5.6E+01 & 98 & 3.4 & 387/5 & 1.4E-01 & -8.9E-01 & -1.1E-01 & 5.5E+01 & 78 & 2.7 & 390/7 & 1.3E-01 \\
IMODE  & 9.2E-01 & 2.0E-01 & 5.8E+01 & 103 & 3.6 & 386/15 & 1.2E-01 & -8.0E-01 & -8.2E-02 & 6.0E+01 & 96 & 3.3 & 392/9 & 1.1E-01 \\
hHGS  & 1.7E+06 & 4.0E+03 & 1.6E+06 & 437 & 15.1 & 100/370 & 4.9E-02 & 2.7E+08 & 7.0E+03 & 1.2E+08 & 431 & 14.9 & 105/377 & 3.7E-02 \\
HGS  & 7.7E+06 & 3.4E+03 & 2.3E+07 & 372 & 12.8 & 160/299 & 4.4E-02 & 3.2E+06 & 6.5E+03 & 5.6E+06 & 355 & 12.2 & 187/274 & 4.9E-02 \\
hHHO  & 2.2E+06 & 4.0E+03 & 1.2E+06 & 501 & 17.3 & 44/457 & 1.6E-02 & 1.7E+07 & 7.1E+03 & 8.4E+06 & 489 & 16.9 & 52/451 & 1.6E-02 \\
HHO  & 2.5E+06 & 3.9E+03 & 9.1E+05 & 473 & 16.3 & 58/427 & 3.5E-02 & 1.6E+07 & 6.6E+03 & 2.8E+06 & 462 & 15.9 & 78/423 & 1.6E-02 \\
hSMA  & 1.7E+05 & 3.3E+03 & 7.6E+04 & 312 & 10.8 & 195/244 & 7.5E-02 & 6.2E+05 & 4.9E+03 & 2.6E+05 & 265 & 9.1 & 244/212 & 5.1E-02 \\
SMA  & 1.6E+05 & 3.3E+03 & 7.2E+04 & 305 & 10.5 & 198/235 & 8.2E-02 & 5.9E+05 & 5.1E+03 & 2.0E+05 & 285 & 9.8 & 226/227 & 5.5E-02 \\
hDMSSA  & 8.4E-01 & 1.2E+00 & 5.2E+01 & 103 & 3.6 & 388/6 & 1.2E-01 & -4.5E-01 & -6.3E-02 & 5.1E+01 & 100 & 3.4 & 393/10 & 1.2E-01 \\
DMSSA  & 6.9E-01 & 1.6E+00 & 5.4E+01 & 103 & 3.6 & 388/4 & 1.3E-01 & -2.0E-01 & 4.7E-01 & 5.3E+01 & 115 & 4.0 & 386/8 & 1.2E-01 \\
hAROA  & 4.4E+05 & 3.4E+03 & 2.1E+05 & 356 & 12.3 & 155/294 & 6.2E-02 & 7.4E+06 & 5.6E+03 & 2.5E+06 & 384 & 13.2 & 145/326 & 3.2E-02 \\
AROA  & 1.4E+05 & 3.1E+03 & 6.8E+04 & 258 & 8.9 & 253/207 & 4.1E-02 & 2.3E+06 & 4.4E+03 & 8.7E+05 & 286 & 9.9 & 238/229 & 4.4E-02 \\
\bottomrule
\label{tab1}
\end{tabular}
}
\\[1ex]
\footnotesize{Friedman test p-values: dim10=1.4E-88, dim30=8.0E-85, dim50=1.4E-84, dim100=5.7E-84}
\end{adjustwidth}
\end{table}
\normalsize
Statistical Results Analysis and Algorithmic Performance Summary.\\
For low-dimensional problems ($dim = 10$), the algorithms MPA, hMPA, hSHADE, and hIMODE clearly dominate (see Tab.\ref{tab1}). They achieve not only the lowest mean and median objective values (on the order of $10^3$), but also demonstrate very low standard deviation values (e.g., MPA: Std = 18), indicating 
excellent stability.

The lowest sum of ranks is obtained by hIMODE (Sum rank = 78), which means it was most frequently ranked first or second among all methods. However, its Wilcoxon test balance of $+/- = 415/0$ and p-value = 0.14 indicates no statistically significant superiority - highlighting high consistency, but not sufficient numerical dominance. In contrast, CMA-ES, with an extremely high average of $1.5 \times 10^8$, achieves the worst ranks and a p-value of $3.6 \times 10^{-10}$, confirming its unsuitability for this class of problems.

In 30-dimensional space (Tab.\ref{tab1}), the effectiveness trend of MPA and hMPA continues (Mean MPA = $4.5 \times 10^3$, hMPA = $1.9 \times 10^4$), though their statistical advantage weakens (e.g., hMPA p-value = 0.037). A noteworthy observation is the persistent strong performance of hSHADE and hIMODE, which, despite low objective function values, exhibit relatively high p-values (above 0.1), likely due to result parity with several other competitive methods. It is also worth noting the performance decline of HGS and HHO, which in this space yield higher mean values and relatively worse rankings, indicating poorer adaptability to mid-dimensional optimization tasks.

For more complex problems ($dim = 50$), the performance gaps between algorithms narrow  (Tab.\ref{tab1}). Most Wilcoxon p-values fall within the (0.05 - 0.15) range, confirming the decline of clear dominance. hSHADE and hIMODE still stand out for their low objective values and high stability; however, even they do not show statistically significant superiority (e.g., hSHADE: $+/- = 382/12, p = 0.13$). A significant degradation is visible for PSO-based algorithms (SPSO2011 and hSPSO2011), which yield much higher average values (on the order of $10^5$), lower ranks, and high standard deviations - suggesting that classical PSO loses its exploratory capabilities in higher-dimensional search spaces.

At the highest tested dimensionality ($dim = 100$, Tab.\ref{tab1}), the greatest performance deterioration is observed for CMA-ES, HHO, and hHHO, which produce extremely high mean values (e.g., CMA-ES: $8.4 \times 10^{14}$, Std = $1.9 \times 10^{15}$) and achieve the worst ranks (Sum rank = 551). Their p-values remain below $10^{-10}$, confirming the statistical superiority of other algorithms. In this setting, hIMODE, hSHADE, and hDMSSA perform exceptionally well, maintaining very low objective function values and minimal variance despite the increased dimensionality - making them the most stable and scalable algorithms.

The comparative analysis across all four decision space sizes reveals a performance degradation of PSO- and HHO-based methods with increasing dimensionality, whereas hSHADE and hIMODE consistently maintain high effectiveness and robustness. The Wilcoxon test effectively identifies statistically dominant pairs of algorithms, although its sensitivity diminishes at higher dimensions. Meanwhile, the Friedman test confirms global significance of differences among methods ($p < 0.05$), providing strong support for reliable statistical inference.

In the subsequent sections, the study conducts analysis of Critical Difference (CD) diagrams and Bayesian statistical testing, forming the basis for generalized conclusions regarding the overall superiority of selected algorithms. This is further complemented by an assessment of boxplots and identification of outlier values as practical indicators of algorithmic robustness and reliability.

\vspace{-0.21cm}
\section{Analysis of Boxplots and Outcome Distributions for Functions f1, f6, f12, and f23}

Boxplots \cite{0aaa},\cite{4ccd} provide a graphical representation of the distribution of outcomes achieved by individual algorithms on a given benchmark function, offering an intuitive insight into the quality, stability, and variability of the results. This analysis focuses on four representative functions: f1, f6, f12, and f23, each evaluated across four decision space dimensionalities: $dim = 10, 30, 50, 100$. 

It is worth emphasizing that these plots serve as a geometric interpretation of the results previously summarized in the reporting Tab.\ref{tab1}, depicting not only central tendency (median) but also the spread of results (quartiles), potential outliers, and the use of a logarithmic Y-axis to accommodate the wide dynamic range of objective function values.

\begin{figure}[H]
  \centering
\begin{adjustwidth}{-2cm}{-2.0cm}
  \begin{subfigure}{0.4\textwidth}
     \includegraphics[height=5.5cm,width=6.9cm]{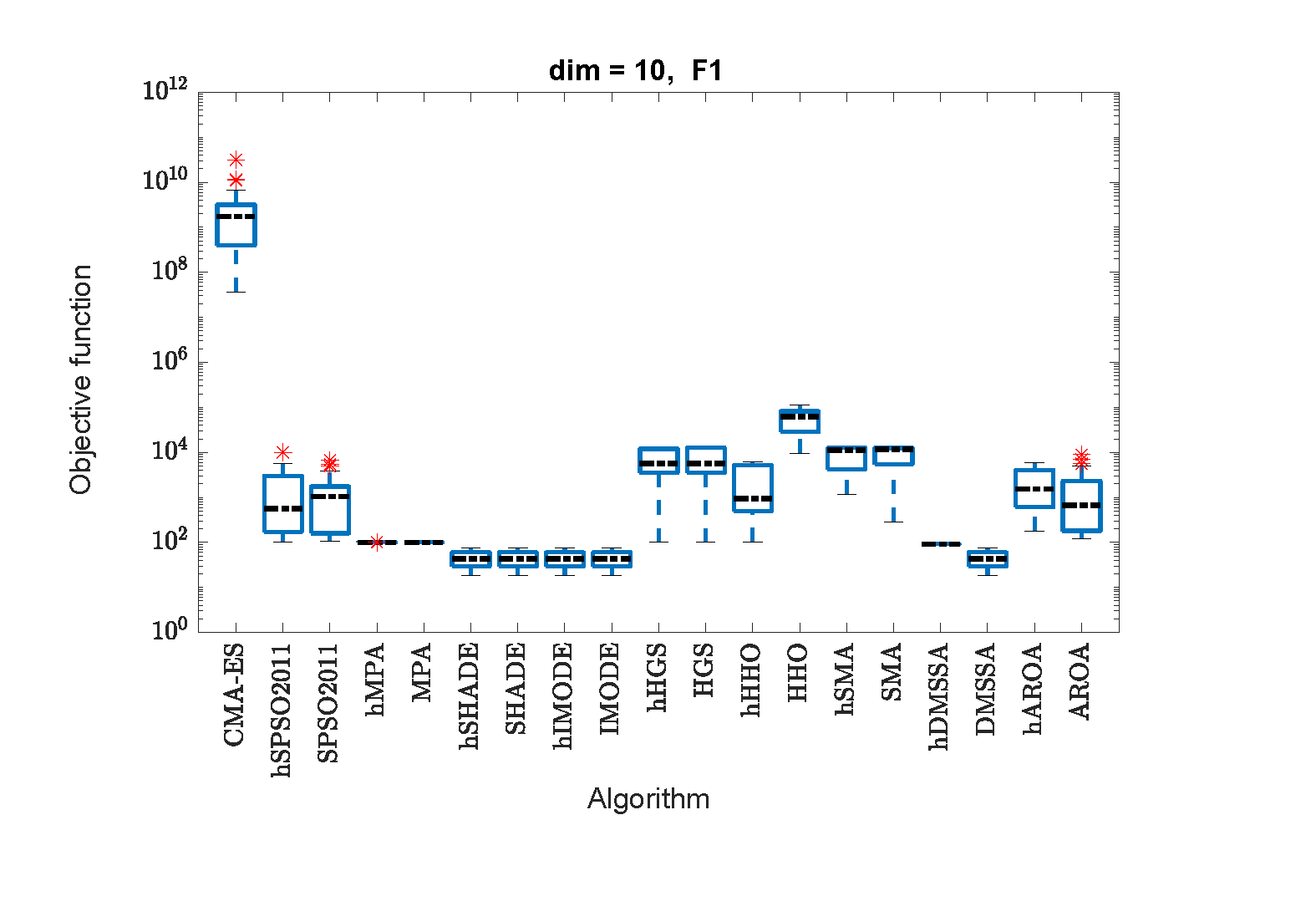}
  \end{subfigure}\hfill
  \begin{subfigure}{0.4\textwidth}
    \includegraphics[height=5.5cm,width=6.9cm]{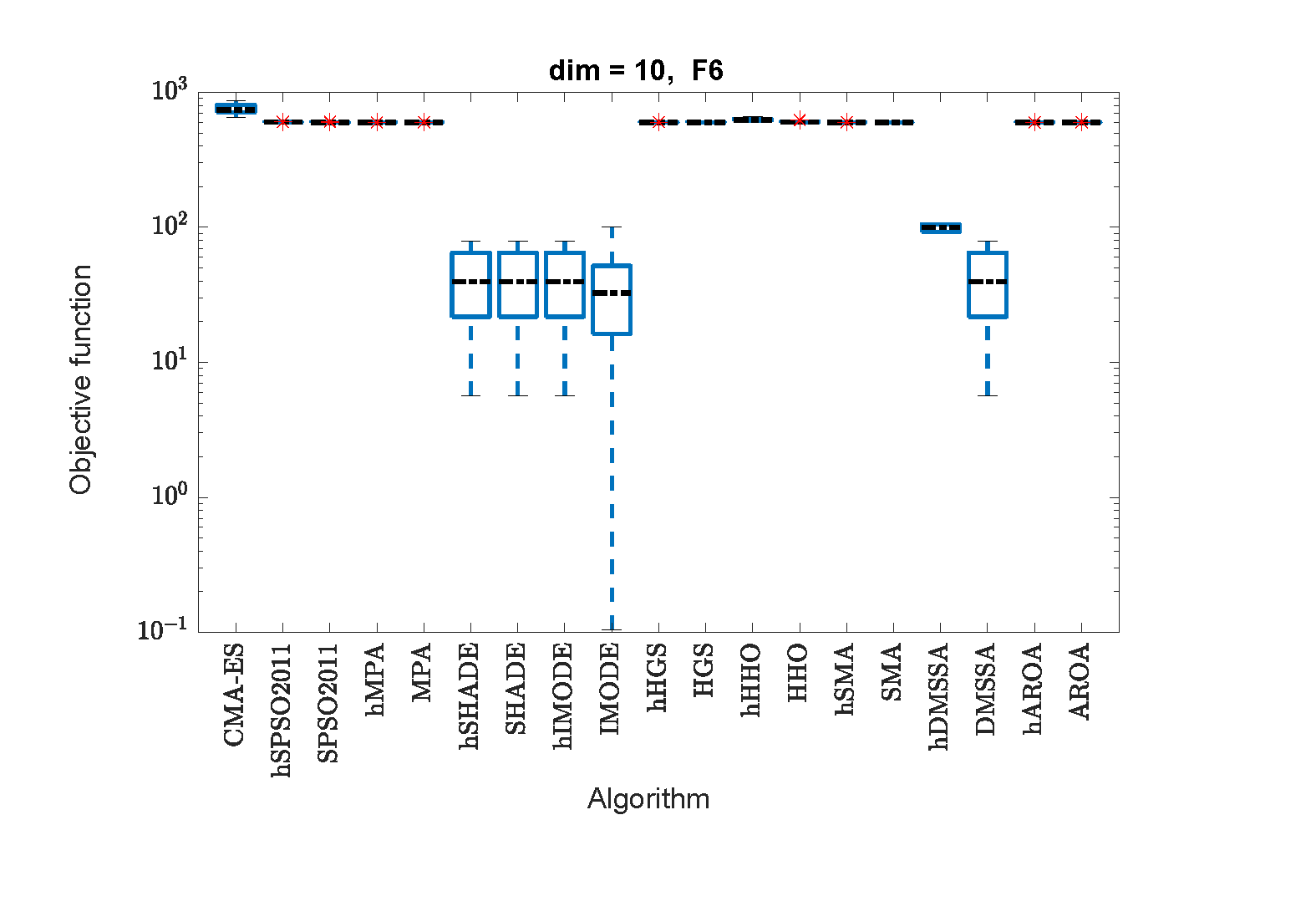}
  \end{subfigure}\hfill
  \begin{subfigure}{0.4\textwidth}
    \includegraphics[height=5.5cm,width=6.9cm]{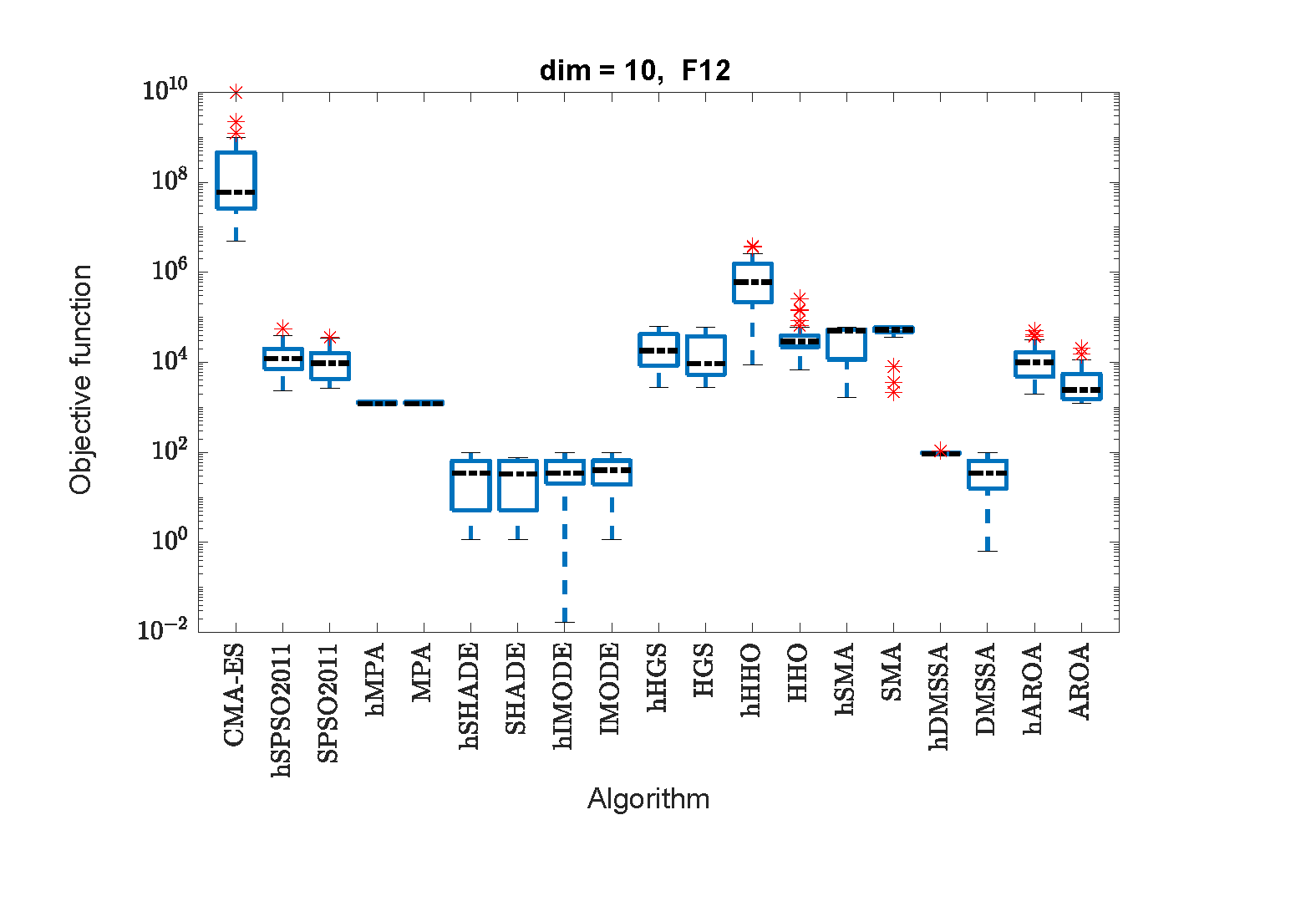}
  \end{subfigure}
\end{adjustwidth}
\end{figure} 
\begin{figure}[H]
 \centering
\begin{adjustwidth}{-2cm}{-2.0cm}
   \begin{subfigure}{0.4\textwidth}
    \includegraphics[height=5.1cm,width=6.9cm]{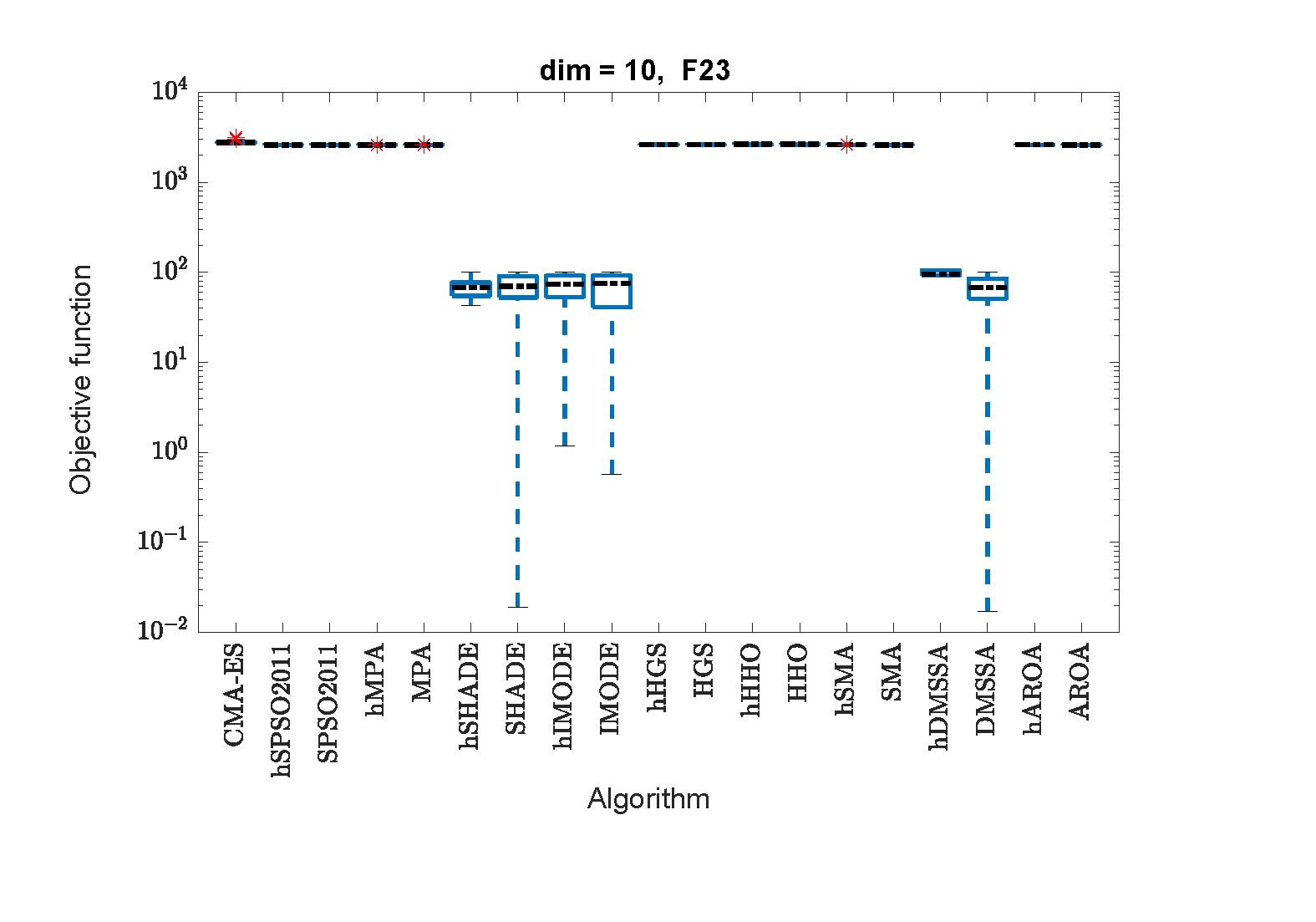}
  \end{subfigure}\hfill
  \begin{subfigure}{0.4\textwidth}
    \includegraphics[height=5.1cm,width=6.9cm]{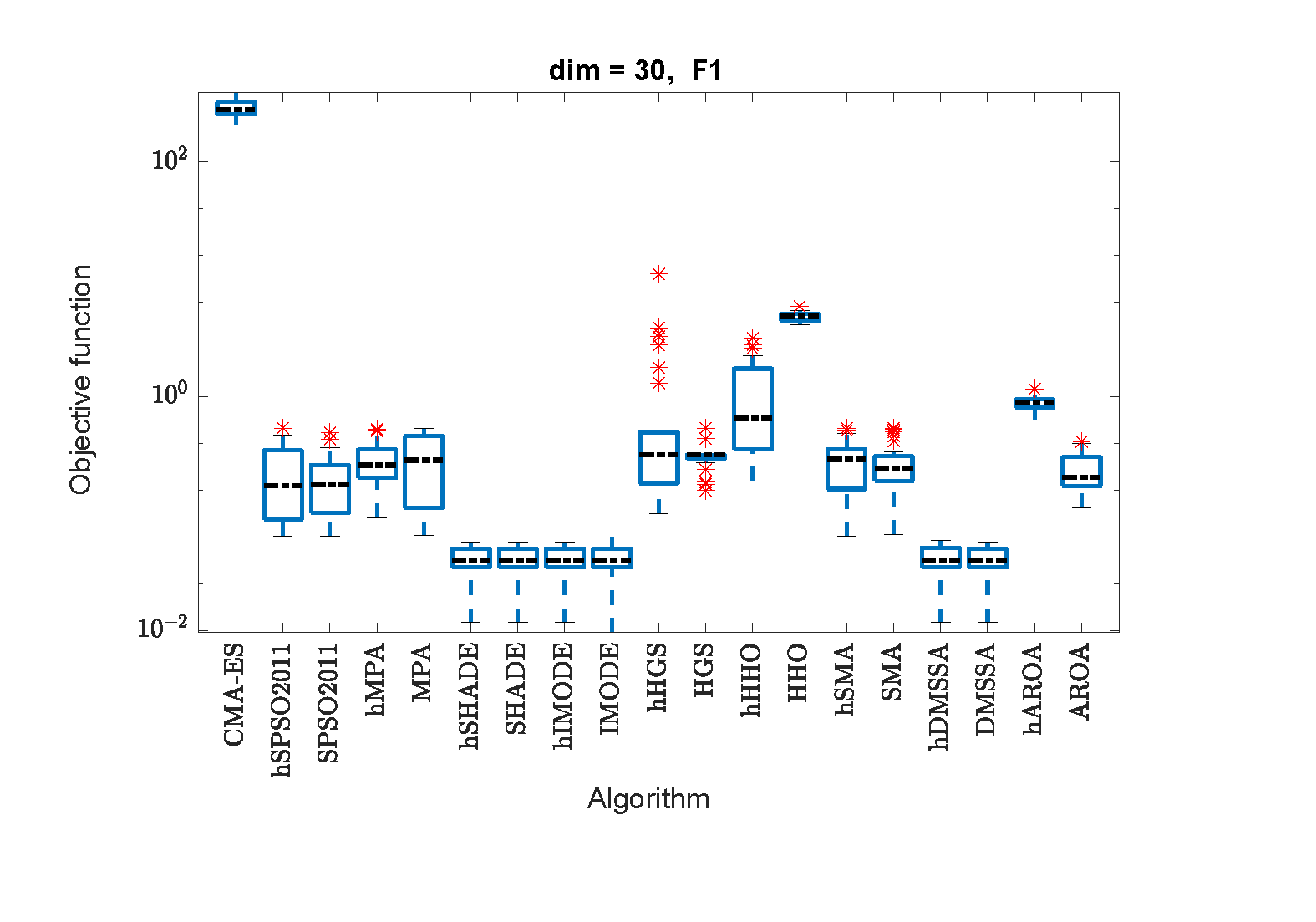}
\end{subfigure}\hfill
  \begin{subfigure}{0.4\textwidth} 
        \includegraphics[height=5.1cm,width=6.9cm]{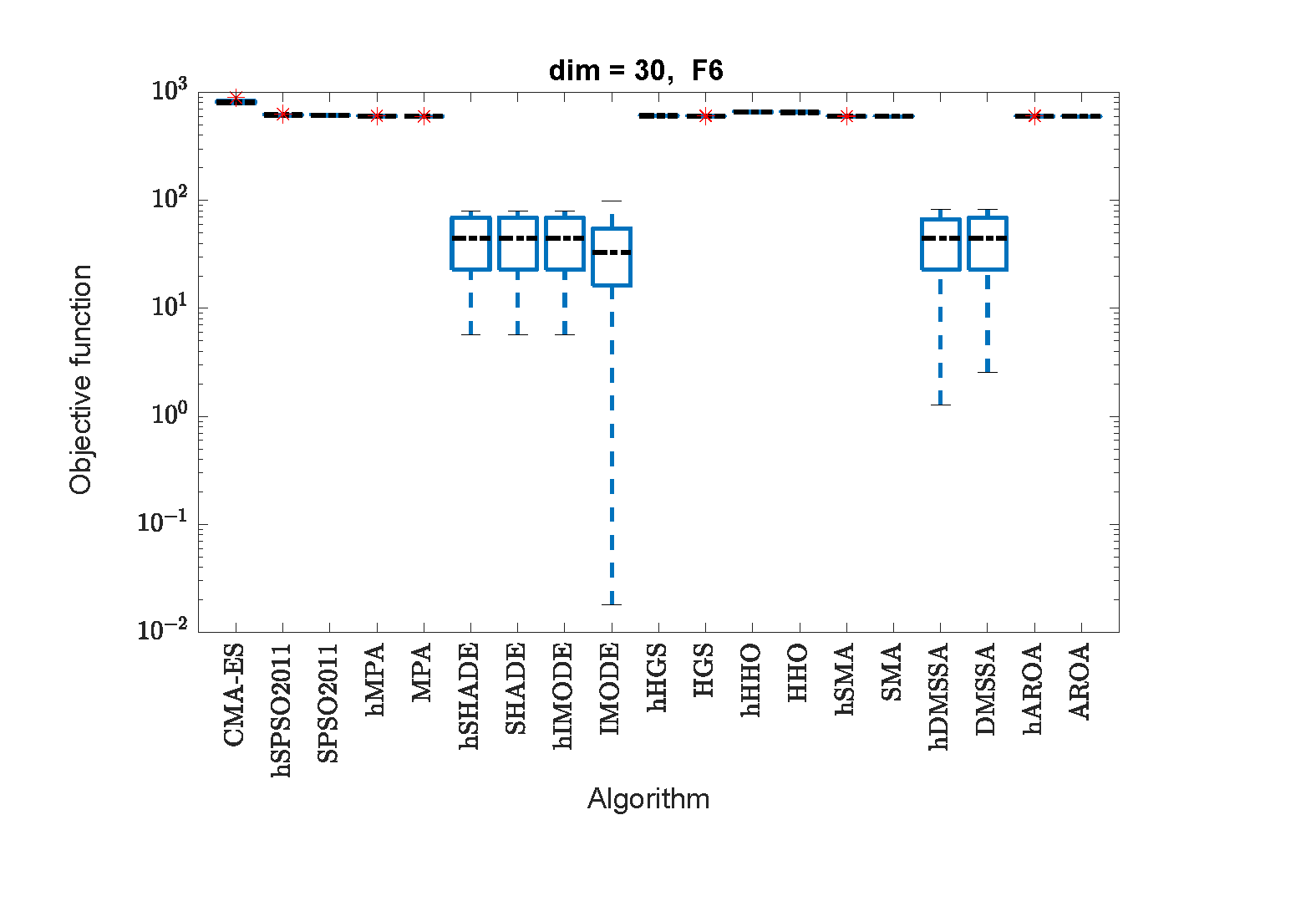}
   \end{subfigure}
   \end{adjustwidth}
\end{figure}
\begin{figure}[H]
  \centering
\begin{adjustwidth}{-2cm}{-2.0cm}
   \begin{subfigure}{0.4\textwidth}
        \includegraphics[height=5.1cm,width=6.9cm]{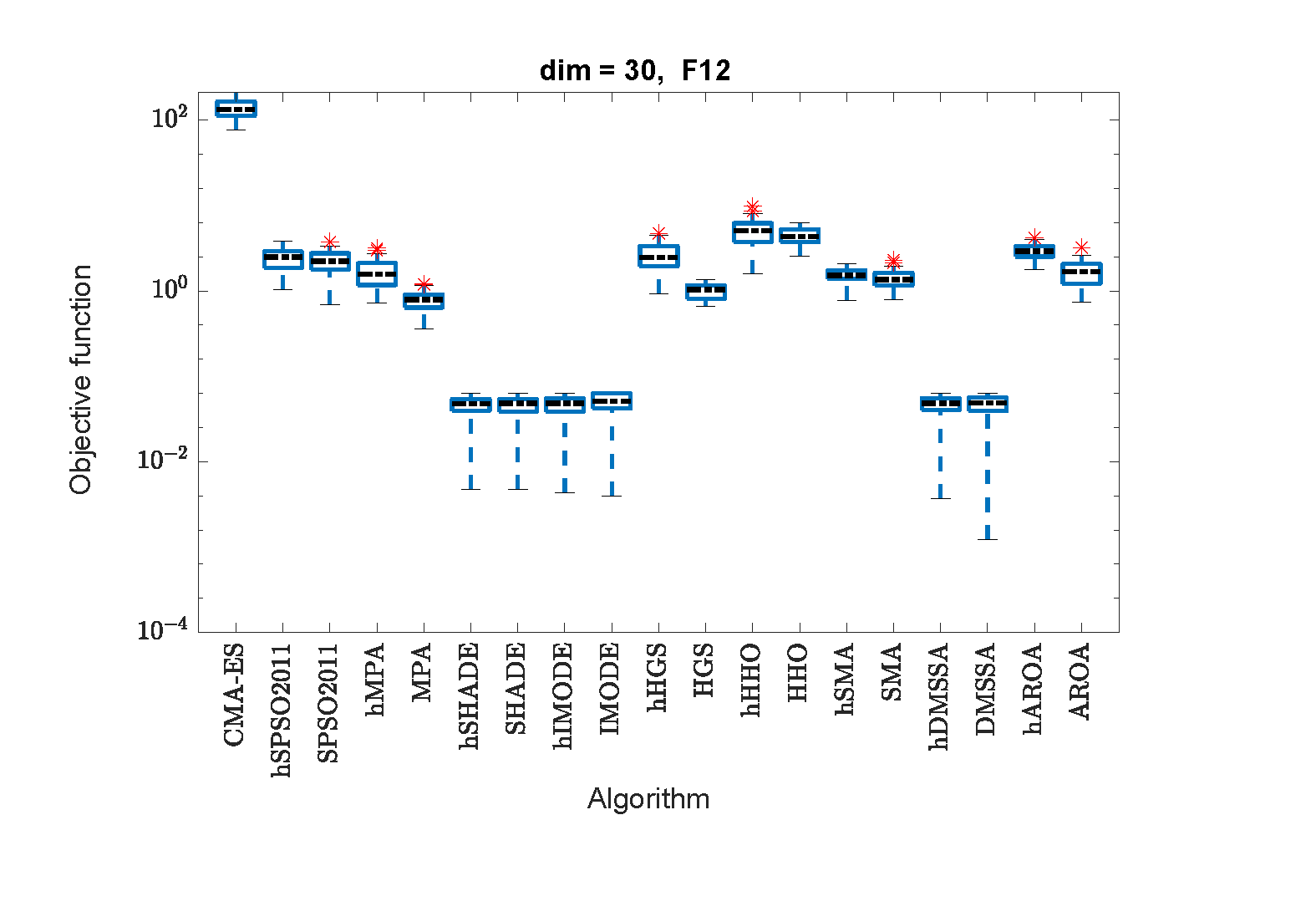}
  \end{subfigure}\hfill
  \begin{subfigure}{0.4\textwidth} 
        \includegraphics[height=5.1cm,width=6.9cm]{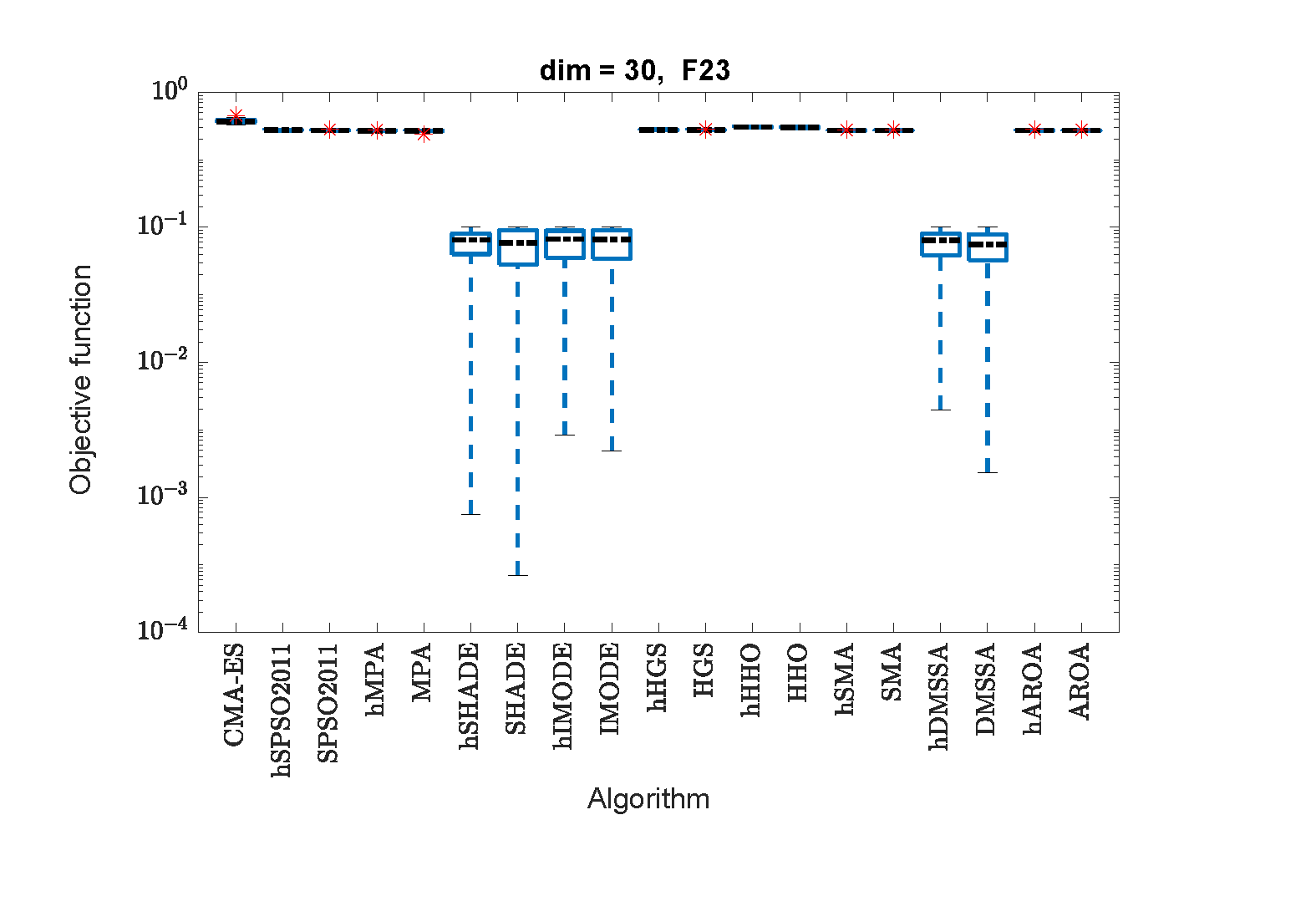}
  \end{subfigure}\hfill
  \begin{subfigure}{0.4\textwidth}
        \includegraphics[height=5.1cm,width=6.9cm]{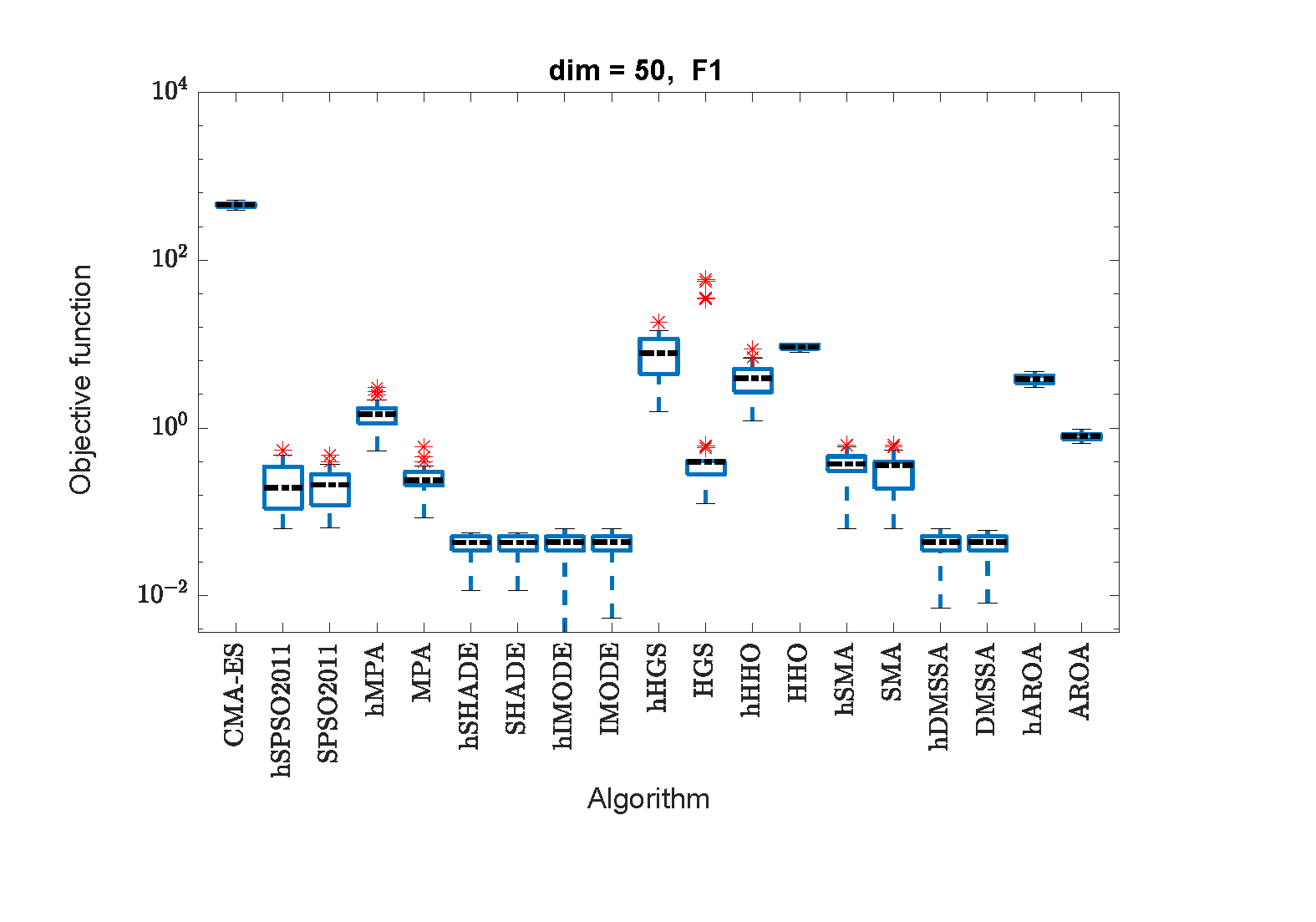}
  \end{subfigure}
  \end{adjustwidth}
\end{figure}

\begin{figure}[H]
  \centering
\begin{adjustwidth}{-2cm}{-2.0cm}
  \begin{subfigure}{0.4\textwidth} 
        \includegraphics[height=5.1cm,width=6.9cm]{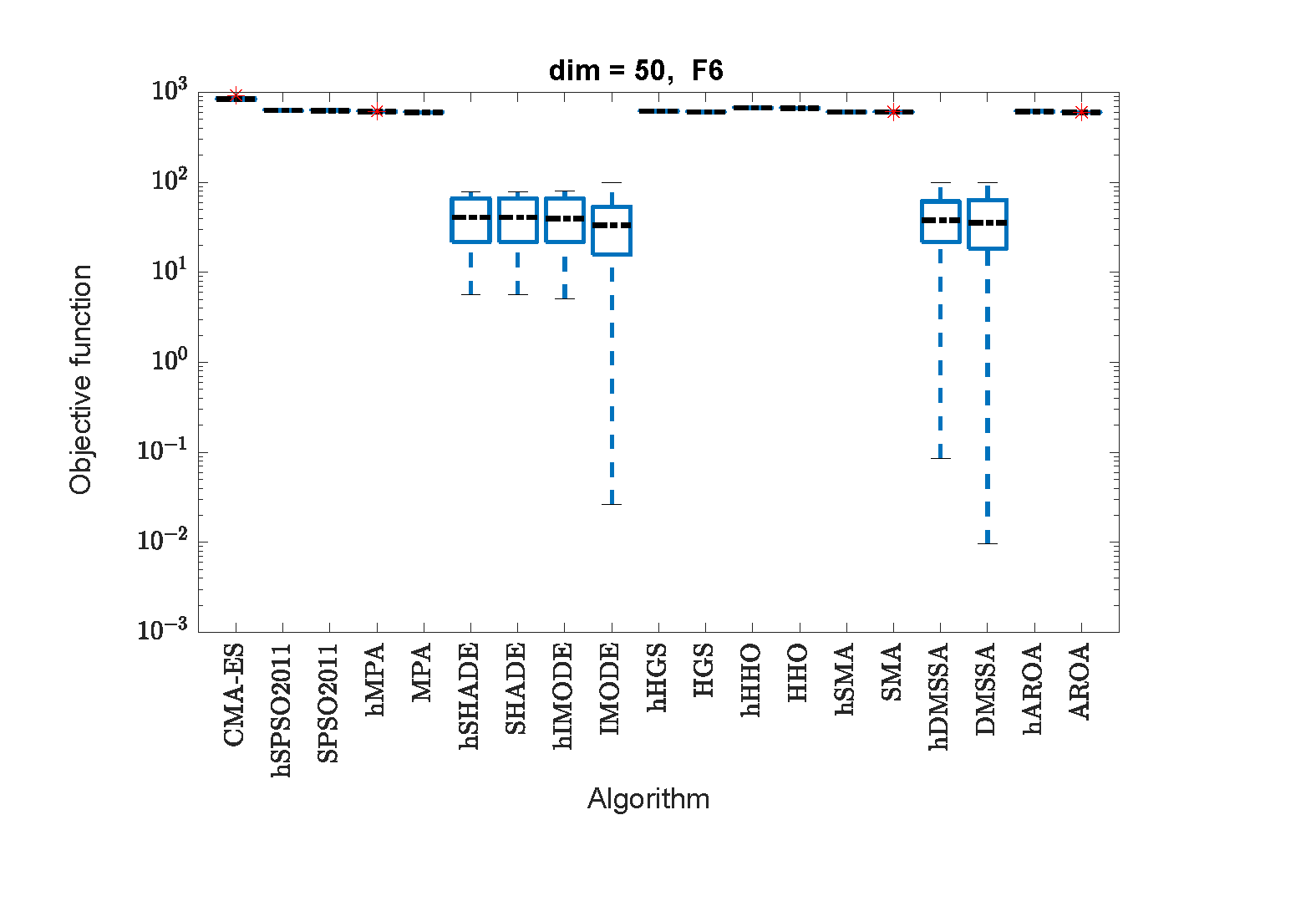}
  \end{subfigure}\hfill
  \begin{subfigure}{0.4\textwidth}
    \includegraphics[height=5.1cm,width=6.9cm]{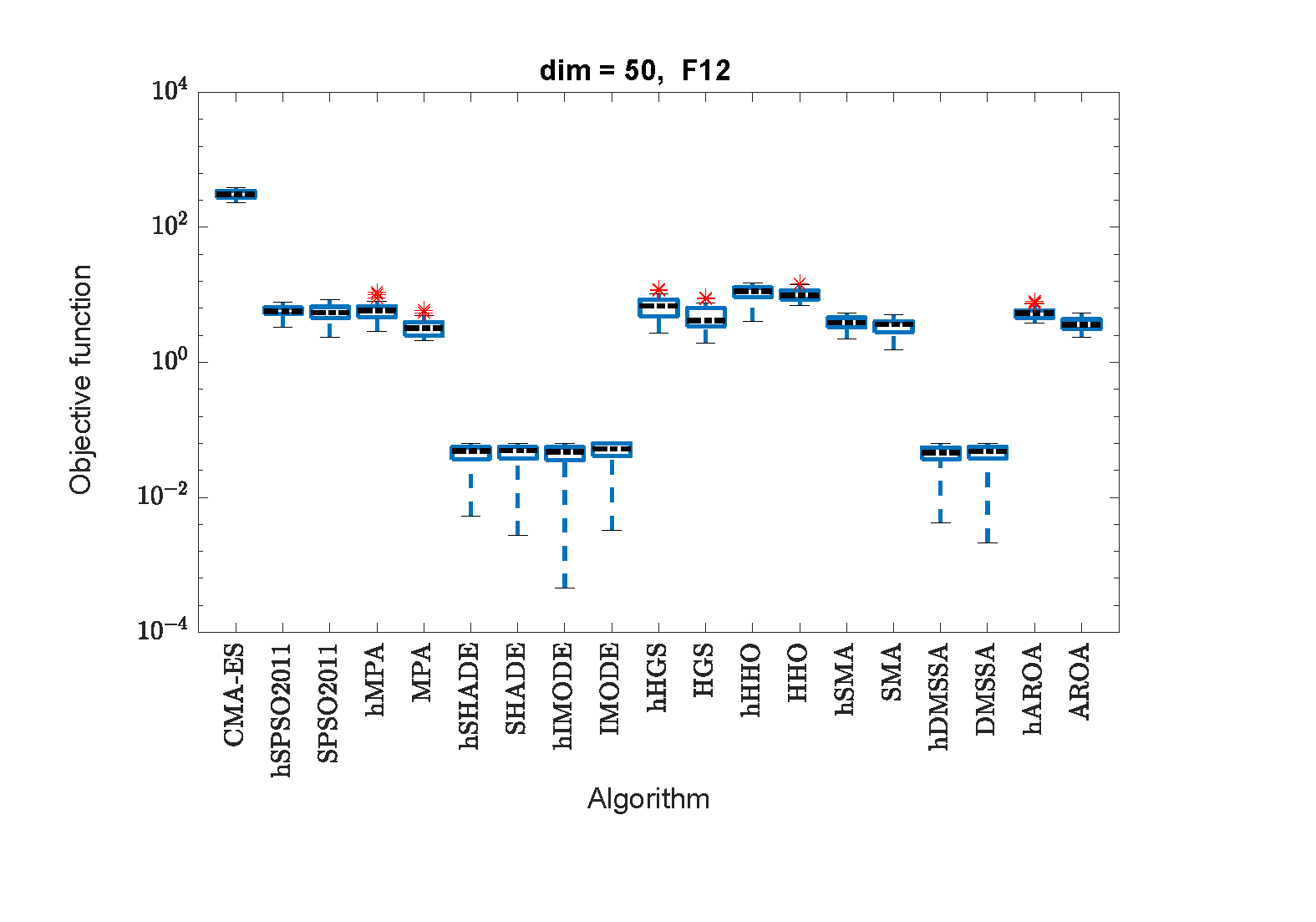}
     \end{subfigure}\hfill
 \begin{subfigure}{0.4\textwidth}
    \includegraphics[height=5.1cm,width=6.9cm]{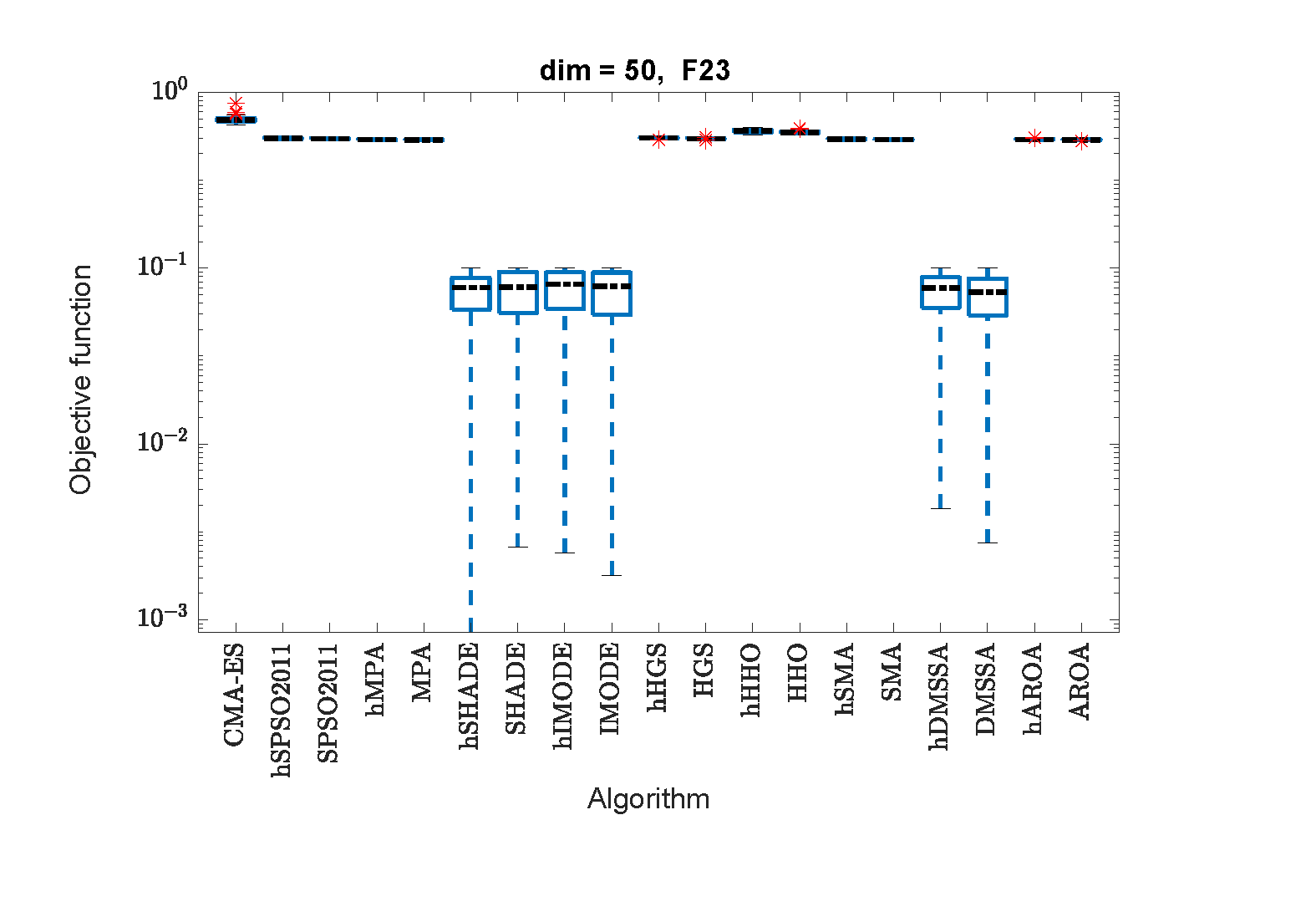}
  \end{subfigure}
\end{adjustwidth}
\end{figure}
 
  \begin{figure}[H]
  \centering
\begin{adjustwidth}{-2cm}{-2.0cm}
  \begin{subfigure}{0.4\textwidth}
    \includegraphics[height=5.1cm,width=6.9cm]{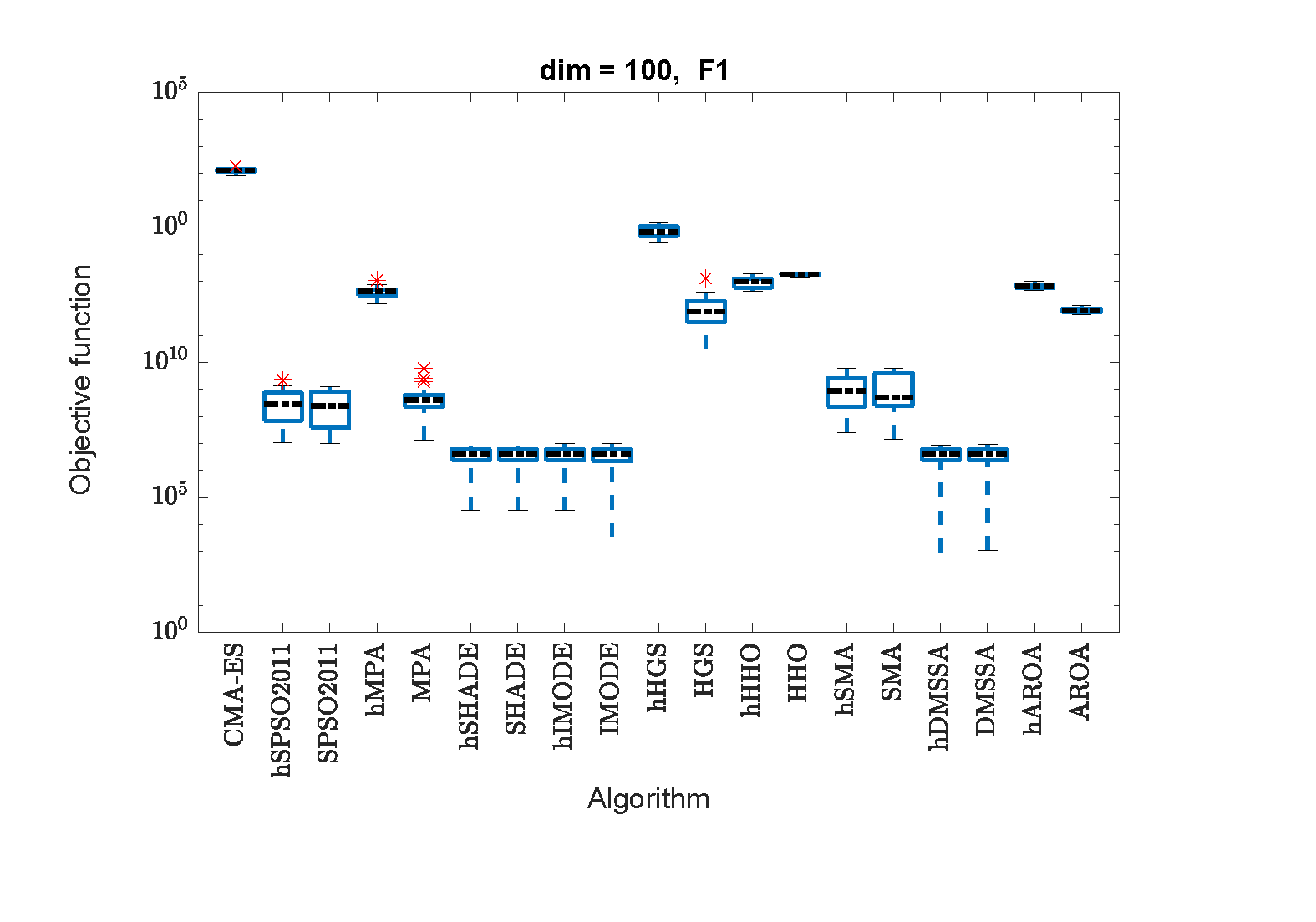}
  \end{subfigure}\hfill
  \begin{subfigure}{0.4\textwidth}
        \includegraphics[height=5.1cm,width=6.9cm]{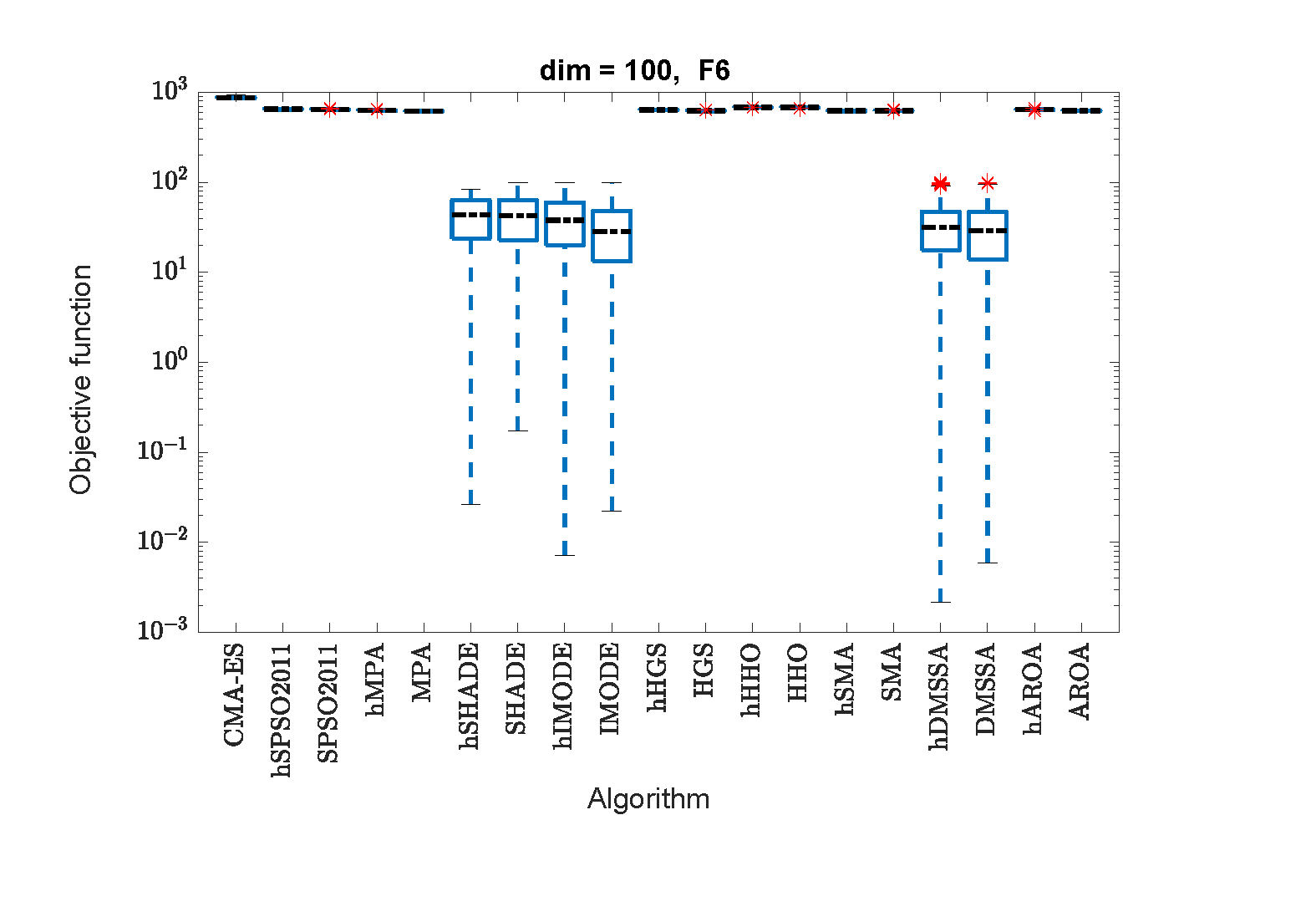}
     \end{subfigure}\hfill
   \begin{subfigure}{0.4\textwidth}
        \includegraphics[height=5.1cm,width=6.9cm]{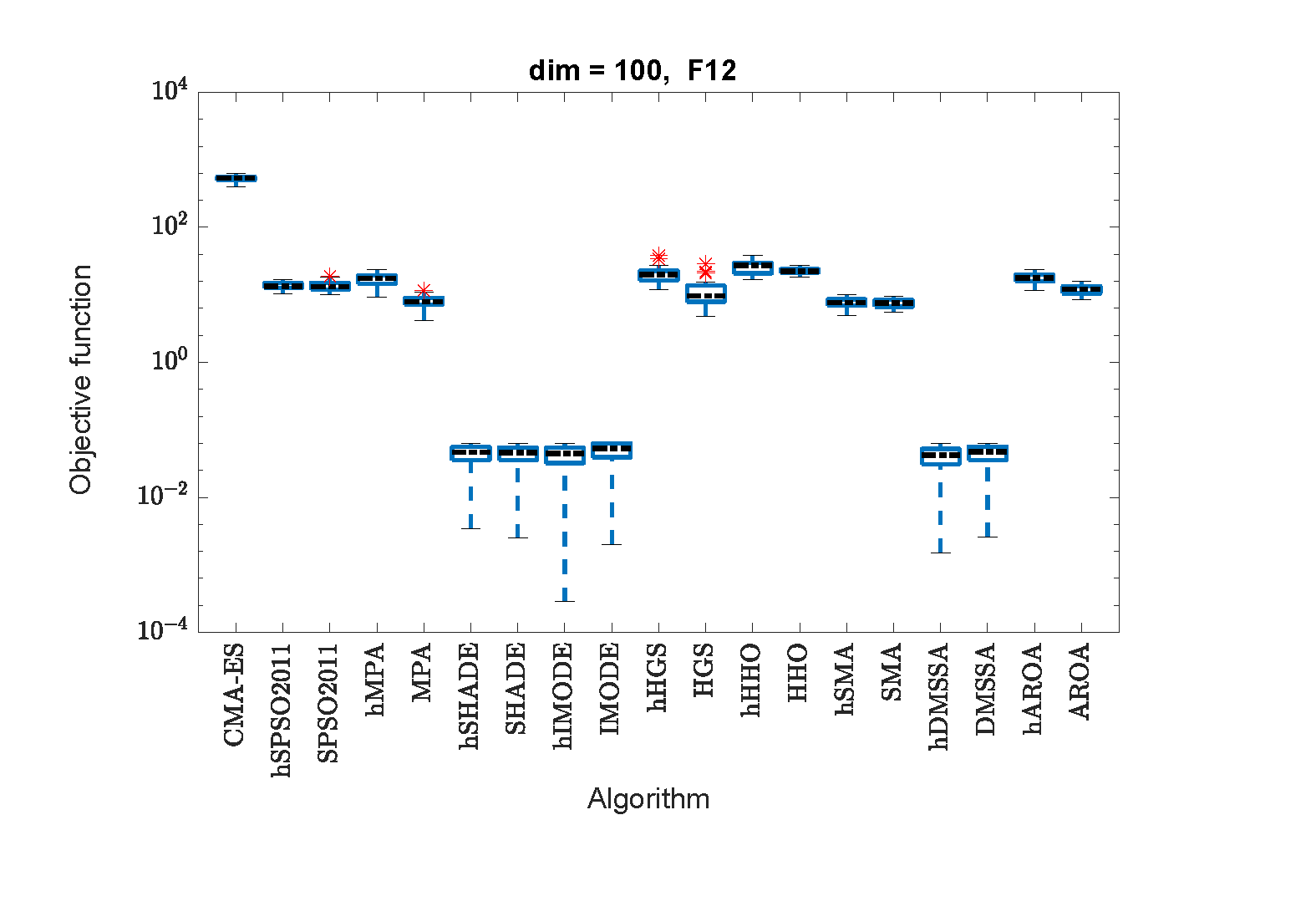}
    \end{subfigure}
    \end{adjustwidth}
    \end{figure}

\begin{figure}[H]
 \centering
\begin{adjustwidth}{-2cm}{-2.0cm}
\begin{center}
  \begin{subfigure}{0.4\textwidth}
   \centering
        \includegraphics[height=5.1cm,width=6.9cm]{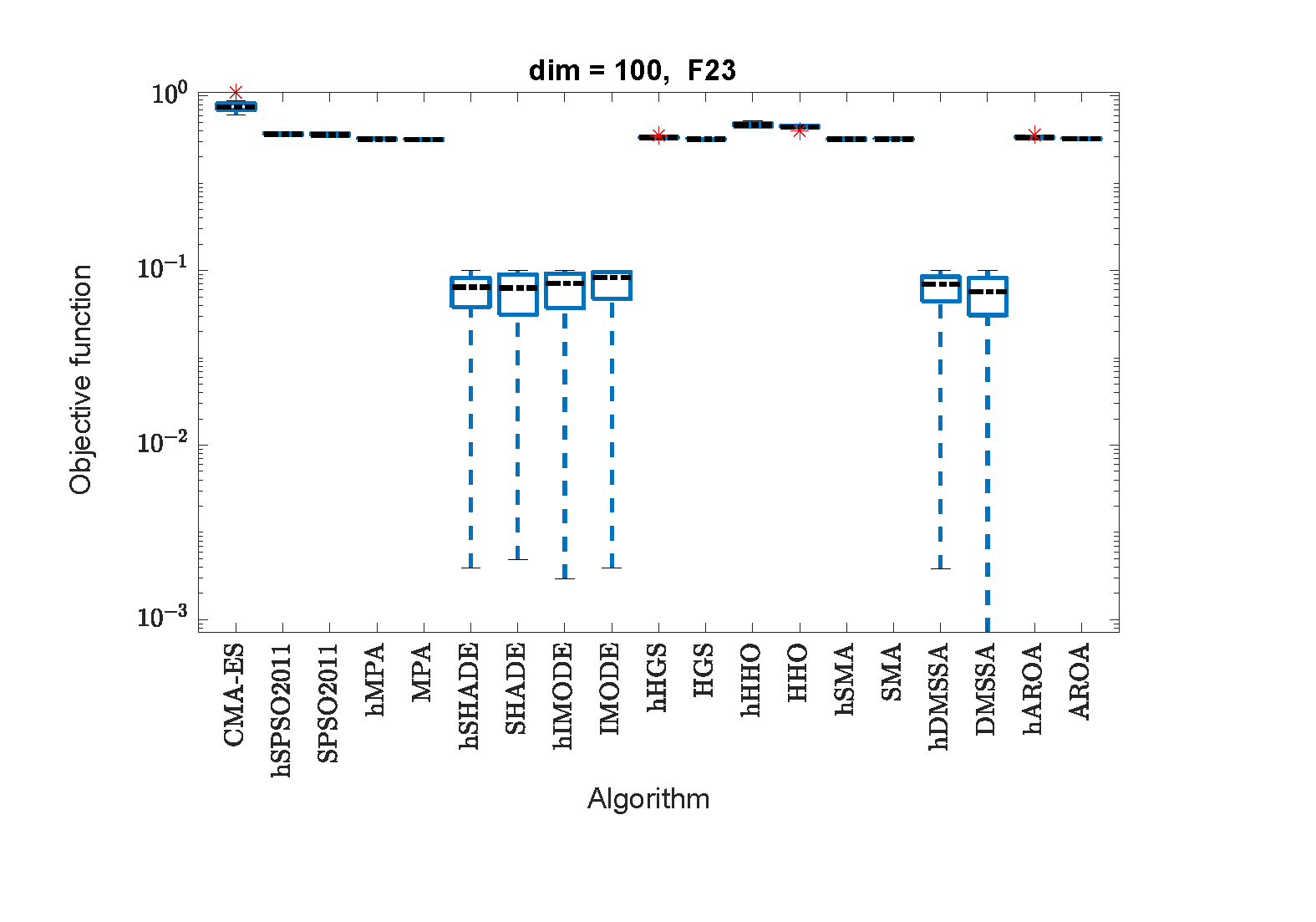}   
  \end{subfigure}
  \end{center}
  \begin{center}
  \begin{subfigure}{0.4\textwidth}
   \centering
    \includegraphics[width=\linewidth]{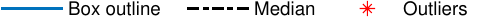}     
  \end{subfigure}
  \end{center}
  \caption{\footnotesize{Boxplots of final results for 10 algorithms and their 9 hybrids on original CEC-2017 functions across 4 dims.}}
  \label{fig1}
\end{adjustwidth}
\end{figure}

\vspace{0.25cm}
\subsection{Analysis of Boxplot Distributions for Representative Functions}
\vspace{0.27cm}
For function f1 (a unimodal function with a smooth landscape), the hybrid algorithms based on differential strategies, particularly hSHADE and hIMODE, consistently produce the best results across all dimensions (Fig.\ref{fig1}). The dispersion of their results remains minimal, emphasizing their robustness and stability. 
In contrast, CMA-ES yields extremely high objective function values across all tested dimensions, frequently accompanied by numerous outliers, which confirms its unsuitability for this type of problem. As the dimensionality $dim$ increases, methods such as MPA and hMPA exhibit wider boxes, indicating an increase in performance variance.

In the case of function f6 (a multimodal function with many local optima), substantial variation in result spread is evident (Fig. \ref{fig1}). Algorithms such as hIMODE and hDMSSA are characterized by compact boxplots with no visible outliers, demonstrating their ability to perform exploration while maintaining stability. At dimension 100, algorithms like hHGS and hHHO exhibit particularly high variance, often generating extreme values. The variability of results generally increases with growing $dim$, as reflected by extended whiskers and the presence of more frequent outliers.

Function f12, which incorporates variable interaction and rotation (Fig.\ref{fig1}), is known for complex inter-variable correlations. This function accentuates the advantage of algorithms with strong exploratory and adaptive components. Here, hSHADE, hIMODE, and hDMSSA demonstrate minimal dispersion 
and significantly lower medians compared to other methods. Occasional outliers from algorithms such as hAROA or hHHO suggest instances of stochastic instability, particularly in high-dimensional cases. The boxplots for CMA-ES remain exceptionally wide, confirming earlier conclusions regarding its low effectiveness in handling rotated landscapes.

Function f23, representing highly challenging landscapes with many local minima (Fig.\ref{fig1}), reveals the full spectrum of optimization difficulty. We observe extensive dispersion and numerous outliers for weaker algorithms. Only hSHADE, hIMODE, and hDMSSA manage to maintain tight distributions, even at $dim = 100$. Their performance medians are markedly lower, reinforcing their clear superiority. Conversely, traditional methods such as PSO and its hybrids display substantial instability, characterized by broader boxes, longer whiskers, and frequent outliers.

For the complete set of 29 functions (excluding f2), a strong correlation is observed between the quantitative results reported in Tab.\ref{tab1} and the structure of the boxplots in Fig.\ref{fig1}. Algorithms such as hSHADE, hIMODE, and hDMSSA maintain narrow boxes and minimal deviations, indicating their distinct advantage. As dimensionality increases, many algorithms-particularly those based on PSO and non-adaptive population-based heuristics-show increasing instability and inconsistent performance.

\section{Theoretical Foundations and Methodology of Ranking and Bayesian Analyses}
\vspace{0.22cm}
To enable a comprehensive comparative evaluation of the performance of optimization algorithms on the CEC-2017 benchmark suite, two complementary classes of statistical analysis methods were employed: classical non-parametric rank-based tests (Friedman test, Nemenyi post-hoc tests, and Critical Difference (CD) diagrams) and Bayesian analysis based on probability matrices of wins and equivalence (Bayes Signed-Rank Test).

\subsection{Rank-Based Statistical Tests: Friedman, Nemenyi, and CD Diagrams}
\vspace{0.22cm}
The analysis began with the computation of average ranks for all algorithms separately for each decision space dimensionality ($dim = 10, 30, 50, 100$) as well as at the meta-analysis level. For each optimization problem (function), algorithms were ranked based on the median value obtained from 30 independent runs. The algorithm with the lowest median received rank 1, the second lowest rank 2, and so forth. These rankings were aggregated into result matrices on which statistical tests were conducted.

In the first step, the Friedman test was applied, which is a non-parametric counterpart of one-way repeated-measures ANOVA. Its purpose was to determine whether statistically significant differences in performance exist among at least two algorithms for a given dimensionality. For all tested dimensions, a p-value < 0.05 was obtained, which justified the continuation with post-hoc tests.

The subsequent analysis used the Nemenyi test, which allows for pairwise comparisons of algorithms to assess whether differences in their average ranks are statistically significant at a given significance level $\alpha = 0.05$. The results were presented in the form of Critical Difference (CD) diagrams, following the procedure described in \cite{0Ck},\cite{00}. The diagrams were generated separately for each dimensionality, with algorithms ordered by their average ranks. Algorithms connected by a horizontal line are not statistically significantly different. Additionally, for each diagram and dimensionality, the CD value was calculated using the formula:
$$CD = q_{\alpha} \cdot \sqrt{\frac{k(k+1)}{6N}},$$
where $k$ is the number of algorithms, $N$ is the number of benchmark functions, and $q_{\alpha}$ is the critical value from the Nemenyi test.
It is worth emphasizing that this analysis is robust to violations of normality and homogeneity of variance assumptions, making it particularly suitable for comparing heuristic algorithms \cite{0CC},\cite{00}.

\subsection{Bayesian Analysis - Bayes Signed-Rank Test and Probability Maps}
\vspace{0.22cm}

To complement the classical ranking-based analysis, a Bayesian approach was applied based on the distribution of pairwise differences in algorithm performance, following the methodology proposed by Benavoli et al.\cite{0jk}. In this method, each pair of algorithms is compared using the distribution of performance differences across 29 benchmark functions, and instead of p-values, three probability measures are analyzed:
\begin{itemize}
\item $P(a_{i} > a_{j} + \text{ROPE})$ - the probability that algorithm $a_{i}$ significantly outperforms $a_{j}$,
\item $P(|a_{i} - a_{j}| \leq \text{ROPE})$ - the probability of practical equivalence between $a_{i}$ and $a_{j}$,
\item $P(a_{i} < a_{j} - \text{ROPE})$ - the probability that $a_{j}$ significantly outperforms $a_{i}$,
\end{itemize}
where $a_{i}$ and $a_{j}$ denote the algorithms being compared.

A ROPE (Region of Practical Equivalence) threshold of 10 was used, indicating that differences smaller than 10 units of the objective function are considered practically irrelevant.

The results are visualized as Bayes maps-color-coded matrices for each algorithm pair, where the color denotes the dominant relationship type (win, loss, equivalence), and the numbers within the cells represent the corresponding probabilities. For clarity, results are presented separately for each dimensionality level ($dim$).

The advantages of this approach include:
\begin{enumerate}
\item Direct interpretability of results in terms of probabilities (rather than p-values),
\item Incorporation of practical relevance through the ROPE threshold,
\item Robustness against issues arising from multiple comparisons and false discoveries.
\end{enumerate}

Both the classical and Bayesian approaches are complementary. CD and Friedman tests offer a clear ranking perspective and highlight groups of statistically indistinguishable algorithms, whereas Bayesian analysis enables a more nuanced interpretation of algorithm dominance in practical terms. Both analyses were implemented and automated in MATLAB, fully aligned with methodologies described in the literature \cite{0jk},\cite{0Ck},\cite{00}.

\subsection{Analysis of Critical Difference (CD) Diagrams for Average Algorithm Rankings}
\vspace{0.25cm}
To identify statistically significant differences in the performance of the analyzed algorithms, the Friedman test was applied along with the Nemenyi post-hoc test. The results are presented using (CD) diagrams. These plots provide a graphical comparison of the average rankings achieved by the algorithms across the full benchmark function set. Connecting lines between algorithms indicate the absence of statistically significant differences between them at a significance level of $\alpha = 0.05$, with the computed critical difference equal to CD = 5.11.
\vspace{0.25cm}
\begin{figure}[H]
  \centering
  \begin{subfigure}{0.36\textwidth} 
        \includegraphics[height=5cm,width=6.5cm]{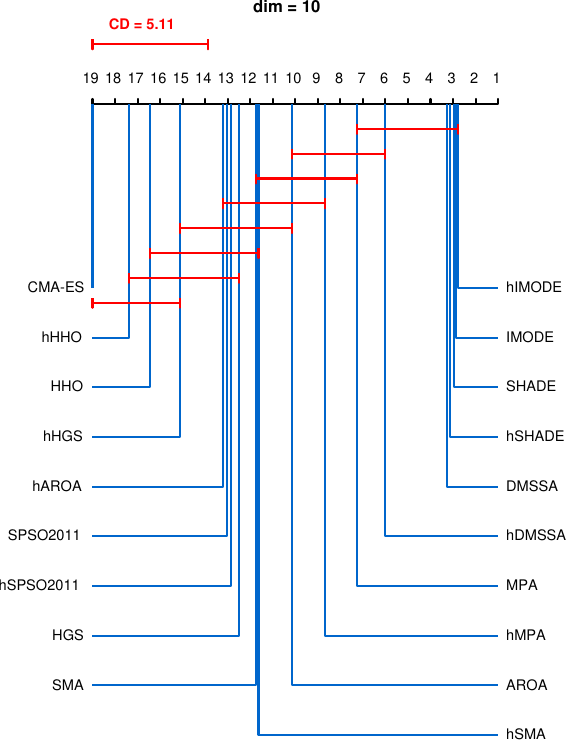}
  \end{subfigure}\hfill
  \begin{subfigure}{0.32\textwidth}
    \includegraphics[height=5cm,width=6.5cm]{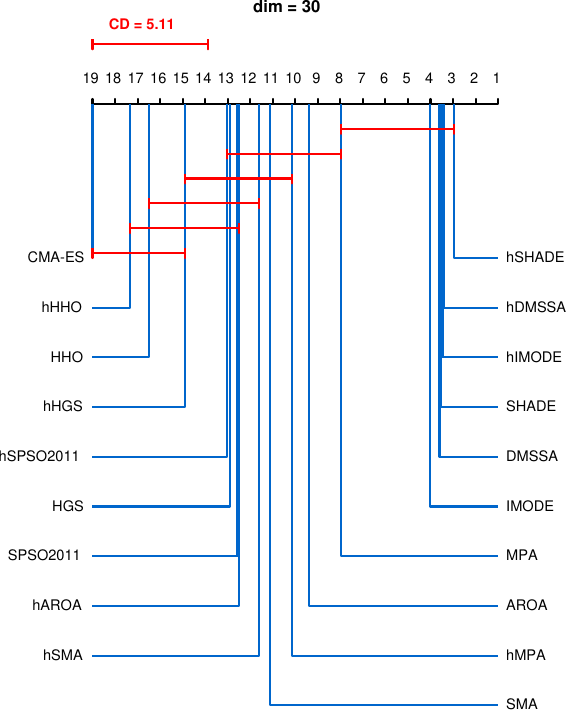}
     \end{subfigure}\hfill
\end{figure}
\begin{figure}[H]
\vspace{0.35cm}
\centering
 \begin{subfigure}{0.36\textwidth}
    \includegraphics[height=5cm,width=6.5cm]{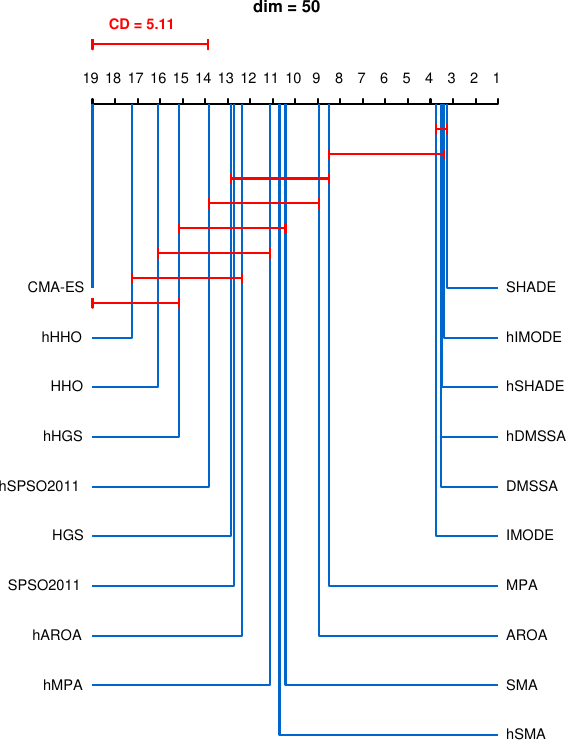}
  \end{subfigure}\hfill
  \begin{subfigure}{0.48\textwidth}
    \includegraphics[height=5cm,width=6.5cm]{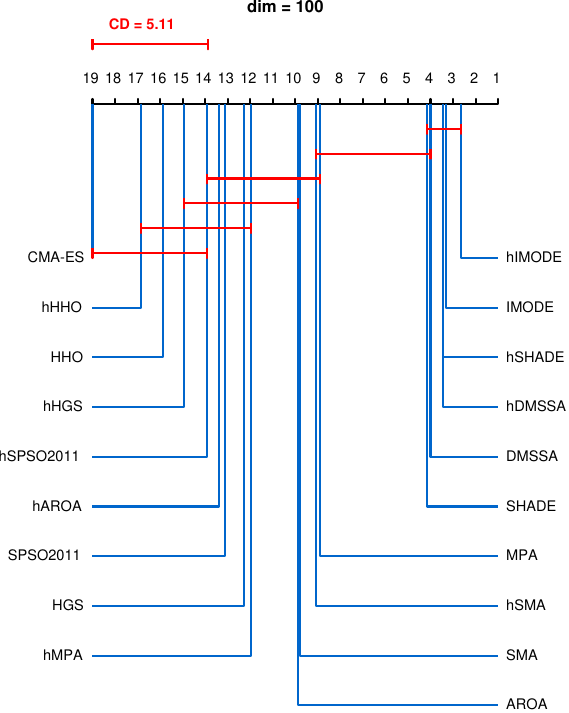}
  \end{subfigure}\hfill
  \caption{\footnotesize{Critical difference diagrams for 10 algorithms and 9 hybrids on original CEC-2017 functions across 4 dims.}}
  \label{fig2}
\end{figure}

\normalsize

Analysis of CD-Friedman Test Results for the Baseline $f(x)$.\\

For $dim = 10$ (Fig.\ref{fig2}), a clear dominance of hybrid algorithms was observed: hIMODE, hSHADE, hDMSSA, and IMODE. These algorithms appear on the left side of the diagram, achieving the lowest average ranks, without any statistical ties to the group of less effective methods. In contrast, the classical CMA-ES, 
along with hHHO and hHGS, ranked at the bottom of the chart, indicating limited effectiveness in low-dimensional decision spaces. The distinct separation between groups points to statistically significant differences.

As the dimensionality increases to $dim = 30$ (Fig.\ref{fig2}), the dominance of hybrid methods persists: hSHADE, hDMSSA, hIMODE, and SHADE consistently occupy the top of the rankings, without any statistical connection to lower-performing methods. CMA-ES and swarm-based algorithms (including hSPSO2011) continue to yield the weakest performance. Algorithms such as MPA, SMA, hMPA, and AROA remain in the middle range of the ranking, forming a homogeneous group without significant advantage over the leading methods.

For $dim = 50$ (Fig.\ref{fig2}), the number of connections between algorithms in the central part of the ranking increases, which may indicate reduced sensitivity of the test in the presence of increased variance in results. Nevertheless, SHADE, hIMODE, and hSHADE retain their top positions, remaining statistically distinct from the group of weaker algorithms. CMA-ES again occupies the lowest rank, with a ranking gap exceeding the CD compared to the top-performing methods.

The highest analyzed dimensionality $dim = 100$ (Fig.\ref{fig2}) confirms the continued superiority of adaptive hybrid approaches. Algorithms such as hIMODE, IMODE, hSHADE, SHADE, and hDMSSA maintain their status as the most competitive methods. Significant differences remain noticeable in comparison to algorithms such as CMA-ES, hHHO, and HGS, which are located on the far right of the diagram and show no statistical links to the top-performing methods. Despite the increased problem difficulty, adaptive methods maintain high stability and effectiveness.

The CD plots across all dimensions clearly indicate the superiority of hybrid algorithms based on differential and adaptive strategies. Their high ranking positions, absence of statistical connections to classical methods, and consistent performance across varying dimensionalities demonstrate their universality and effectiveness. Simultaneously, the results reveal limited scalability and poor adaptability of traditional approaches such as CMA-ES and swarm intelligence - based algorithms (e.g., SPSO2011), particularly in high-dimensional decision spaces.

\subsection{Bayesian Test Result Analysis}
\vspace{0.22cm}
To enable a deeper comparative evaluation of the effectiveness of the analyzed optimization algorithms, Bayesian tests were conducted for four levels of decision space dimensionality: 10, 30, 50, and 100. Heatmap matrices were employed, in which each cell represents the posterior probability that the algorithm in the row outperforms the one in the column, taking into account the Region of Practical Equivalence (ROPE).
\vspace{0.25cm}

\begin{figure}[H] 
\adjustbox{scale=1.25,center}{
\centering
  \begin{subfigure}[t]{0.495\textwidth}
    \centering
\includegraphics[width=\linewidth]{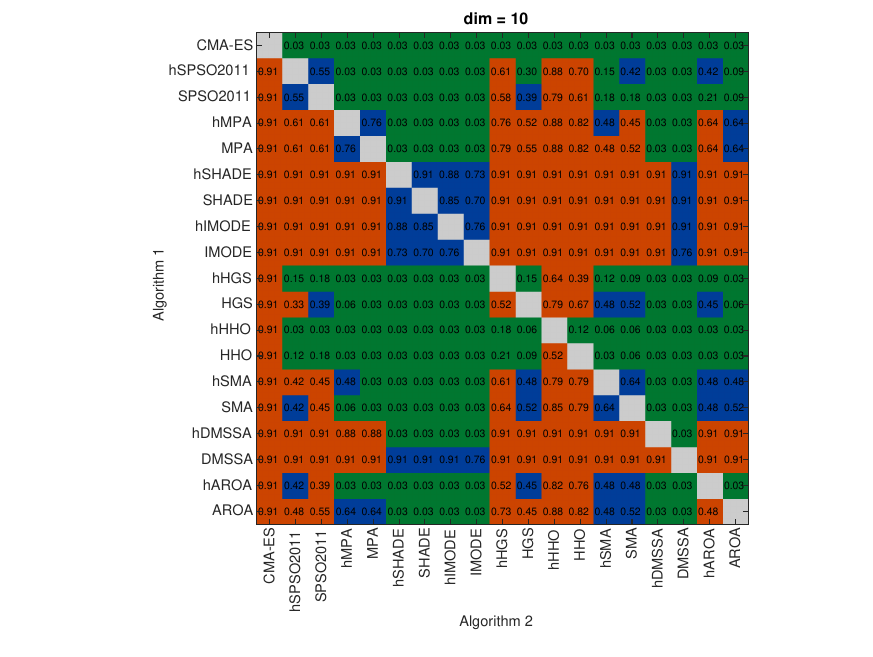}
  \end{subfigure}
  \begin{subfigure}[t]{0.495\textwidth}
    \centering
\includegraphics[width=\linewidth]{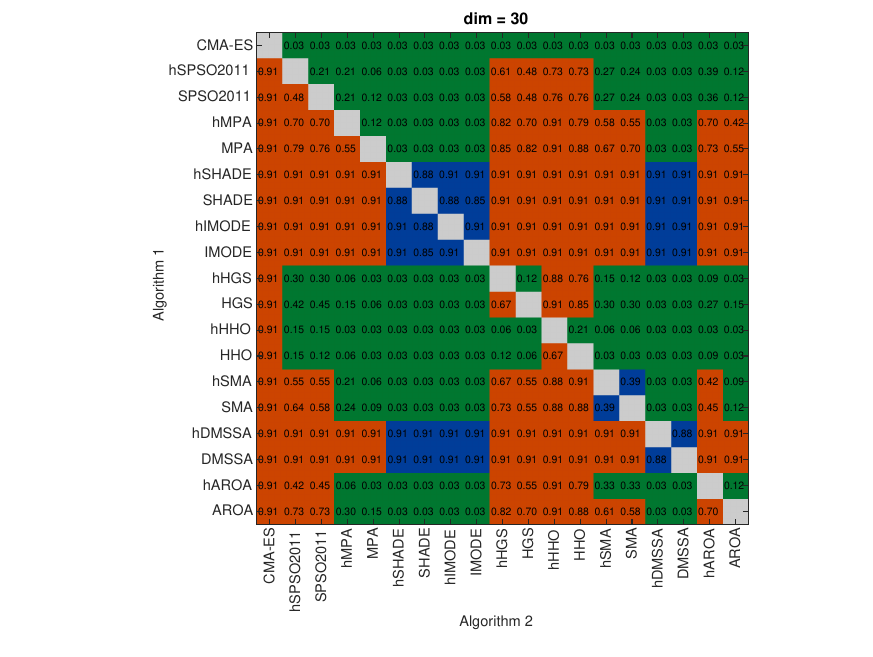}
  \end{subfigure}
  }
  \end{figure}
  \vspace{0.2cm}
  \begin{figure}[H] 
\adjustbox{scale=1.25,center}{
\begin{subfigure}[t]{0.495\textwidth}
    \centering
\includegraphics[width=\linewidth]{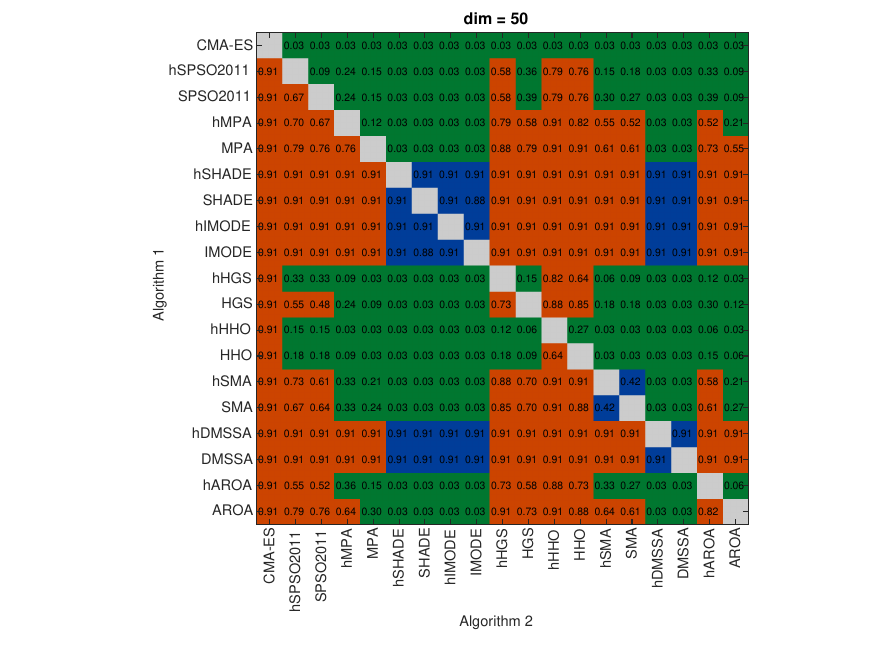}
  \end{subfigure}
 \begin{subfigure}[t]{0.495\textwidth}
    \centering
\includegraphics[width=\linewidth]{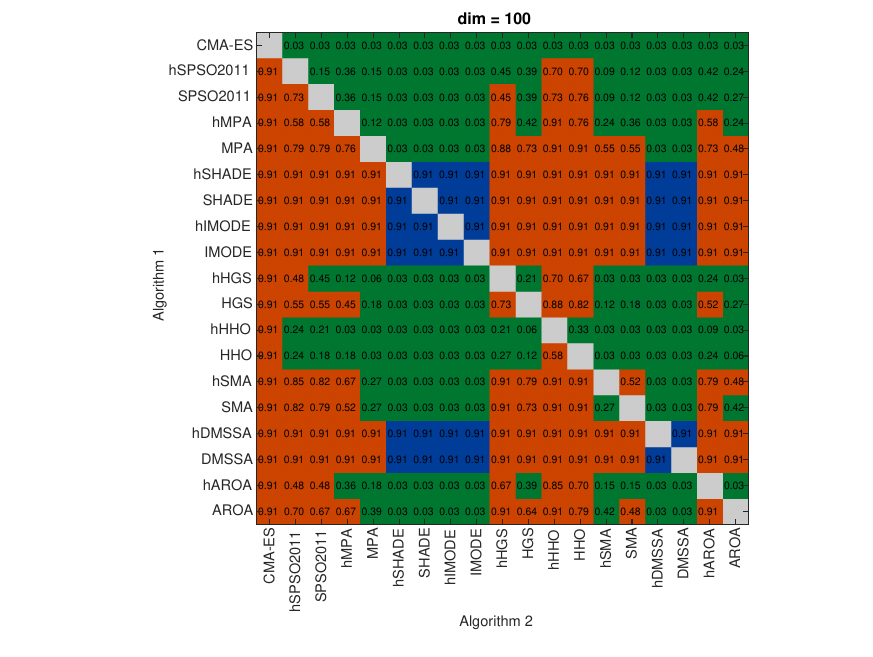}
  \end{subfigure}
  }
  \vspace{0.2cm}
  \adjustbox{scale=1.1,center}{
\begin{subfigure}{0.6\textwidth}
    \centering
    \includegraphics[width=\linewidth]{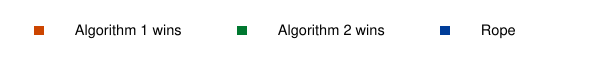}
  \end{subfigure}
  }
  \vspace{-0.75cm}
 \caption{\small Bayesian test results for 10 algorithms and their 9 hybrids on original CEC-2017 functions across 4 dims.}
  \label{fig3}
  \end{figure}

For the lowest dimensionality $dim = 10$ (Fig.\ref{fig3}), a clear dominance of the group of hybrid algorithms was observed: hIMODE, hSHADE, hMPA, hSMA, and hDMSSA consistently outperformed the remaining methods, achieving performance values around  $\sim\!0.03$ against most competitors. Their superiority was particularly evident when compared to their classical counterparts (IMODE, SHADE, MPA), strongly indicating the effectiveness of hybridization. In contrast, CMA-ES and HHO proved highly inefficient, losing in almost every comparison.
As the dimensionality increased to 30 (Fig.\ref{fig3}), the dominance of hybrid methods persisted. hIMODE and hSHADE continued to exhibit the highest winning probabilities, often without suffering a single loss. It is worth noting a slight flattening of performance differences compared to the lower dimension - for example, MPA and DMSSA improved their relative positions, whereas classical methods remained significantly less effective. CMA-ES and HHO again occupied the lowest ranks.
At dimensionality 50 (Fig.\ref{fig3}), performance differences between algorithms began to narrow slightly, yet the overall dominance pattern remained consistent with lower-dimensional cases. hSHADE, hIMODE, and hDMSSA retained their leading positions, winning the vast majority of pairwise comparisons, often with probabilities close to 0.03. MPA showed growing competitiveness, particularly in comparisons with classical representatives of the PSO and ES families (namely SPSO2011 and CMA-ES). Meanwhile, CMA-ES, HHO, and SPSO2011 continued to demonstrate very poor effectiveness. Hybridization remains a consistently effective strategy, even under increasing decision space complexity.

For the highest tested dimensionality $dim = 100$ (Fig.\ref{fig3}), the persistent superiority of hybrid variants was further confirmed. In particular, hIMODE, hSHADE, and hDMSSA continued to be the most effective algorithms, achieving the lowest values in nearly all pairwise comparisons. Importantly, even in this challenging configuration, no significant drop in their performance was observed - a strong indication of good scalability. CMA-ES and HHO maintained their declining trend, consistently underperforming relative to the majority of methods, regardless of dimensionality.

The Bayesian analysis strongly reinforces the conclusions drawn from the CD test: hybrid methods not only achieve superior results, but do so in a statistically and probabilistically consistent manner. The dominance of hIMODE, hSHADE, hDMSSA, and hMPA remains robust across the entire examined range of dimensionalities. Moreover, the Bayesian test highlights the stability of algorithm behavior as problem complexity increases, making it a valuable tool for drawing conclusions about the practical reliability and applicability of the evaluated approaches.

\vspace{-0.14cm}
\subsection{Convergence Trajectory Analysis of the Algorithms}
\vspace{-0.12cm}
To complement the comparative evaluation of the optimization algorithms, an assessment of their temporal behavior was conducted using convergence plots (Fig.\ref{fig4}). This analysis allows for the evaluation not only of the final solution quality but also of the exploration dynamics of the search space, the rate of convergence, and the stability of performance in the context of increasing dimensionality and the diverse nature of the objective functions (f1, f6, f12, f23).

\vspace{0.4cm}
\begin{figure}[H]
  \centering
\begin{adjustwidth}{-1cm}{-2.0cm}
  \begin{subfigure}{0.35\textwidth}
     \includegraphics[height=4cm,width=5.5cm]{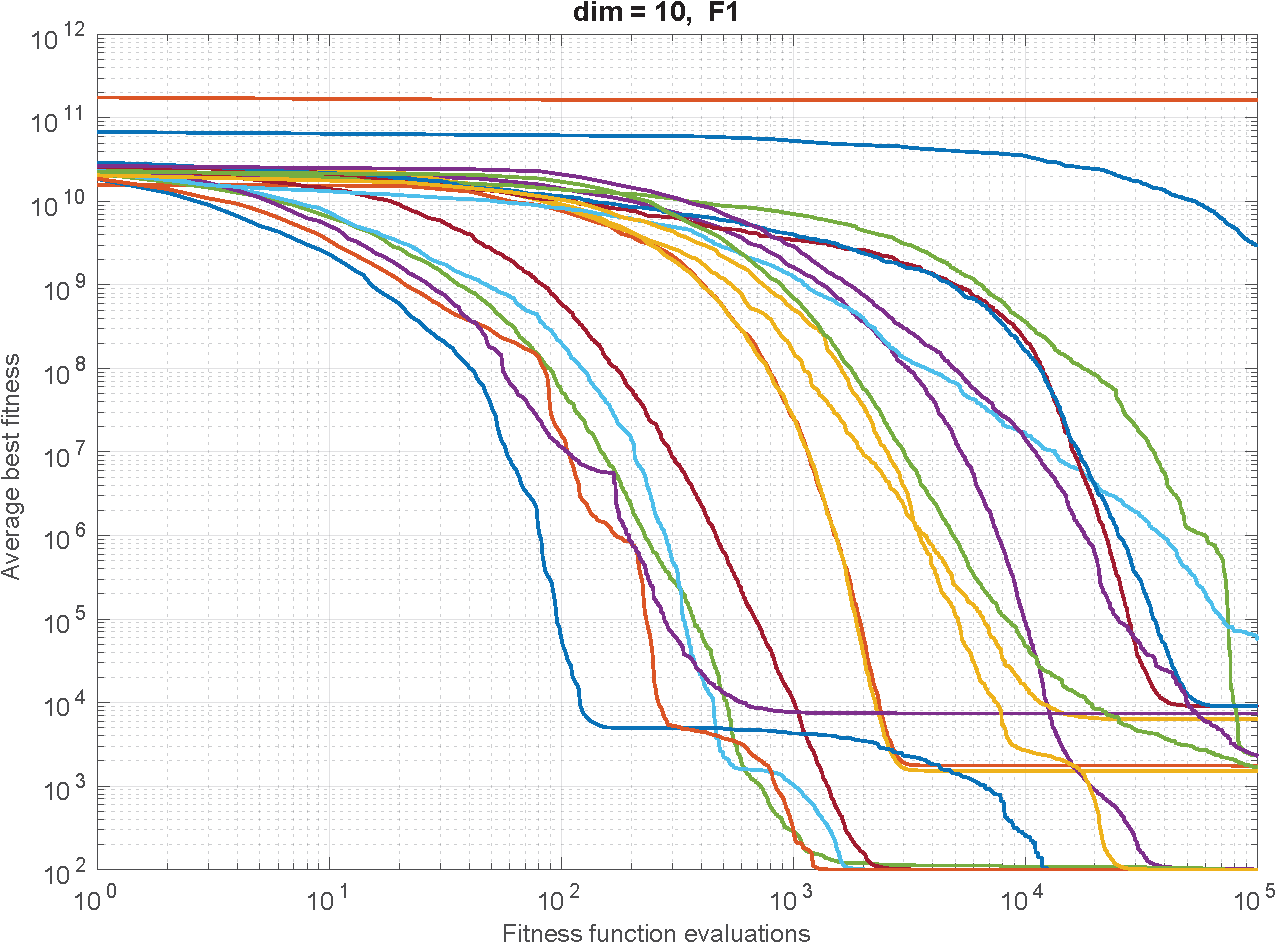}
  \end{subfigure}\hfill
  \begin{subfigure}{0.35\textwidth}
    \includegraphics[height=4cm,width=5.5cm]{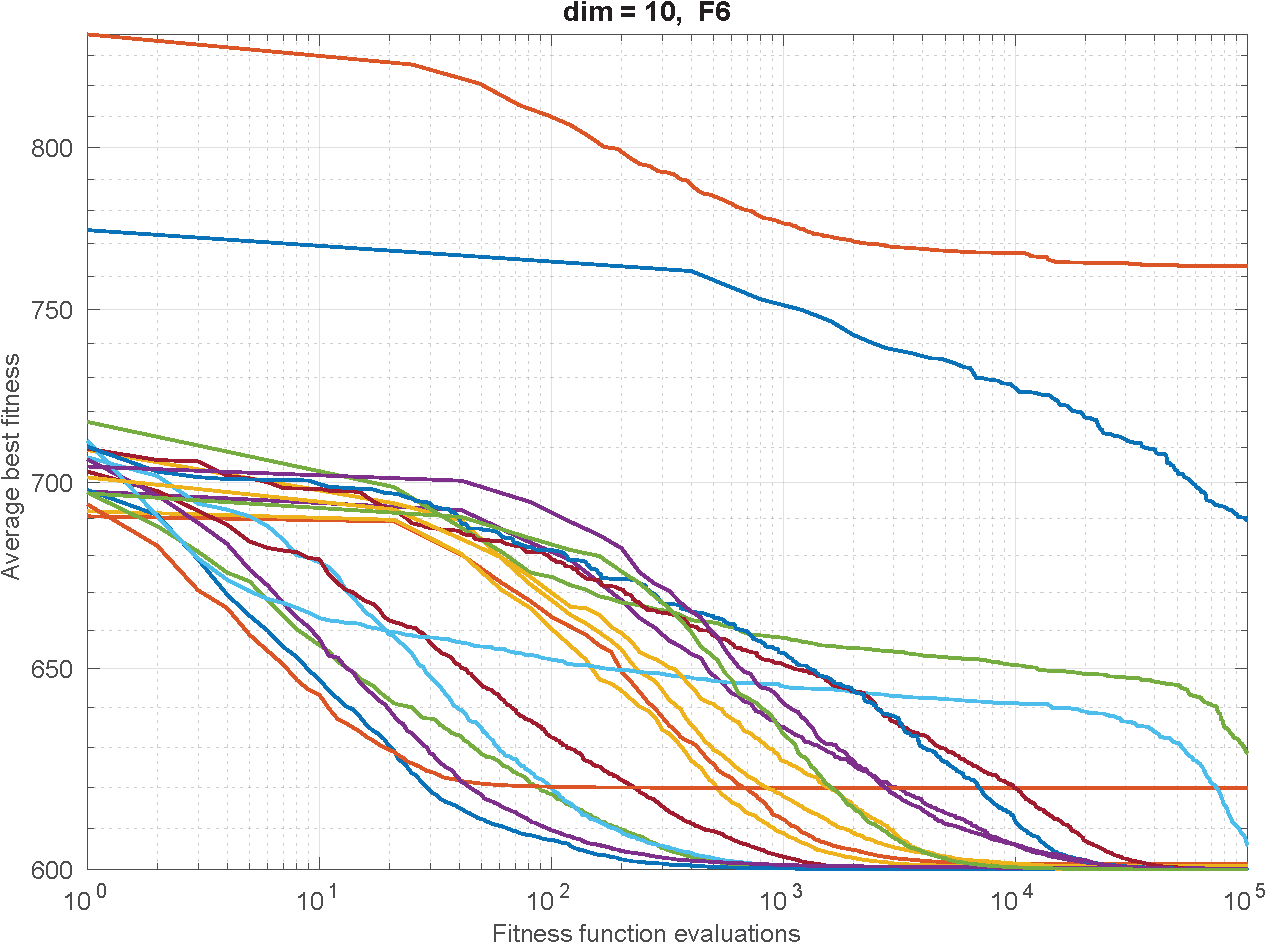}
  \end{subfigure}\hfill
  \begin{subfigure}{0.35\textwidth}
    \includegraphics[height=4cm,width=5.5cm]{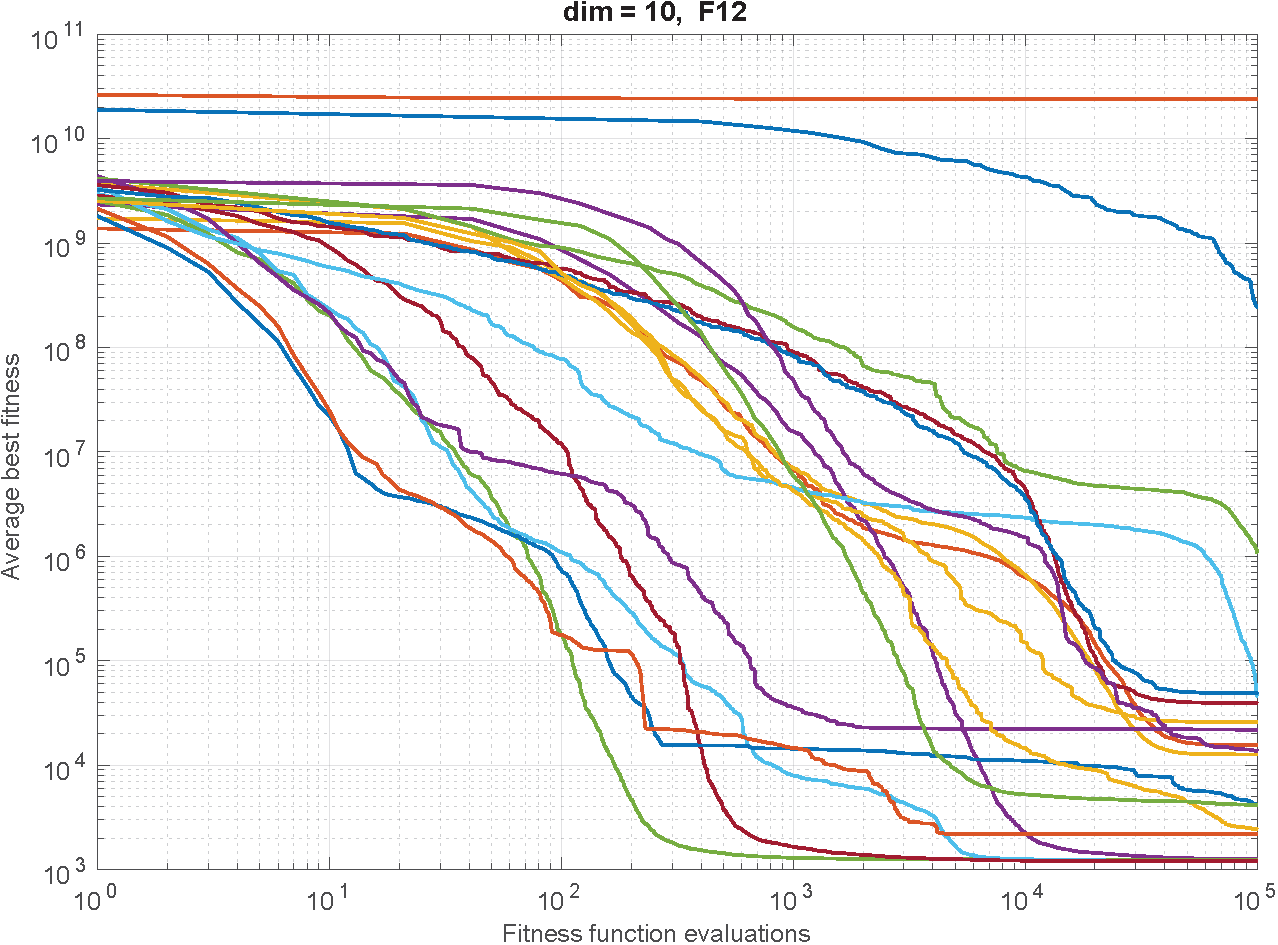}
  \end{subfigure}
 \end{adjustwidth}
\end{figure} 

\begin{figure}[H]
  \centering
\begin{adjustwidth}{-1cm}{-2.0cm}
   \begin{subfigure}{0.35\textwidth}
    \includegraphics[height=4cm,width=5.5cm]{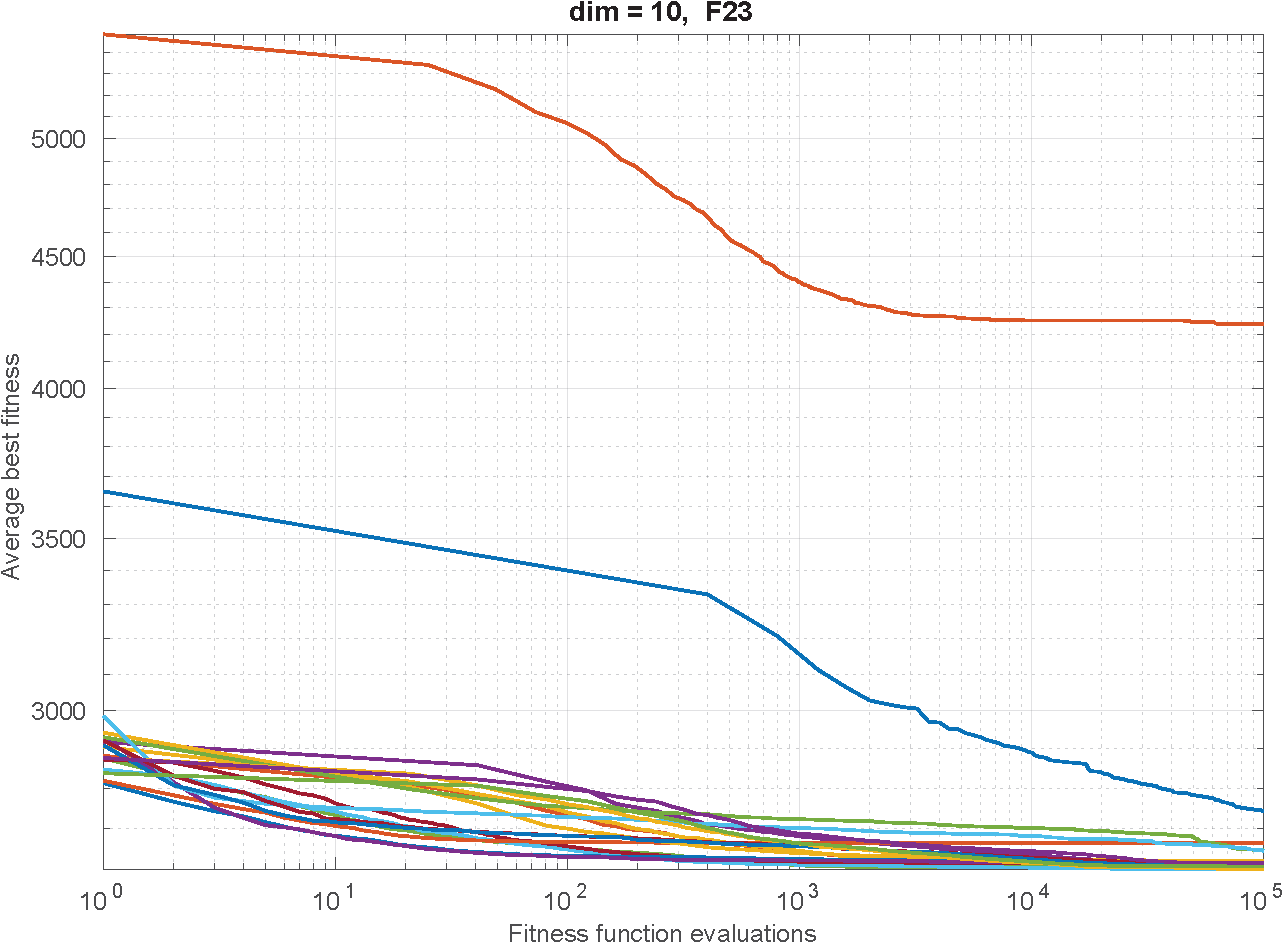}
  \end{subfigure}\hfill
  \begin{subfigure}{0.35\textwidth}
    \includegraphics[height=4cm,width=5.5cm]{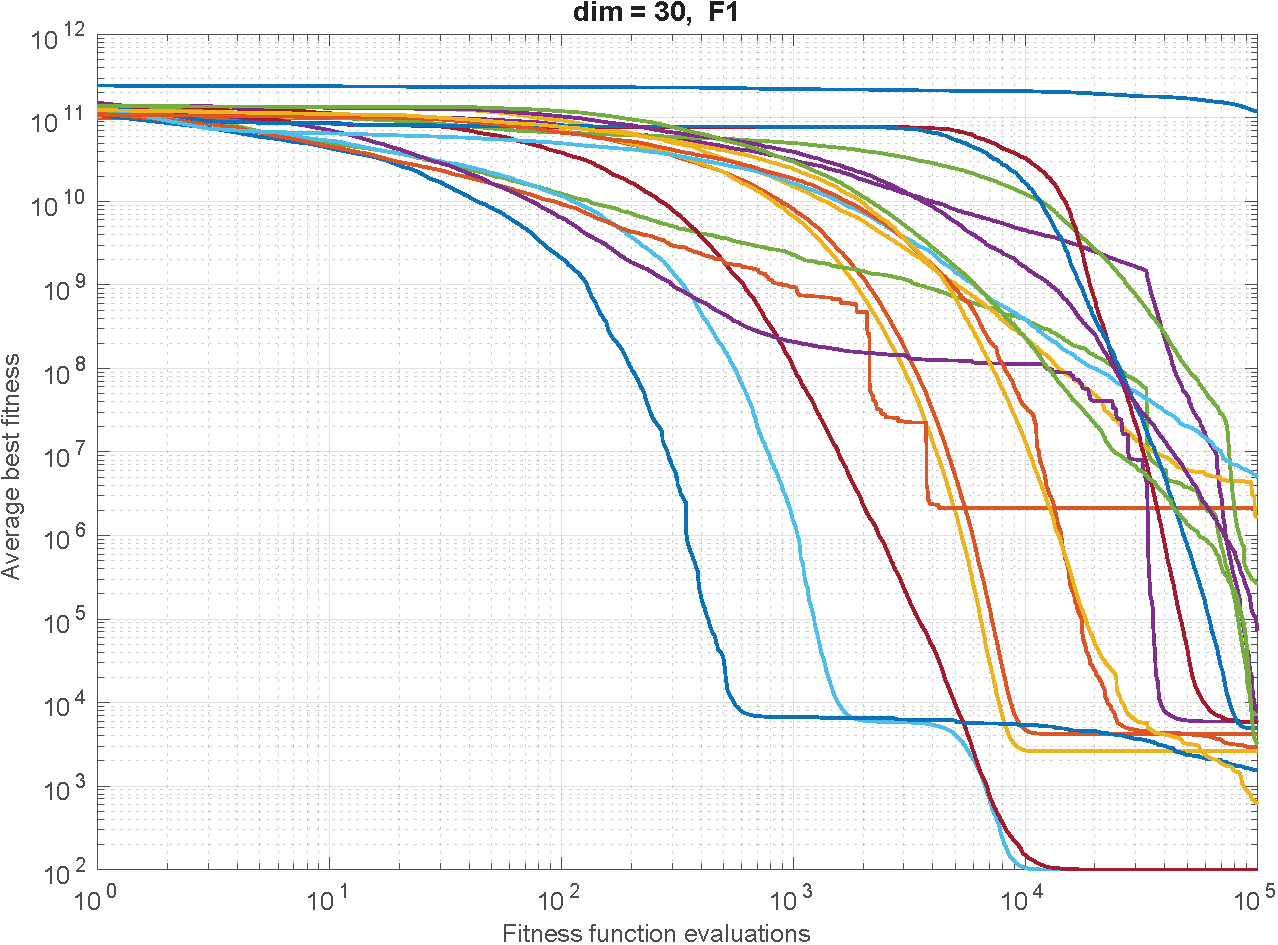}
\end{subfigure}\hfill
  \begin{subfigure}{0.35\textwidth} 
        \includegraphics[height=4cm,width=5.5cm]{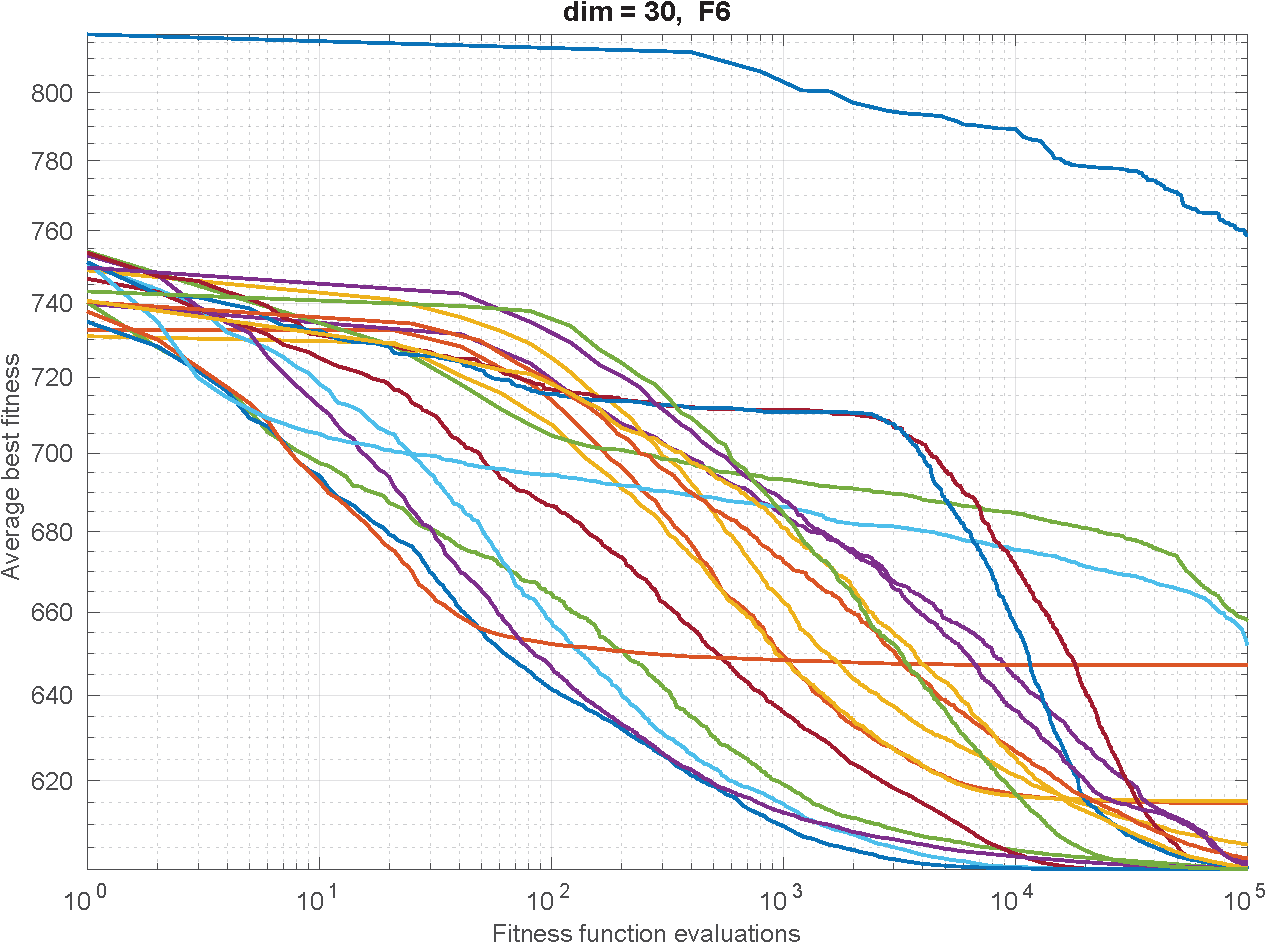}
   \end{subfigure}
   \end{adjustwidth}
\end{figure}
   
 \begin{figure}[H]
  \centering
\begin{adjustwidth}{-1cm}{-2.0cm}
   \begin{subfigure}{0.35\textwidth}
        \includegraphics[height=4cm,width=5.5cm]{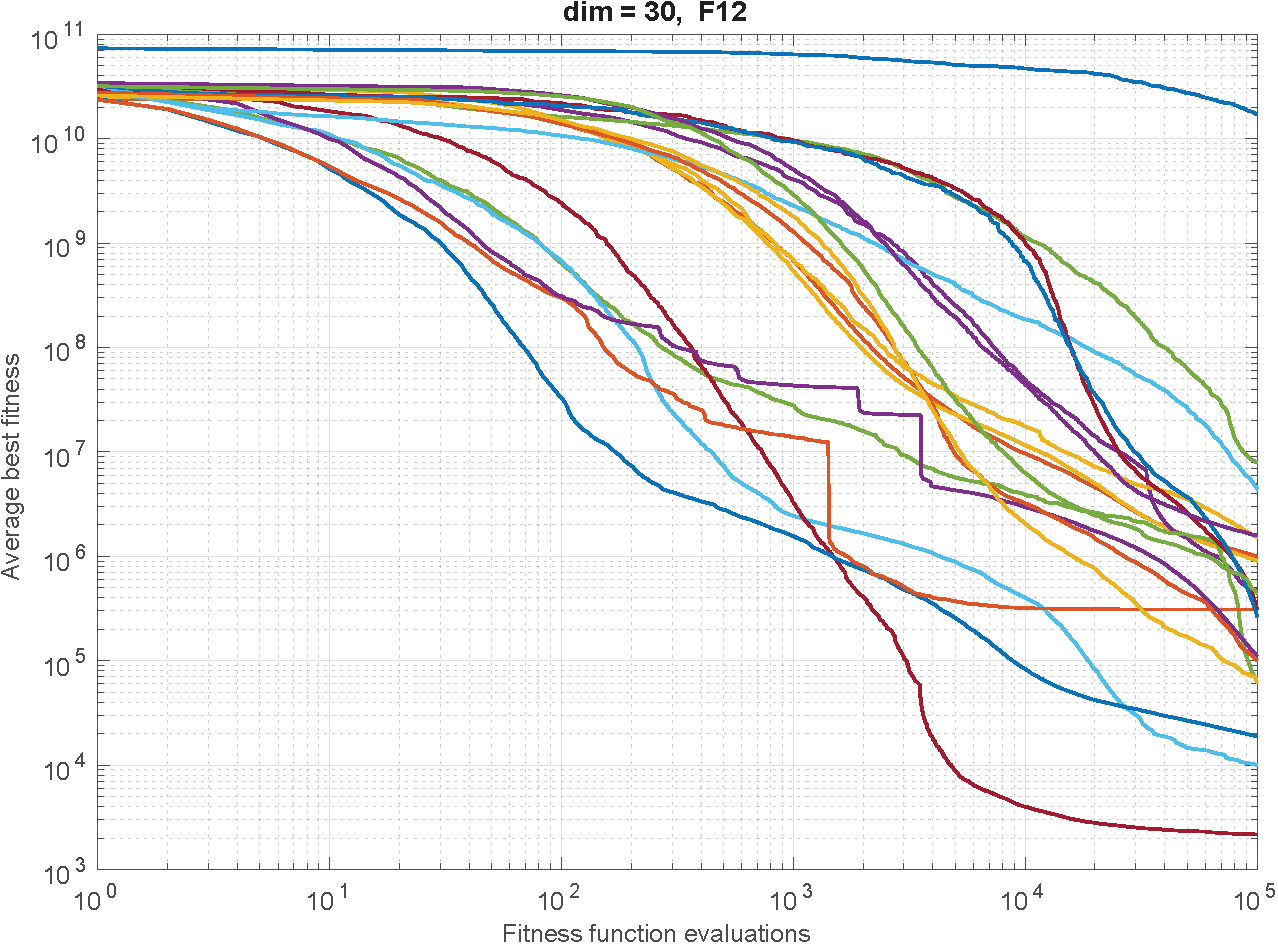}
  \end{subfigure}\hfill
  \begin{subfigure}{0.35\textwidth} 
        \includegraphics[height=4cm,width=5.5cm]{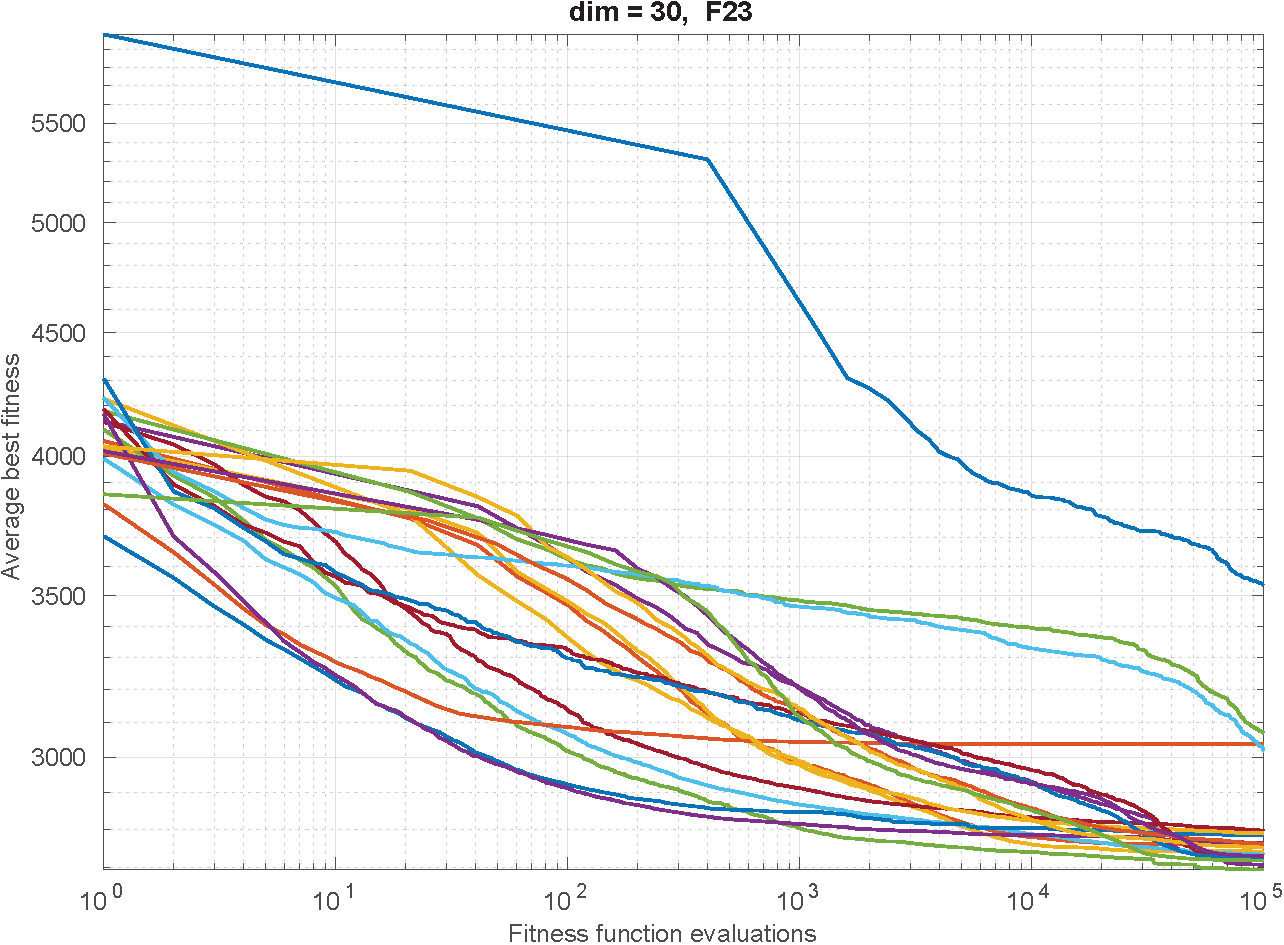}
  \end{subfigure}\hfill
  \begin{subfigure}{0.35\textwidth}
        \includegraphics[height=4cm,width=5.5cm]{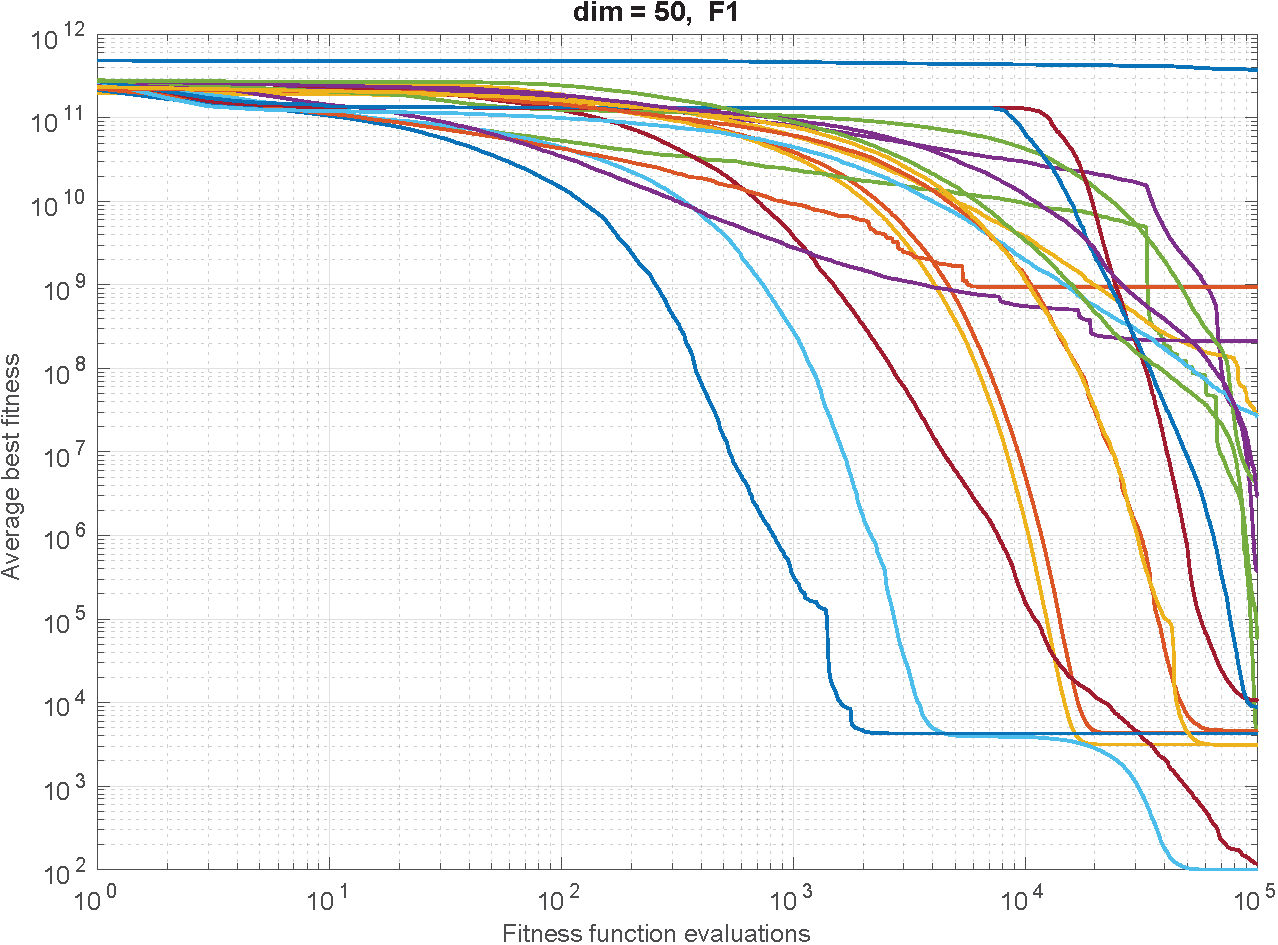}
  \end{subfigure}
  \end{adjustwidth}
\end{figure}

  \begin{figure}[H]
  \centering
\begin{adjustwidth}{-1cm}{-2.0cm}
  \begin{subfigure}{0.35\textwidth} 
        \includegraphics[height=4cm,width=5.5cm]{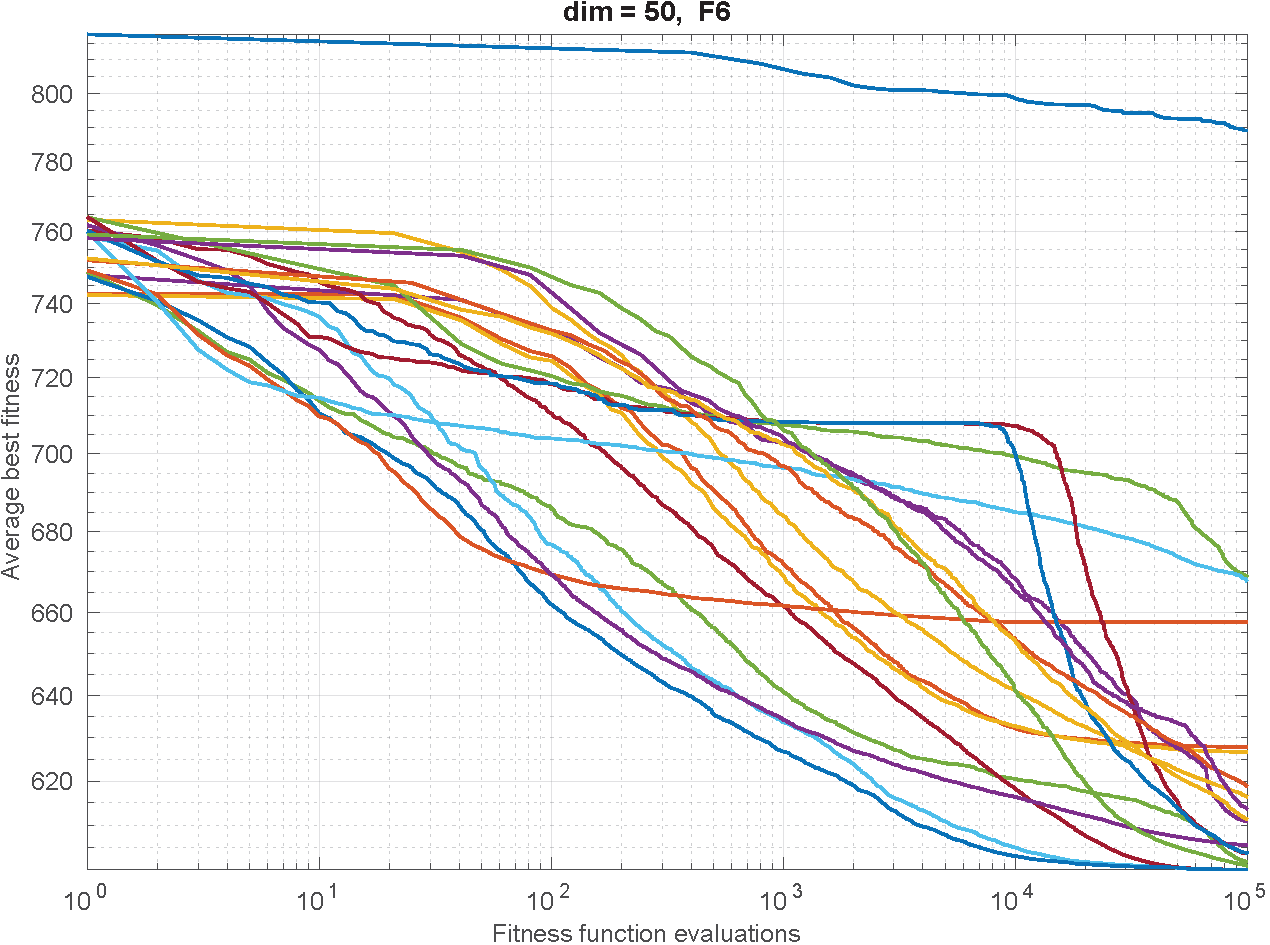}
  \end{subfigure}\hfill
  \begin{subfigure}{0.36\textwidth}
    \includegraphics[height=4cm,width=5.5cm]{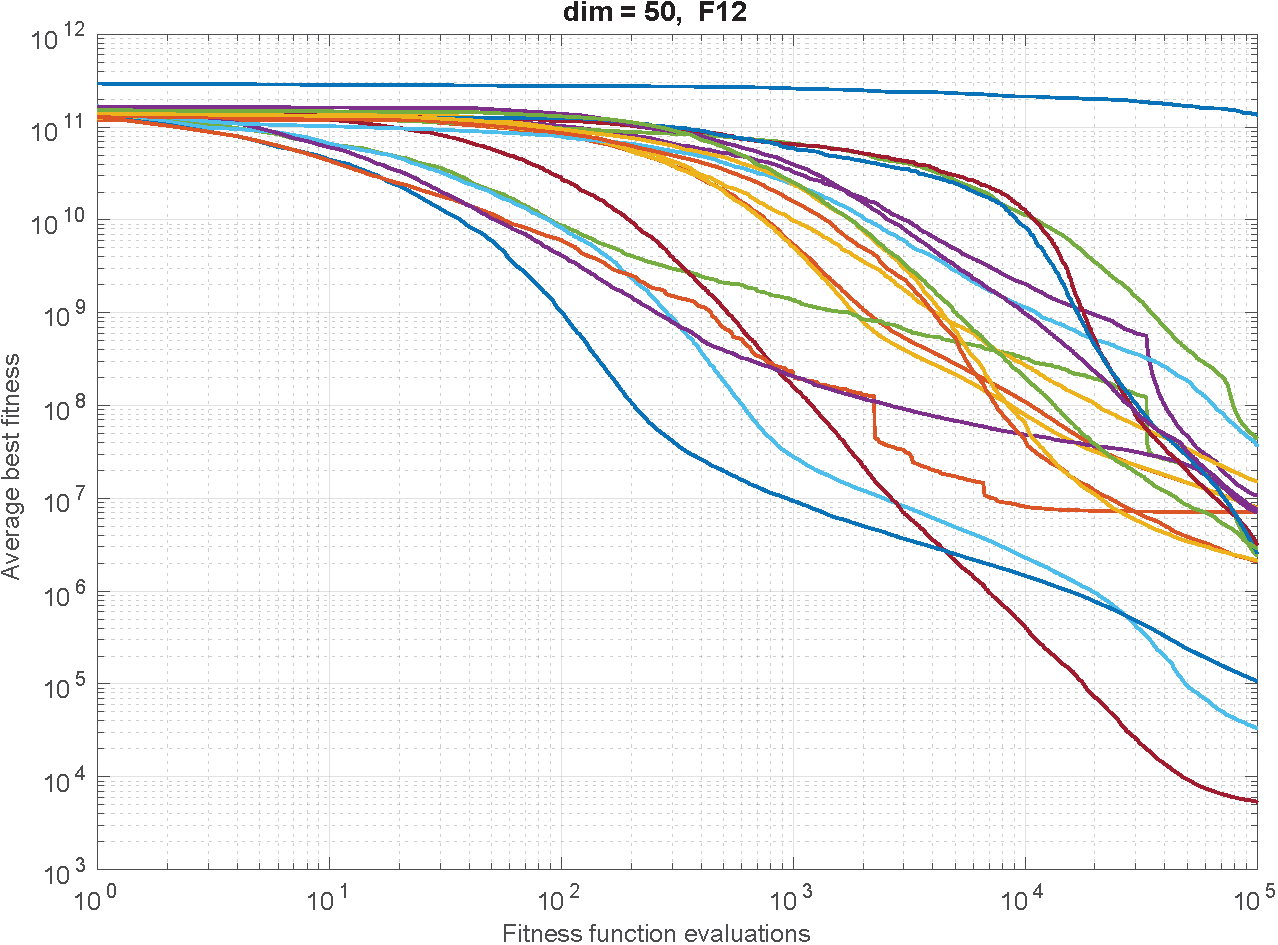}
     \end{subfigure}\hfill
 \begin{subfigure}{0.35\textwidth}
    \includegraphics[height=4cm,width=5.5cm]{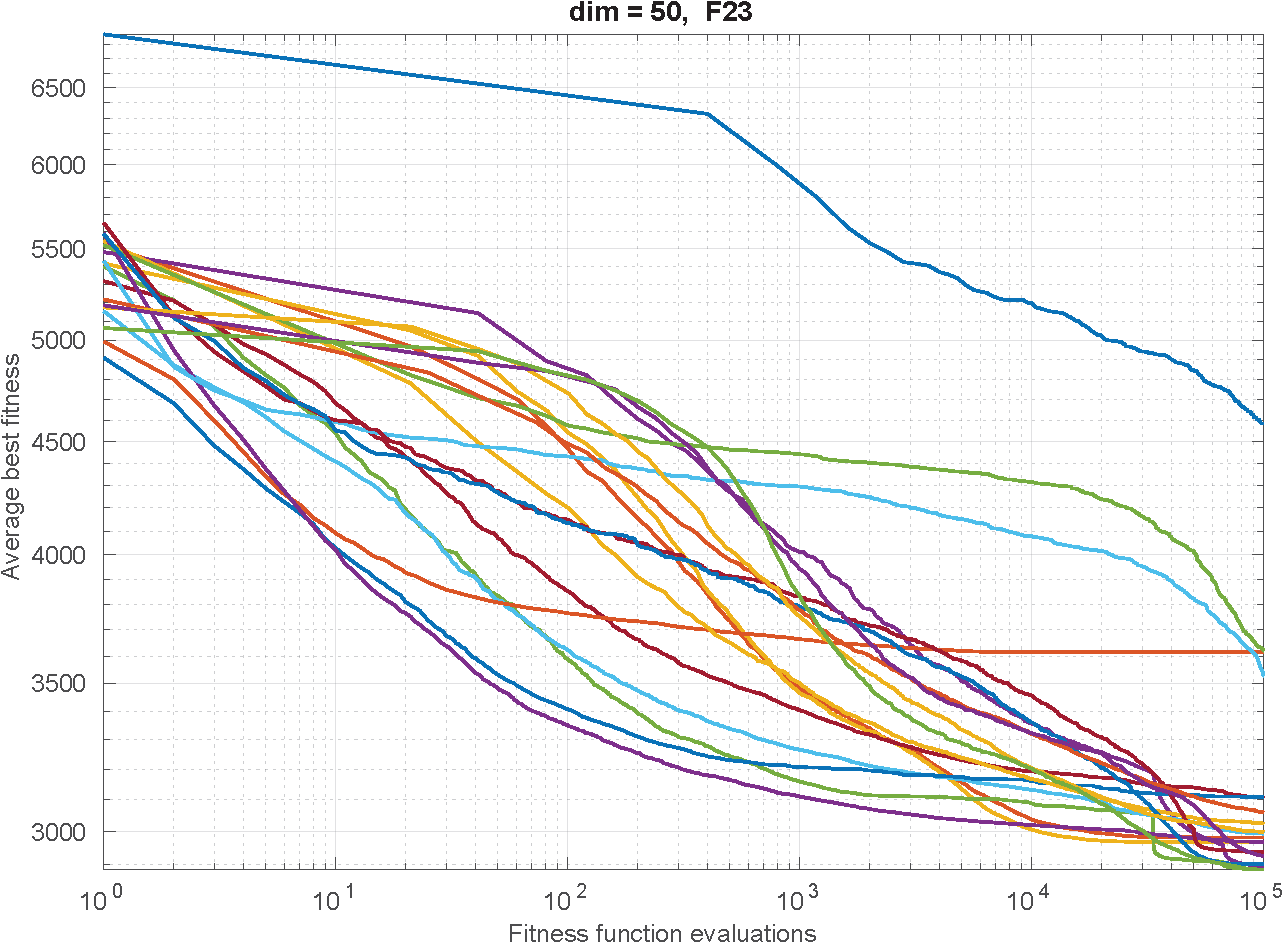}
  \end{subfigure}
\end{adjustwidth}
\end{figure}
 
  \begin{figure}[H]
  \centering
\begin{adjustwidth}{-1cm}{-2.0cm}
  \begin{subfigure}{0.35\textwidth}
    \includegraphics[height=4cm,width=5.5cm]{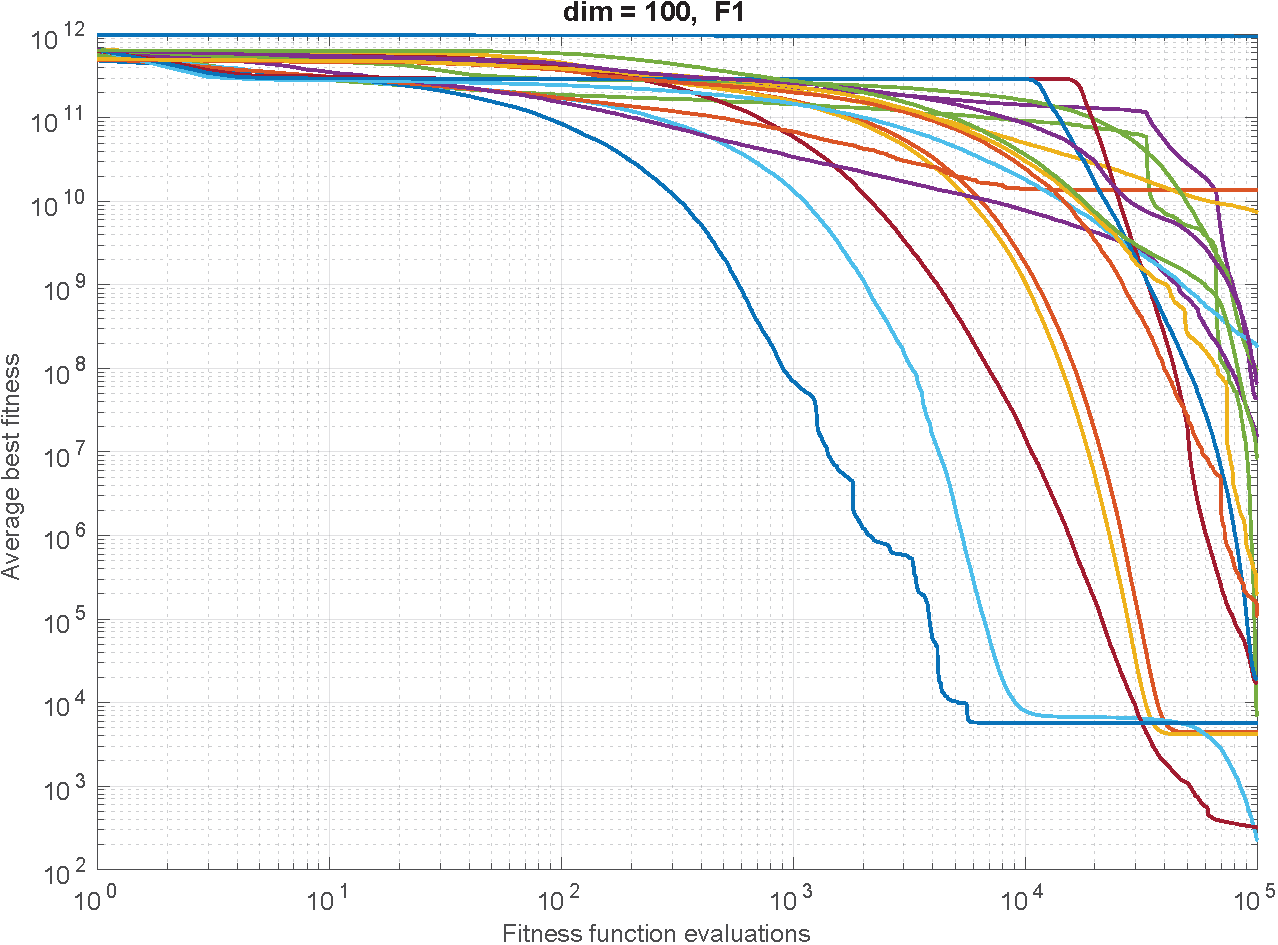}
  \end{subfigure}\hfill
  \begin{subfigure}{0.35\textwidth}
        \includegraphics[height=4cm,width=5.5cm]{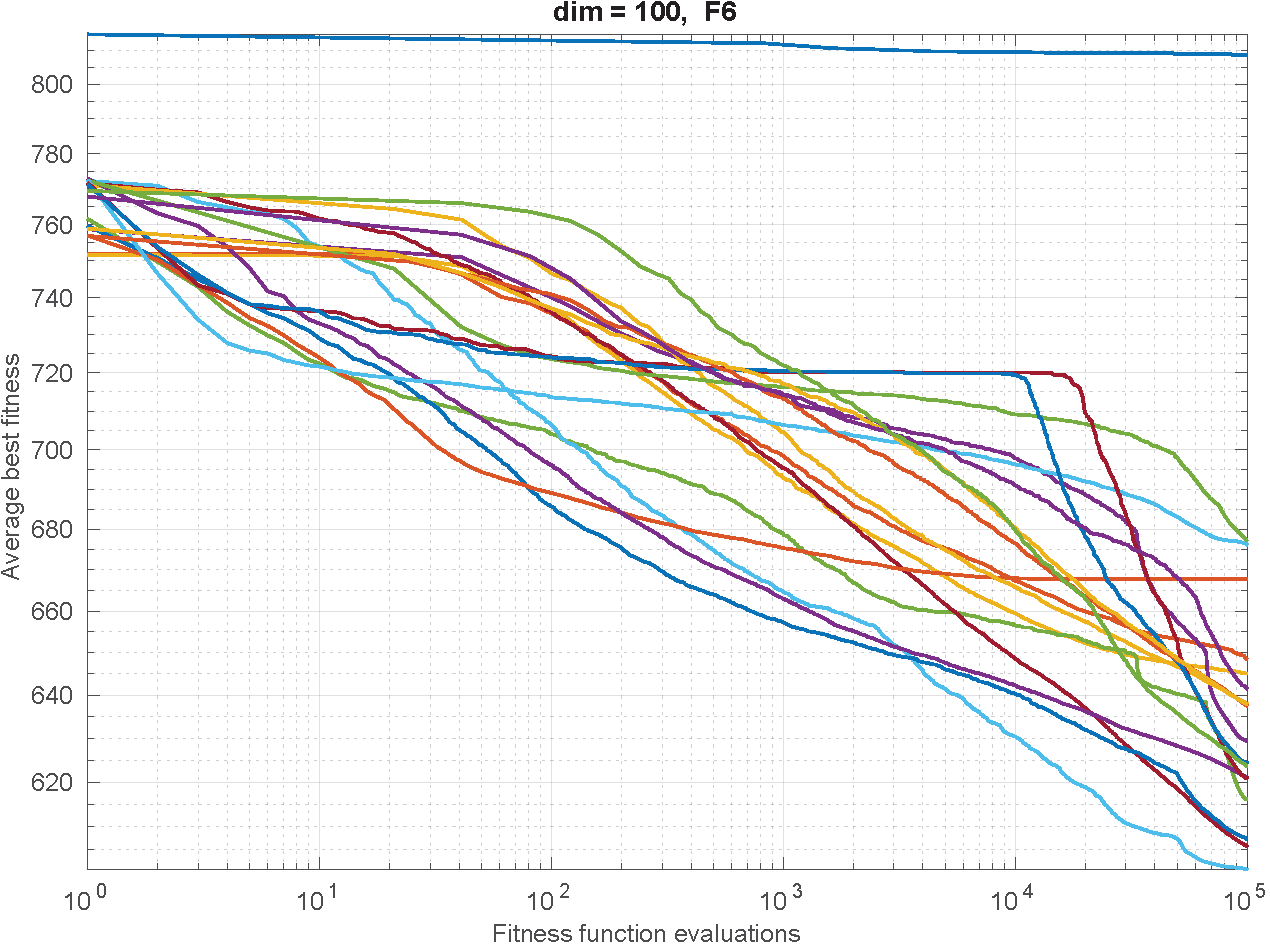}
     \end{subfigure}\hfill
   \begin{subfigure}{0.35\textwidth}
        \includegraphics[height=4cm,width=5.5cm]{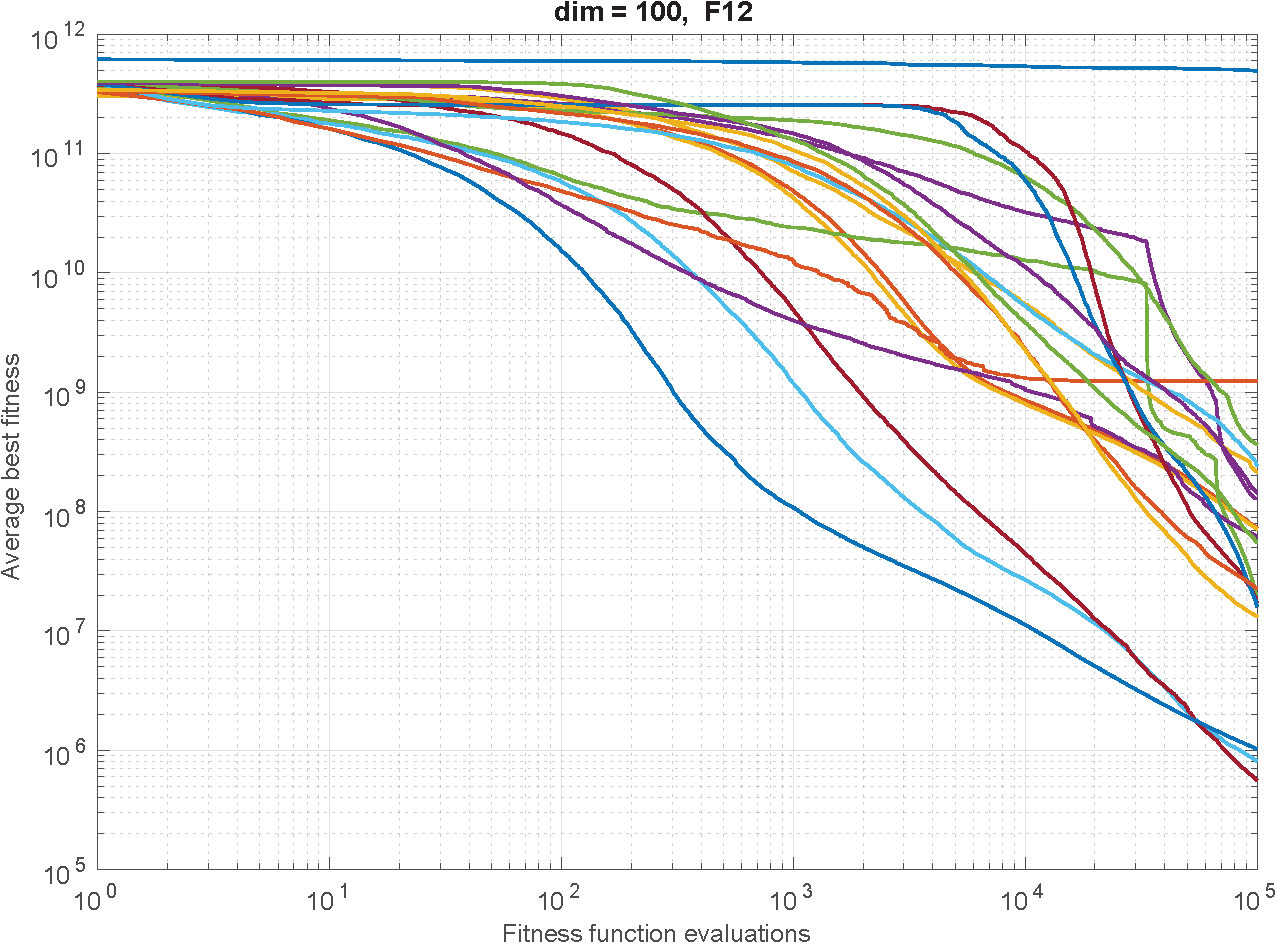}
    \end{subfigure}
    \end{adjustwidth}
    \end{figure}

\begin{figure}[H]
 \centering
\begin{adjustwidth}{-1.4cm}{-2.0cm}
\begin{center}
  \begin{subfigure}{0.35\textwidth}
   \centering
        \includegraphics[height=4cm,width=5.5cm]{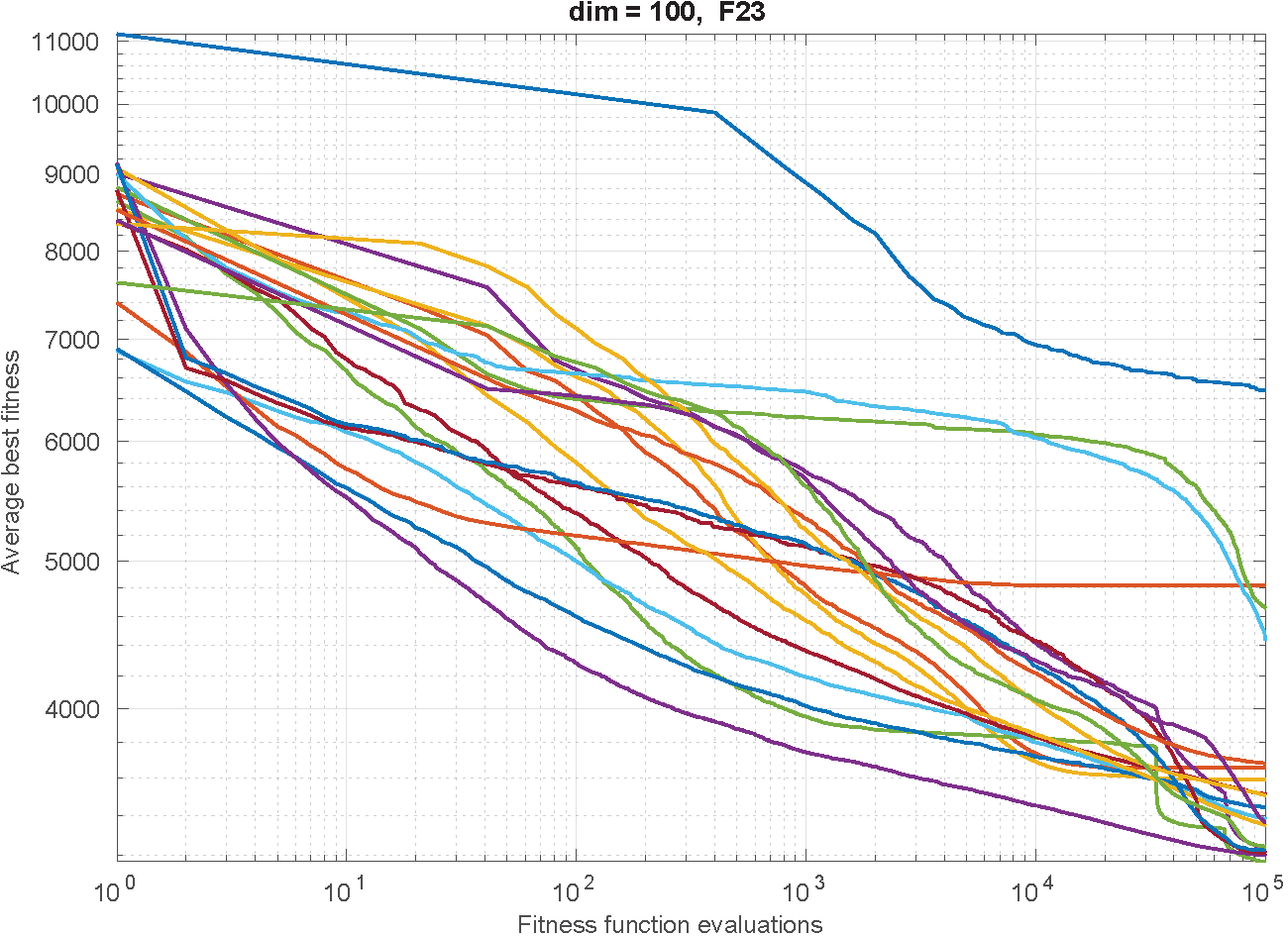}   
  \end{subfigure}
  \end{center}
  \begin{center}
  \begin{subfigure}{0.85\textwidth}
   \centering
    \includegraphics[height=0.75cm, width=\linewidth]{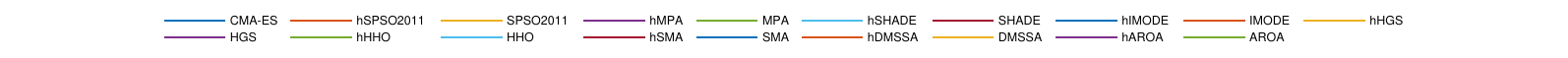}     
  \end{subfigure}
  \end{center}
  \vspace{-0.15cm}
  \caption{\small{Convergence trajectories of metaheuristic algorithms on a representative CEC-2017 benchmark function.}}
  \label{fig4}
\end{adjustwidth}
\end{figure}

\vspace{0.12cm}
Convergence Characteristics and Analysis.
\vspace{0.12cm}

In the case of f1 (Fig.\ref{fig4}), a very rapid convergence rate is observed for most hybrid algorithms, particularly hSHADE, hIMODE, and hMPA, which reach very low objective function values already in the initial phase of the run. Their trajectories are smooth, monotonic, and stable. Increasing the dimensionality to 100 naturally slows down the convergence rate; however, the overall curve layout remains intact - the most effective methods remain the same. In contrast, CMA-ES and HHO converge significantly more slowly and often settle into a plateau, indicating poor exploration efficiency for this class of functions.

For f6 (Fig.\ref{fig4}), the differences in the algorithms’ ability to exploit the valley-shaped surface become pronounced. hIMODE and hSHADE display a longer initial exploration phase, but their convergence accelerates after overcoming the initial stagnation period. hDMSSA and hMPA also achieve low objective values, though their trajectories exhibit greater variability. In dimensions 50 and 100, some methods (e.g., MPA, SMA) show signs of difficulty in navigating the topology - becoming trapped in local minima. Classical variants such as IMODE, SHADE, and CMA-ES converge very slowly or fail to break out of stagnation altogether.

For f12 (Fig.\ref{fig4}), the analyzed algorithms confront challenges stemming from intensive locality and disrupted gradients. hSHADE and hIMODE retain the capacity for stable exploration and relatively smooth convergence, albeit not as consistently as in the f1 case. Increased instability is evident in the trajectories of hMPA and DMSSA, which display oscillations or stalls in some dimensions. Meanwhile, CMA-ES and HHO tend to ''drown'' early in local minima - their plots are flat or even flattened, with only marginal progress even over extended iterations. Dimensionality 100 significantly degrades the performance of non-hybrid methods.

Function f23 (Fig.\ref{fig4}) represents a case of a highly complex surface with many local minima and irregularities. Here, differences between algorithms become most pronounced. hIMODE and hSHADE are the only ones demonstrating continued convergence even at dimension 100, while most other methods show early stagnation or chaotic jumps. SMA, HHO, CMA-ES, and SPSO2011 fail to overcome the landscape structure, and their plots remain nearly flat. hDMSSA and hMPA retain partial exploratory capability, but their convergence is significantly less stable and predictable compared to the top-performing methods.

\vspace{0.1cm}
Conclusions from Convergence Analysis.
\vspace{0.041cm}
The analysis of convergence trajectories reveals significant differences in the operational behavior of individual algorithms. Hybrid methods exhibit not only higher final optimization efficiency but also more consistent and predictable convergence dynamics-regardless of the type of objective function or its dimensionality. Their trajectories are smooth, monotonic, and resilient to increased topological complexity.

In contrast, classical variants (including CMA-ES, HHO, SMA, and SPSO2011) frequently demonstrate stagnation, instability, or premature convergence-particularly on functions with rich local structure. Convergence analysis not only complements the statistical performance picture but also uncovers qualitative differences in the algorithms’ ability to explore and regulate the pace of optimization.

\FloatBarrier

\section{Invariance of Optimization Algorithms - Theoretical and Practical Perspective}
In practical applications of metaheuristics, a critical property of optimization algorithms is their invariance with respect to transformations of the decision space and the objective function. Invariance implies that the algorithm’s performance should remain unaffected by elementary transformations of the optimization problem. Algorithms exhibiting high universality are expected to be independent of the coordinate system used, the location of the global minimum, and the scale of objective function values.

According to the literature \cite{4aijk},\cite{6atz}, the most frequently distinguished types of transformations are as follows:\\
a) Translation (shift of the reference system): $f(x) \rightarrow f(x+a)$, where $a \in \mathbb{R}^n$ (where $\mathbb{R}$ is the set of real numbers, and $n \in \mathbb{N} \setminus \{0\}$ is the set of natural numbers excluding zero) - assessment of the algorithm’s robustness to relocation of the optimum within the search space,\\
b) Scaling (change of scale): $f(x) \rightarrow f(\alpha x)$, where $\alpha \in \mathbb{R}$ - evaluation of sensitivity to unit differences across dimensions,\\
c) Rotation (rotation of the decision space): $f(x) \rightarrow f(xM)$, where $M \in \mathbb{R}^{n \times n}$ is an orthonormal matrix - assessment of independence from the orientation of coordinate axes,\\
d) Vertical shift (addition of a constant to the objective function): $f(x)\rightarrow f(x) + c$, where $c \in \mathbb{R}$ - verification of whether absolute function values influence the search strategy.\\

As previous studies indicate, many widely used algorithms - including PSO and classical variants of DE - do not exhibit full invariance, which can result in substantial differences in optimization performance following even simple geometric or scaling transformations. For instance, standard PSO is not invariant to rotation of the decision space, leading to performance degradation when optimizing non-separable functions \cite{4aijk}.

In this chapter, a detailed analysis is conducted on the influence of the four aforementioned transformations on the behavior of the studied algorithms. Each case (translation, scaling, rotation, and vertical shift of the objective function) is presented in a separate subsection and subjected to a comprehensive evaluation, including comparisons of objective function values, ranking analysis CD, statistical testing (Bayesian methods), convergence plots, and distribution visualization (boxplots).

The purpose of this analysis is to assess the robustness of algorithms in the face of decision space and objective function deformations, and to identify the mechanisms that determine the stability or sensitivity of a given approach.

\vspace{0.2cm}
\subsection{Translation Invariance}
\vspace{0.1cm}
Translation of the decision space, defined as a shift of the objective function argument by a constant vector $a \in \mathbb{R}^n$ ($f(x) \rightarrow f(x + a)$), does not alter the structure of the objective surface but changes the location of the optimum within the search space. Algorithms that fulfill translation invariance are expected to maintain consistent performance regardless of such a modification.

Although the literature suggests that methods based on internal population relationships (e.g., DE, CMA-ES, PSO) exhibit robustness to this type of transformation \cite{4aijk},\cite{6atz}, practical implementations may reveal subtle deviations due to factors such as initialization procedures, memory components, or predefined reference points. This subsection evaluates the translation invariance of the examined methods by analyzing their performance under the transformation $f(x)\rightarrow f(x + a)$, with $a = 8$, in order to assess the impact of constant decision space displacement on optimization quality and dynamics.

\scriptsize
\setlength{\tabcolsep}{4pt}

\begin{table}[H]
\centering
\begin{adjustwidth}{-1.5cm}{-2.4cm}
\caption{\footnotesize{Statistical results for 9 hybrid algorithms on CEC-2017 functions and their translated variants across 4 dims.}}
\scalebox{0.67}{
\renewcommand{\arraystretch}{0.9}
\begin{tabular}{@{}lrrrrrrrrrrrrrr@{}}
\toprule
\multirow{2}{*}{Algorithm} & \multicolumn{7}{c}{dim=10} & \multicolumn{7}{c}{dim=30} \\
\cmidrule(lr){2-8} \cmidrule(lr){9-15}
 & Avg & Med & Std & Sum Rank & Mean Rank & +/- & p-value & Avg & Med & Std & Sum Rank & Mean Rank & +/- & p-value \\
\midrule
hSPSO2011\_f(x+8)  & 9.0E+03 & 1.9E+03 & 1.2E+04 & 259.5 & 8.9 & 239/199 & 4.4E-02 & 8.4E+04 & 2.9E+03 & 5.1E+04 & 301 & 10.4 & 195/245 & 5.4E-02 \\
hSPSO2011  & 1.6E+04 & 1.9E+03 & 2.0E+04 & 253.5 & 8.7 & 247/194 & 3.8E-02 & 5.9E+04 & 2.8E+03 & 3.9E+04 & 289 & 10.0 & 206/239 & 5.0E-02 \\
hMPA\_f(x+8)  & 2.3E+10 & 2.9E+04 & 1.3E+01 & 401.5 & 13.8 & 100/360 & 1.2E-02 & 1.7E+15 & 1.6E+05 & 7.4E+10 & 454.5 & 15.7 & 52/400 & 3.8E-02 \\
hMPA  & 2.6E+03 & 1.6E+03 & 5.2E+03 & 184 & 6.3 & 325/153 & 1.4E-02 & 1.9E+04 & 2.3E+03 & 1.7E+04 & 235 & 8.1 & 270/196 & 2.0E-02 \\
hSHADE\_f(x+8)  & 9.7E+01 & 9.3E+01 & 5.3E+00 & 119 & 4.1 & 383/59 & 6.9E-02 & 9.7E+01 & 9.2E+01 & 5.8E+00 & 140.5 & 4.8 & 354/91 & 6.2E-02 \\
hSHADE  & -6.1E-01 & 1.4E+00 & 4.7E+01 & 47 & 1.6 & 464/2 & 4.0E-02 & -2.2E-01 & 4.5E-01 & 5.2E+01 & 55 & 1.9 & 436/2 & 6.5E-02 \\
hIMODE\_f(x+8)  & 9.7E+01 & 9.3E+01 & 5.3E+00 & 113 & 3.9 & 377/63 & 7.1E-02 & 9.7E+01 & 9.2E+01 & 5.8E+00 & 144.5 & 5.0 & 353/91 & 6.5E-02 \\
hIMODE  & -1.5E+00 & 1.0E+00 & 5.1E+01 & 40 & 1.4 & 466/0 & 4.0E-02 & -2.3E-01 & 1.3E+00 & 5.5E+01 & 62 & 2.1 & 438/0 & 6.1E-02 \\
hHGS\_f(x+8)  & 2.3E+10 & 2.9E+04 & 3.3E+01 & 444 & 15.3 & 70/379 & 2.7E-02 & 1.7E+15 & 1.6E+05 & 1.5E+02 & 445 & 15.3 & 64/401 & 2.6E-02 \\
hHGS  & 1.1E+04 & 2.6E+03 & 1.4E+04 & 301 & 10.4 & 211/250 & 2.6E-02 & 1.4E+05 & 3.2E+03 & 3.3E+05 & 319 & 11.0 & 197/271 & 1.9E-02 \\
hHHO\_f(x+8)  & 2.3E+10 & 2.9E+04 & 6.1E+01 & 475.5 & 16.4 & 38/427 & 1.7E-02 & 1.7E+15 & 1.6E+05 & 1.8E+12 & 516 & 17.8 & 12/466 & 1.7E-02 \\
hHHO  & 5.8E+04 & 2.3E+03 & 7.1E+04 & 340 & 11.7 & 179/299 & 1.1E-02 & 3.5E+05 & 3.3E+03 & 3.1E+05 & 367 & 12.7 & 148/329 & 1.6E-02 \\
hSMA\_f(x+8)  & 2.3E+10 & 2.9E+04 & 1.4E+01 & 413 & 14.2 & 79/378 & 1.4E-02 & 1.7E+15 & 1.6E+05 & 7.1E+10 & 449 & 15.5 & 45/422 & 2.2E-02 \\
hSMA  & 3.6E+03 & 1.9E+03 & 1.1E+03 & 231 & 8.0 & 265/194 & 2.8E-02 & 2.1E+04 & 2.9E+03 & 9.6E+03 & 257 & 8.9 & 240/209 & 4.5E-02 \\
hDMSSA\_f(x+8)  & 3.3E+10 & 3.2E+04 & 1.8E+09 & 522 & 18.0 & 0/493 & 1.9E-11 & 9.7E+01 & 9.2E+01 & 5.8E+00 & 150 & 5.2 & 351/93 & 6.3E-02 \\
hDMSSA  & 9.7E+01 & 9.2E+01 & 5.3E+00 & 116 & 4.0 & 380/61 & 7.1E-02 & 3.0E-01 & 2.2E-01 & 5.2E+01 & 57 & 2.0 & 436/3 & 6.0E-02 \\
hAROA\_f(x+8)  & 2.3E+10 & 2.9E+04 & 3.2E+01 & 441 & 15.2 & 58/396 & 2.3E-02 & 1.7E+15 & 1.6E+05 & 2.1E+02 & 445.5 & 15.4 & 52/396 & 3.8E-02 \\
hAROA  & 5.4E+03 & 1.7E+03 & 9.2E+03 & 258 & 8.9 & 242/216 & 2.6E-02 & 6.7E+04 & 2.9E+03 & 3.5E+04 & 273 & 9.4 & 227/222 & 3.9E-02 \\
\bottomrule
\end{tabular}
}
\vspace{1.1mm}
\scalebox{0.67}{
\renewcommand{\arraystretch}{0.9}
\begin{tabular}{@{}l
ccccccc  
@{\hspace{0.5cm}}
ccccccc
@{}}
\toprule
\multirow{2}{*}{Algorithm} & \multicolumn{7}{c}{dim=50} & \multicolumn{7}{c}{dim=100} \\
\cmidrule(lr){2-8} \cmidrule(lr){9-15}
 & Avg & Med & Std & Sum Rank & Mean Rank & +/- & p-value & Avg & Med & Std & Sum Rank & Mean Rank & +/- & p-value \\
\midrule
hSPSO2011\_f(x+8)  & 8.2E+05 & 3.3E+03 & 2.3E+05 & 314 & 10.8 & 185/260 & 4.6E-02 & 3.7E+06 & 6.3E+03 & 1.7E+06 & 323 & 11.1 & 185/276 & 3.2E-02 \\
hSPSO2011  & 8.8E+05 & 3.3E+03 & 2.4E+05 & 301 & 10.4 & 190/254 & 4.6E-02 & 3.1E+06 & 6.0E+03 & 1.0E+06 & 303 & 10.4 & 198/263 & 3.1E-02 \\
hMPA\_f(x+8)  & 7.1E+11 & 2.0E+05 & 6.6E+10 & 463 & 16.0 & 42/423 & 2.5E-02 & 6.8E+17 & 3.5E+05 & 1.1E+16 & 475 & 16.4 & 33/437 & 2.1E-02 \\
hMPA  & 4.3E+05 & 3.2E+03 & 3.3E+05 & 243 & 8.4 & 263/199 & 2.2E-02 & 6.0E+06 & 5.3E+03 & 2.8E+06 & 260 & 9.0 & 248/218 & 2.0E-02 \\
hSHADE\_f(x+8)  & 9.7E+01 & 9.4E+01 & 5.9E+00 & 130 & 4.5 & 354/95 & 5.2E-02 & 9.7E+01 & 9.2E+01 & 5.9E+00 & 132 & 4.6 & 360/91 & 5.2E-02 \\
hSHADE  & 1.1E+00 & 1.9E+00 & 5.3E+01 & 59 & 2.0 & 435/4 & 5.8E-02 & -6.2E-01 & -7.9E-02 & 5.2E+01 & 60 & 2.1 & 437/5 & 5.8E-02 \\
hIMODE\_f(x+8)  & 9.7E+01 & 9.4E+01 & 5.9E+00 & 151 & 5.2 & 357/92 & 5.6E-02 & 9.7E+01 & 9.2E+01 & 5.9E+00 & 150 & 5.2 & 360/92 & 4.1E-02 \\
hIMODE  & 8.0E-01 & 5.1E-01 & 5.6E+01 & 57 & 2.0 & 438/3 & 5.7E-02 & -8.9E-01 & -1.1E-01 & 5.5E+01 & 51 & 1.8 & 440/4 & 5.8E-02 \\
hHGS\_f(x+8)  & 6.7E+11 & 2.0E+05 & 2.6E+02 & 437.5 & 15.1 & 75/394 & 1.4E-02 & 6.5E+17 & 3.5E+05 & 2.8E+04 & 429 & 14.8 & 87/396 & 6.8E-03 \\
hHGS  & 1.7E+06 & 4.0E+03 & 1.6E+06 & 321 & 11.1 & 195/268 & 2.8E-02 & 2.7E+08 & 7.0E+03 & 1.2E+08 & 314 & 10.8 & 202/269 & 2.1E-02 \\
hHHO\_f(x+8)  & 1.1E+12 & 2.0E+05 & 2.8E+11 & 512 & 17.7 & 10/474 & 9.9E-03 & 7.3E+17 & 3.6E+05 & 2.3E+16 & 511 & 17.6 & 14/470 & 9.3E-03 \\
hHHO  & 2.2E+06 & 4.0E+03 & 1.2E+06 & 364 & 12.6 & 154/327 & 9.7E-03 & 1.7E+07 & 7.1E+03 & 8.4E+06 & 355 & 12.2 & 161/320 & 1.1E-02 \\
hSMA\_f(x+8)  & 6.7E+11 & 2.0E+05 & 1.3E+09 & 446 & 15.4 & 56/413 & 2.2E-02 & 6.5E+17 & 3.5E+05 & 9.3E+14 & 441 & 15.2 & 67/413 & 9.5E-03 \\
hSMA  & 1.7E+05 & 3.3E+03 & 7.6E+04 & 243 & 8.4 & 261/198 & 3.0E-02 & 6.2E+05 & 4.9E+03 & 2.6E+05 & 217 & 7.5 & 292/182 & 1.5E-02 \\
hDMSSA\_f(x+8)  & 9.7E+01 & 9.4E+01 & 6.0E+00 & 154 & 5.3 & 356/97 & 4.5E-02 & 9.8E+01 & 9.2E+01 & 6.0E+00 & 153 & 5.3 & 351/105 & 4.6E-02 \\
hDMSSA  & 8.4E-01 & 1.2E+00 & 5.2E+01 & 58 & 2.0 & 440/1 & 5.4E-02 & -4.5E-01 & -6.3E-02 & 5.1E+01 & 63 & 2.2 & 441/4 & 5.3E-02 \\
hAROA\_f(x+8)  & 6.7E+11 & 2.0E+05 & 3.3E+02 & 441.5 & 15.2 & 70/393 & 2.4E-02 & 6.5E+17 & 3.5E+05 & 4.0E+14 & 441 & 15.2 & 76/400 & 1.5E-02 \\
hAROA  & 4.4E+05 & 3.4E+03 & 2.1E+05 & 265 & 9.1 & 232/218 & 3.4E-02 & 7.4E+06 & 5.6E+03 & 2.5E+06 & 283 & 9.8 & 228/235 & 1.8E-02 \\
\bottomrule
\label{tab2}
\end{tabular}
}
\vspace{1.5mm}
\footnotesize{Friedman test p-values: dim10=1.7E-90, dim30=5.7E-88, dim50=2.1E-87, dim100=2.3E-87}
\end{adjustwidth}
\end{table}

\normalsize

\begin{figure}[H]
  \centering
\begin{adjustwidth}{-1.8cm}{-2.0cm}
  \begin{subfigure}{0.4\textwidth}
     \includegraphics[height=5.5cm,width=6.9cm]{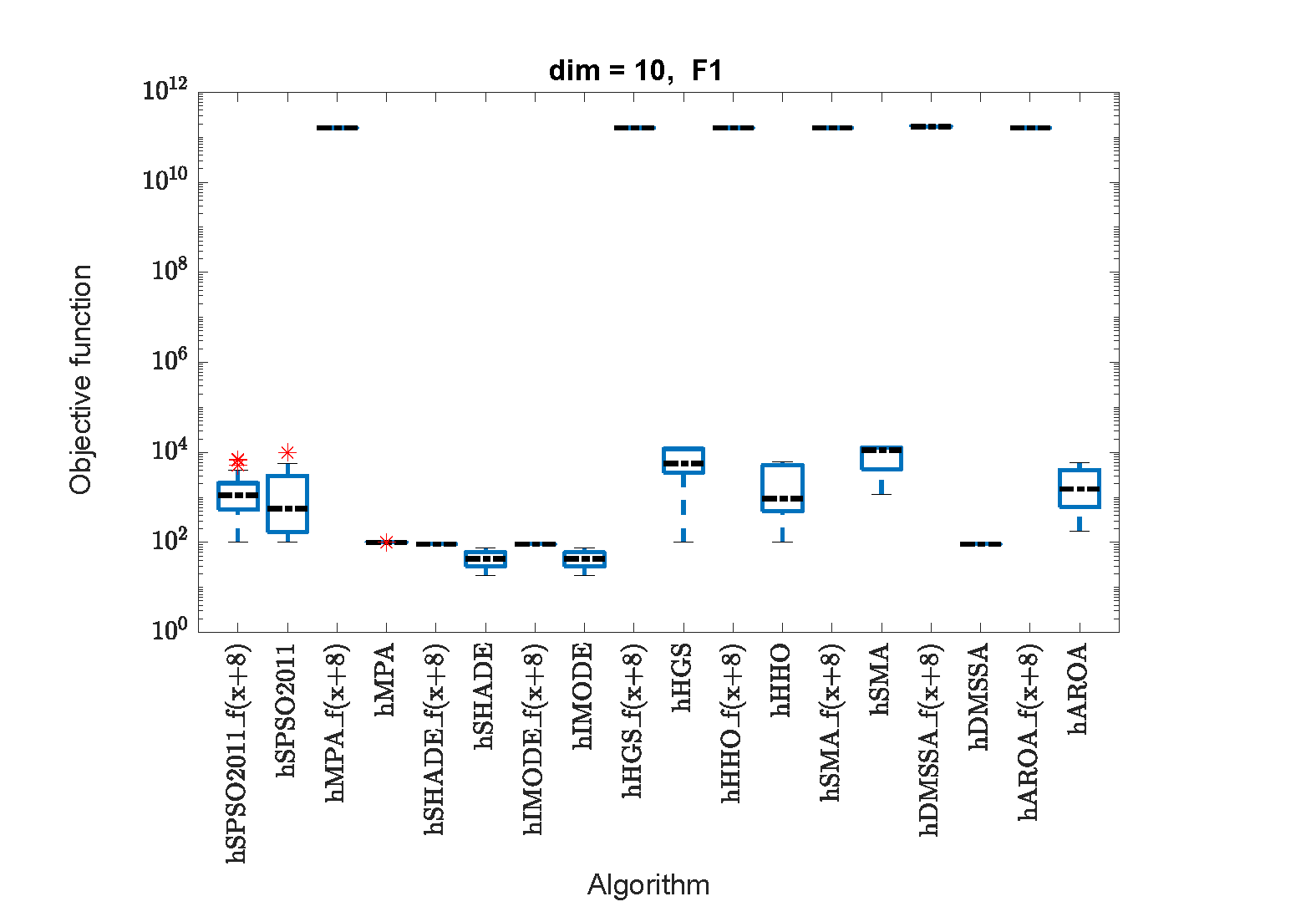}
  \end{subfigure}\hfill
  \begin{subfigure}{0.4\textwidth}
    \includegraphics[height=5.5cm,width=6.9cm]{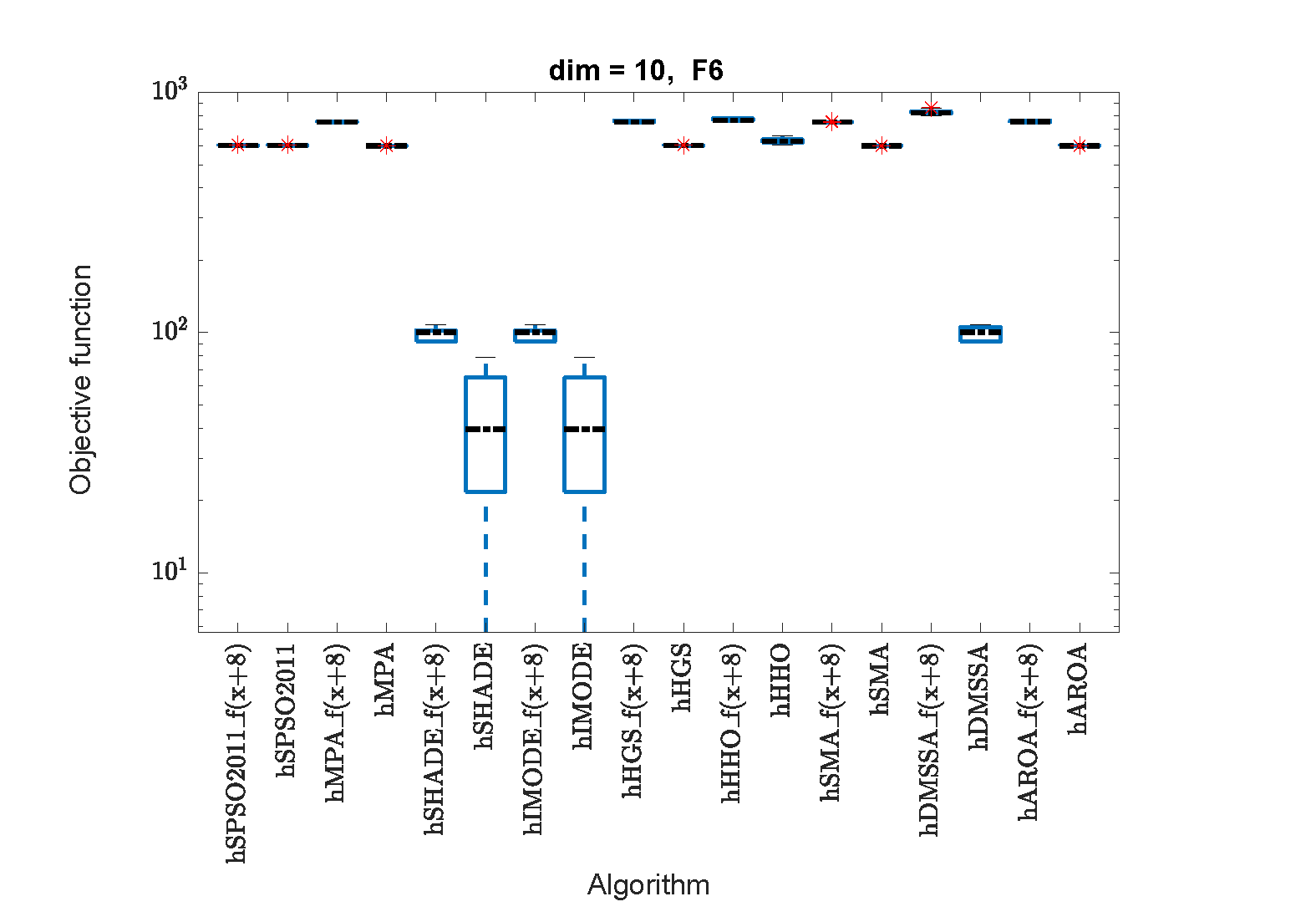}
  \end{subfigure}\hfill
  \begin{subfigure}{0.4\textwidth}
    \includegraphics[height=5.5cm,width=6.9cm]{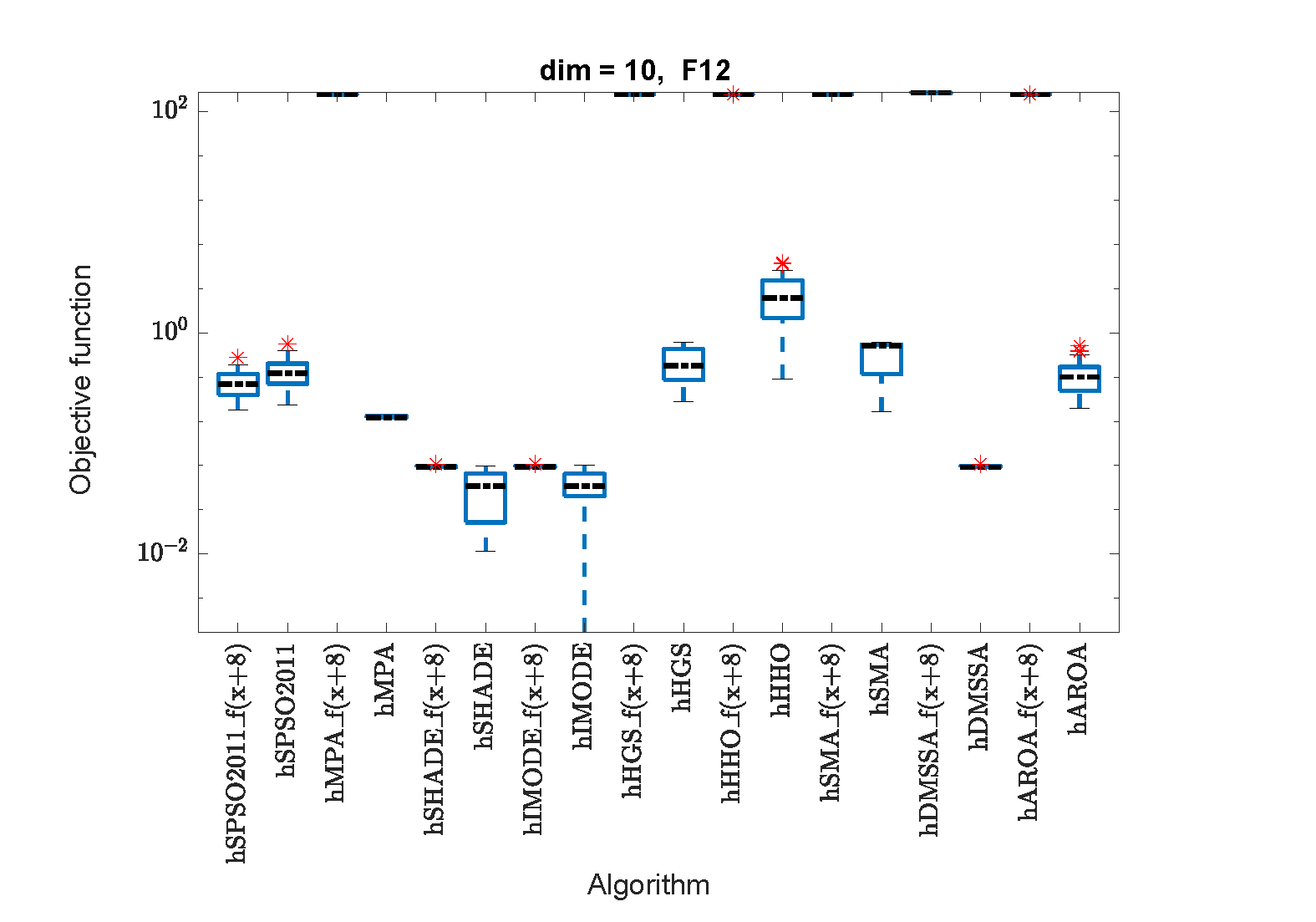}
  \end{subfigure}
 \end{adjustwidth}
\end{figure}

\begin{figure}[H]
  \centering
\begin{adjustwidth}{-2cm}{-2.0cm}
   \begin{subfigure}{0.4\textwidth}
    \includegraphics[height=5.1cm,width=6.9cm]{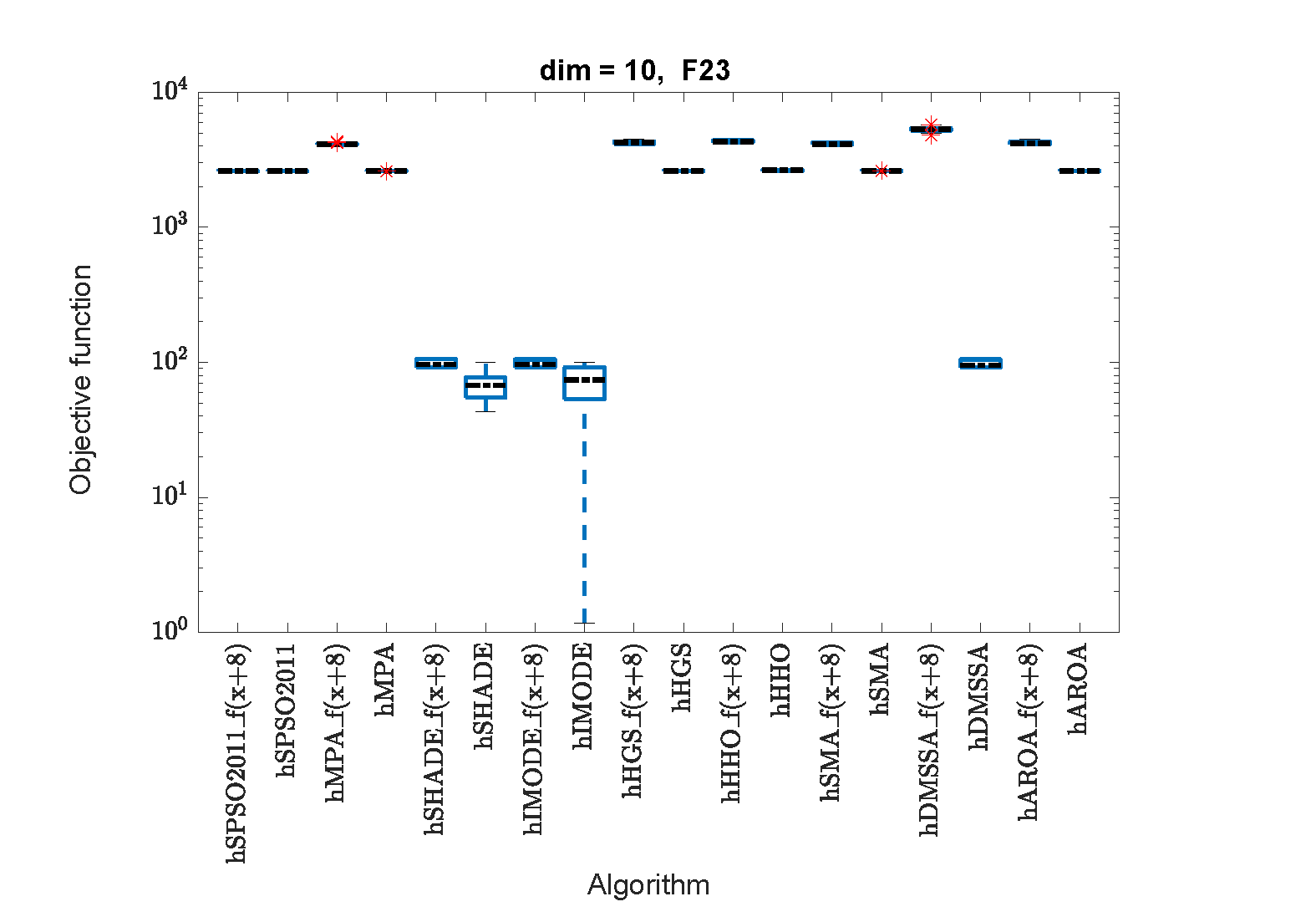}
  \end{subfigure}\hfill
  \begin{subfigure}{0.4\textwidth}
    \includegraphics[height=5.1cm,width=6.9cm]{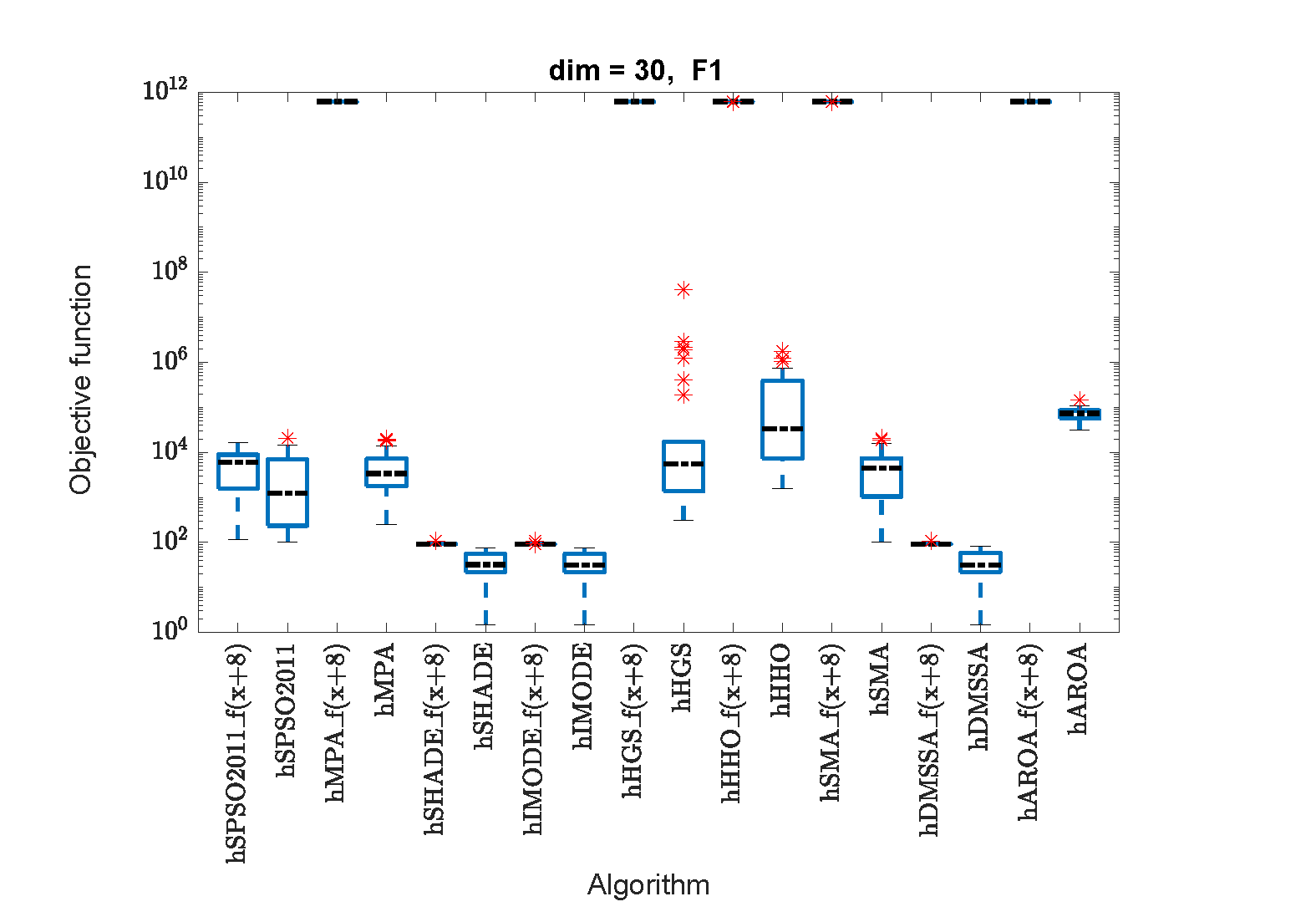}
\end{subfigure}\hfill
  \begin{subfigure}{0.4\textwidth} 
        \includegraphics[height=5.1cm,width=6.9cm]{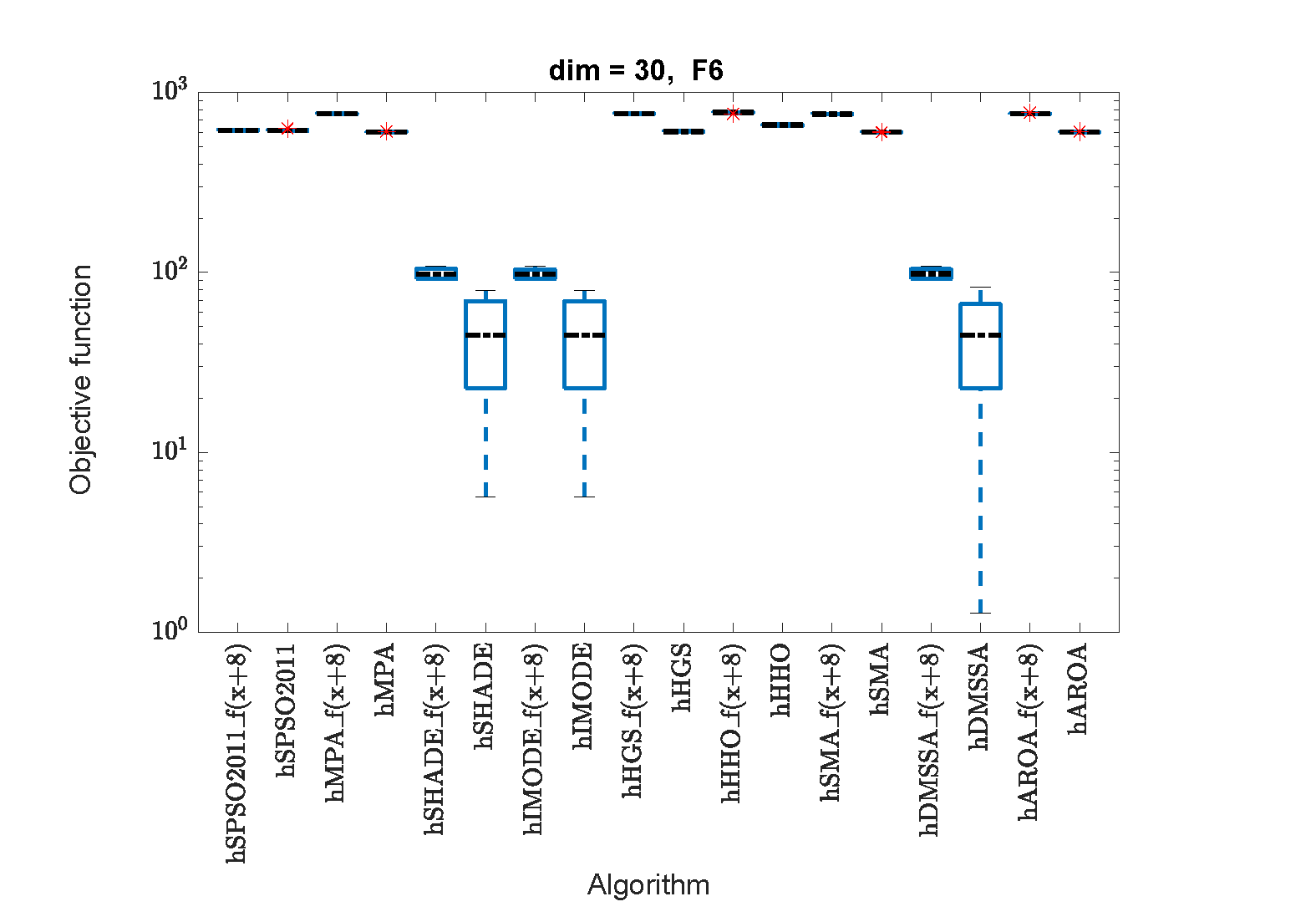}
   \end{subfigure}
   \end{adjustwidth}
\end{figure}

 \begin{figure}[H]
  \centering
\begin{adjustwidth}{-2cm}{-2.0cm}
   \begin{subfigure}{0.4\textwidth}
        \includegraphics[height=5.1cm,width=6.9cm]{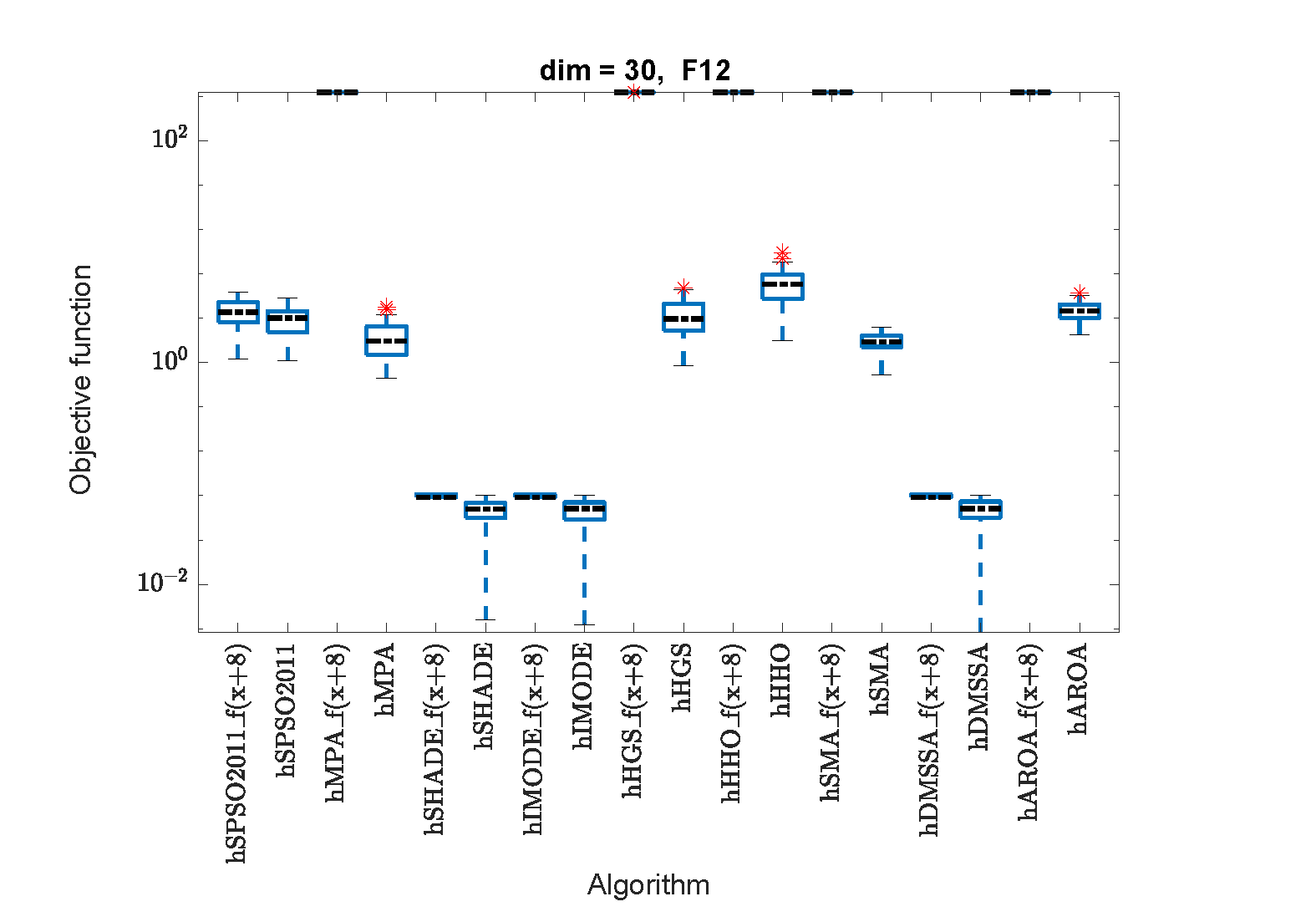}
  \end{subfigure}\hfill
  \begin{subfigure}{0.4\textwidth} 
        \includegraphics[height=5.1cm,width=6.9cm]{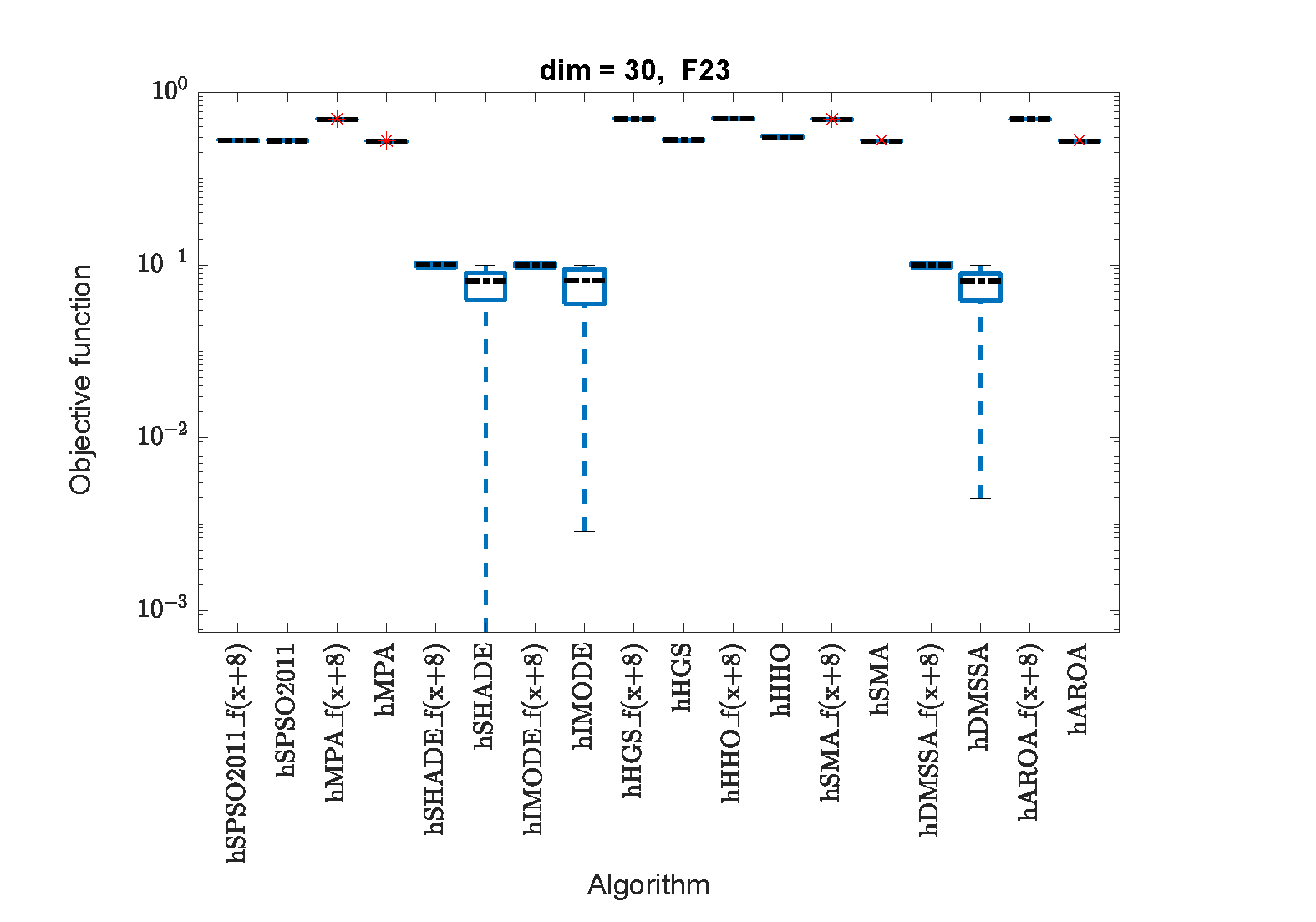}
  \end{subfigure}\hfill
  \begin{subfigure}{0.4\textwidth}
        \includegraphics[height=5.1cm,width=6.9cm]{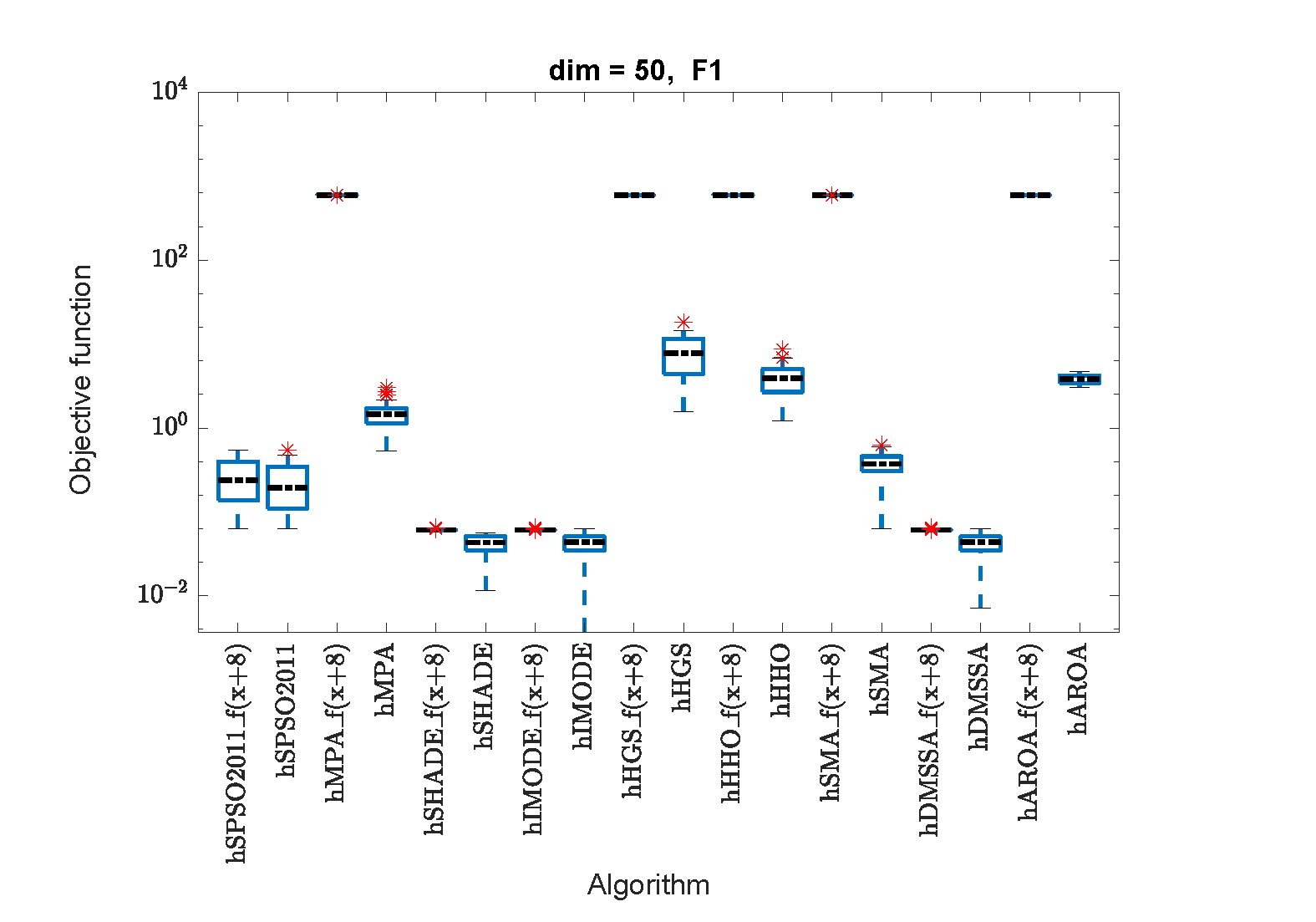}
  \end{subfigure}
  \end{adjustwidth}
\end{figure}

  \begin{figure}[H]
  \centering
\begin{adjustwidth}{-2cm}{-2.0cm}
  \begin{subfigure}{0.4\textwidth} 
        \includegraphics[height=5.1cm,width=6.9cm]{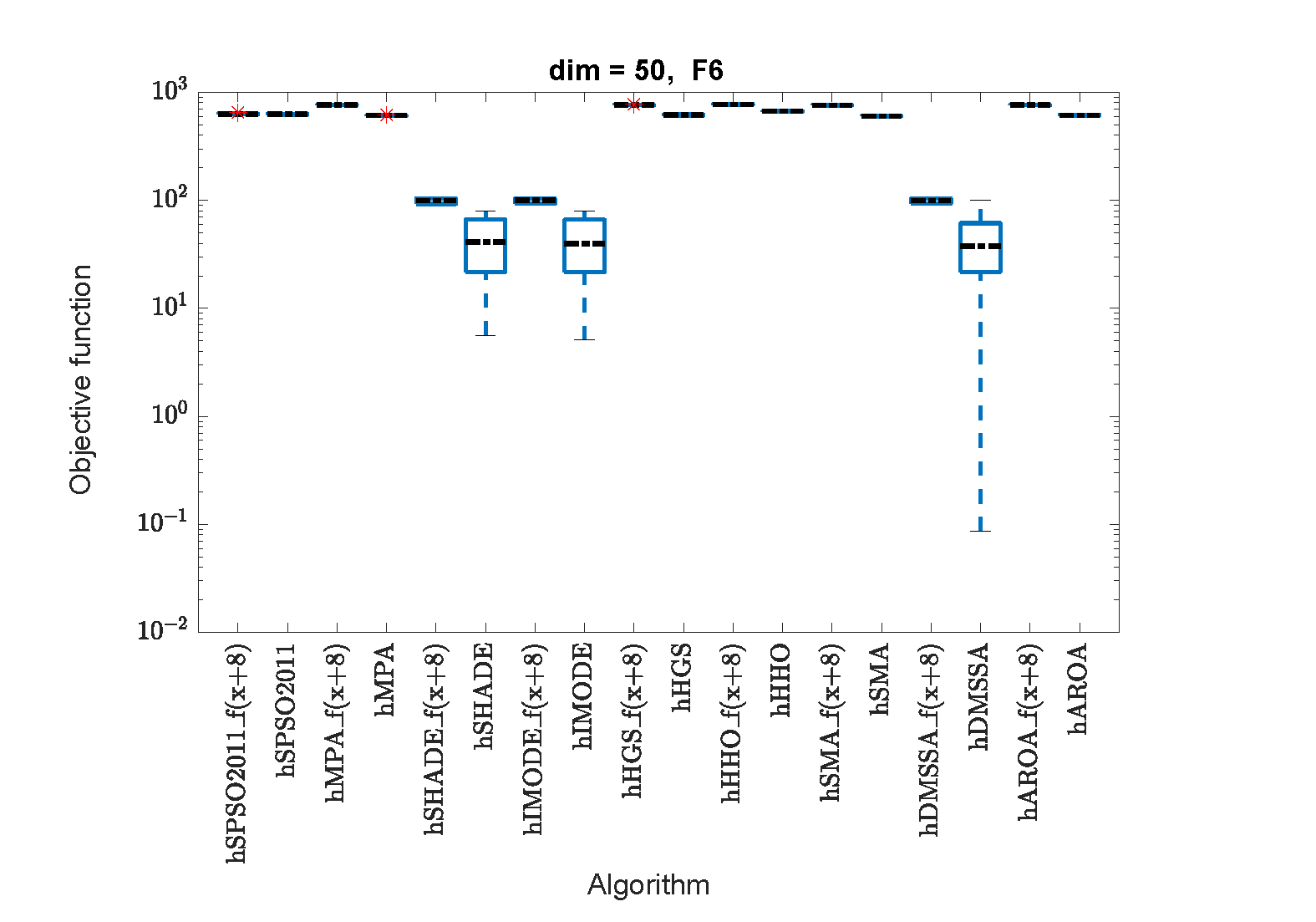}
  \end{subfigure}\hfill
  \begin{subfigure}{0.4\textwidth}
    \includegraphics[height=5.1cm,width=6.9cm]{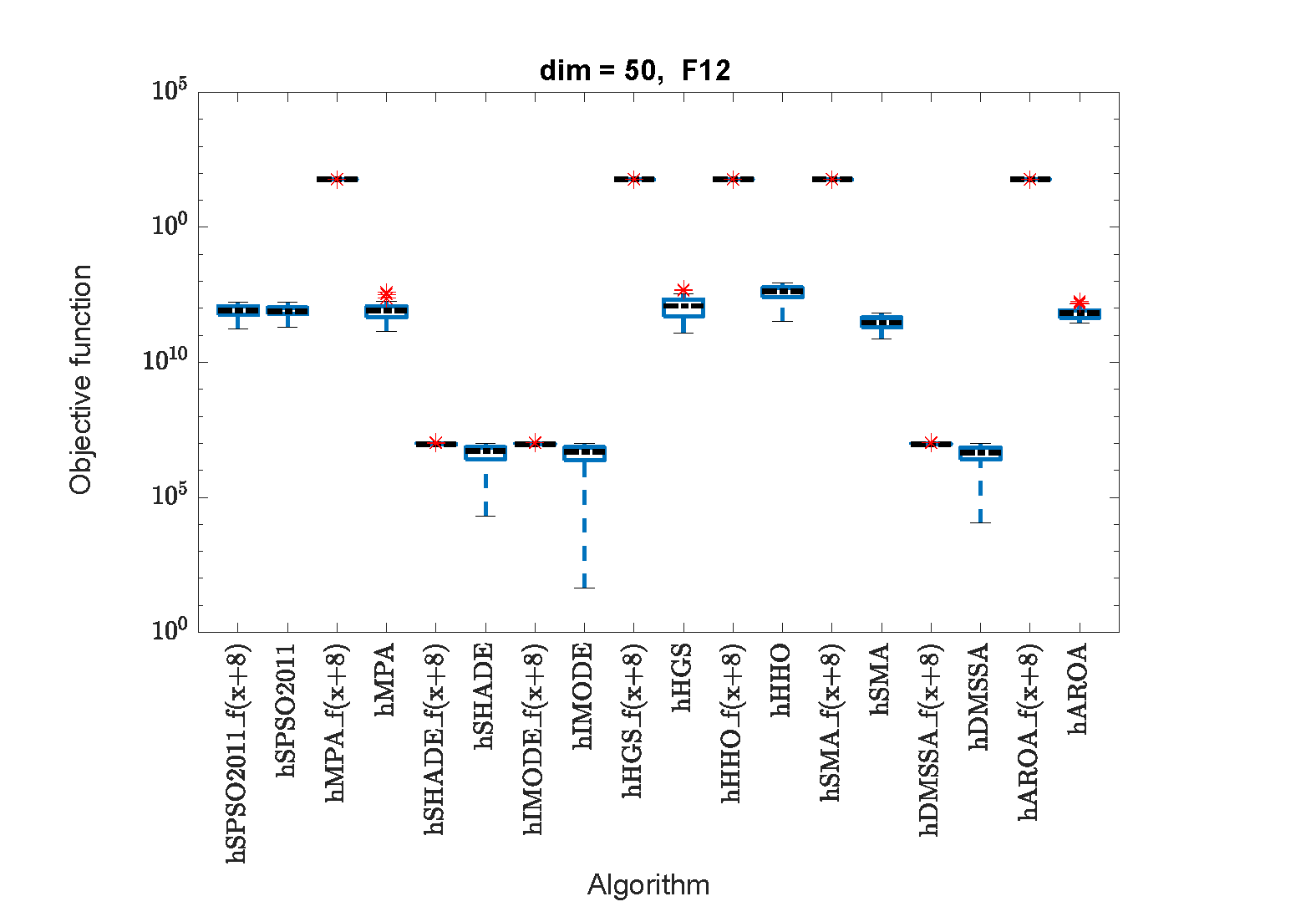}
     \end{subfigure}\hfill
 \begin{subfigure}{0.4\textwidth}
    \includegraphics[height=5.1cm,width=6.9cm]{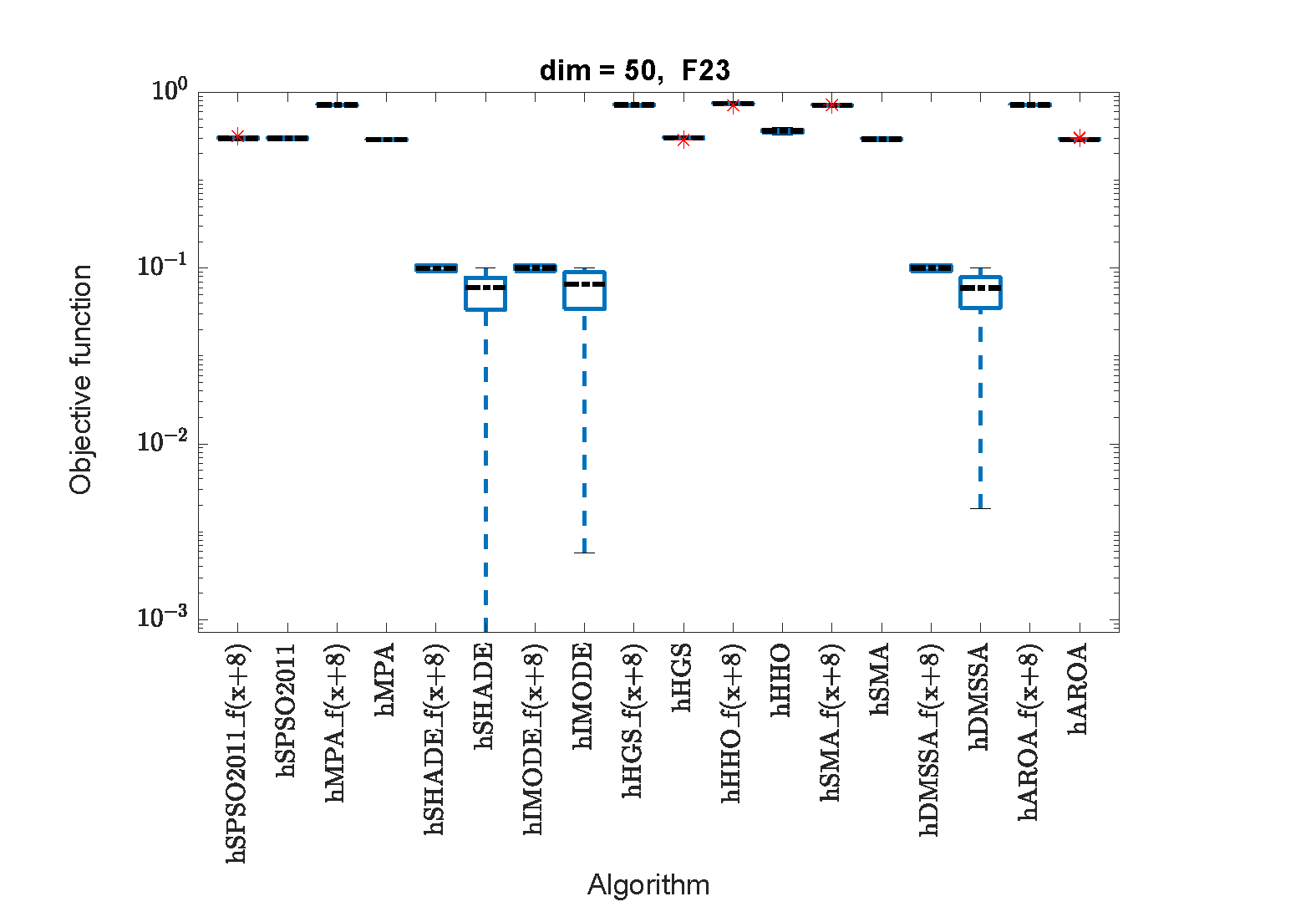}
  \end{subfigure}
\end{adjustwidth}
\end{figure}
 
  \begin{figure}[H]
  \centering
\begin{adjustwidth}{-2cm}{-2.0cm}
  \begin{subfigure}{0.4\textwidth}
    \includegraphics[height=5.1cm,width=6.9cm]{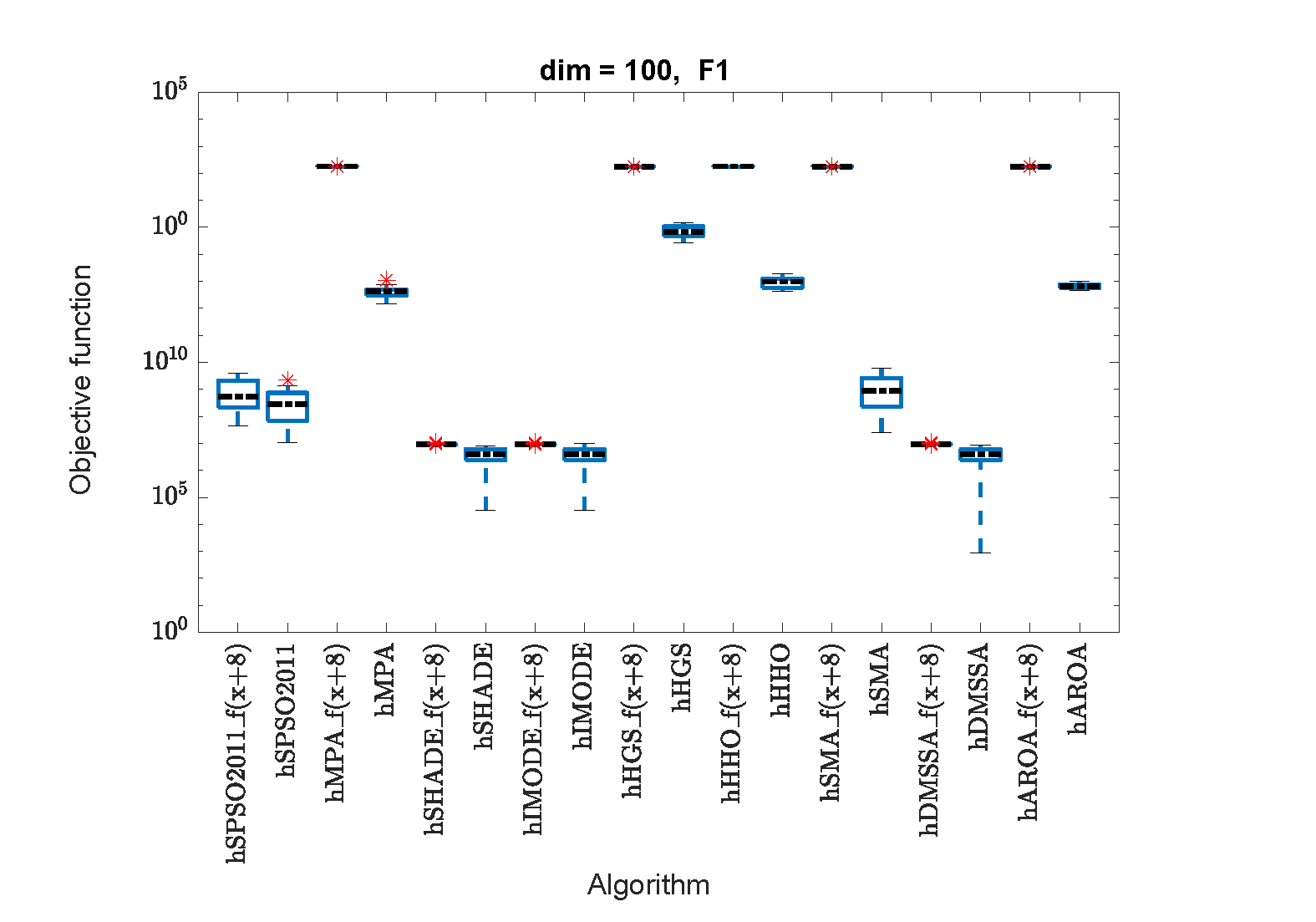}
  \end{subfigure}\hfill
  \begin{subfigure}{0.4\textwidth}
        \includegraphics[height=5.1cm,width=6.9cm]{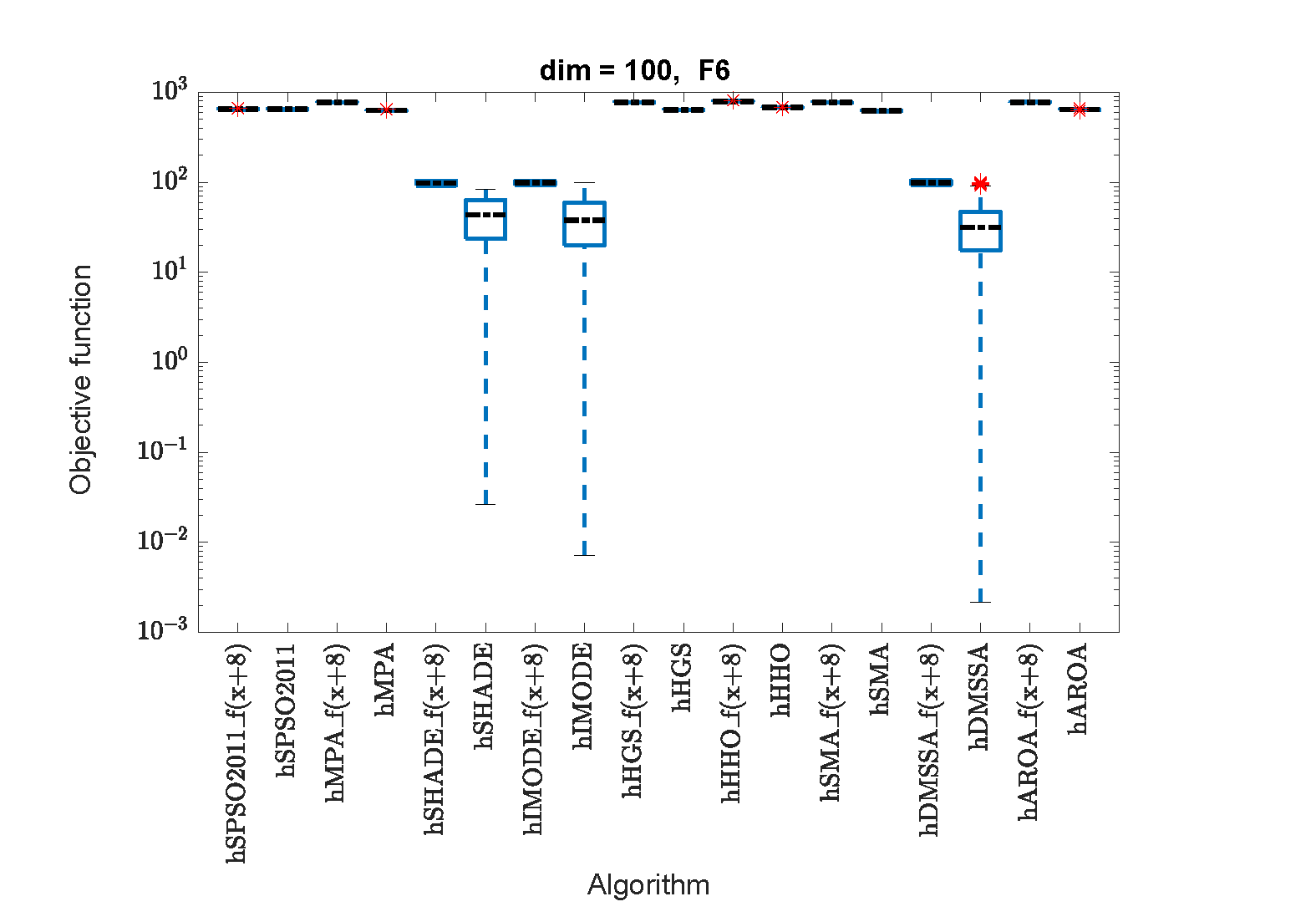}
     \end{subfigure}\hfill
   \begin{subfigure}{0.4\textwidth}
        \includegraphics[height=5.1cm,width=6.9cm]{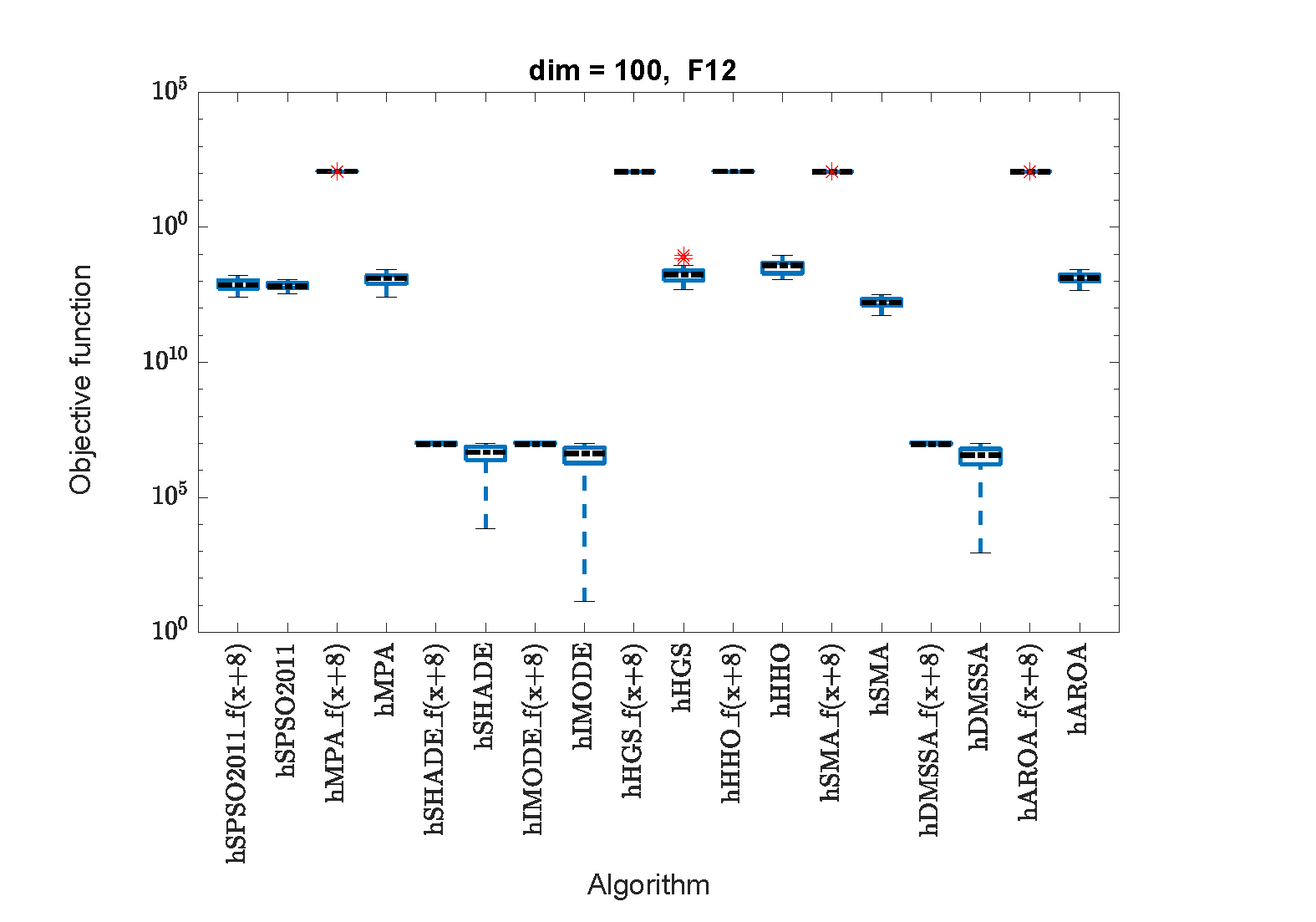}
    \end{subfigure}
    \end{adjustwidth}
    \end{figure}

\begin{figure}[H]
 \centering
\begin{adjustwidth}{-2cm}{-2.0cm}
\begin{center}
  \begin{subfigure}{0.4\textwidth}
   \centering
        \includegraphics[height=5.1cm,width=6.9cm]{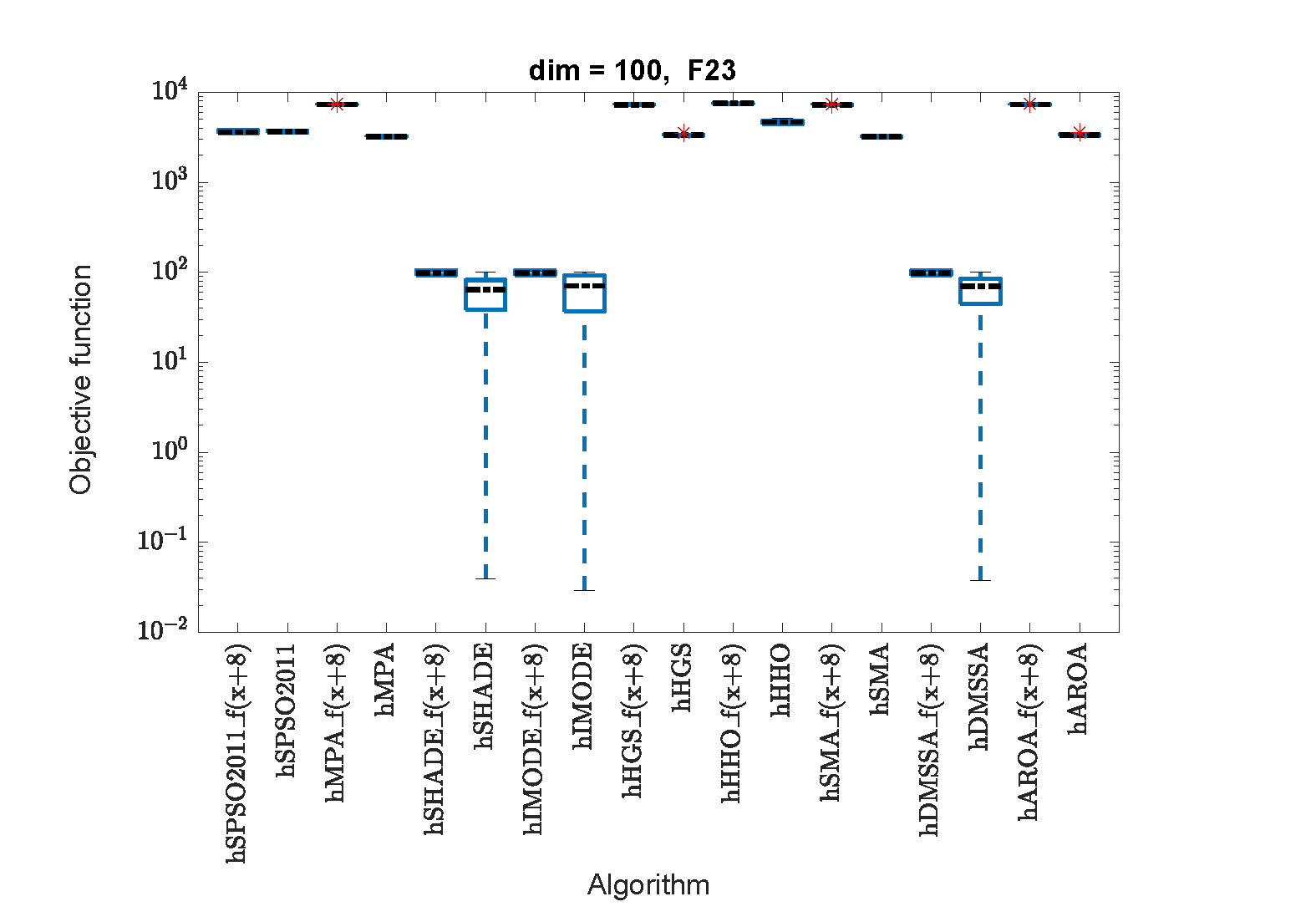}   
  \end{subfigure}
  \end{center}
  \begin{center}
  \begin{subfigure}{0.4\textwidth}
   \centering
   \vspace{0.4cm}
    \includegraphics[width=\linewidth]{boxplotlegend.pdf}     
  \end{subfigure}
  \end{center}
  \caption{\footnotesize{Boxplots of final objective values for 9 hybrid algorithms evaluated on original CEC-2017 functions and their transformed variants across 4 dims.}}
  \label{fig5}
\end{adjustwidth}
\end{figure}

\normalsize

\begin{figure}[H]
\vspace{0.3cm}
  \centering
\begin{adjustwidth}{-1cm}{1.0cm}
  \begin{subfigure}{0.35\textwidth} 
        \includegraphics[height=5.4cm,width=8cm]{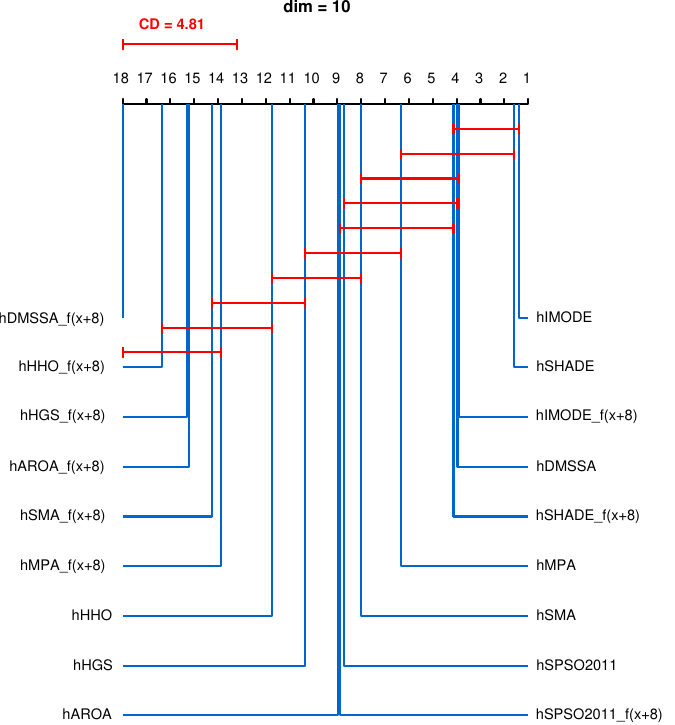}
  \end{subfigure}\hfill
  \begin{subfigure}{0.36\textwidth}
    \includegraphics[height=5.9cm,width=8cm]{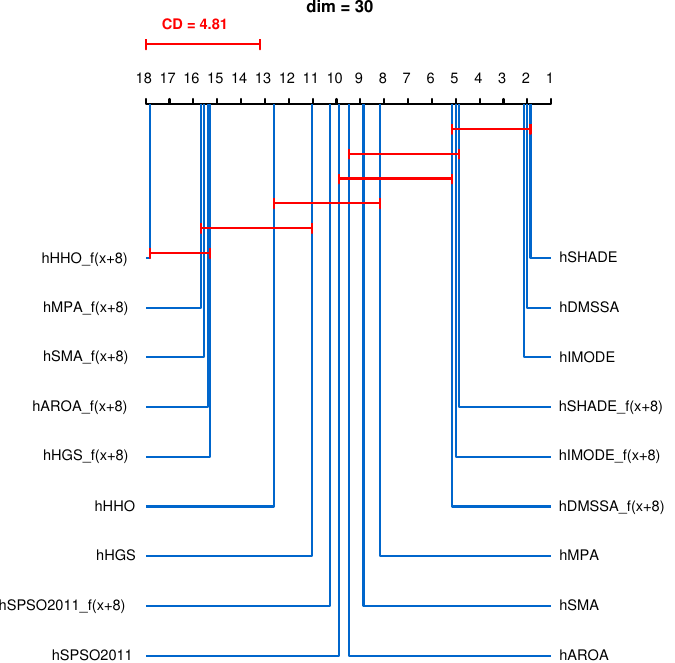}
     \end{subfigure}\hfill
\end{adjustwidth}
\end{figure}
\begin{figure}[H]
\centering
\begin{adjustwidth}{-1cm}{1.0cm}
 \begin{subfigure}{0.35\textwidth}
    \includegraphics[height=5.4cm,width=8cm]{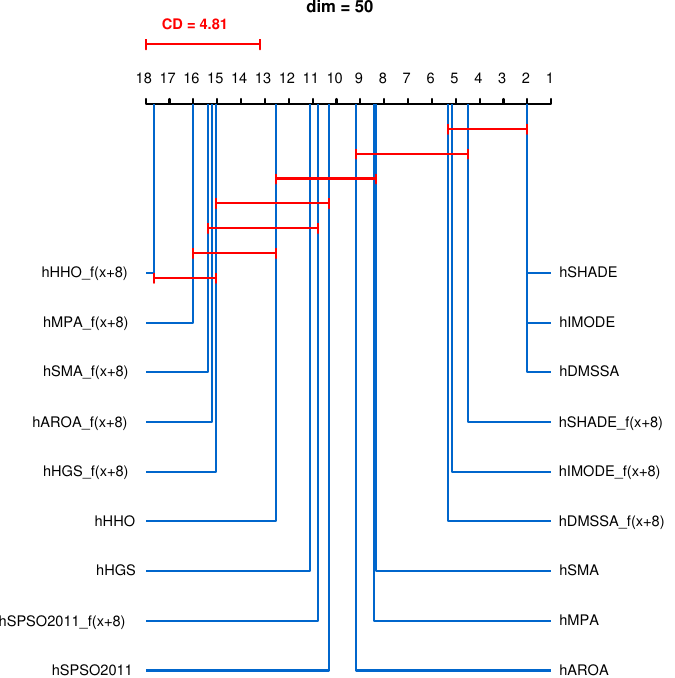}
  \end{subfigure}\hfill
  \begin{subfigure}{0.325\textwidth}
    \vspace{0.15cm}
    \includegraphics[height=5.4cm,width=8cm]{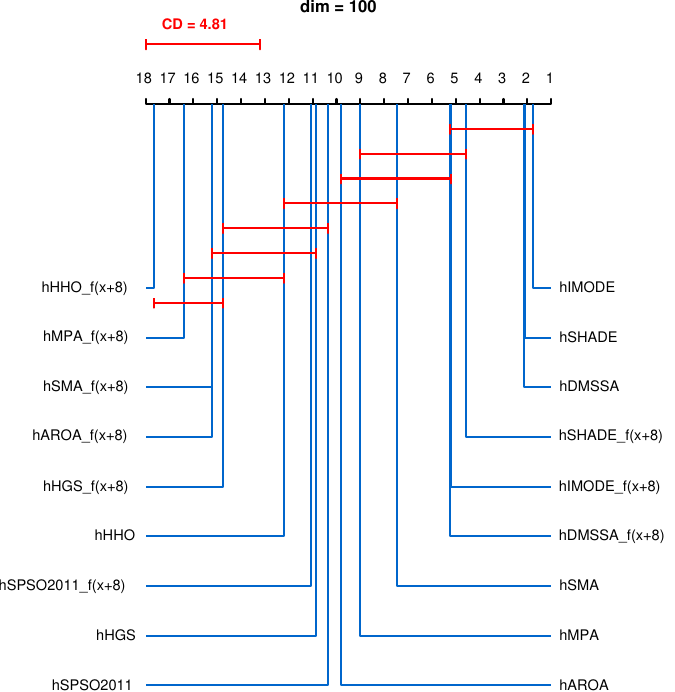}
  \end{subfigure}\hfill
 \caption{\footnotesize{CD diagrams for 9 hybrid algorithms tested on original CEC-2017 functions and their translated variants across 4 dims.}}
  \label{fig6}
\end{adjustwidth}
\end{figure}

\begin{figure}[H] 
\adjustbox{scale=1.25,center}{
\centering
  \begin{subfigure}[t]{0.495\textwidth}
    \centering
\includegraphics[width=\linewidth]{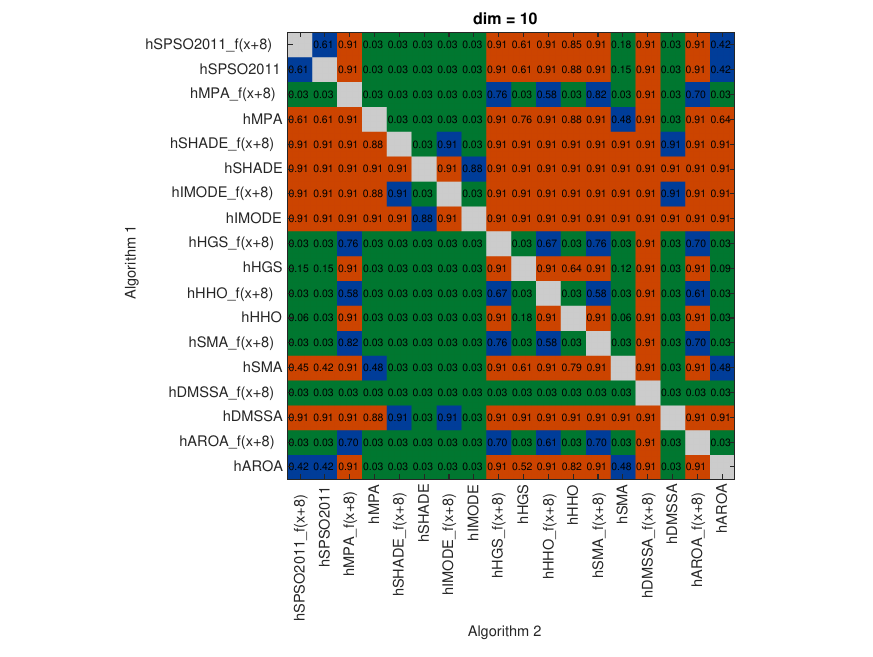}
  \end{subfigure}
  \begin{subfigure}[t]{0.495\textwidth}
    \centering
\includegraphics[width=\linewidth]{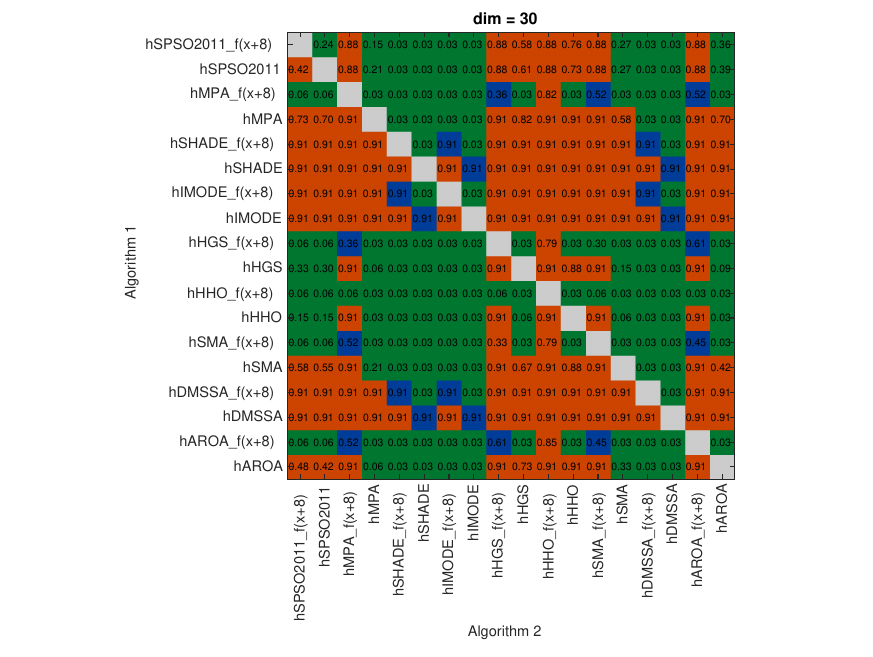}
  \end{subfigure}
  }
  \end{figure}
  \vspace{0.2cm}
  \begin{figure}[H] 
\adjustbox{scale=1.25,center}{
\begin{subfigure}[t]{0.495\textwidth}
    \centering
\includegraphics[width=\linewidth]{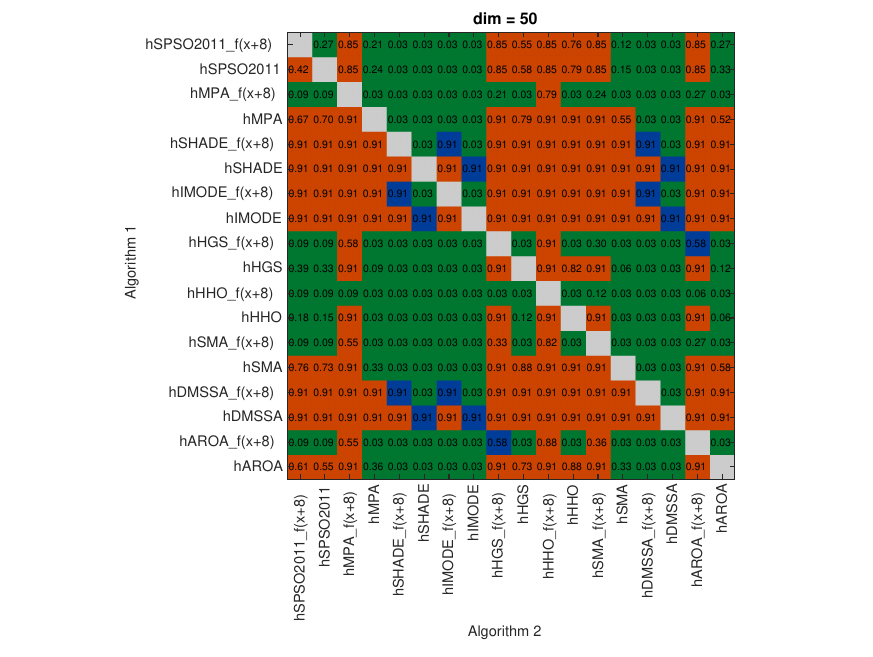}
  \end{subfigure}
 \begin{subfigure}[t]{0.495\textwidth}
    \centering
\includegraphics[width=\linewidth]{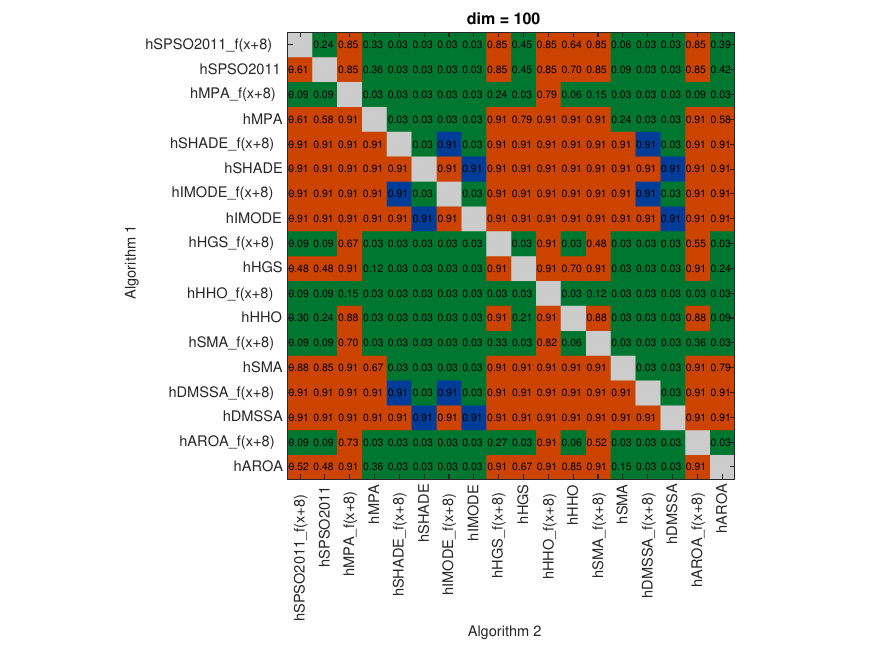}
  \end{subfigure}
  }
  \vspace{0.2cm}
  \adjustbox{scale=1.1,center}{
\begin{subfigure}{0.6\textwidth}
    \centering
    \includegraphics[width=\linewidth]{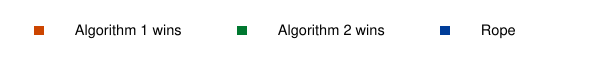}
  \end{subfigure}
  }
  \vspace{-0.75cm}
 \caption{\small Bayesian heatmaps for 9 hybrid algorithms evaluated on original CEC-2017 functions and their translated variants across 4 dims.}
  \label{fig7}
  \end{figure}

\begin{figure}[H]
  \centering
\begin{adjustwidth}{-1cm}{-2.0cm}
  \begin{subfigure}{0.35\textwidth}
     \includegraphics[height=4cm,width=5.5cm]{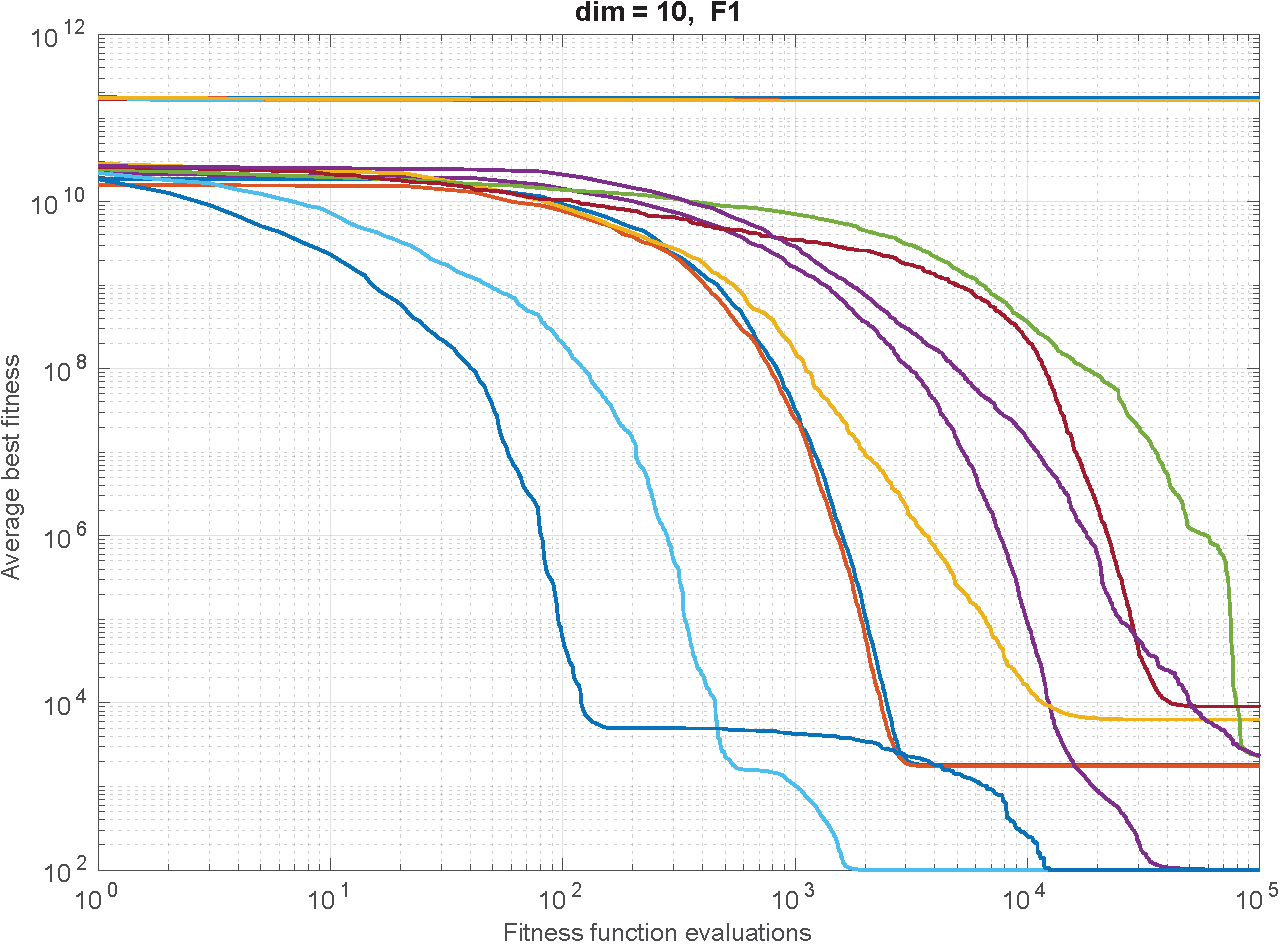}
  \end{subfigure}\hfill
  \begin{subfigure}{0.35\textwidth}
    \includegraphics[height=4cm,width=5.5cm]{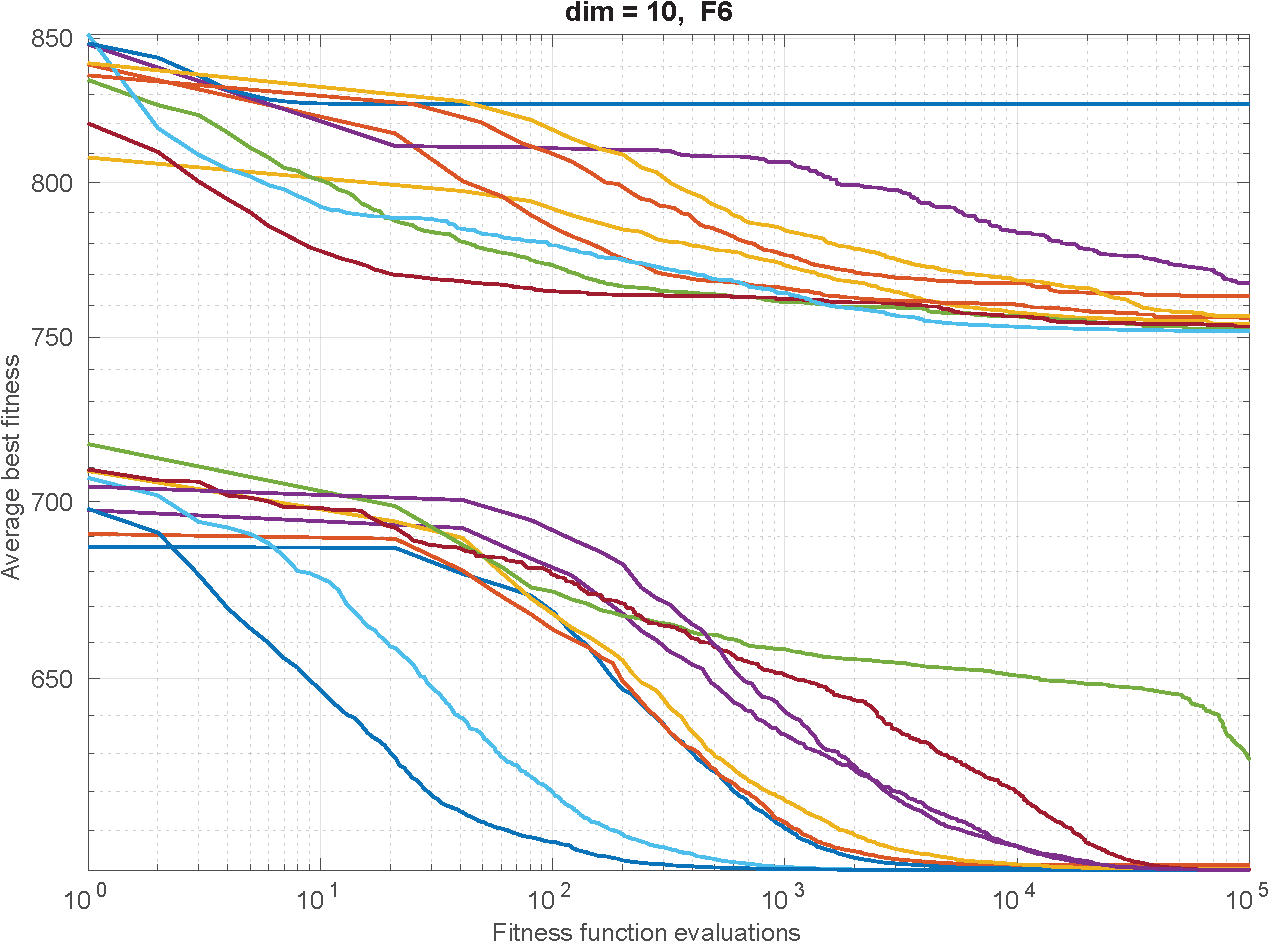}
  \end{subfigure}\hfill
  \begin{subfigure}{0.35\textwidth}
    \includegraphics[height=4cm,width=5.5cm]{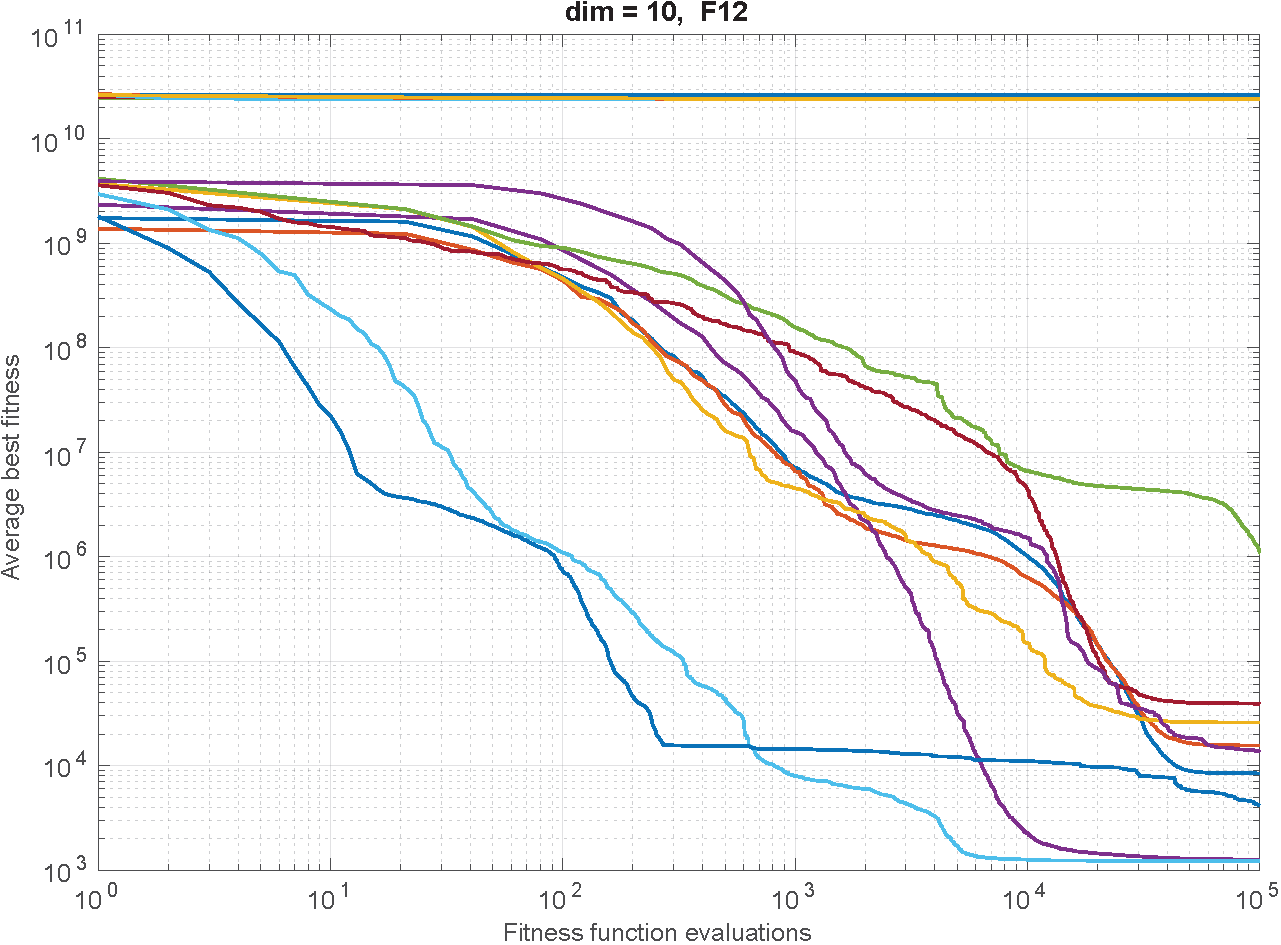}
  \end{subfigure}
 \end{adjustwidth}
\end{figure} 

\begin{figure}[H]
  \centering
\begin{adjustwidth}{-1cm}{-2.0cm}
   \begin{subfigure}{0.35\textwidth}
    \includegraphics[height=4cm,width=5.5cm]{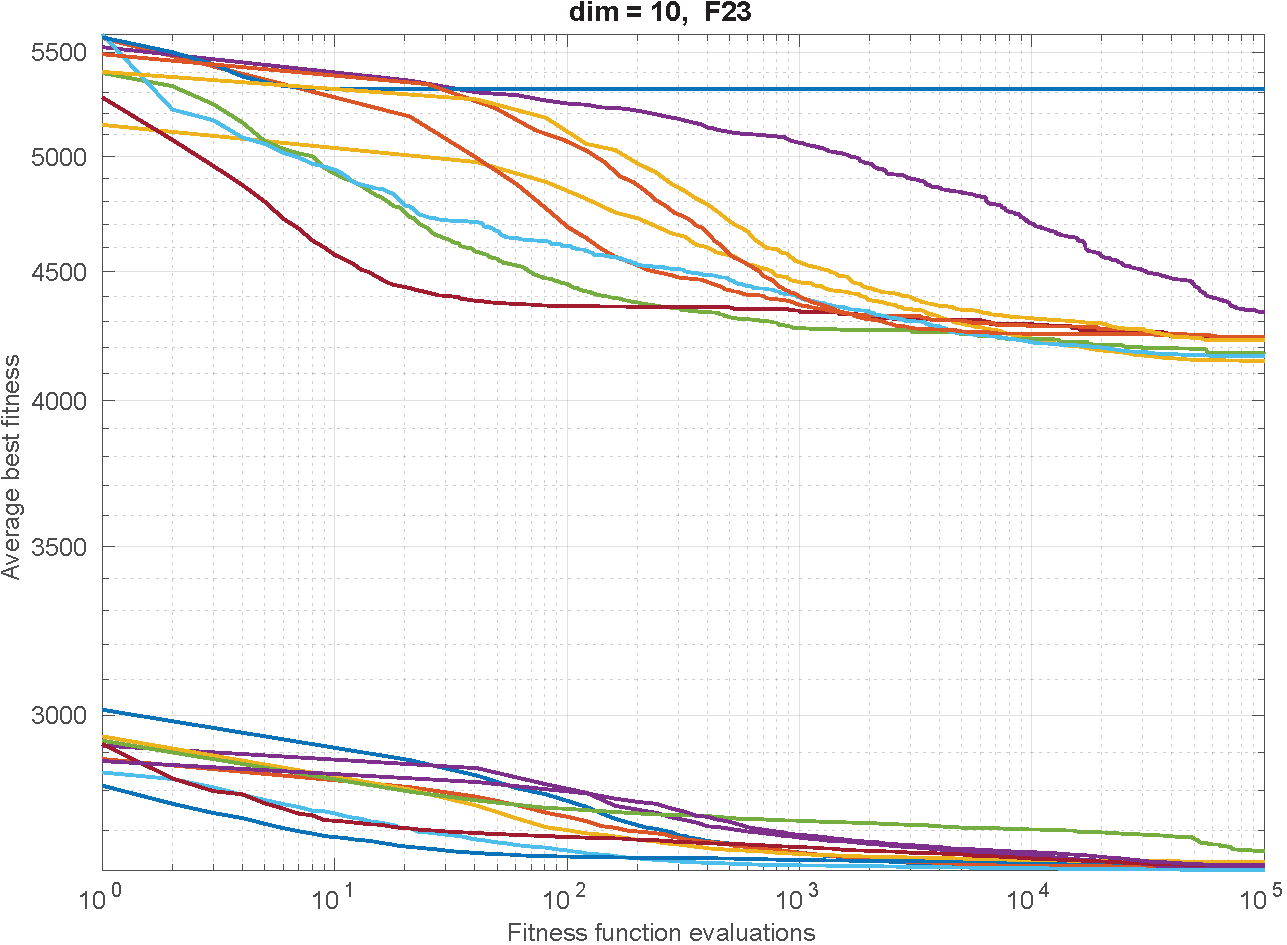}
  \end{subfigure}\hfill
  \begin{subfigure}{0.35\textwidth}
    \includegraphics[height=4cm,width=5.5cm]{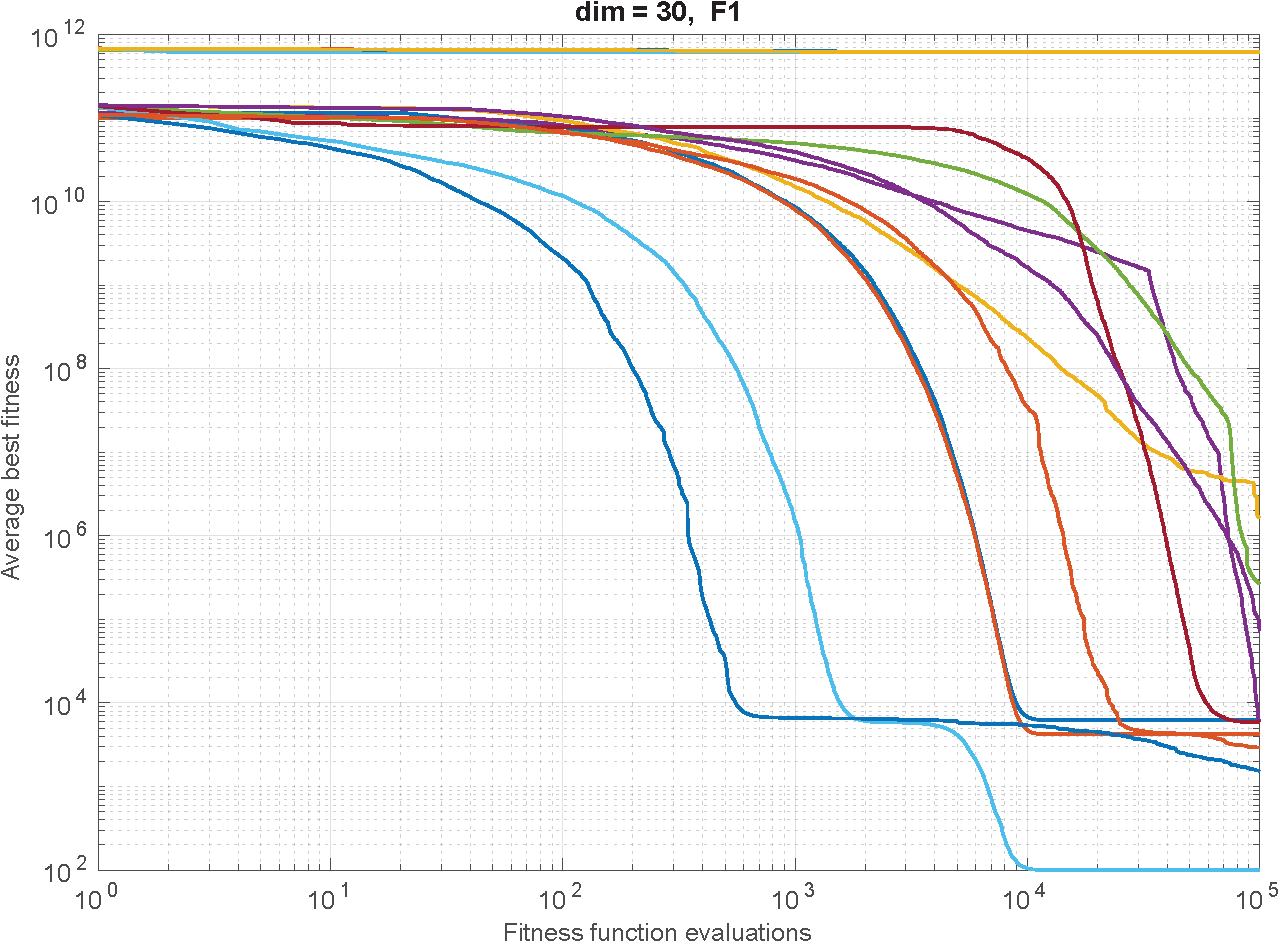}
\end{subfigure}\hfill
  \begin{subfigure}{0.35\textwidth} 
        \includegraphics[height=4cm,width=5.5cm]{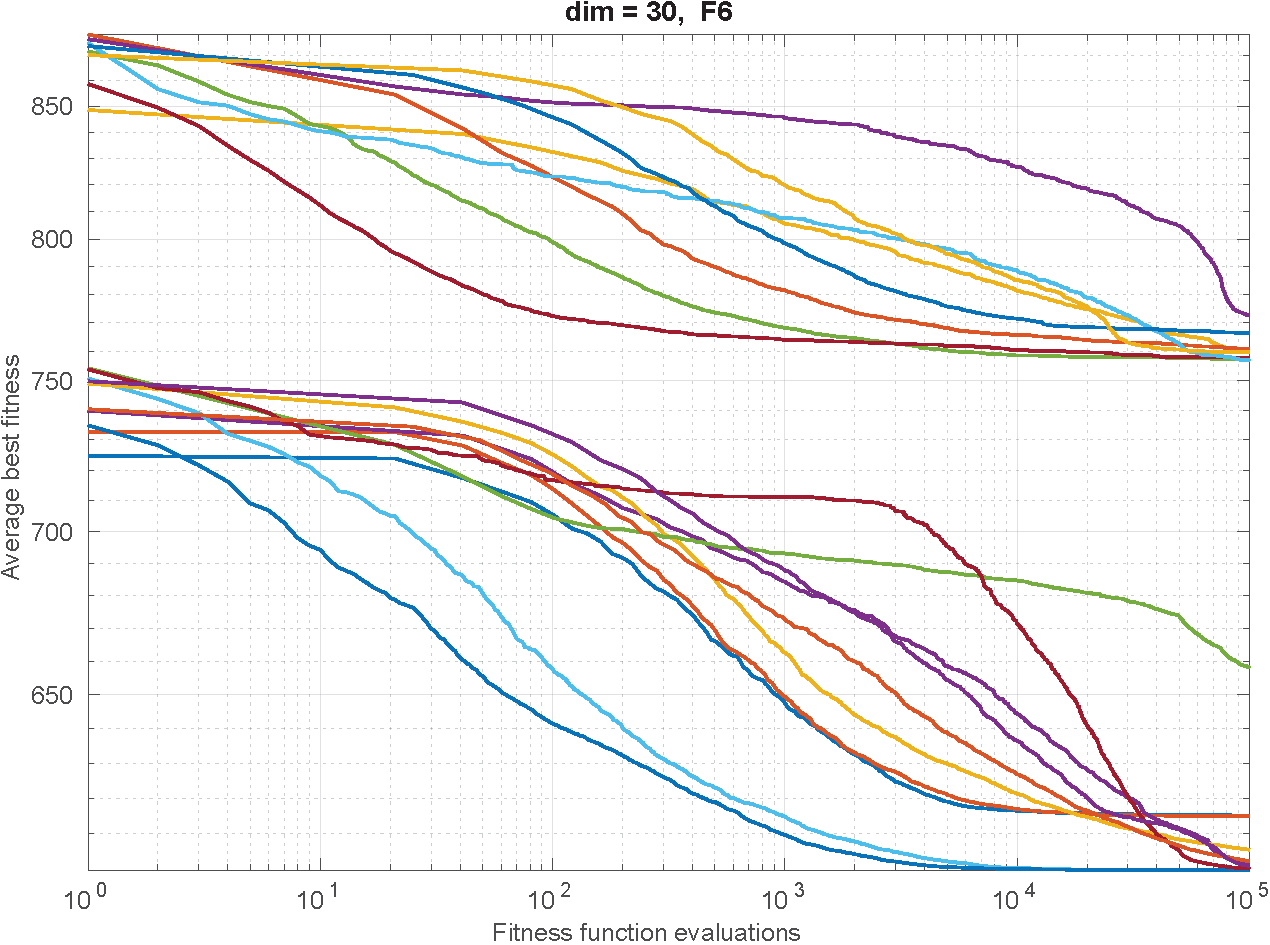}
   \end{subfigure}
   \end{adjustwidth}
\end{figure}
\begin{figure}[H]
  \centering
\begin{adjustwidth}{-1cm}{-2.0cm}
   \begin{subfigure}{0.35\textwidth}
        \includegraphics[height=4cm,width=5.5cm]{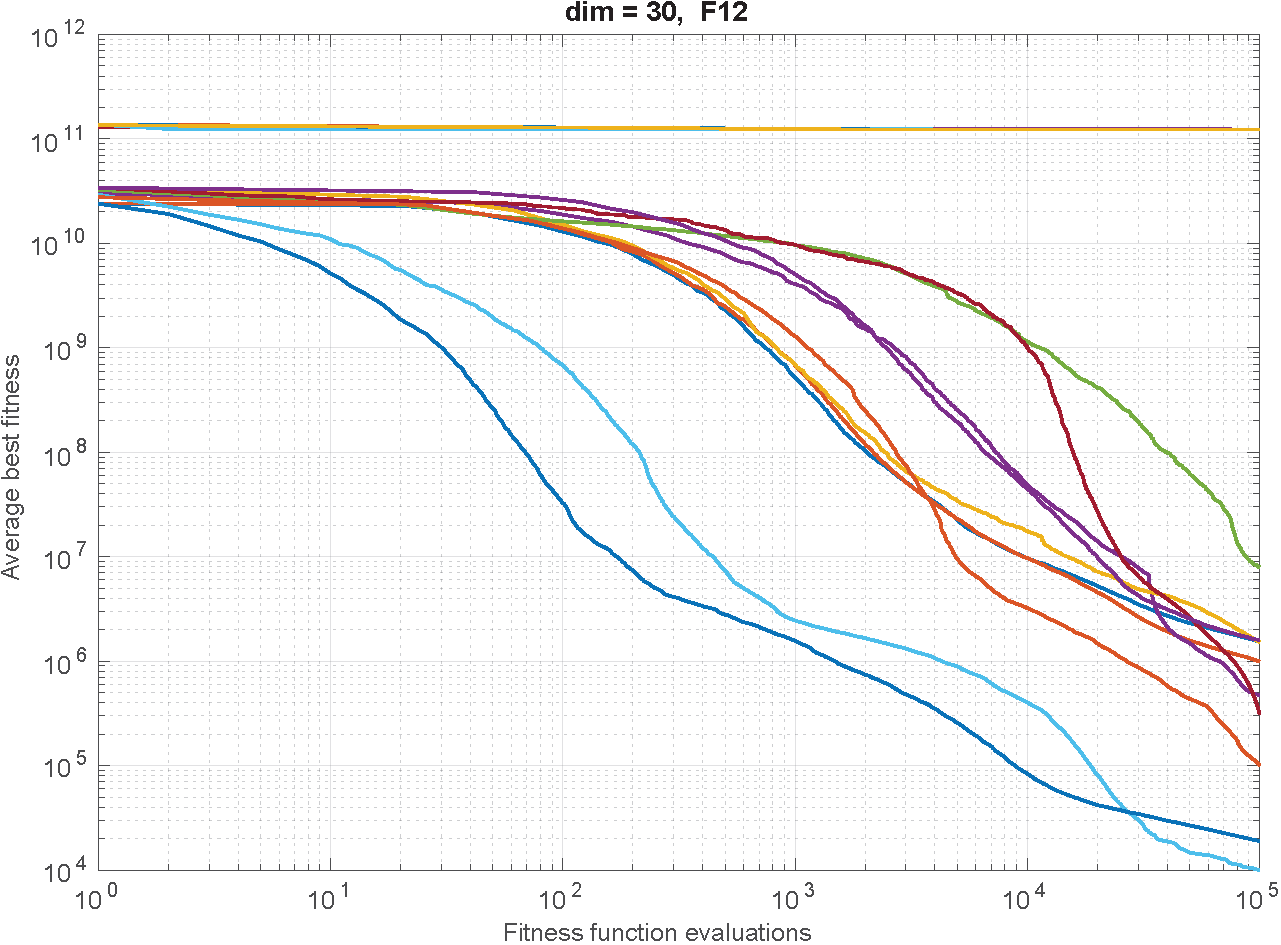}
  \end{subfigure}\hfill
  \begin{subfigure}{0.35\textwidth} 
        \includegraphics[height=4cm,width=5.5cm]{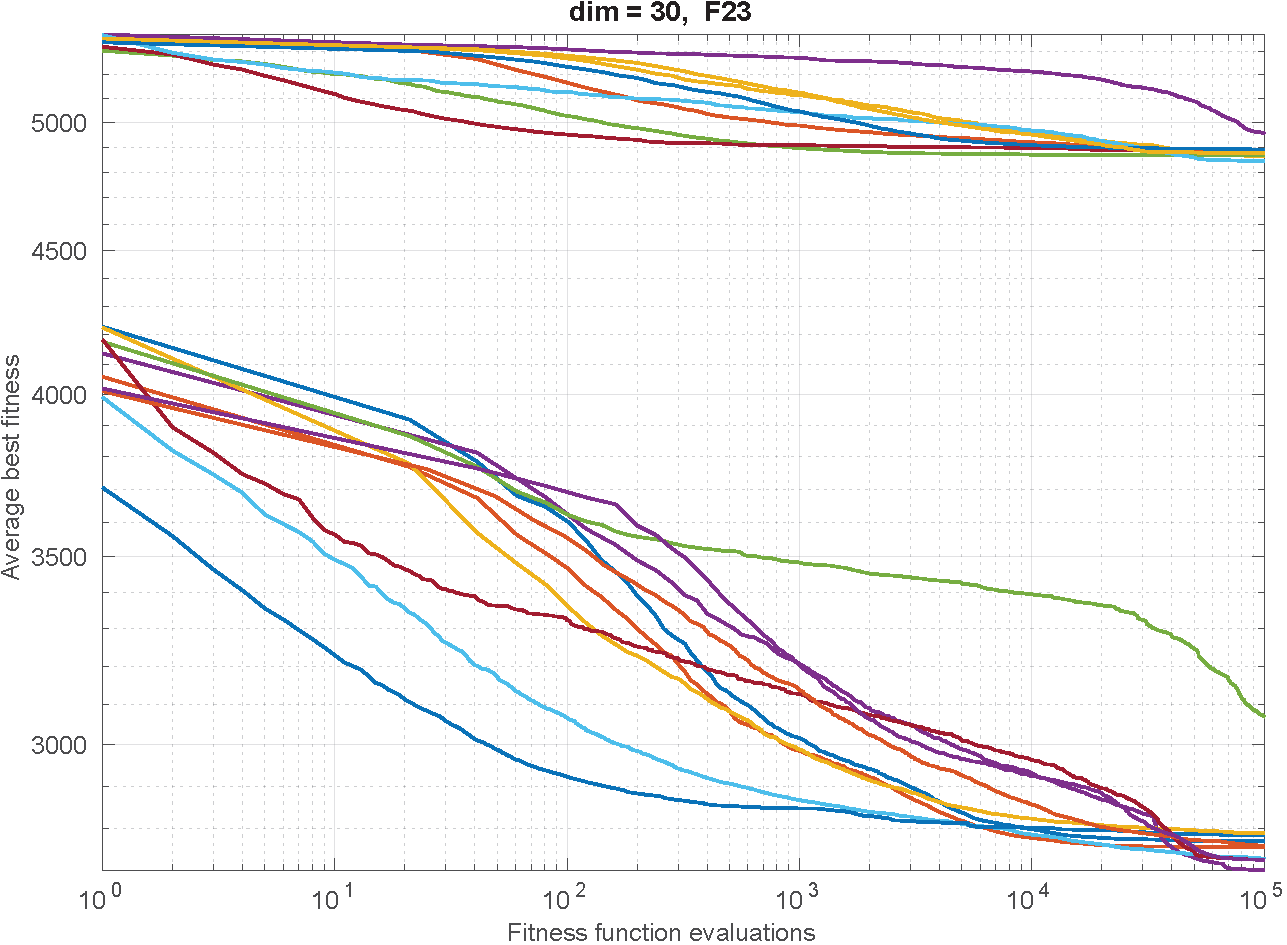}
  \end{subfigure}\hfill
  \begin{subfigure}{0.35\textwidth}
        \includegraphics[height=4cm,width=5.5cm]{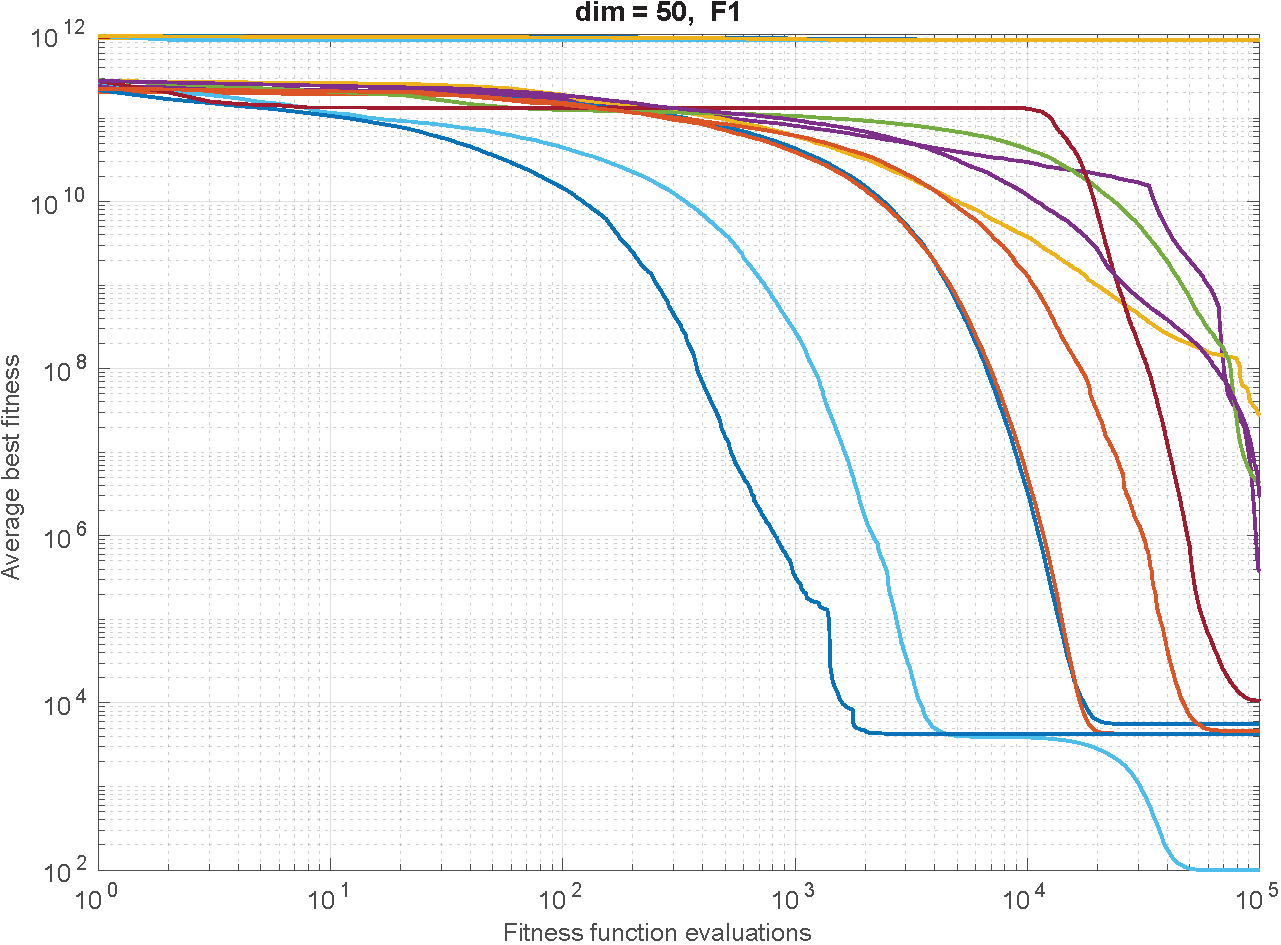}
  \end{subfigure}
  \end{adjustwidth}
\end{figure}

\begin{figure}[H]
  \centering
\begin{adjustwidth}{-1cm}{-2.0cm}
  \begin{subfigure}{0.35\textwidth} 
        \includegraphics[height=4cm,width=5.5cm]{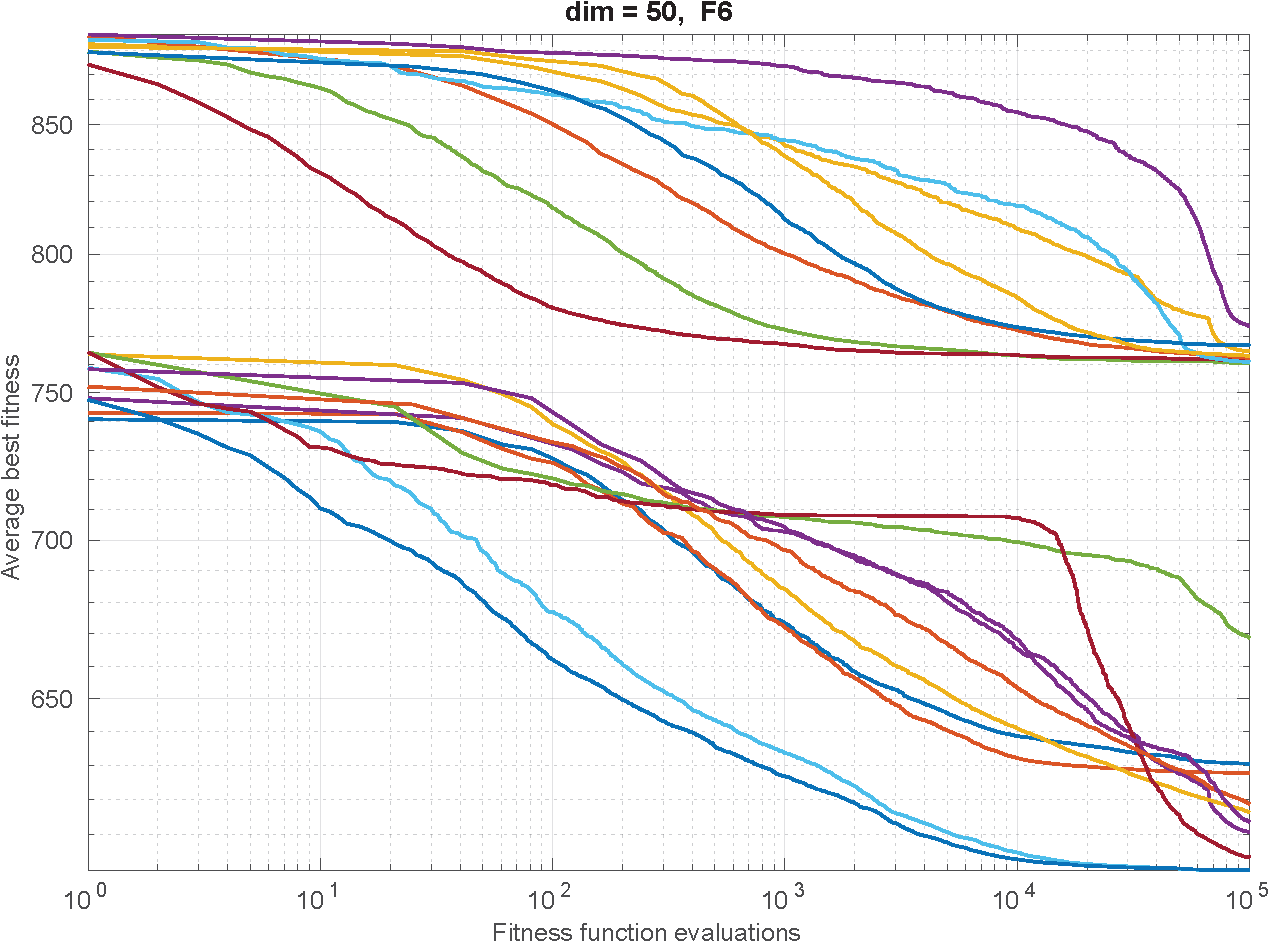}
  \end{subfigure}\hfill
  \begin{subfigure}{0.36\textwidth}
    \includegraphics[height=4cm,width=5.5cm]{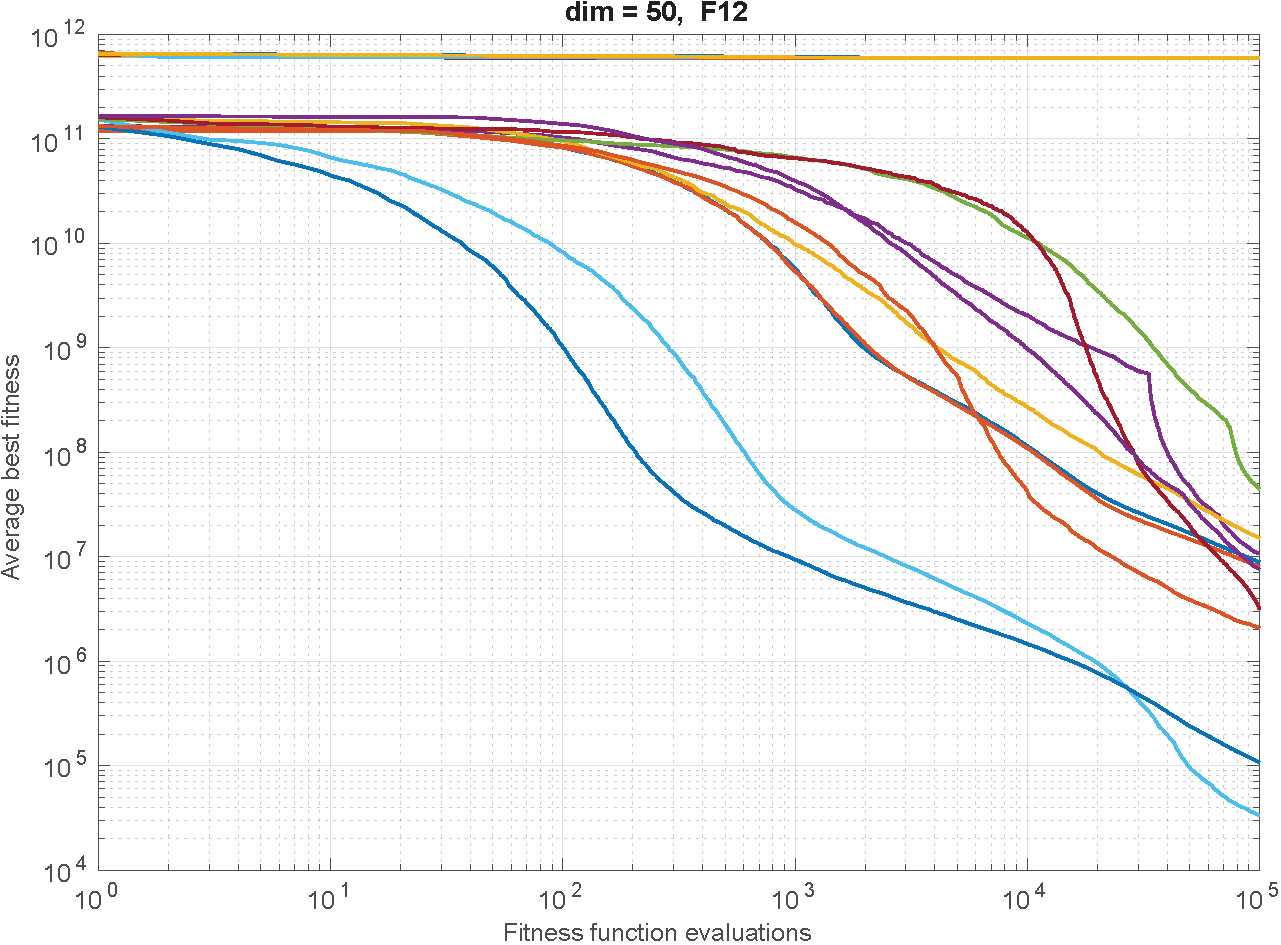}
     \end{subfigure}\hfill
 \begin{subfigure}{0.35\textwidth}
    \includegraphics[height=4cm,width=5.5cm]{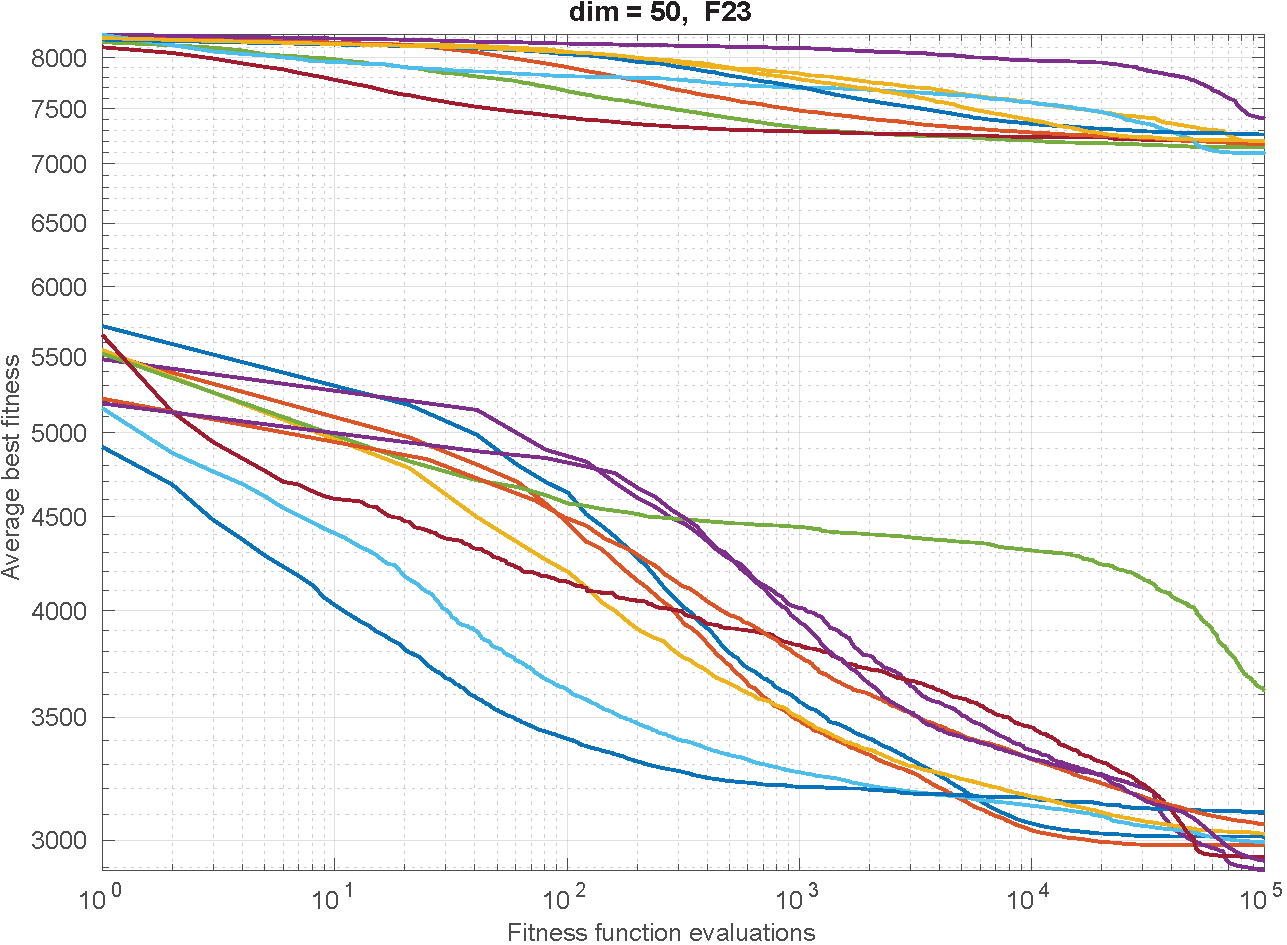}
  \end{subfigure}
\end{adjustwidth}
\end{figure}

\begin{figure}[H]
  \centering
\begin{adjustwidth}{-1cm}{-2.0cm}
  \begin{subfigure}{0.35\textwidth}
    \includegraphics[height=4cm,width=5.5cm]{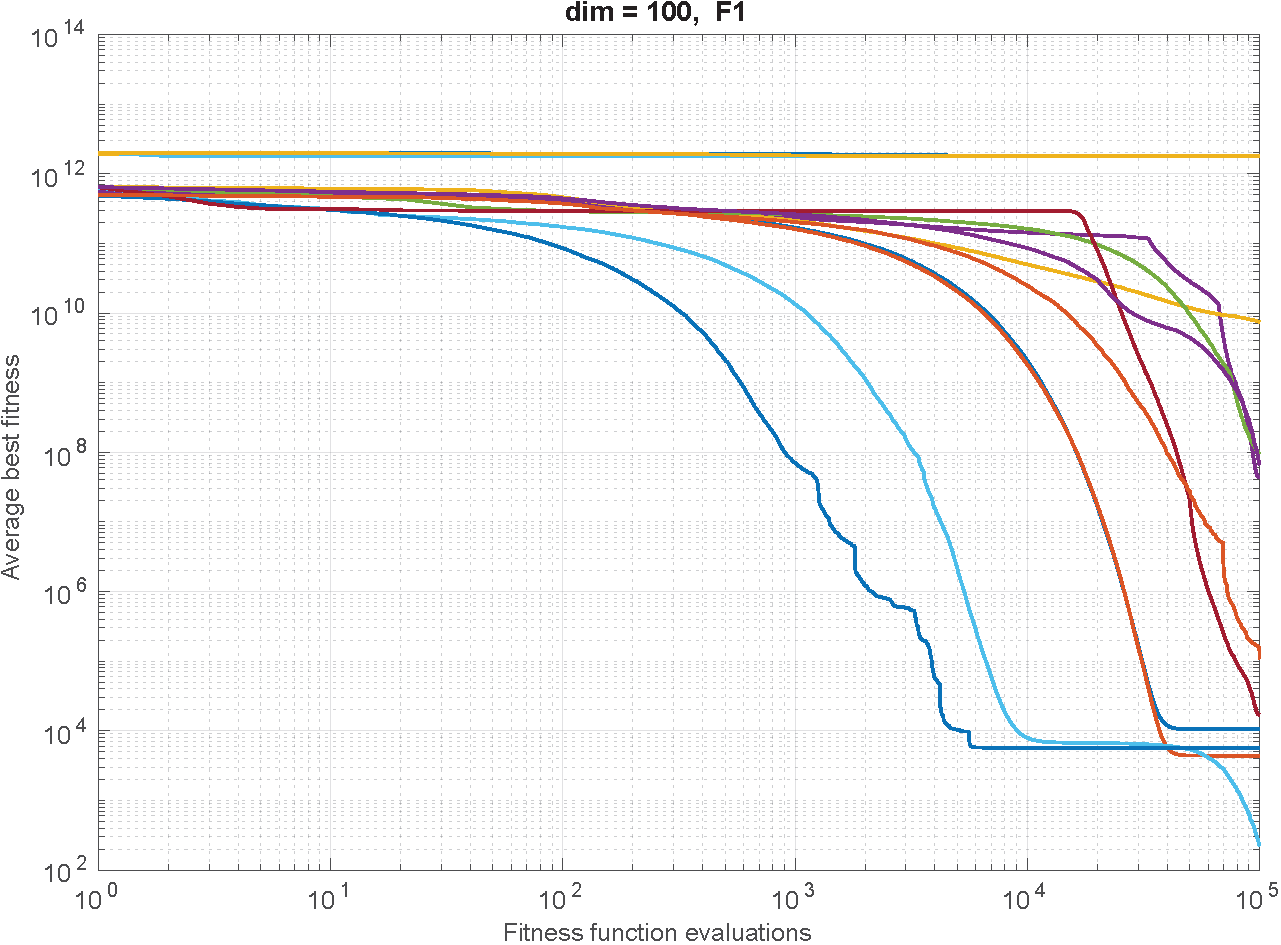}
  \end{subfigure}\hfill
  \begin{subfigure}{0.35\textwidth}
        \includegraphics[height=4cm,width=5.5cm]{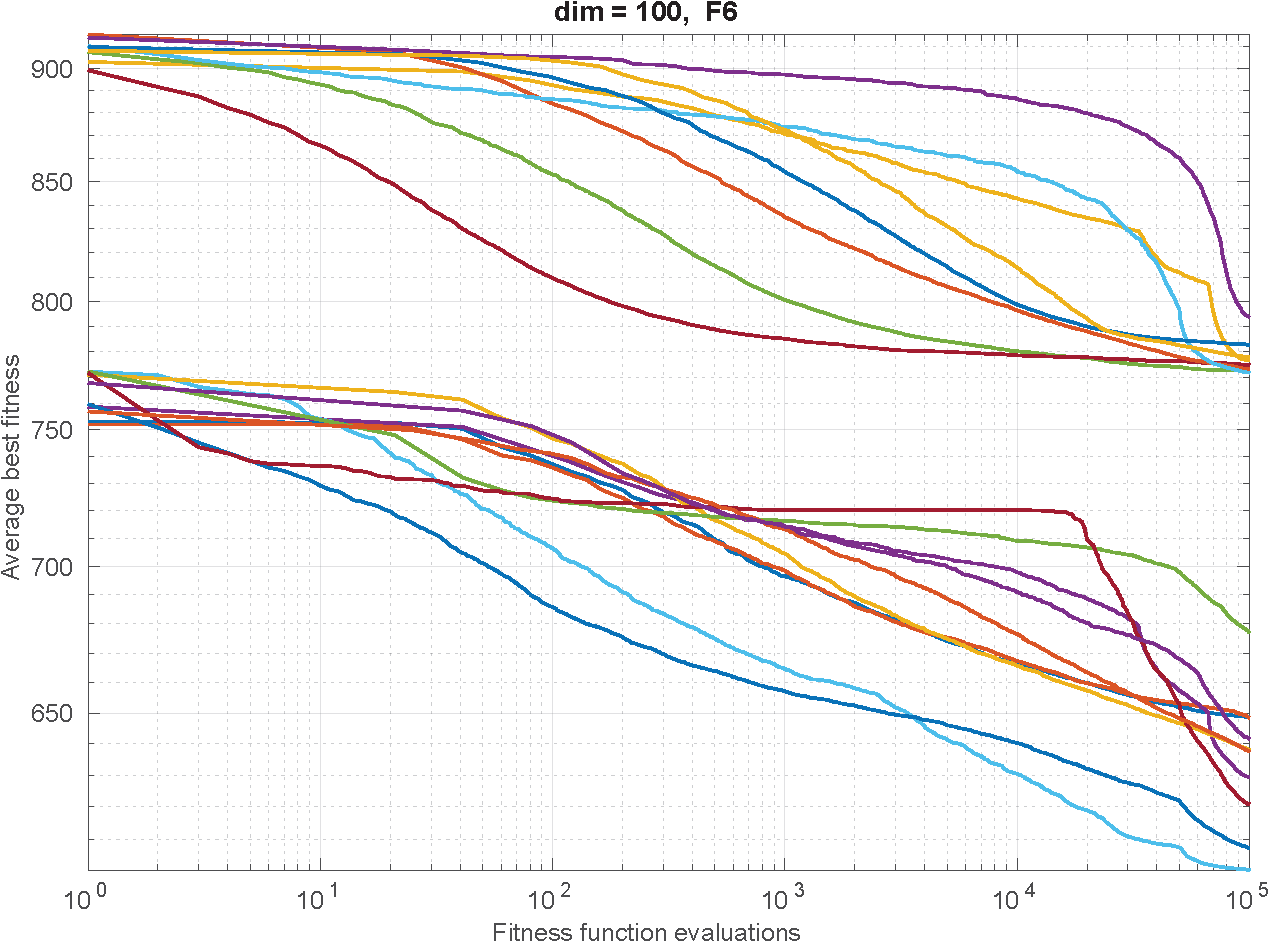}
     \end{subfigure}\hfill
   \begin{subfigure}{0.35\textwidth}
        \includegraphics[height=4cm,width=5.5cm]{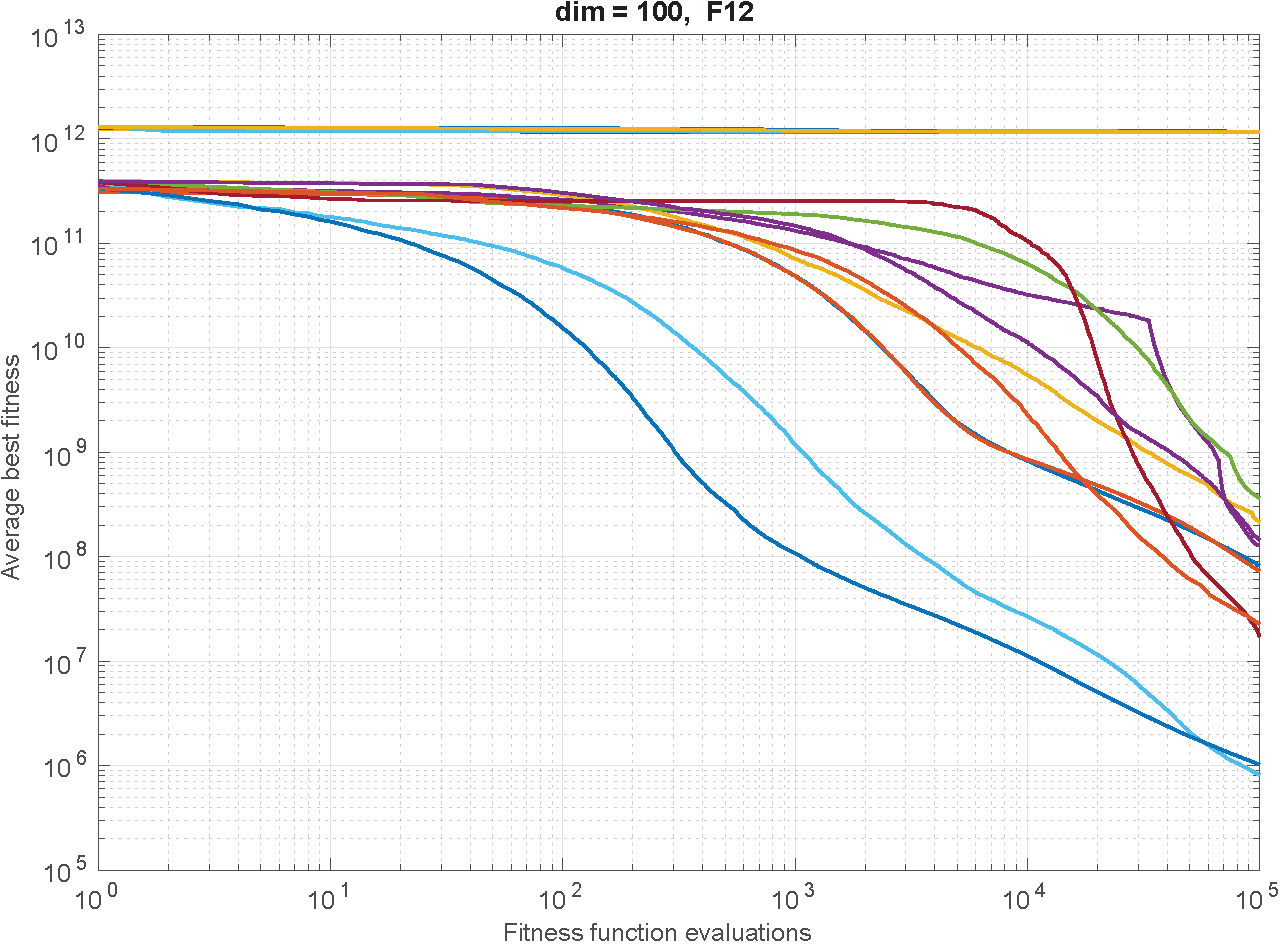}
    \end{subfigure}
    \end{adjustwidth}
\end{figure}

\begin{figure}[H]
 \centering
\begin{adjustwidth}{-1.4cm}{-2.0cm}
\begin{center}
  \begin{subfigure}{0.35\textwidth}
   \centering
        \includegraphics[height=3.8cm,width=5.2cm]{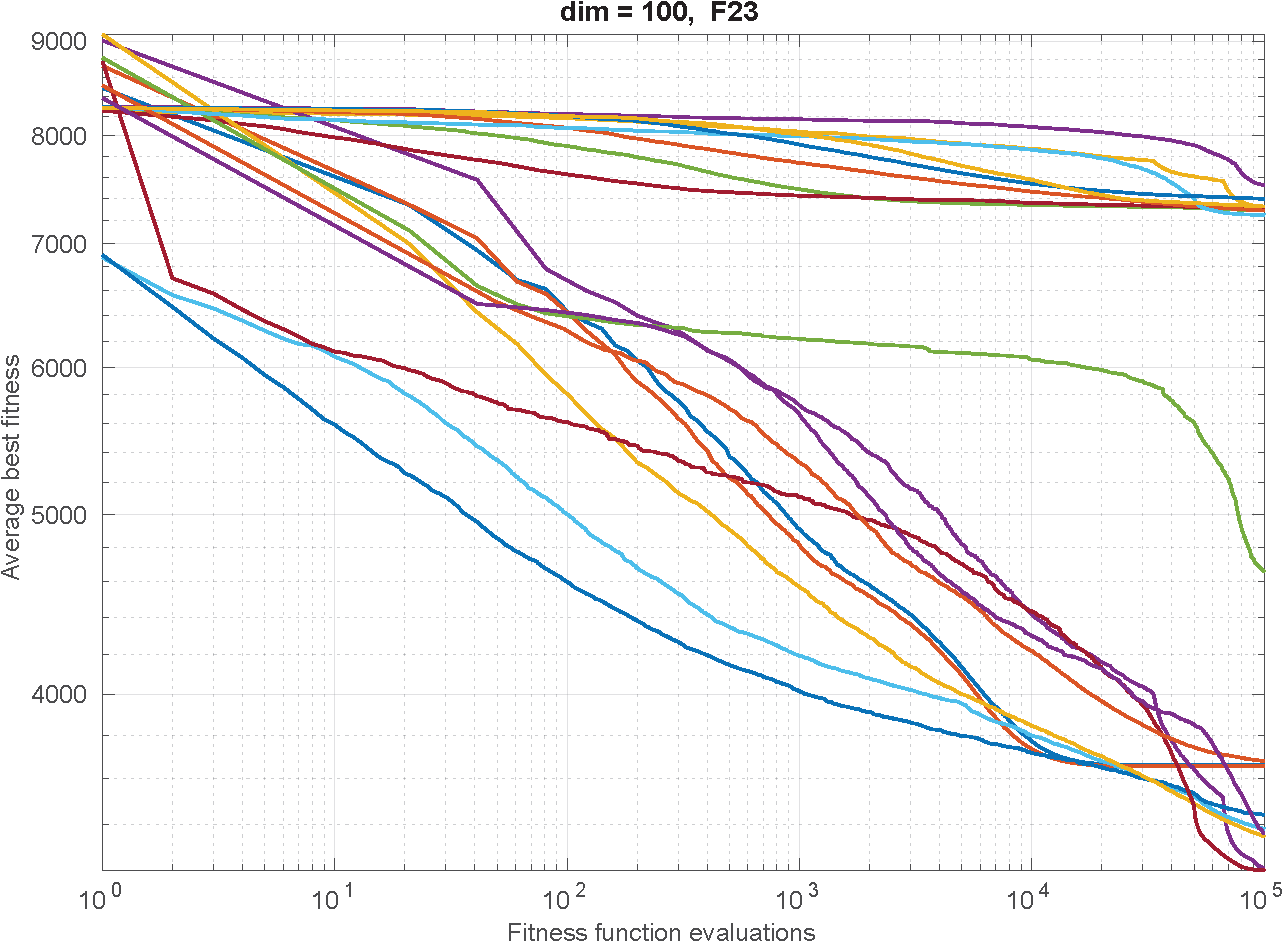}   
  \end{subfigure}
  \end{center}
  \begin{center}
  \begin{subfigure}{0.85\textwidth}
   \centering
   \vspace{-0.2cm}
    \includegraphics[height=0.75cm, width=\linewidth]{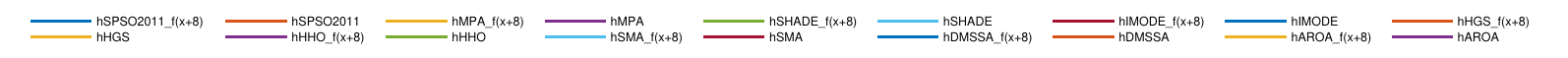}     
  \end{subfigure}
  \end{center}
   \vspace{-0.2cm}
  \caption{\small{Convergence trajectories of metaheuristic algorithms on a representative CEC-2017 benchmark function.}}
  \label{fig8}
\end{adjustwidth}
\end{figure}

\vspace{-0.55cm}
\subsection{Conclusions from the Analysis of Objective Function Translation.}

Following the translation transformation, many algorithms exhibit a drastic decline in optimization performance. For instance, the hybrid MPA version recorded an increase in the mean objective value from approximately $2.6 \times 10^3$ to $2.3 \times 10^{10}$, with the median shifting from about $1.6 \times 10^3$ to $2.9 \times 10^4$. Similar trends are observed in algorithms such as hHGS, hHHO, hSMA, and hAROA, whose average values escalated to the order of $10^{10}$, compared to baseline values in the low thousands.

An opposite pattern is observed for more robust methods like hSHADE, hIMODE, and hDMSSA, which, despite the translated decision space, maintain relatively low medians (around $\sim\!100$). Although some of these methods show slight increases in the mean and shifts in central tendency, their spread remains significantly smaller than that of less resilient algorithms. Moreover, many methods exhibit increased variance - for example, hMPA shows an apparent drop in standard deviation (from $5.2 \times 10^3$ to $1.3 \times 10^1$), while simultaneously the median increases, indicating instability and distribution skew.

Overall, the tabular analysis  (Tab.\ref{tab2}) shows that translation strongly differentiates method performance: some maintain their previous ranks, while others suffer dramatic drops. These differences are especially pronounced in higher dimensions ($dim = 50, 100$), where the effect of translation intensifies due to greater dispersion in the decision space.

The boxplots (Fig.\ref{fig5}) confirm these observations, highlighting reduced stability and an increase in extreme values. For representative functions (f1, f6, f12, f23), algorithms lacking translation invariance show significantly higher medians and broader dispersion compared to the baseline (Tab.\ref{tab1}). 
Numerous outliers frequently appear, which - according to prior analysis - is a clear sign of instability. In contrast, translation-invariant algorithms (e.g., hSHADE, hIMODE) yield compact distributions centered around new but stable values, indicating effective adaptation to the transformed space.

Friedman test results (Fig.\ref{fig6}) also confirm statistically significant differences across algorithms for all dimensions considered ($p \ll 0.05$). In the critical difference (CD $\approx $ 4.81) plots, hybrids like hIMODE and hSHADE consistently occupy top positions, while others shift beyond the threshold of statistical significance. For example, hMPA\_f(x+8) drops in ranking (more losses than wins), while hSPSO2011\_f(x+8) shows modest improvement (mean rank changing from $\sim\!8.9$  to $\sim\!8.7$ ). Overall, translation weakens the position of non-invariant methods, whereas invariant algorithms maintain or enhance their competitiveness.

Bayesian analysis (probability heatmap, (Fig.\ref{fig7}) offers consistent support. Top-performing algorithms such as hIMODE and hSHADE still achieve win probabilities of 90-100\% against nearly all rivals-both before and after translation. Meanwhile, weaker methods (e.g., HHO, HGS) dominate only in isolated comparisons. This confirms that translation affects not only rankings but also the structure of statistical dominance. The structural advantage of top algorithms is therefore not accidental but stems from their robustness to decision space deformation.

Convergence curve analysis (Fig.\ref{fig8}) also reveals clear delays and optimization slowdown after translation. For f1 (Sphere function), most algorithms slow down significantly in reaching the optimum. For f6, f12, and f23, plateau phenomena and reduced convergence rates are frequently observed. In particular, non-invariant algorithms (e.g., hHHO, CMA-ES) exhibit nearly flat trajectories, indicating loss of exploratory capability. Conversely, methods like hSHADE and hIMODE, though slower, steadily approach the shifted optimum ($\sim\!100$ instead of $\sim\!0$), demonstrating preserved adaptive ability. Algorithms like hMPA and hDMSSA show irregular convergence - fluctuations and stalls suggest difficulty in exploring the new optimum location.

These phenomena have solid theoretical foundations: the lack of translation invariance implies that most classical mutation, recombination, and update operators (e.g., in HHO, HGS) operate relative to a static coordinate system. After translation, the reference point is shifted, disrupting the exploration process - the population ''loses'' track of the new optimum, and search efficiency drops. Algorithms employing translation-invariant operators (e.g., SBX, DE, CMA-ES) exhibit greater resilience to such perturbations, as confirmed in \cite{2},\cite{6atz}. Importantly, this effect intensifies with increasing dimensionality ($dim = 50, 100$), as the sparser decision space makes it more difficult to accurately locate the optimum, thereby significantly limiting exploration effectiveness.

The statistical, graphical, and convergence analyses consistently demonstrate that objective function translation significantly affects algorithm performance, and only a few hybrid methods exhibit structural resistance to such distortions, making them more reliable for applications involving unknown optimum locations.

\vspace{-0.18cm}
\subsection{Scale Invariance}
\vspace{-0.1cm}
Scaling of the decision space, defined as the transformation of the form
 $f(x) \rightarrow f(\alpha x)$, where $\alpha \in \mathbb{R}$ alters the rate of change of the objective function values with respect to the coordinates, without modifying the topology of the objective surface. Although, in theory, such a transformation should not affect the optimization outcome, in practice, it may disrupt the balance between exploration and exploitation - especially for algorithms that rely on scale-sensitive operators whose parameters depend on the search range (e.g., vector differences, learning coefficients). This subsection evaluates the resilience of the studied methods to scaling by analyzing their performance under the transformation $f(x) \rightarrow f(5x)$, with particular attention to how dimensionality influences the stability and efficiency of the optimization process.

\FloatBarrier

\scriptsize
\setlength{\tabcolsep}{4pt}

\begin{table}[H]
\centering
\begin{adjustwidth}{-1.5cm}{-2.4cm}
\caption{\footnotesize{Statistical results for 9 hybrid algorithms on CEC-2017 functions and their scaled variants across 4 dims.}}
\scalebox{0.65}{
\renewcommand{\arraystretch}{0.9}
\begin{tabular}{@{}lrrrrrrrrrrrrrrr@{}}
\toprule
\multirow{2}{*}{Algorithm} & \multicolumn{7}{c}{dim=10} & \multicolumn{7}{c}{dim=30} \\
\cmidrule(lr){2-8} \cmidrule(lr){9-15}
 & Avg & Med & Std & Sum Rank & Mean Rank & +/- & p-value & Avg & Med & Std & Sum Rank & Mean Rank & +/- & p-value \\
\midrule
hSPSO2011\_f(5x)  & 1.8E+04 & 1.9E+03 & 2.3E+04 & 351.5 & 12.1 & 138/275 & 7.4E-02 & 5.3E+04 & 2.8E+03 & 3.8E+04 & 349 & 12.0 & 122/290 & 6.6E-02 \\
hSPSO2011  & 1.6E+04 & 1.9E+03 & 2.0E+04 & 352.5 & 12.2 & 140/270 & 7.3E-02 & 5.9E+04 & 2.8E+03 & 3.9E+04 & 361 & 12.4 & 121/292 & 7.2E-02 \\
hMPA\_f(5x)  & 1.7E+03 & 1.6E+03 & 2.5E+01 & 235 & 8.1 & 255/187 & 6.7E-02 & 2.5E+04 & 2.3E+03 & 2.3E+04 & 270 & 9.3 & 214/202 & 8.0E-02 \\
hMPA  & 2.6E+03 & 1.6E+03 & 5.2E+03 & 232 & 8.0 & 257/186 & 6.8E-02 & 1.9E+04 & 2.3E+03 & 1.7E+04 & 269 & 9.3 & 215/206 & 7.5E-02 \\
hSHADE\_f(5x)  & -1.6E-01 & 2.3E-01 & 9.5E+00 & 94 & 3.2 & 401/17 & 1.0E-01 & -1.2E-01 & -5.2E-02 & 1.0E+01 & 105 & 3.6 & 376/23 & 1.0E-01 \\
hSHADE  & -6.1E-01 & 1.4E+00 & 4.7E+01 & 90 & 3.1 & 398/36 & 7.9E-02 & -2.2E-01 & 4.5E-01 & 5.2E+01 & 94 & 3.2 & 374/28 & 9.2E-02 \\
hIMODE\_f(5x)  & -3.3E-01 & -1.5E-01 & 1.0E+01 & 77 & 2.7 & 402/14 & 1.1E-01 & -6.3E-02 & 1.6E-01 & 1.1E+01 & 105 & 3.6 & 376/23 & 1.0E-01 \\
hIMODE  & -1.5E+00 & 1.0E+00 & 5.1E+01 & 80 & 2.8 & 400/33 & 7.5E-02 & -2.3E-01 & 1.3E+00 & 5.5E+01 & 97 & 3.3 & 372/23 & 9.8E-02 \\
hHGS\_f(5x)  & 1.2E+04 & 2.6E+03 & 1.5E+04 & 422 & 14.6 & 66/350 & 6.9E-02 & 1.5E+05 & 3.2E+03 & 3.2E+05 & 414 & 14.3 & 84/335 & 6.3E-02 \\
hHGS  & 1.1E+04 & 2.6E+03 & 1.4E+04 & 425 & 14.7 & 65/349 & 6.7E-02 & 1.4E+05 & 3.2E+03 & 3.3E+05 & 413 & 14.2 & 82/337 & 6.7E-02 \\
hHHO\_f(5x)  & 3.9E+04 & 2.1E+03 & 5.3E+04 & 480 & 16.6 & 18/430 & 3.5E-02 & 4.5E+05 & 3.3E+03 & 4.5E+05 & 499 & 17.2 & 7/450 & 2.7E-02 \\
hHHO  & 5.8E+04 & 2.3E+03 & 7.1E+04 & 496 & 17.1 & 12/438 & 3.4E-02 & 3.5E+05 & 3.3E+03 & 3.1E+05 & 494 & 17.0 & 15/448 & 2.8E-02 \\
hSMA\_f(5x)  & 3.7E+03 & 1.9E+03 & 1.1E+03 & 312 & 10.8 & 161/257 & 7.2E-02 & 1.9E+04 & 2.9E+03 & 1.0E+04 & 307 & 10.6 & 161/233 & 8.4E-02 \\
hSMA  & 3.6E+03 & 1.9E+03 & 1.1E+03 & 318 & 11.0 & 160/258 & 7.0E-02 & 2.1E+04 & 2.9E+03 & 9.6E+03 & 306 & 10.6 & 158/234 & 9.4E-02 \\
hDMSSA\_f(5x)  & -1.0E-01 & 2.0E-01 & 9.9E+00 & 94 & 3.2 & 400/16 & 1.1E-01 & 2.5E-02 & 3.6E-01 & 1.0E+01 & 110 & 3.8 & 376/25 & 9.8E-02 \\
hDMSSA  & 9.7E+01 & 9.2E+01 & 5.3E+00 & 174 & 6.0 & 348/145 & 3.6E-20 & 3.0E-01 & 2.2E-01 & 5.2E+01 & 96 & 3.3 & 370/34 & 8.1E-02 \\
hAROA\_f(5x)  & 5.4E+03 & 1.8E+03 & 9.3E+03 & 372 & 12.8 & 122/305 & 6.2E-02 & 7.2E+04 & 2.7E+03 & 5.7E+04 & 341 & 11.8 & 138/259 & 8.2E-02 \\
hAROA  & 5.4E+03 & 1.7E+03 & 9.2E+03 & 356 & 12.3 & 119/296 & 6.2E-02 & 6.7E+04 & 2.9E+03 & 3.5E+04 & 327 & 11.3 & 139/258 & 9.3E-02 \\
\bottomrule
\end{tabular}
}
\vspace{1.1mm}
\scalebox{0.65}{
\renewcommand{\arraystretch}{0.9}
\begin{tabular}{@{}l
ccccccc  
@{\hspace{0.5cm}}
ccccccc
@{}}
\toprule
\multirow{2}{*}{Algorithm} & \multicolumn{7}{c}{dim=50} & \multicolumn{7}{c}{dim=100} \\
\cmidrule(lr){2-8} \cmidrule(lr){9-15}
 & Avg & Med & Std & Sum Rank & Mean Rank & +/- & p-value & Avg & Med & Std & Sum Rank & Mean Rank & +/- & p-value \\
\midrule
hSPSO2011\_f(5x)  & 8.6E+05 & 3.3E+03 & 3.0E+05 & 362 & 12.5 & 107/302 & 8.5E-02 & 3.2E+06 & 5.9E+03 & 1.2E+06 & 378 & 13.0 & 111/331 & 5.0E-02 \\
hSPSO2011  & 8.8E+05 & 3.3E+03 & 2.4E+05 & 378 & 13.0 & 102/306 & 8.4E-02 & 3.1E+06 & 6.0E+03 & 1.0E+06 & 390 & 13.4 & 110/332 & 5.3E-02 \\
hMPA\_f(5x)  & 5.0E+05 & 3.1E+03 & 3.4E+05 & 277 & 9.6 & 195/221 & 7.1E-02 & 6.3E+06 & 5.3E+03 & 2.7E+06 & 320 & 11.0 & 164/254 & 6.7E-02 \\
hMPA  & 4.3E+05 & 3.2E+03 & 3.3E+05 & 288 & 9.9 & 192/215 & 7.2E-02 & 6.0E+06 & 5.3E+03 & 2.8E+06 & 308 & 10.6 & 171/244 & 6.8E-02 \\
hSHADE\_f(5x)  & 2.0E-01 & 5.0E-01 & 1.1E+01 & 104 & 3.6 & 381/21 & 1.0E-01 & -1.4E-01 & 1.3E-01 & 1.0E+01 & 99 & 3.4 & 376/22 & 9.2E-02 \\
hSHADE  & 1.1E+00 & 1.9E+00 & 5.3E+01 & 104 & 3.6 & 368/40 & 8.7E-02 & -6.2E-01 & -7.9E-02 & 5.2E+01 & 109 & 3.8 & 374/29 & 9.5E-02 \\
hIMODE\_f(5x)  & 1.8E-01 & -6.6E-02 & 1.1E+01 & 104 & 3.6 & 382/20 & 1.0E-01 & -1.3E-01 & 9.0E-02 & 1.1E+01 & 101 & 3.5 & 375/25 & 9.9E-02 \\
hIMODE  & 8.0E-01 & 5.1E-01 & 5.6E+01 & 96 & 3.3 & 369/34 & 9.8E-02 & -8.9E-01 & -1.1E-01 & 5.5E+01 & 87 & 3.0 & 375/30 & 9.2E-02 \\
hHGS\_f(5x)  & 3.2E+06 & 4.0E+03 & 5.2E+06 & 413 & 14.2 & 73/336 & 6.7E-02 & 2.8E+08 & 6.7E+03 & 1.2E+08 & 410 & 14.1 & 79/333 & 6.8E-02 \\
hHGS  & 1.7E+06 & 4.0E+03 & 1.6E+06 & 418 & 14.4 & 76/339 & 7.0E-02 & 2.7E+08 & 7.0E+03 & 1.2E+08 & 410 & 14.1 & 87/339 & 6.3E-02 \\
hHHO\_f(5x)  & 4.1E+06 & 4.2E+03 & 3.0E+06 & 497 & 17.1 & 10/444 & 3.2E-02 & 7.0E+07 & 8.2E+03 & 3.1E+07 & 489 & 16.9 & 20/451 & 2.3E-02 \\
hHHO  & 2.2E+06 & 4.0E+03 & 1.2E+06 & 489 & 16.9 & 18/439 & 3.4E-02 & 1.7E+07 & 7.1E+03 & 8.4E+06 & 468 & 16.1 & 38/426 & 2.8E-02 \\
hSMA\_f(5x)  & 1.8E+05 & 3.3E+03 & 9.0E+04 & 299 & 10.3 & 187/223 & 7.9E-02 & 5.7E+05 & 5.1E+03 & 2.5E+05 & 248 & 8.6 & 242/191 & 4.9E-02 \\
hSMA  & 1.7E+05 & 3.3E+03 & 7.6E+04 & 289 & 10.0 & 189/222 & 8.0E-02 & 6.2E+05 & 4.9E+03 & 2.6E+05 & 247 & 8.5 & 242/190 & 5.1E-02 \\
hDMSSA\_f(5x)  & 8.0E-02 & -9.5E-02 & 1.1E+01 & 96 & 3.3 & 383/22 & 8.8E-02 & -7.8E-02 & 4.2E-02 & 1.0E+01 & 105 & 3.6 & 377/25 & 8.5E-02 \\
hDMSSA  & 8.4E-01 & 1.2E+00 & 5.2E+01 & 105 & 3.6 & 371/29 & 8.4E-02 & -4.5E-01 & -6.3E-02 & 5.1E+01 & 107 & 3.7 & 373/31 & 8.5E-02 \\
hAROA\_f(5x)  & 4.7E+05 & 3.4E+03 & 2.2E+05 & 312 & 10.8 & 152/240 & 9.6E-02 & 6.5E+06 & 5.0E+03 & 2.3E+06 & 341 & 11.8 & 146/271 & 7.1E-02 \\
hAROA  & 4.4E+05 & 3.4E+03 & 2.1E+05 & 324 & 11.2 & 148/250 & 8.8E-02 & 7.4E+06 & 5.6E+03 & 2.5E+06 & 350 & 12.1 & 139/275 & 6.2E-02 \\
\bottomrule
\label{tab3}
\end{tabular}
}
\vspace{0.15Cm}\\
\footnotesize{Friedman test p-values: dim10=6.2E-81, dim30=1.2E-76, dim50=1.1E-76, dim100=3.1E-76}
\end{adjustwidth}
\end{table}

\normalsize

\begin{figure}[H]
  \centering
\begin{adjustwidth}{-1.8cm}{-2.0cm}
  \begin{subfigure}{0.4\textwidth}
     \includegraphics[height=5.5cm,width=6.9cm]{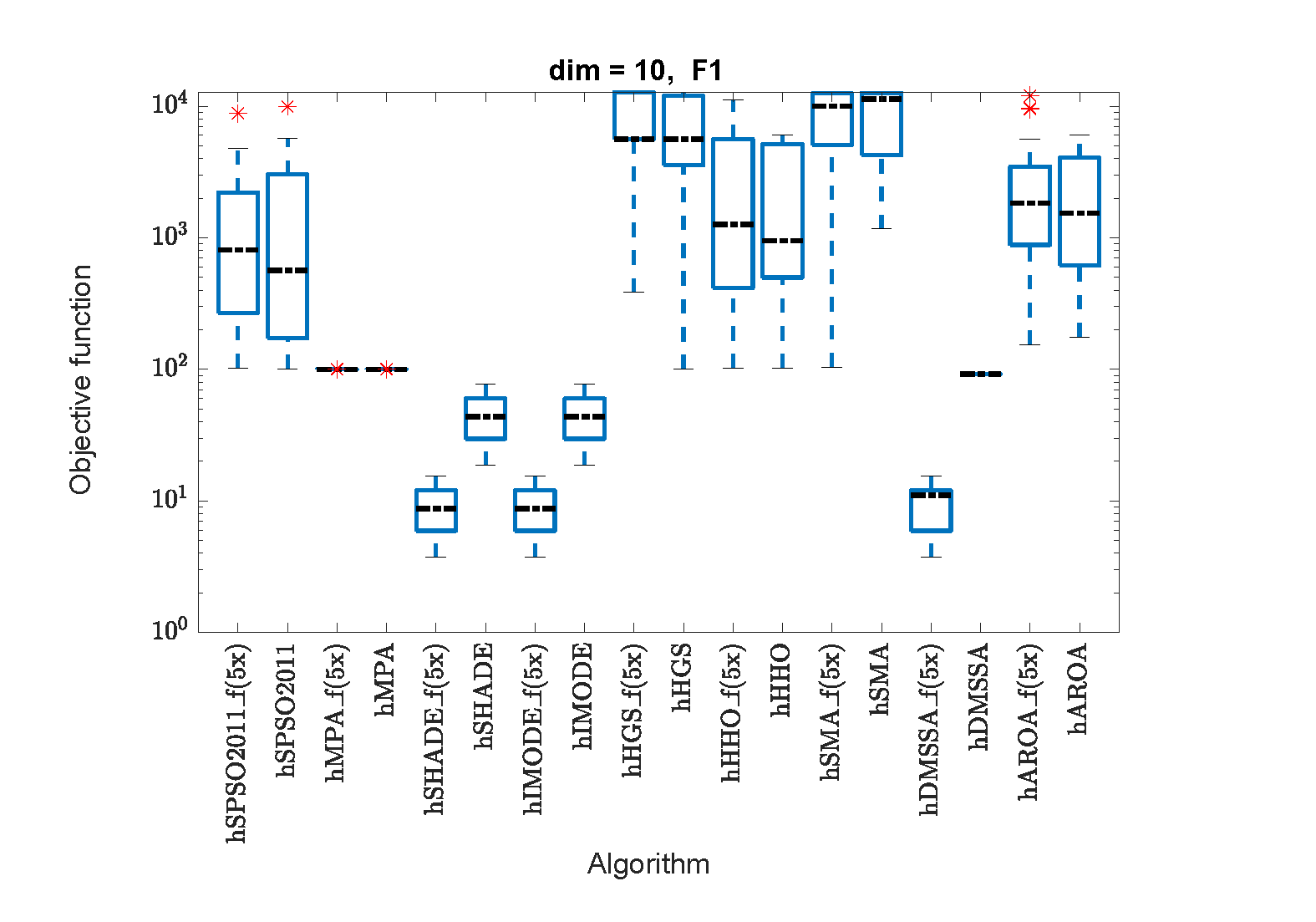}
  \end{subfigure}\hfill
  \begin{subfigure}{0.4\textwidth}
    \includegraphics[height=5.5cm,width=6.9cm]{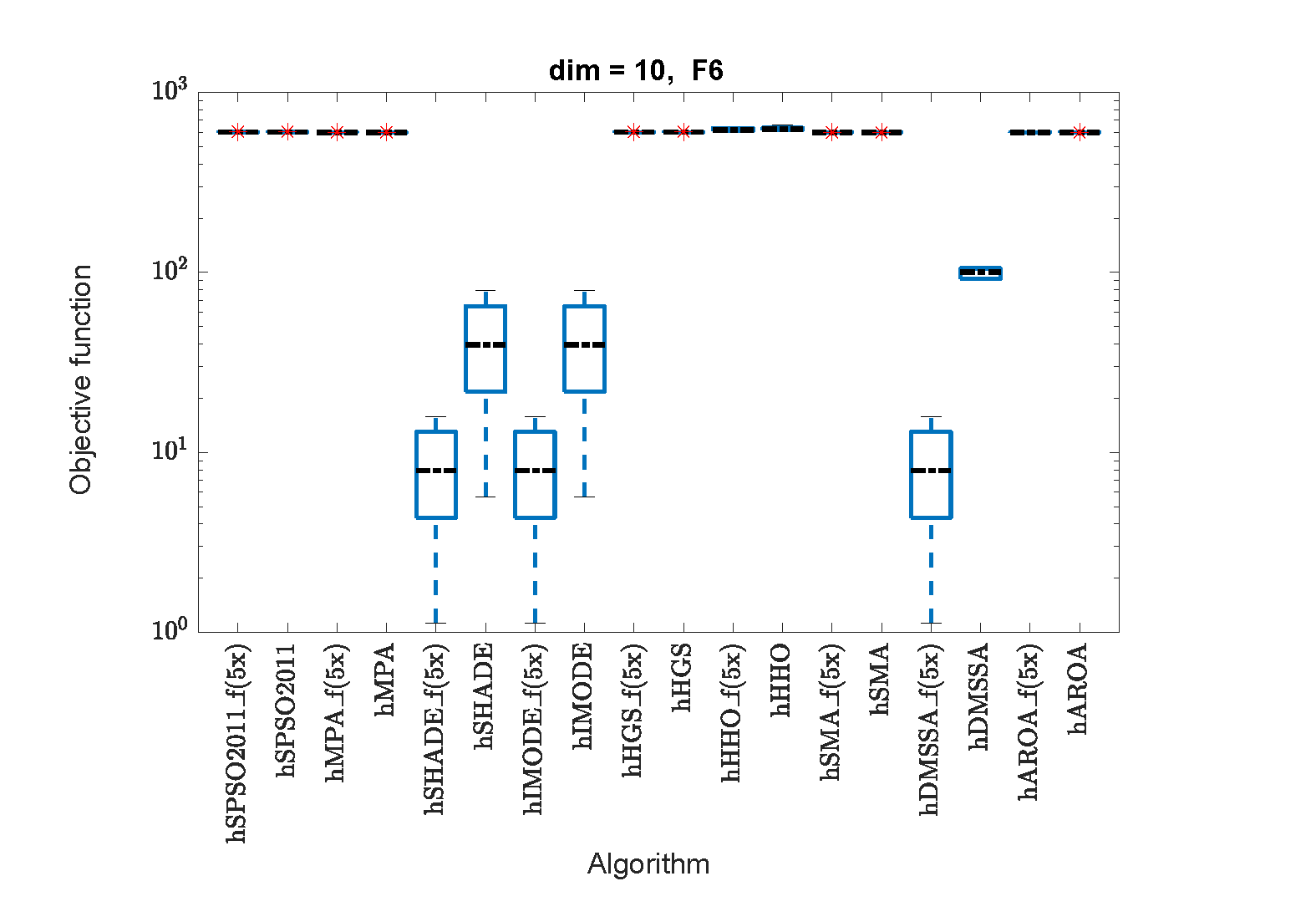}
  \end{subfigure}\hfill
  \begin{subfigure}{0.4\textwidth}
    \includegraphics[height=5.5cm,width=6.9cm]{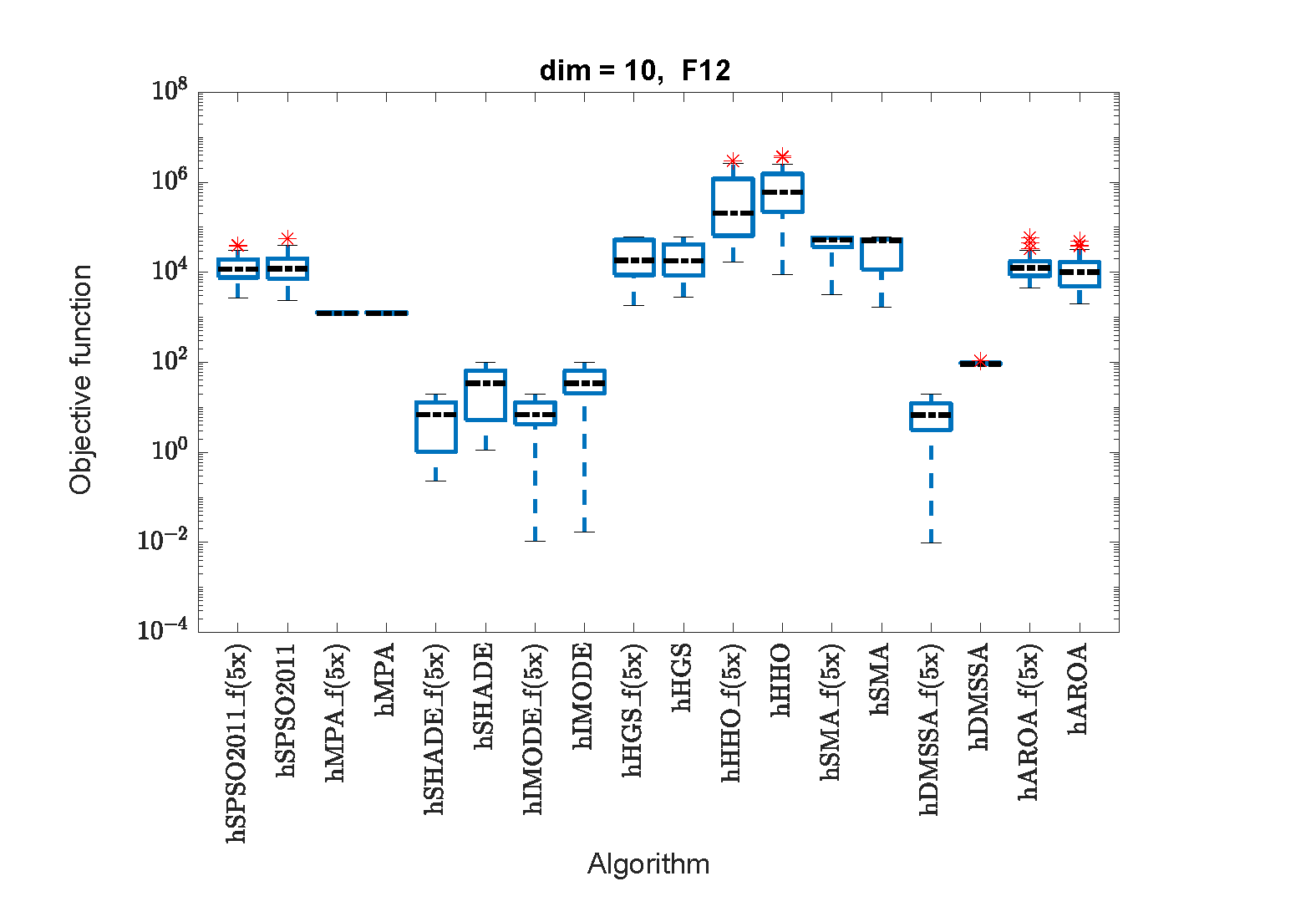}
  \end{subfigure}
 \end{adjustwidth}
\end{figure}

\begin{figure}[H]
  \centering
\begin{adjustwidth}{-2cm}{-2.0cm}
   \begin{subfigure}{0.4\textwidth}
    \includegraphics[height=5.1cm,width=6.9cm]{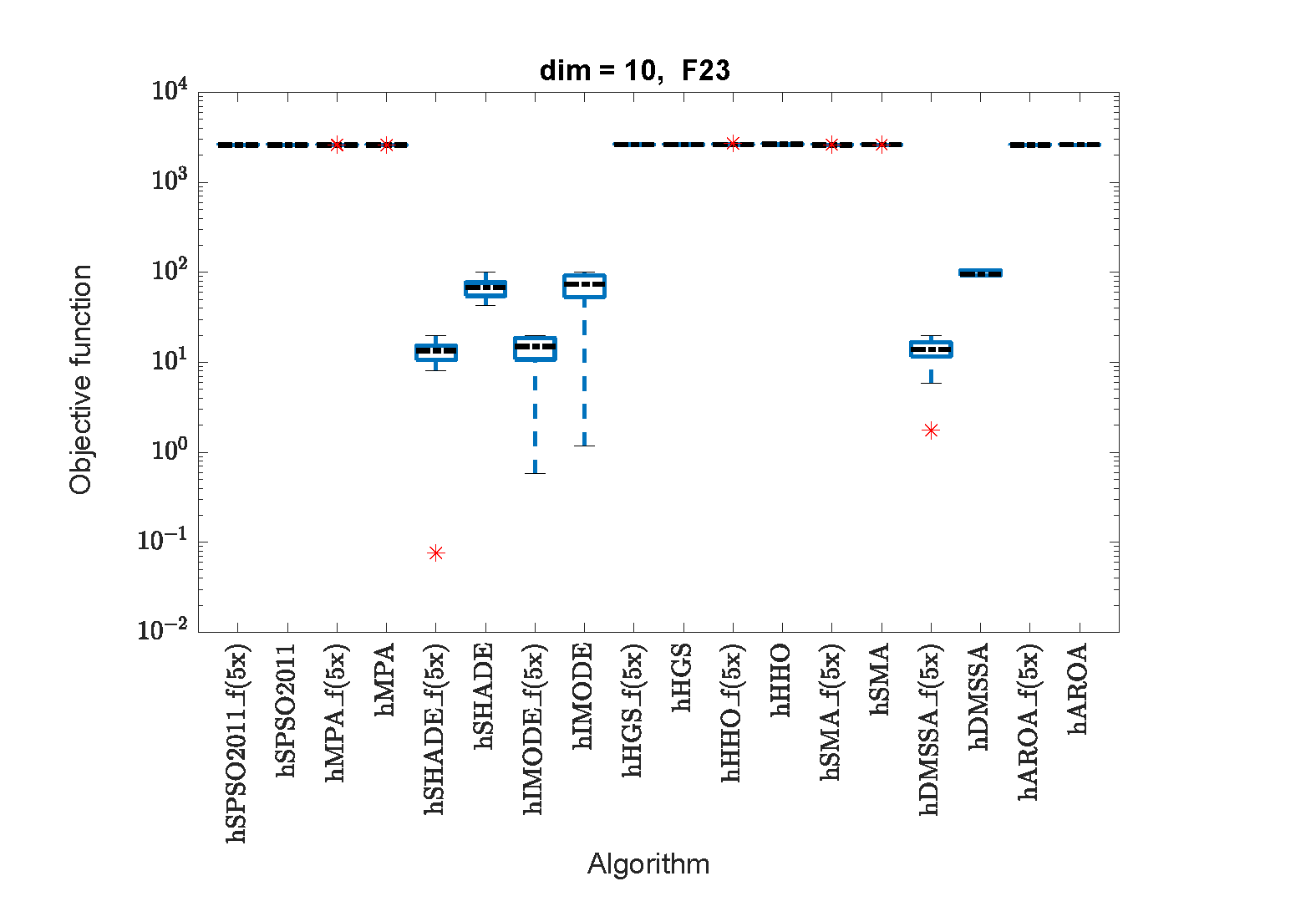}
  \end{subfigure}\hfill
  \begin{subfigure}{0.4\textwidth}
    \includegraphics[height=5.1cm,width=6.9cm]{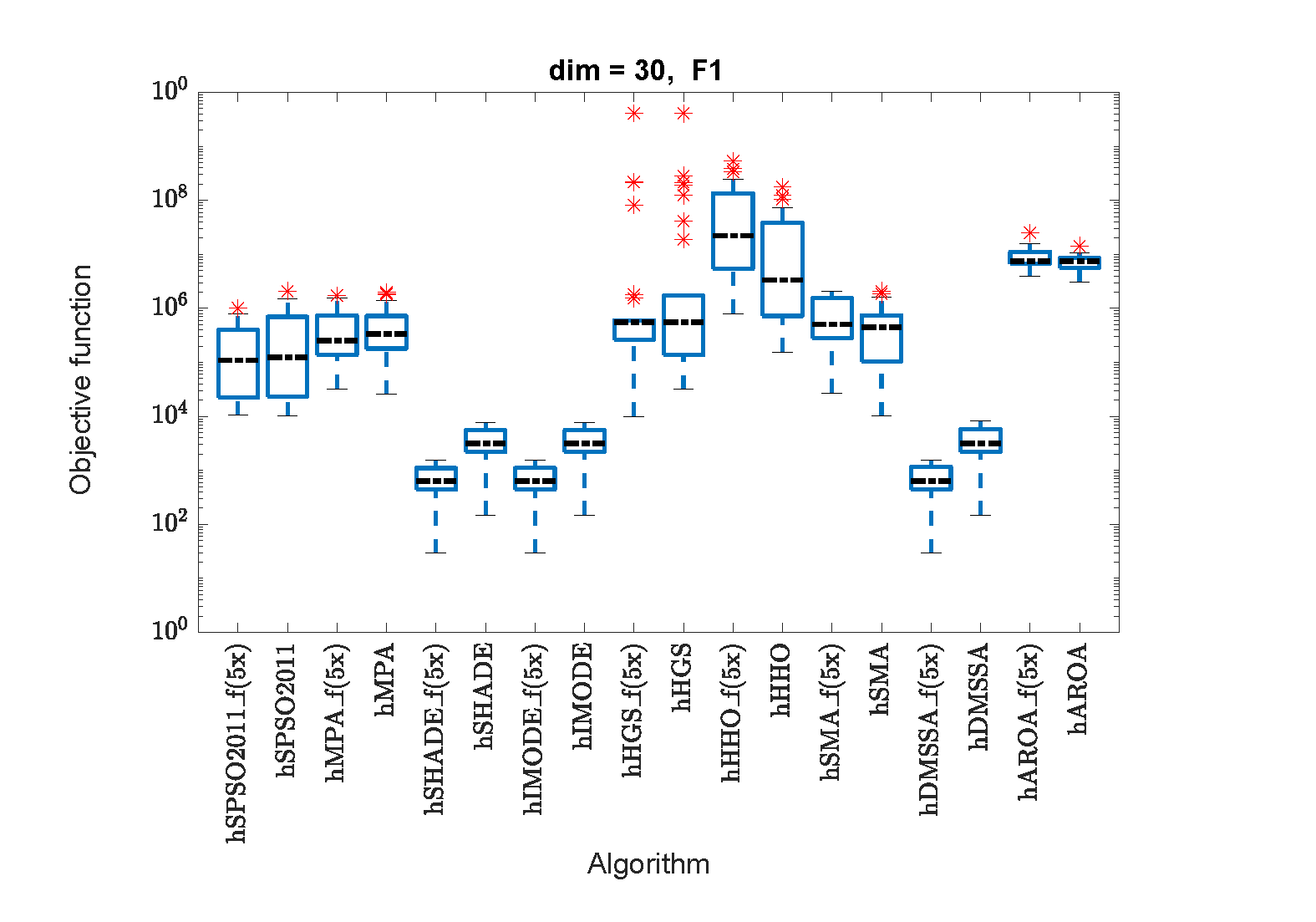}
\end{subfigure}\hfill
  \begin{subfigure}{0.4\textwidth} 
        \includegraphics[height=5.1cm,width=6.9cm]{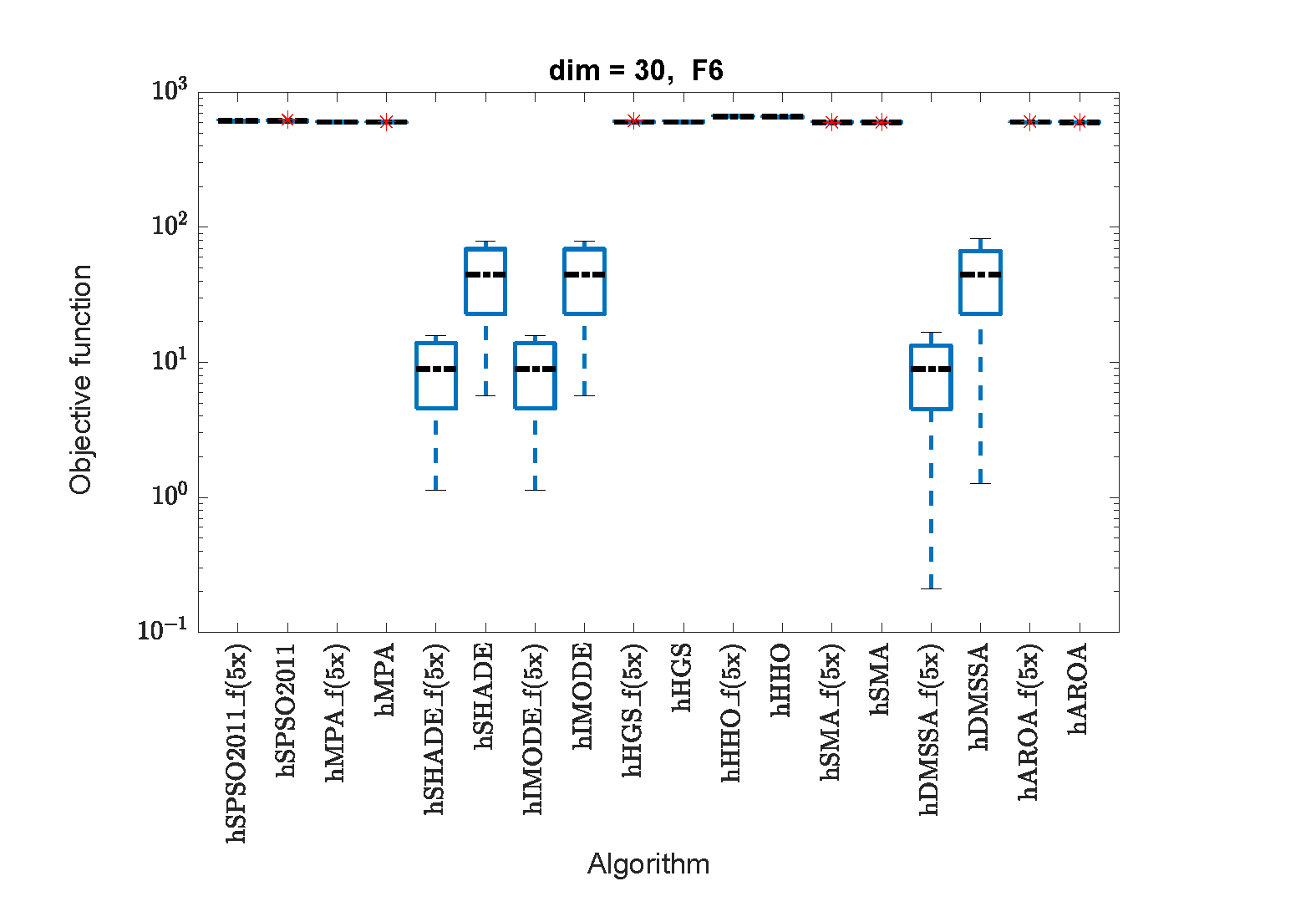}
   \end{subfigure}
   \end{adjustwidth}
\end{figure}

 \begin{figure}[H]
  \centering
\begin{adjustwidth}{-2cm}{-2.0cm}
   \begin{subfigure}{0.4\textwidth}
        \includegraphics[height=5.1cm,width=6.9cm]{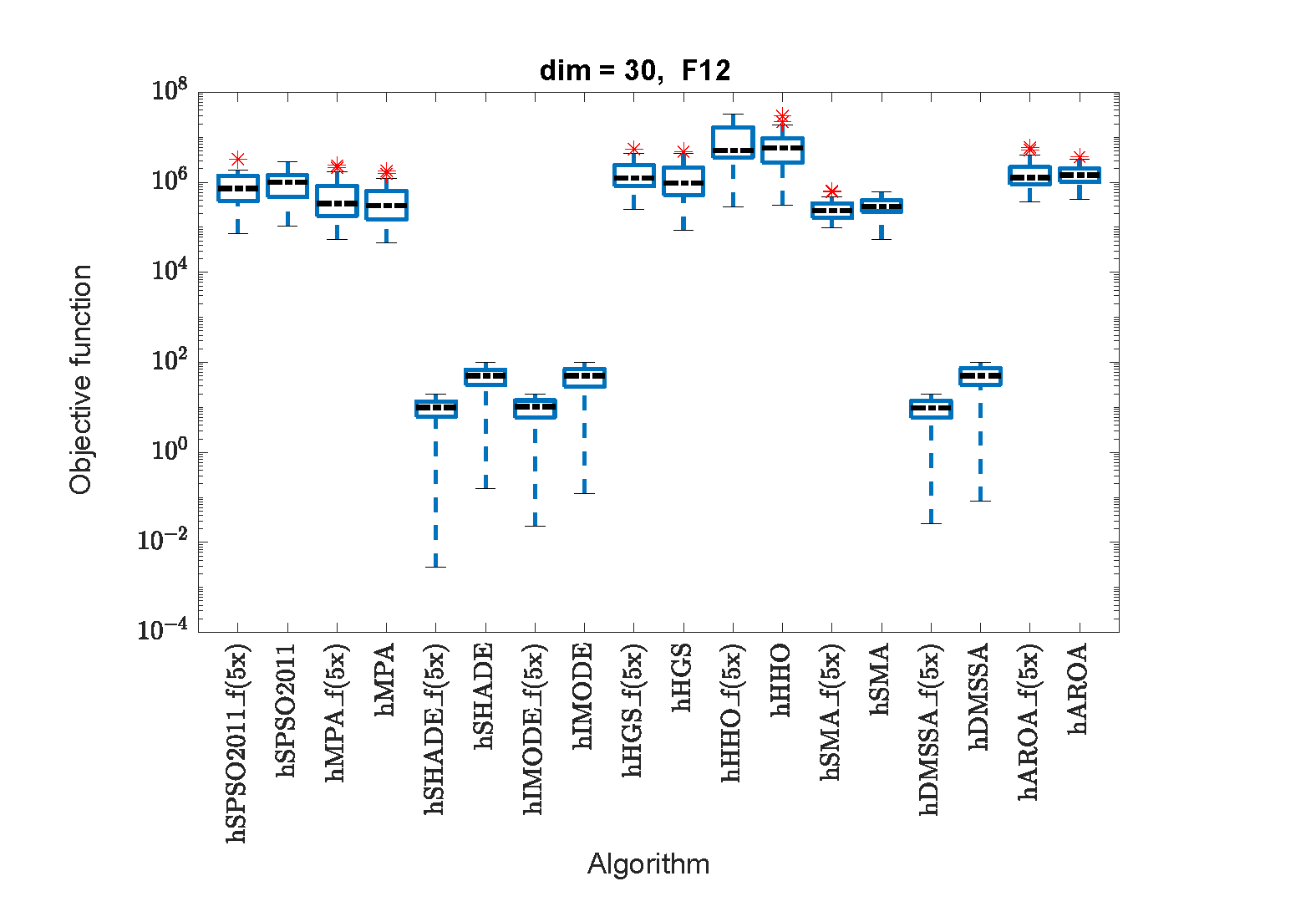}
  \end{subfigure}\hfill
  \begin{subfigure}{0.4\textwidth} 
        \includegraphics[height=5.1cm,width=6.9cm]{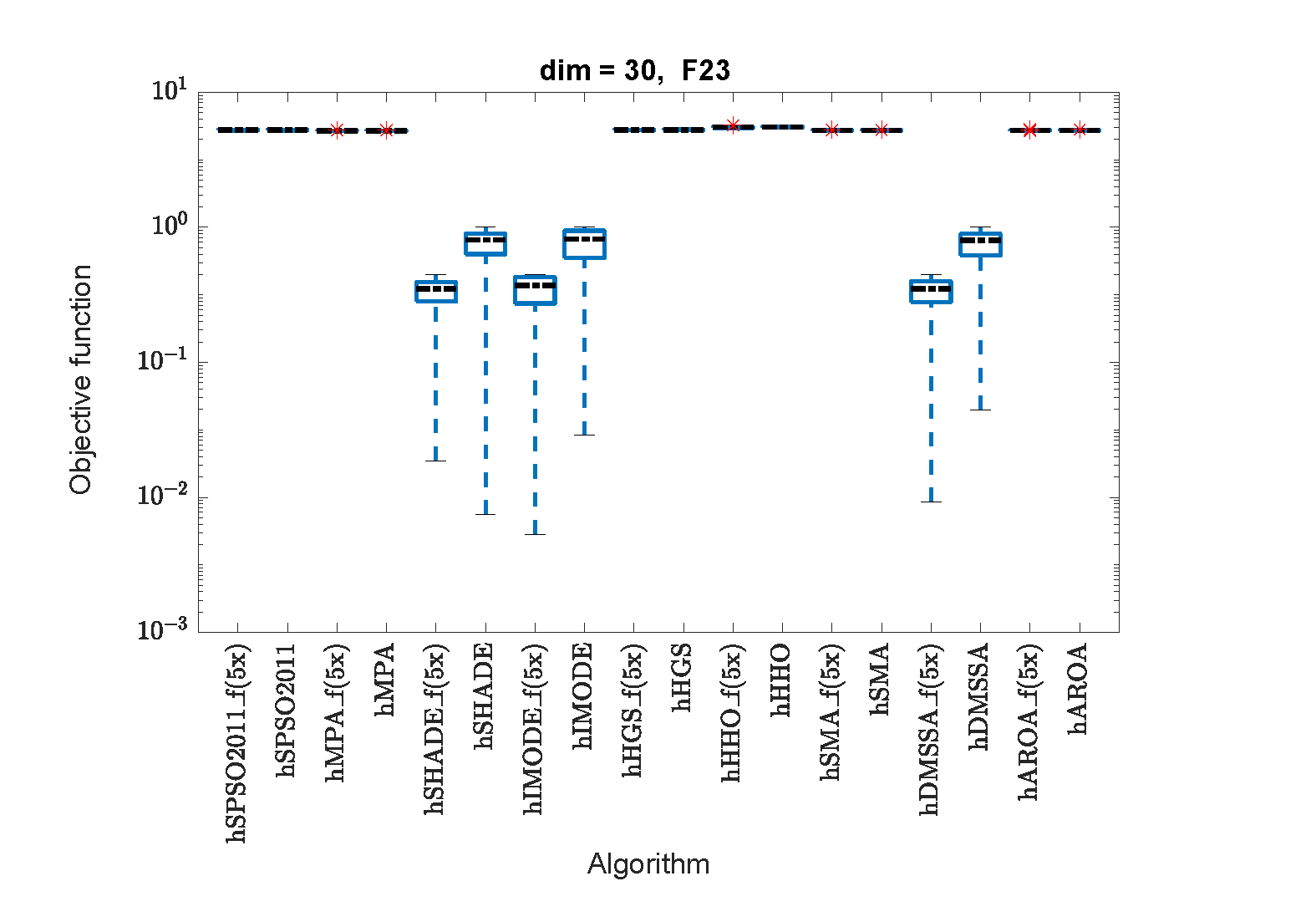}
  \end{subfigure}\hfill
  \begin{subfigure}{0.4\textwidth}
        \includegraphics[height=5.1cm,width=6.9cm]{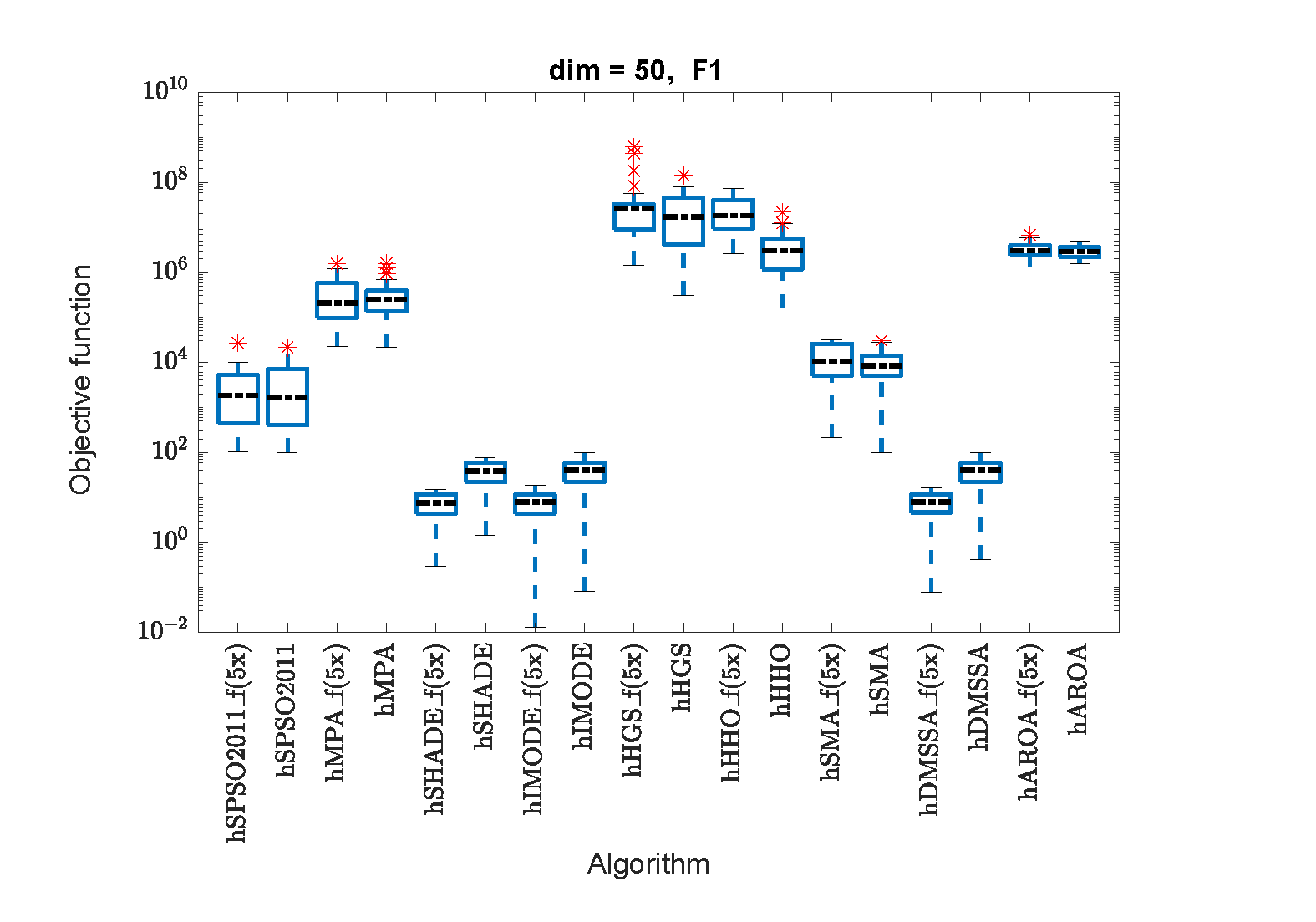}
  \end{subfigure}
  \end{adjustwidth}
\end{figure}

  \begin{figure}[H]
  \centering
\begin{adjustwidth}{-2cm}{-2.0cm}
  \begin{subfigure}{0.4\textwidth} 
        \includegraphics[height=5.1cm,width=6.9cm]{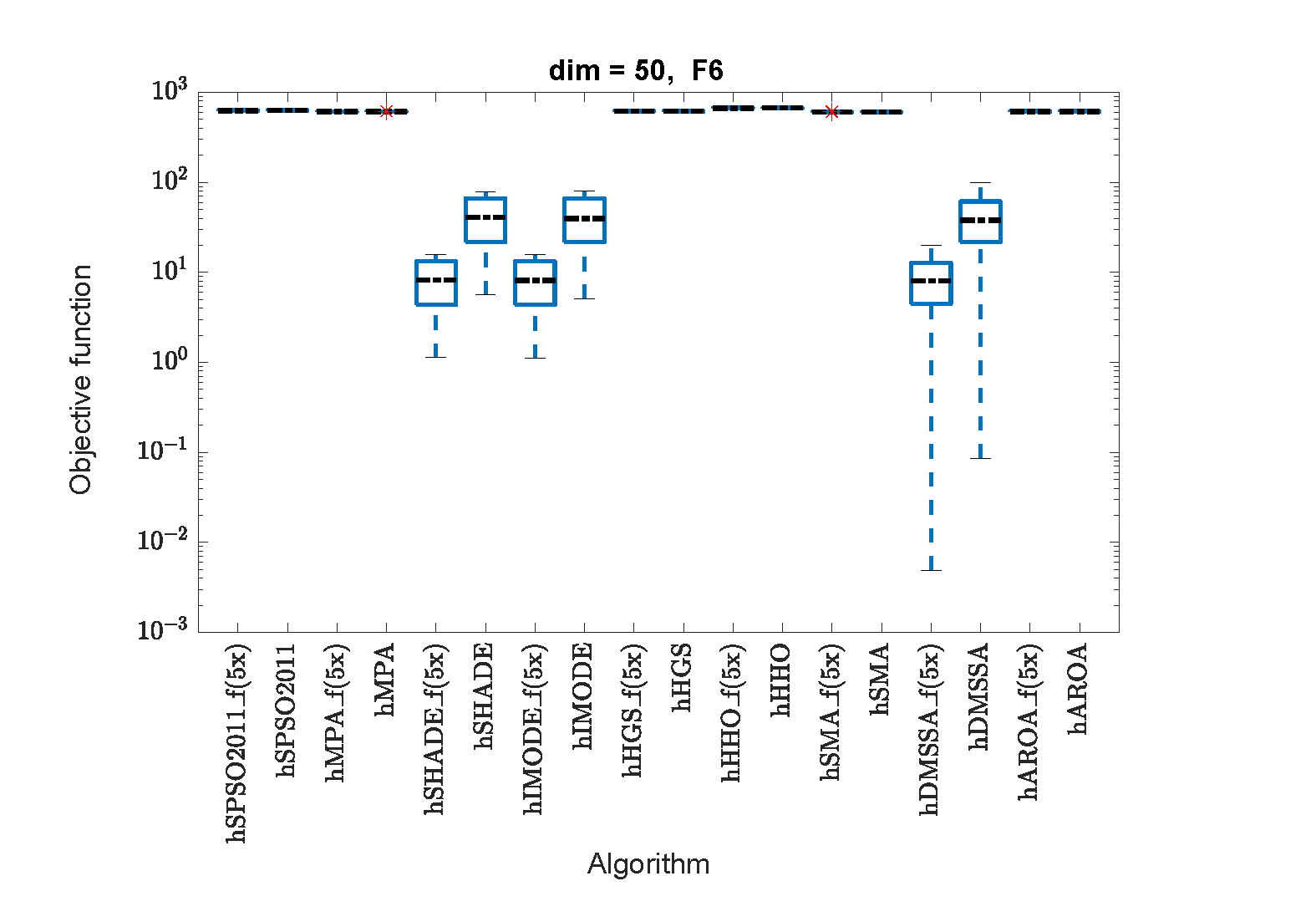}
  \end{subfigure}\hfill
  \begin{subfigure}{0.4\textwidth}
    \includegraphics[height=5.1cm,width=6.9cm]{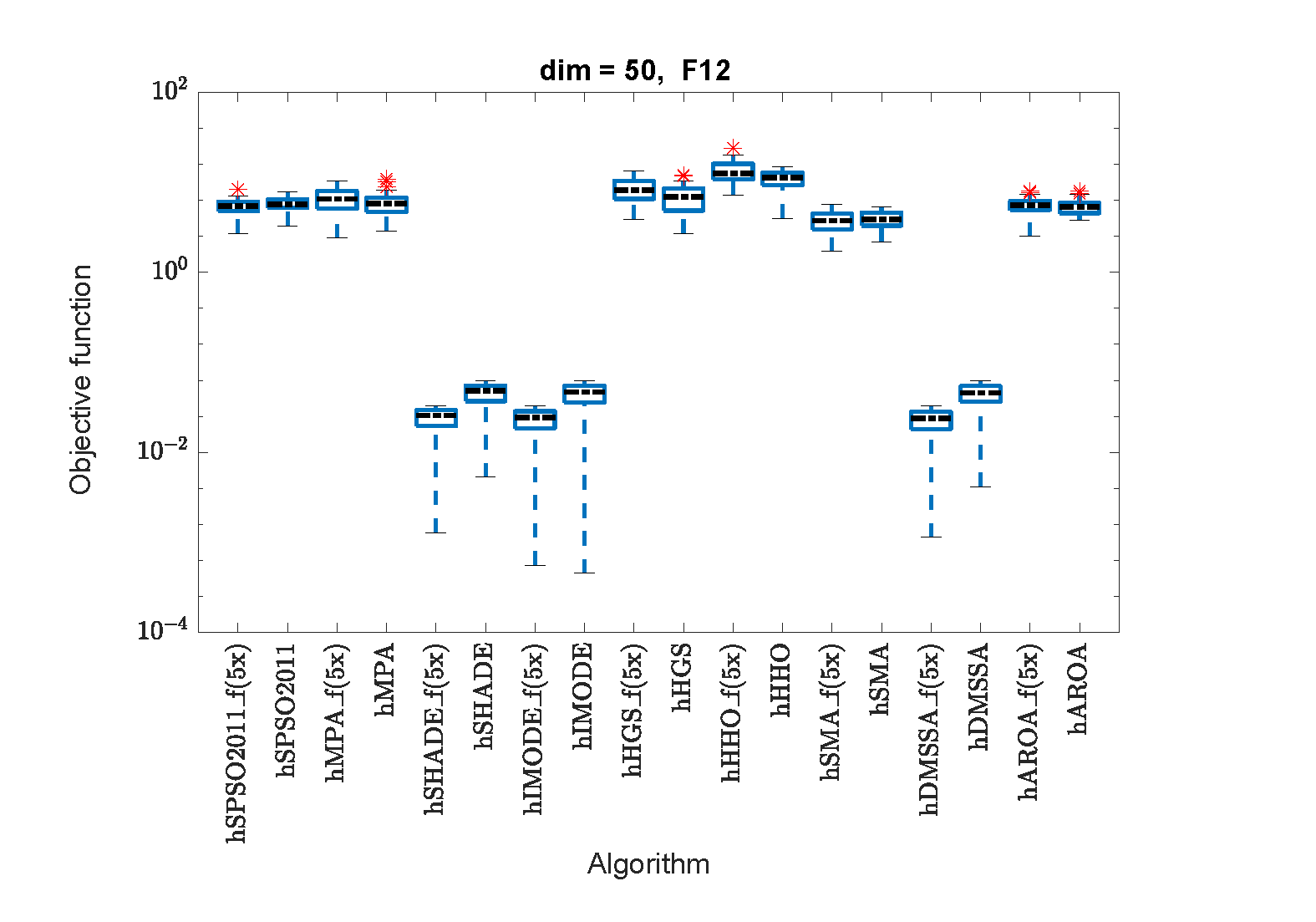}
     \end{subfigure}\hfill
 \begin{subfigure}{0.4\textwidth}
    \includegraphics[height=5.1cm,width=6.9cm]{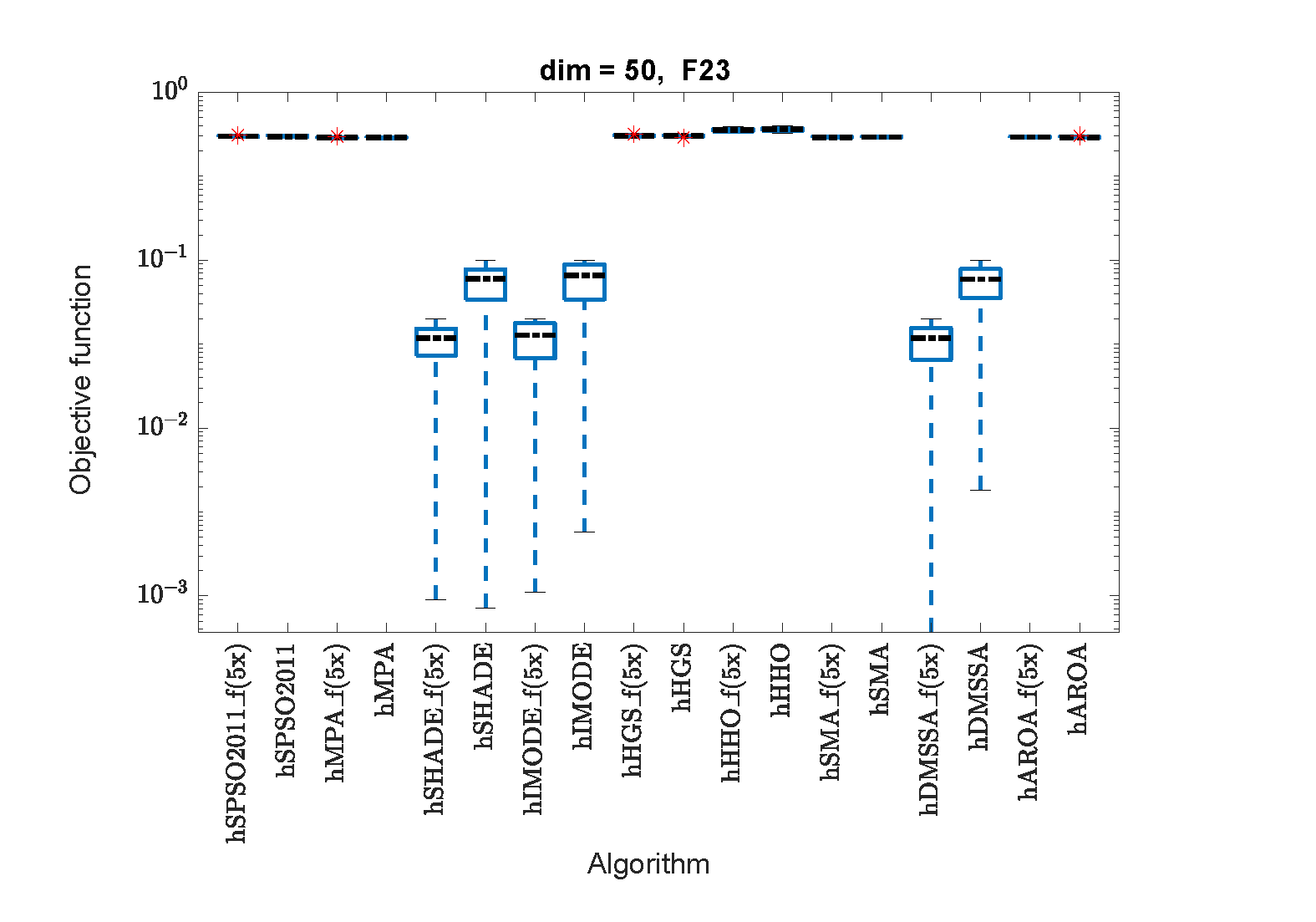}
  \end{subfigure}
\end{adjustwidth}
\end{figure}
 
  \begin{figure}[H]
  \centering
\begin{adjustwidth}{-2cm}{-2.0cm}
  \begin{subfigure}{0.4\textwidth}
    \includegraphics[height=5.1cm,width=6.9cm]{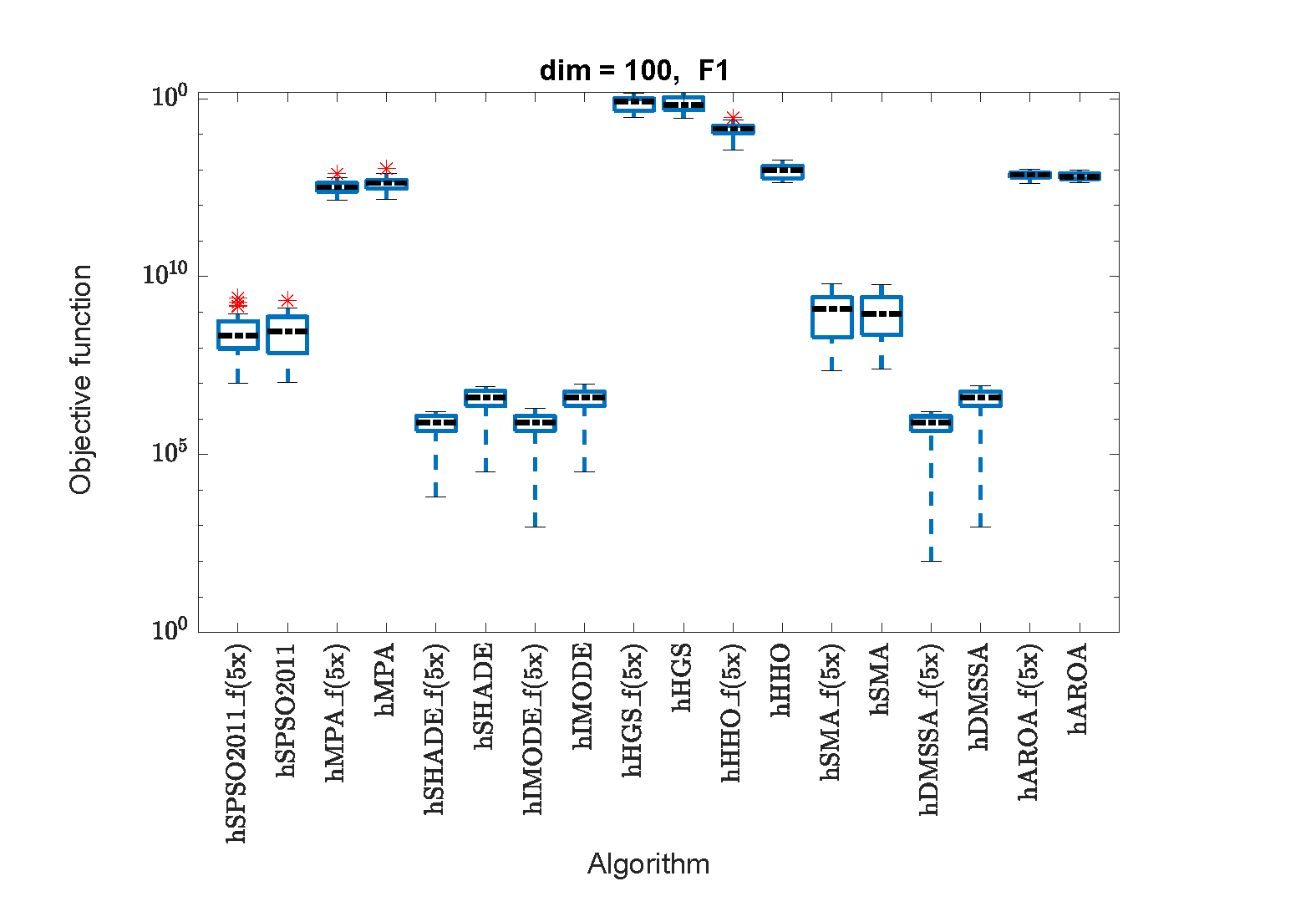}
  \end{subfigure}\hfill
  \begin{subfigure}{0.4\textwidth}
        \includegraphics[height=5.1cm,width=6.9cm]{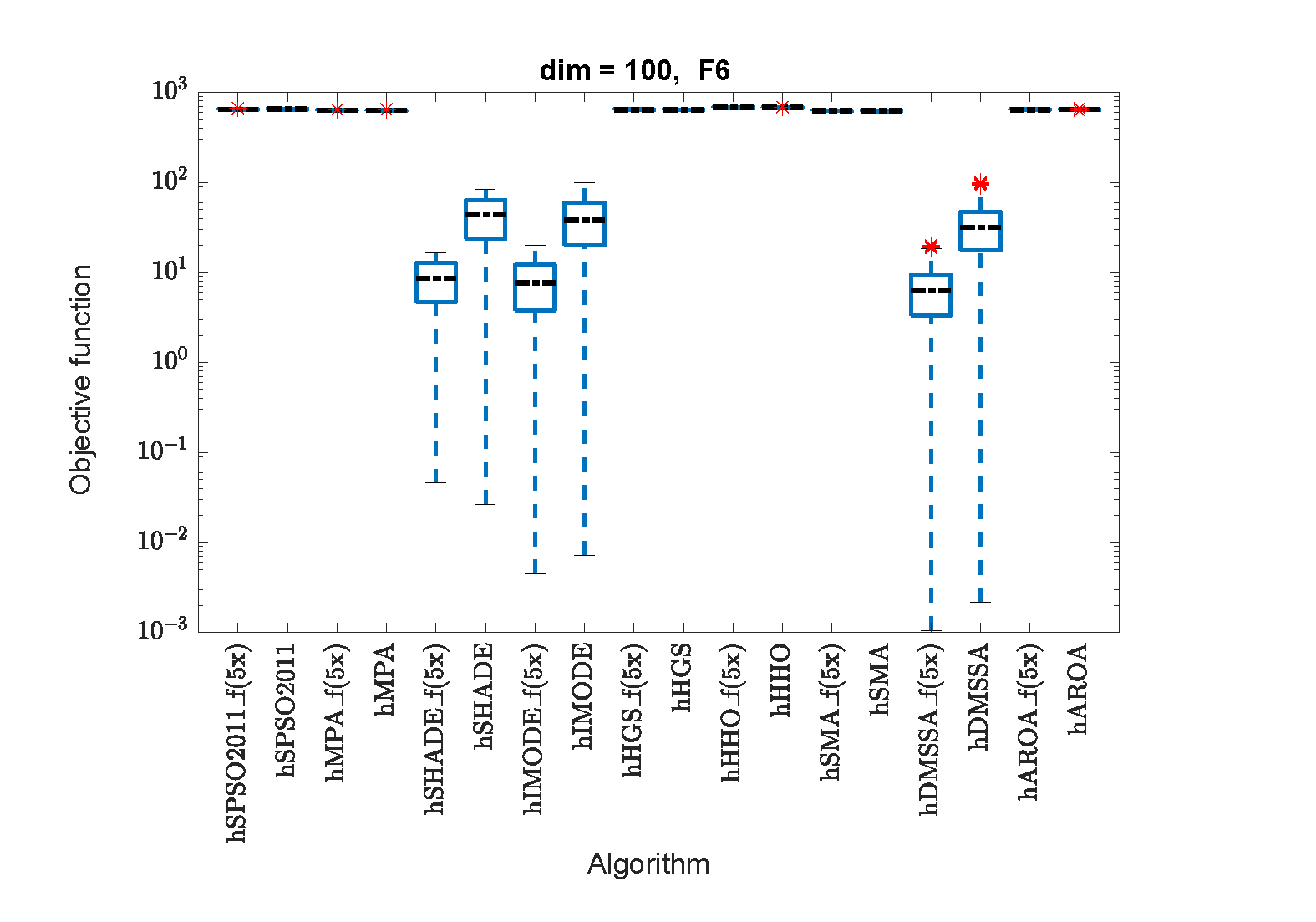}
     \end{subfigure}\hfill
   \begin{subfigure}{0.4\textwidth}
        \includegraphics[height=5.1cm,width=6.9cm]{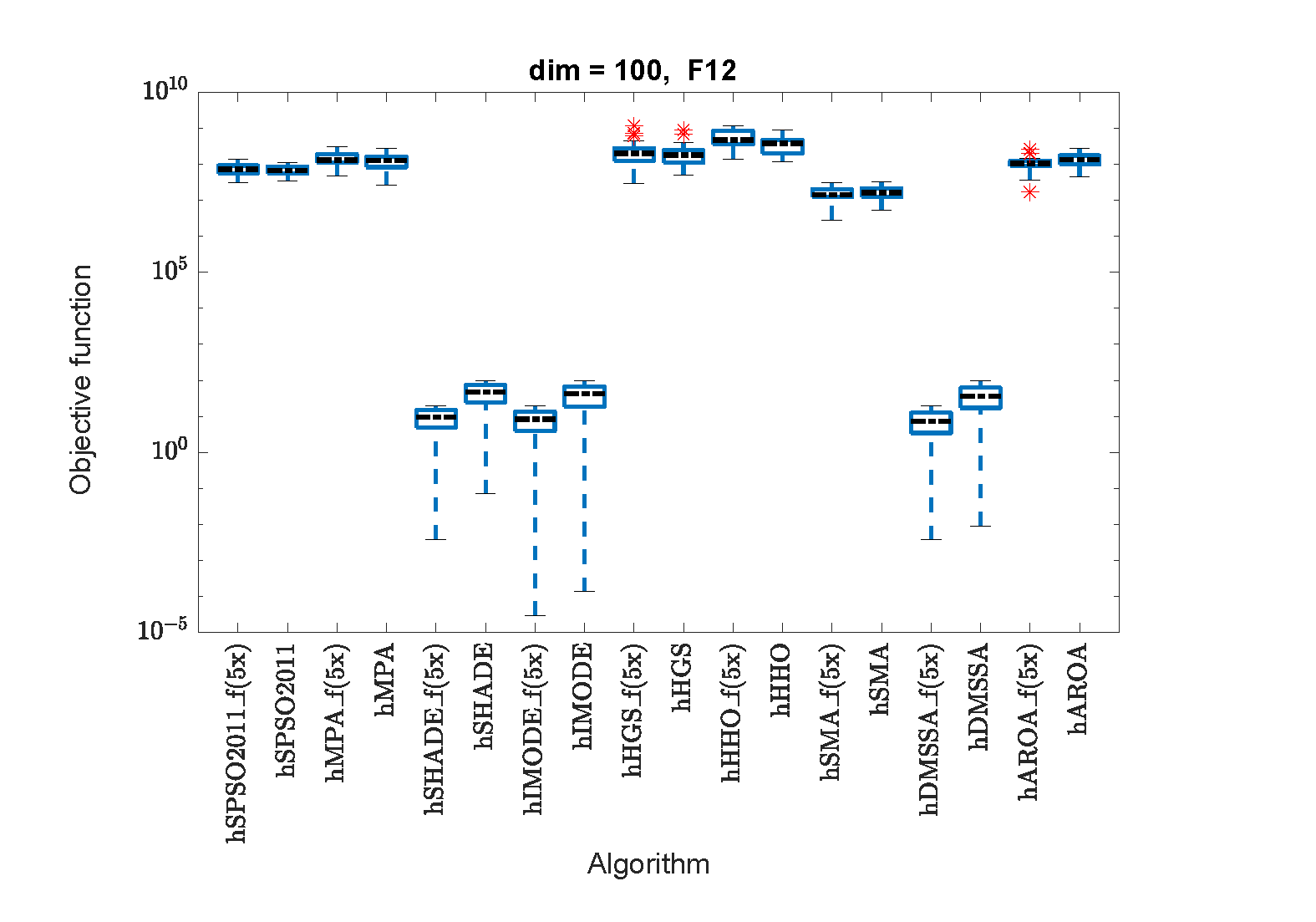}
    \end{subfigure}
    \end{adjustwidth}
    \end{figure}

\begin{figure}[H]
 \centering
\begin{adjustwidth}{-2cm}{-2.0cm}
\begin{center}
  \begin{subfigure}{0.4\textwidth}
   \centering
        \includegraphics[height=5.1cm,width=6.9cm]{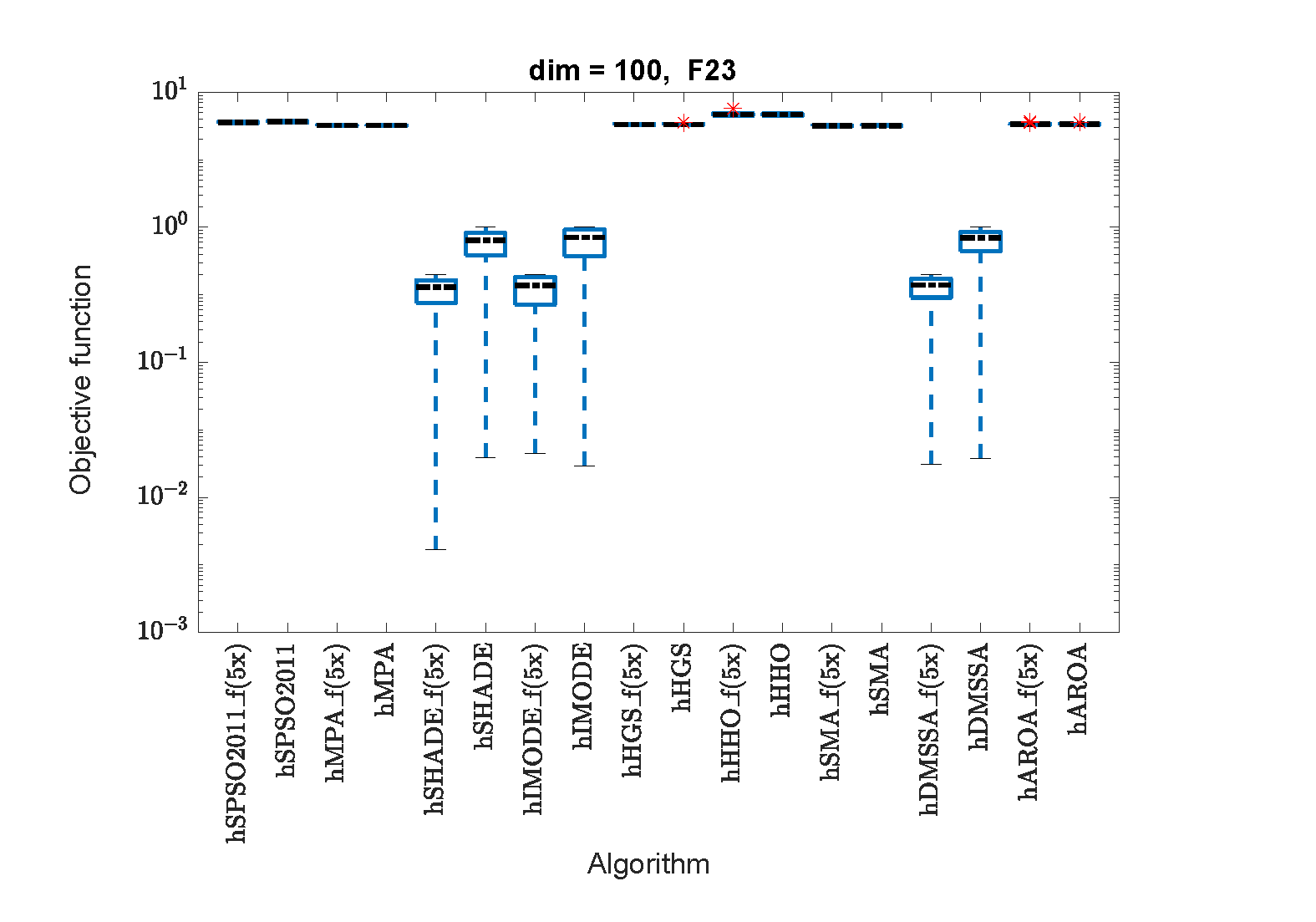}   
  \end{subfigure}
  \end{center}
  \begin{center}
  \begin{subfigure}{0.4\textwidth}
   \centering
    \includegraphics[width=\linewidth]{boxplotlegend.pdf}     
  \end{subfigure}
  \end{center}
  \caption{\footnotesize{Boxplots of final objective values for 9 hybrid algorithms evaluated on original CEC-2017 functions and their transformed variants across 4 dims.}}
  \label{fig9}
\end{adjustwidth}
\end{figure}

\begin{figure}[H]
  \centering
\begin{adjustwidth}{-1cm}{1.0cm}
  \begin{subfigure}{0.35\textwidth} 
        \includegraphics[height=5cm,width=7.5cm]{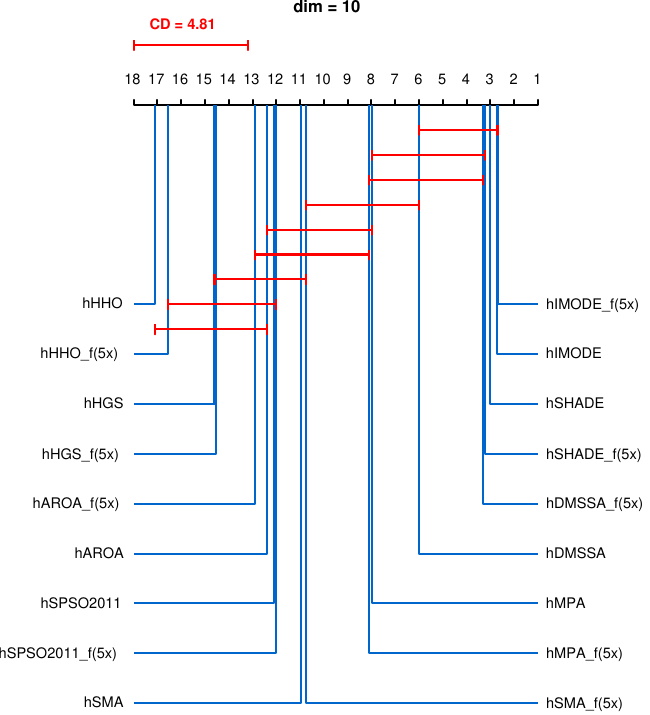}
  \end{subfigure}\hfill
  \begin{subfigure}{0.36\textwidth}
    \includegraphics[height=6cm,width=7.5cm]{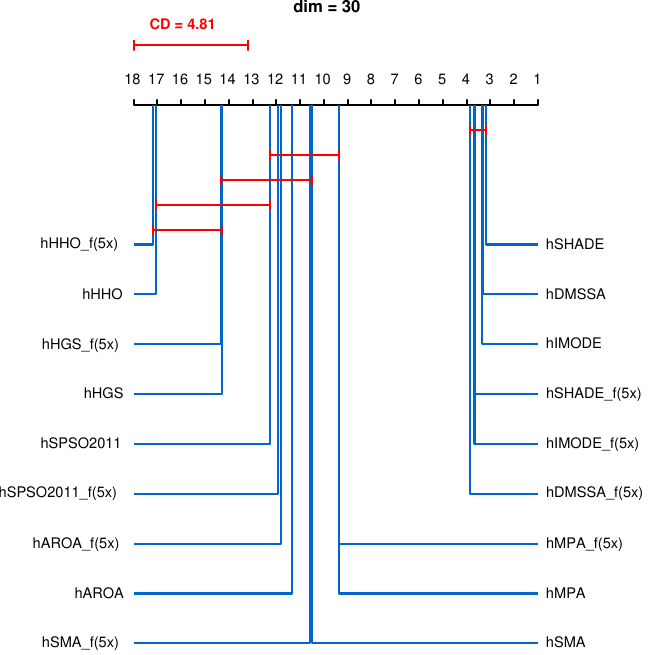}
     \end{subfigure}\hfill
\end{adjustwidth}
\end{figure}

\normalsize

\begin{figure}[H]
\begin{adjustwidth}{-1cm}{1.0cm}
 \begin{subfigure}{0.35\textwidth}
    \includegraphics[height=5cm,width=7.5cm]{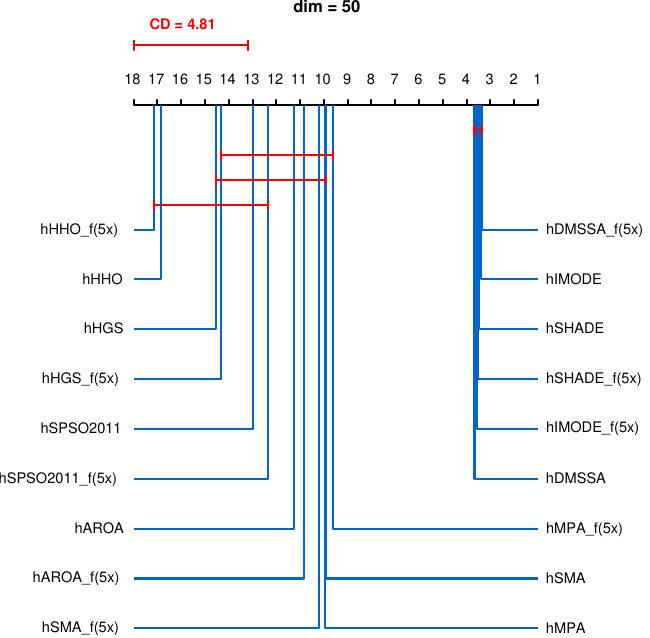}
  \end{subfigure}\hfill
  \begin{subfigure}{0.35\textwidth}
    \includegraphics[height=5cm,width=7.5cm]{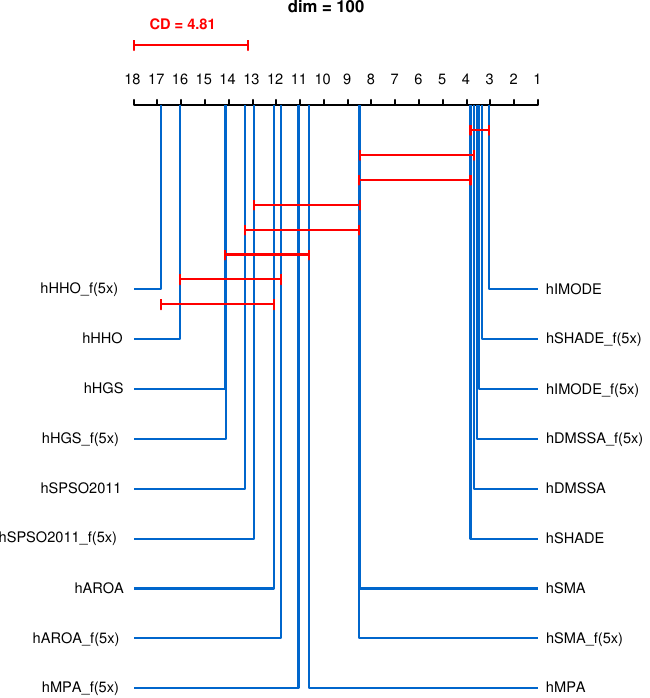}
  \end{subfigure}\hfill
\caption{\footnotesize{CD diagrams for 9 hybrid algorithms tested on original CEC-2017 functions and their scaled variants across 4 dims.}}
 \label{fig10}
\end{adjustwidth}
\end{figure}

\begin{figure}[H] 
\adjustbox{scale=1.25,center}{
\centering
  \begin{subfigure}[t]{0.495\textwidth}
    \centering
\includegraphics[width=\linewidth]{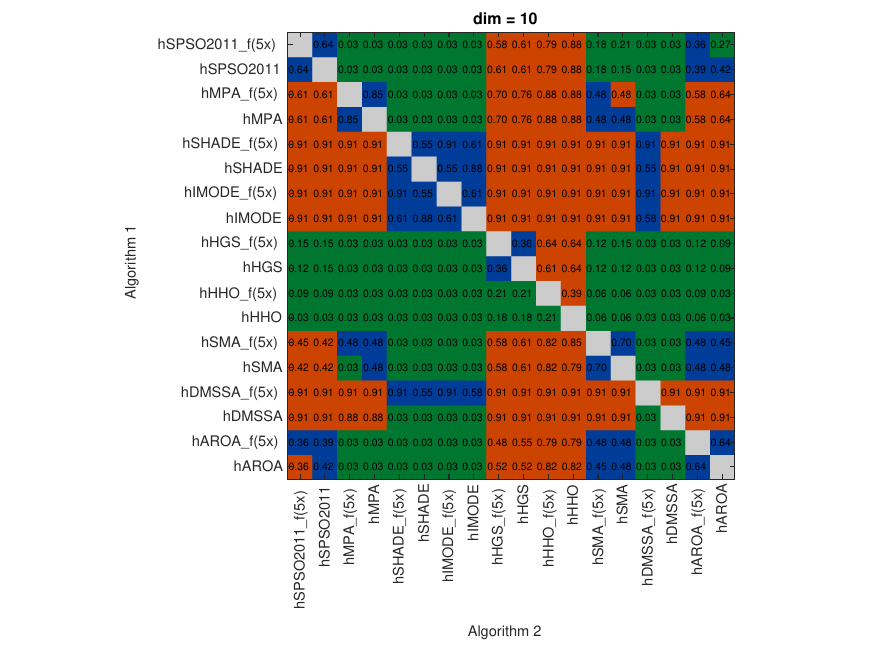}
  \end{subfigure}
  \begin{subfigure}[t]{0.495\textwidth}
    \centering
\includegraphics[width=\linewidth]{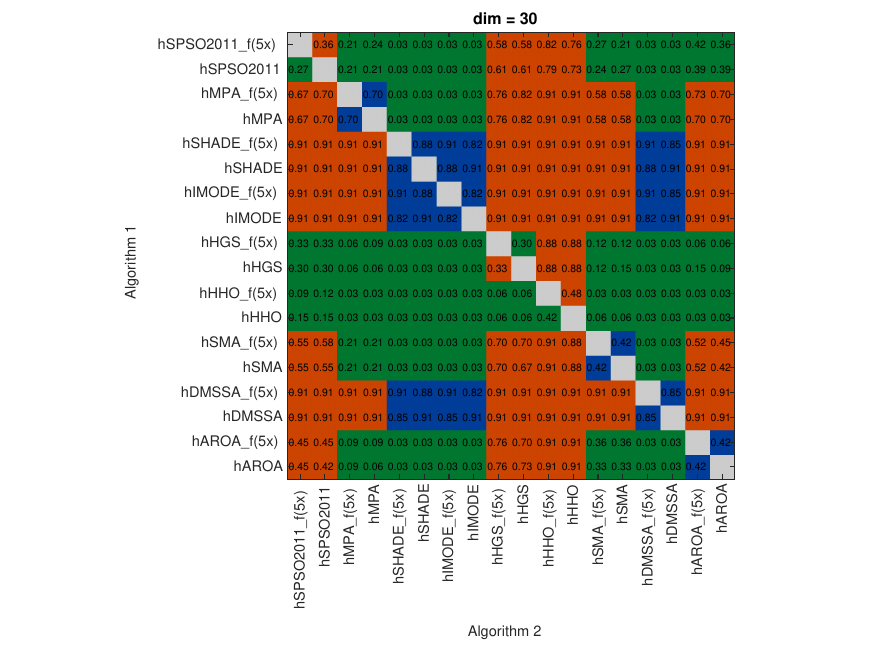}
  \end{subfigure}
  }
  \end{figure}
  \vspace{0.2cm}
  \begin{figure}[H] 
\adjustbox{scale=1.25,center}{
\begin{subfigure}[t]{0.495\textwidth}
    \centering
\includegraphics[width=\linewidth]{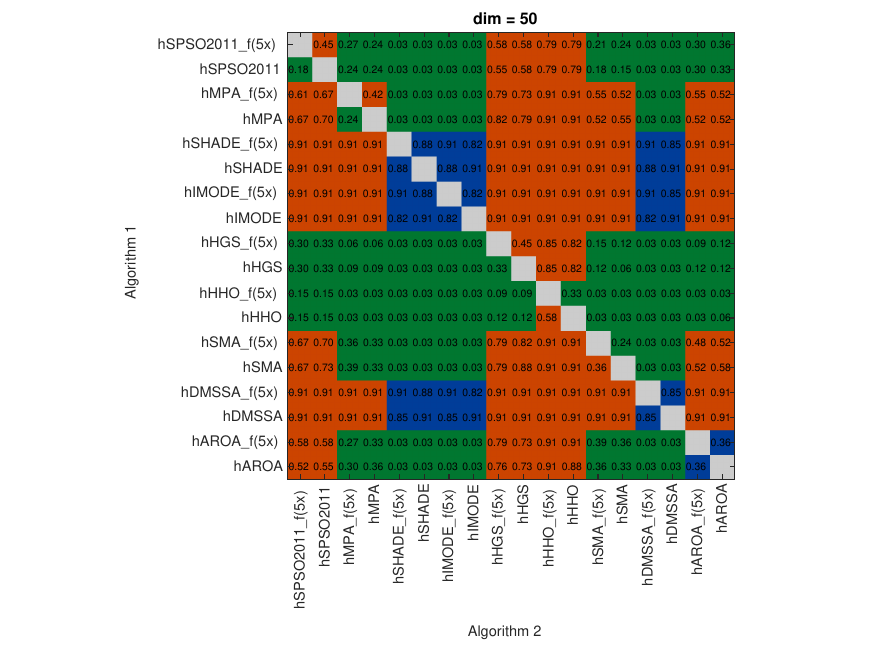}
  \end{subfigure}
 \begin{subfigure}[t]{0.495\textwidth}
    \centering
\includegraphics[width=\linewidth]{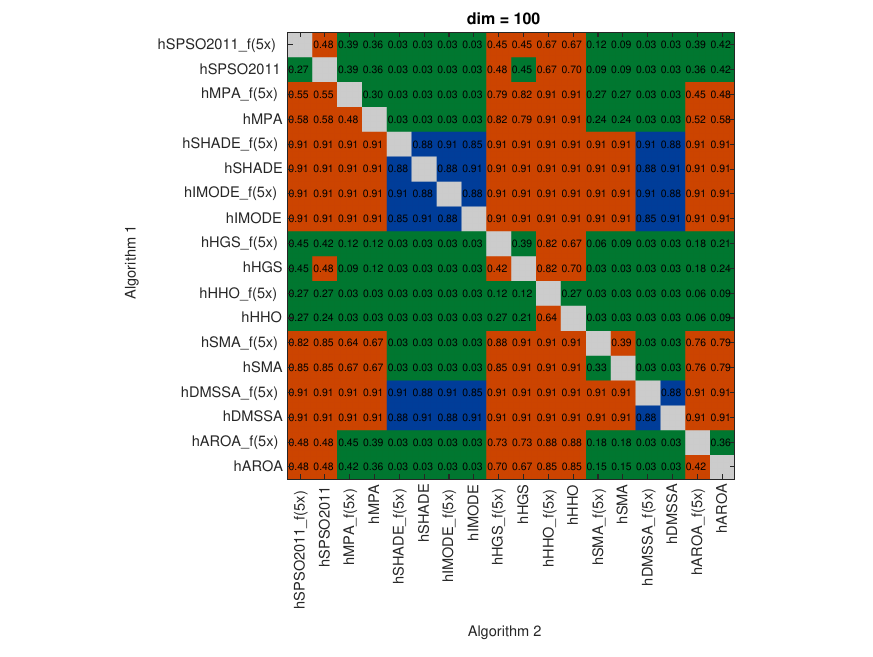}
  \end{subfigure}
  }
  \vspace{0.2cm}
  \adjustbox{scale=1.1,center}{
\begin{subfigure}{0.6\textwidth}
    \centering
    \includegraphics[width=\linewidth]{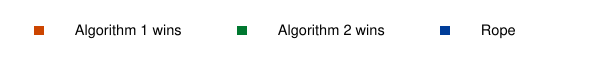}
  \end{subfigure}
  }
  \vspace{-0.75cm}
  \caption{\small Bayesian heatmaps for 9 hybrid algorithms evaluated on original CEC-2017 functions and their scaled variants across 4 dims.}
  \label{fig11}
  \end{figure}

\begin{figure}[H]
  \centering
\begin{adjustwidth}{-1cm}{-2.0cm}
  \begin{subfigure}{0.35\textwidth}
     \includegraphics[height=4cm,width=5.5cm]{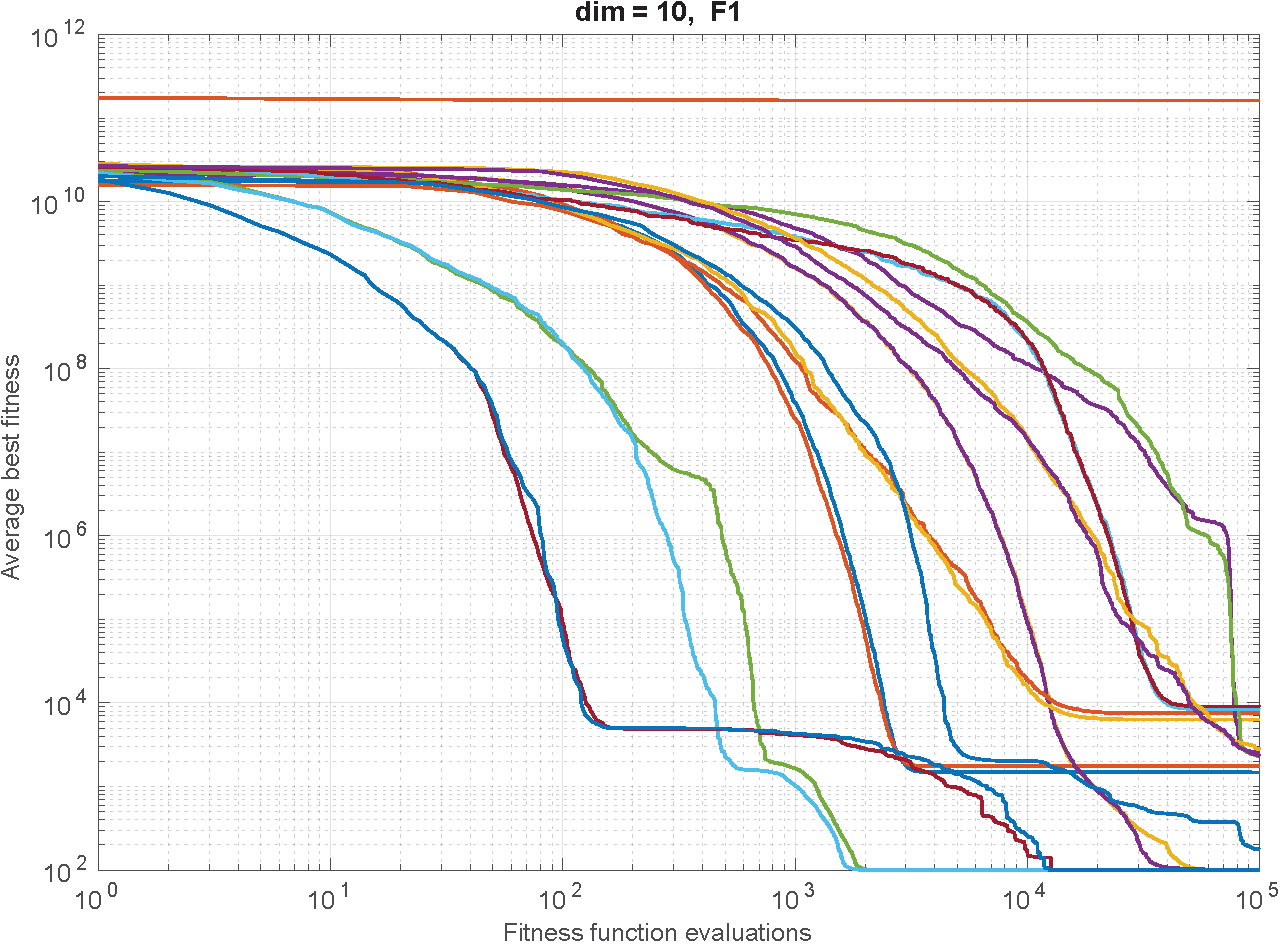}
  \end{subfigure}\hfill
  \begin{subfigure}{0.35\textwidth}
    \includegraphics[height=4cm,width=5.5cm]{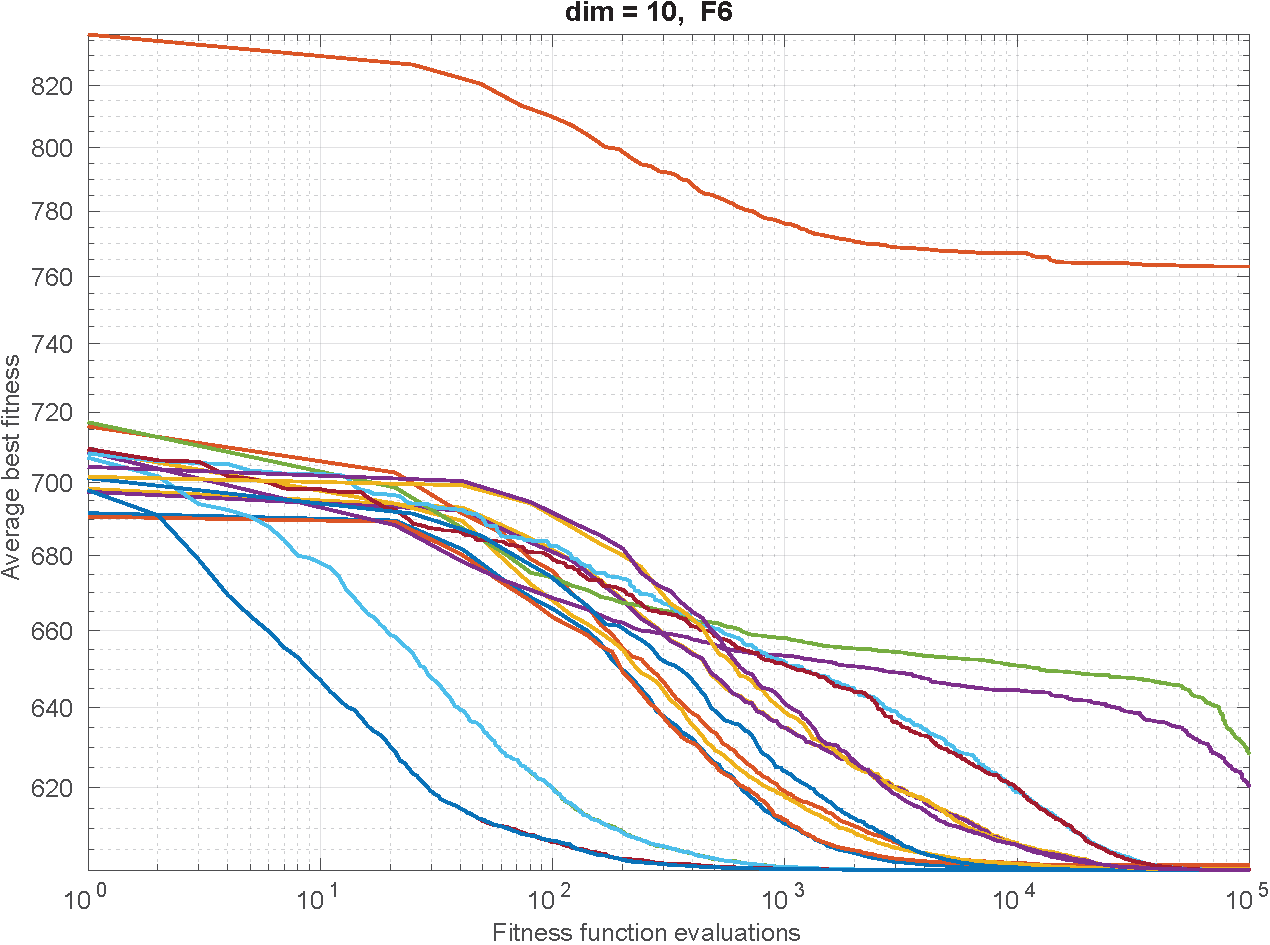}
  \end{subfigure}\hfill
  \begin{subfigure}{0.35\textwidth}
    \includegraphics[height=4cm,width=5.5cm]{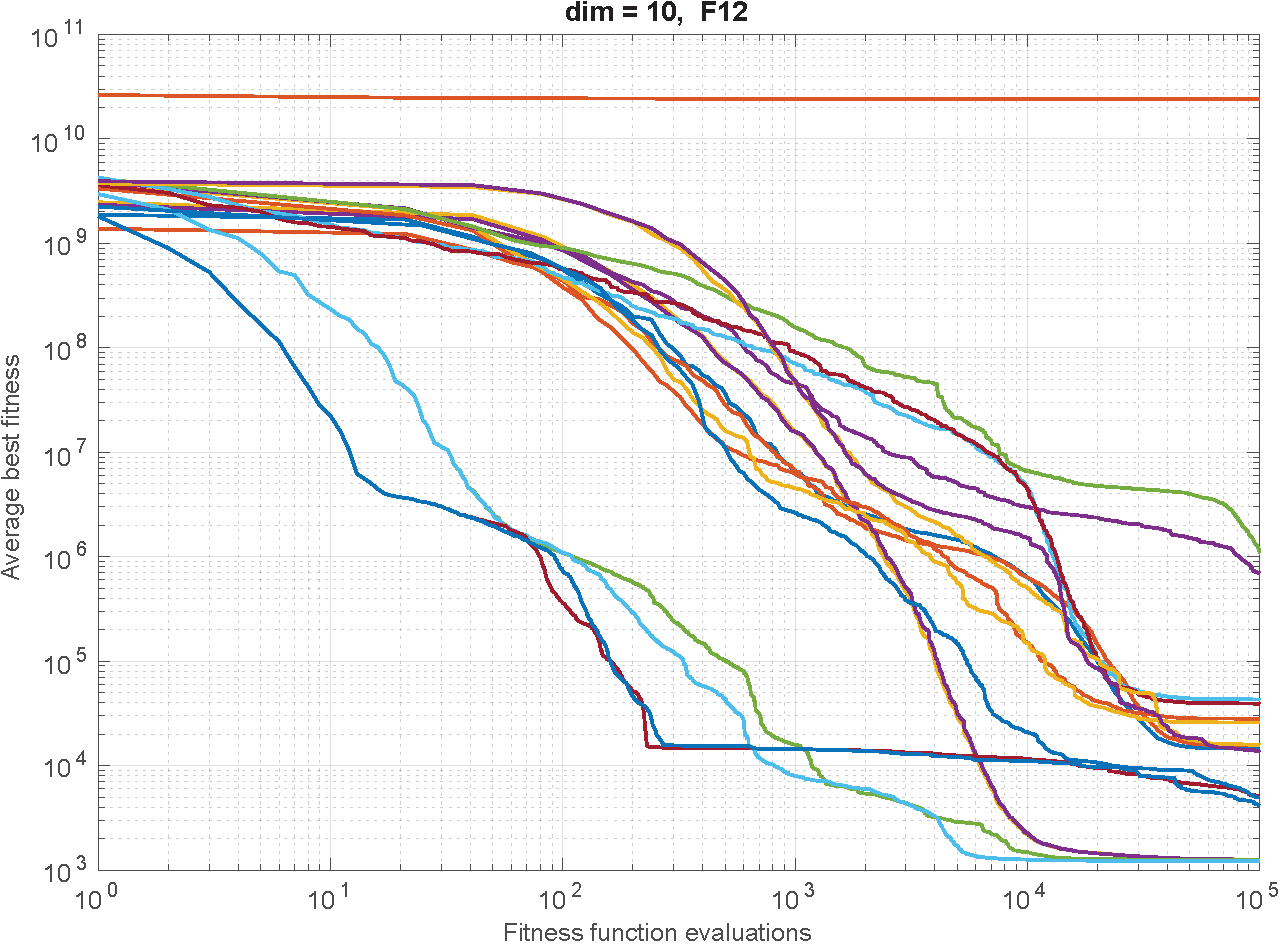}
  \end{subfigure}
 \end{adjustwidth}
\end{figure} 

\begin{figure}[H]
  \centering
\begin{adjustwidth}{-1cm}{-2.0cm}
   \begin{subfigure}{0.35\textwidth}
    \includegraphics[height=4cm,width=5.5cm]{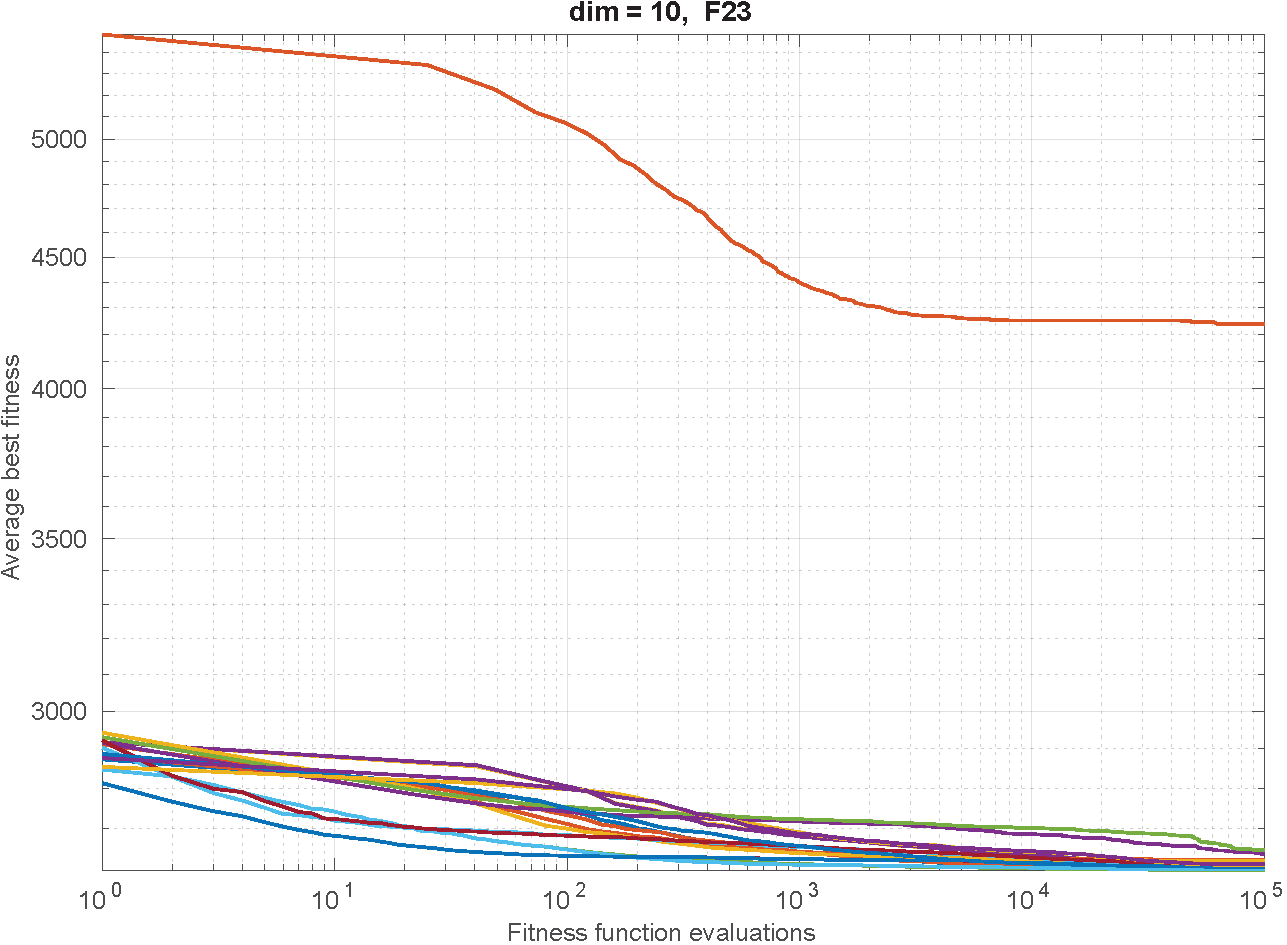}
  \end{subfigure}\hfill
  \begin{subfigure}{0.35\textwidth}
    \includegraphics[height=4cm,width=5.5cm]{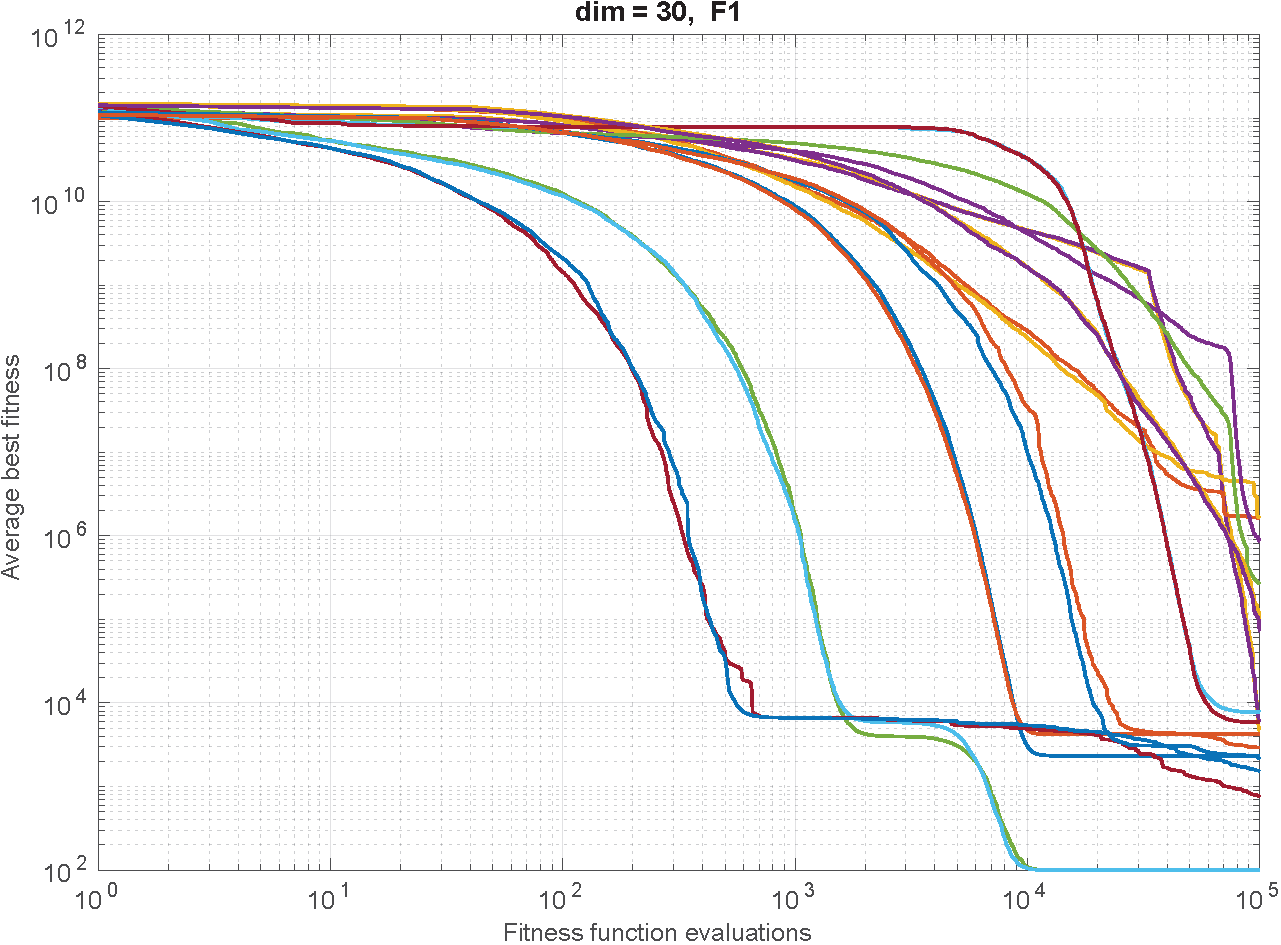}
\end{subfigure}\hfill
  \begin{subfigure}{0.35\textwidth} 
        \includegraphics[height=4cm,width=5.5cm]{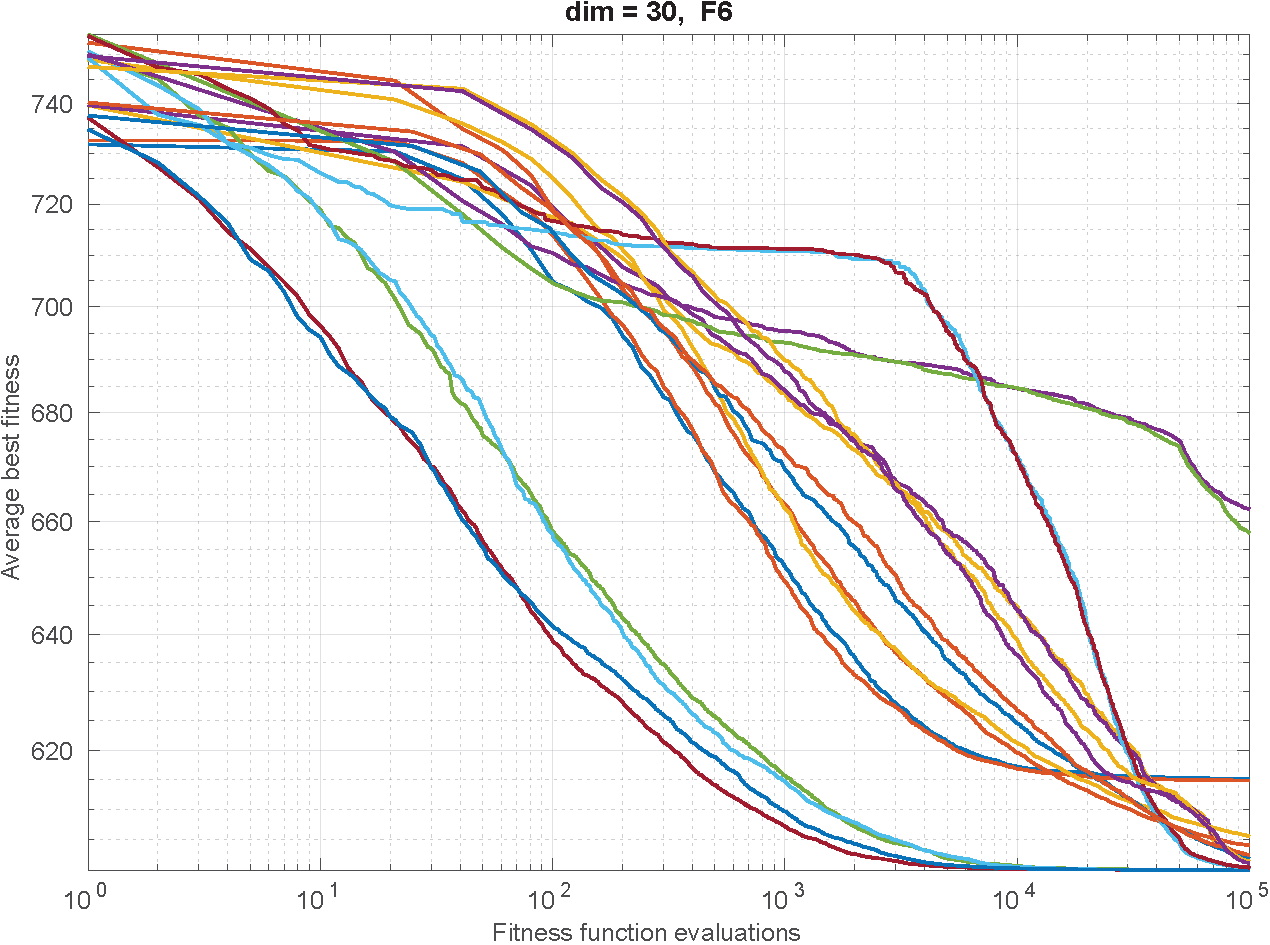}
   \end{subfigure}
   \end{adjustwidth}
\end{figure}
\begin{figure}[H]
  \centering
\begin{adjustwidth}{-1cm}{-2.0cm}
   \begin{subfigure}{0.35\textwidth}
        \includegraphics[height=4cm,width=5.5cm]{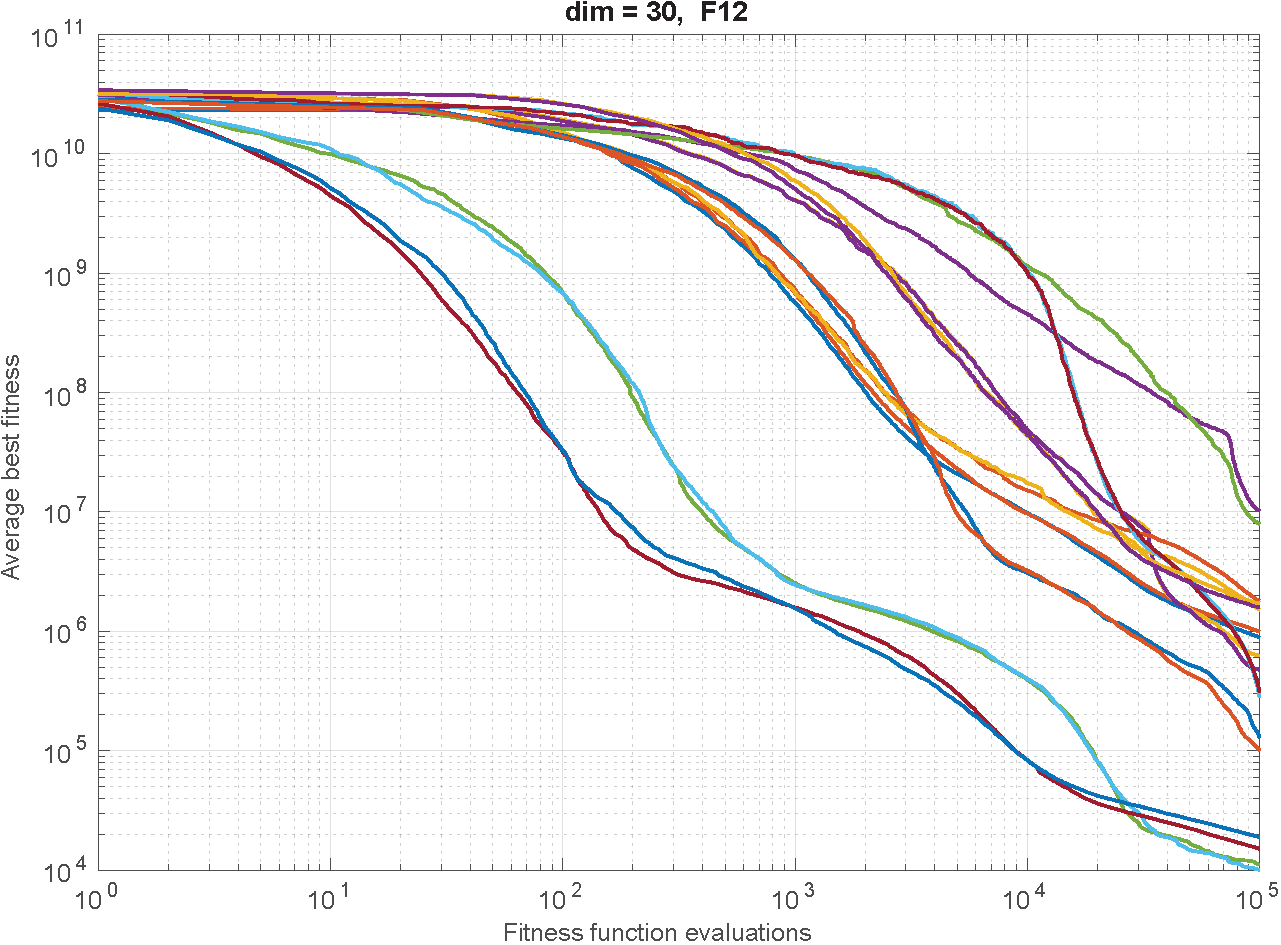}
  \end{subfigure}\hfill
  \begin{subfigure}{0.35\textwidth} 
        \includegraphics[height=4cm,width=5.5cm]{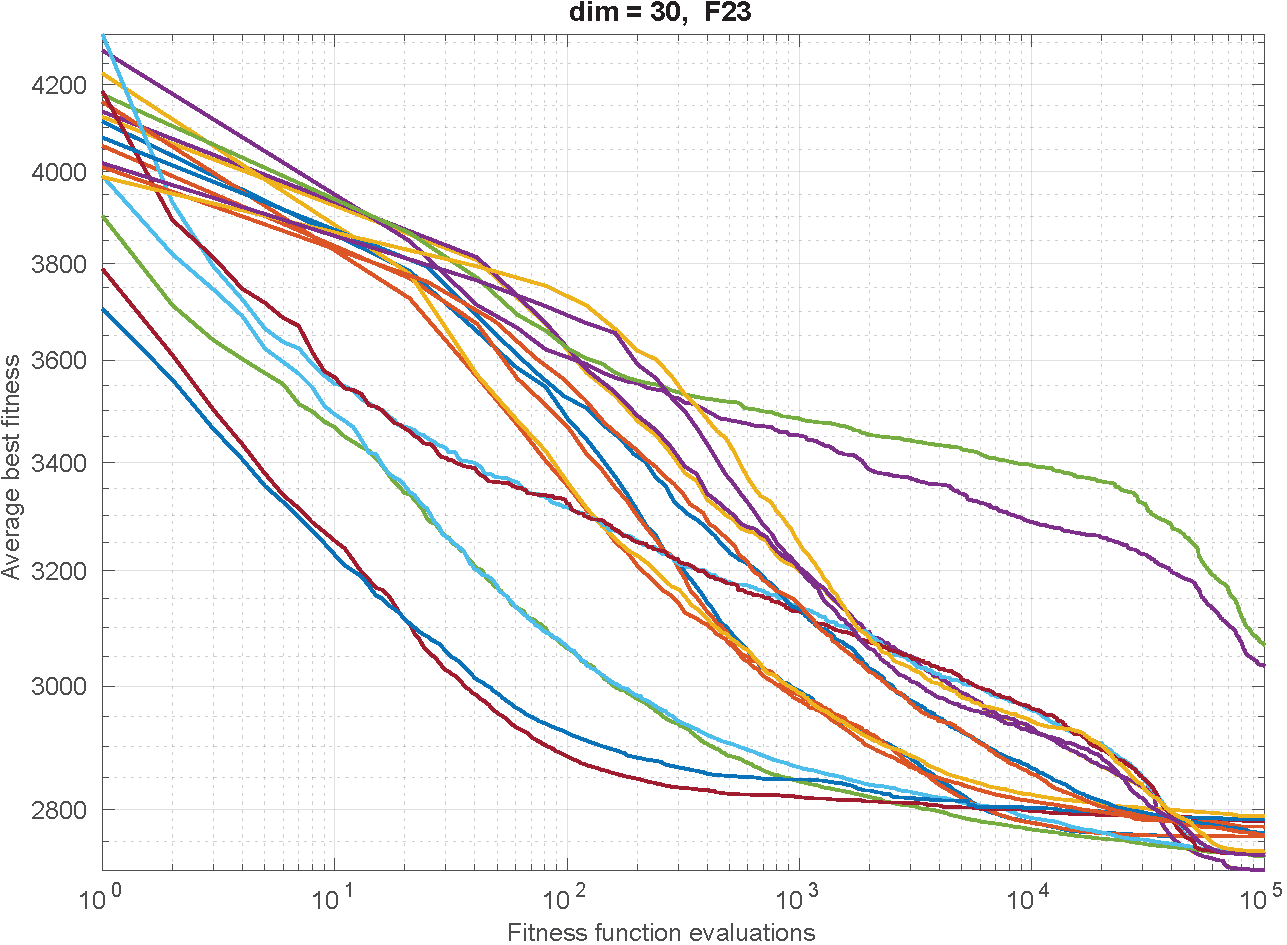}
  \end{subfigure}\hfill
  \begin{subfigure}{0.35\textwidth}
        \includegraphics[height=4cm,width=5.5cm]{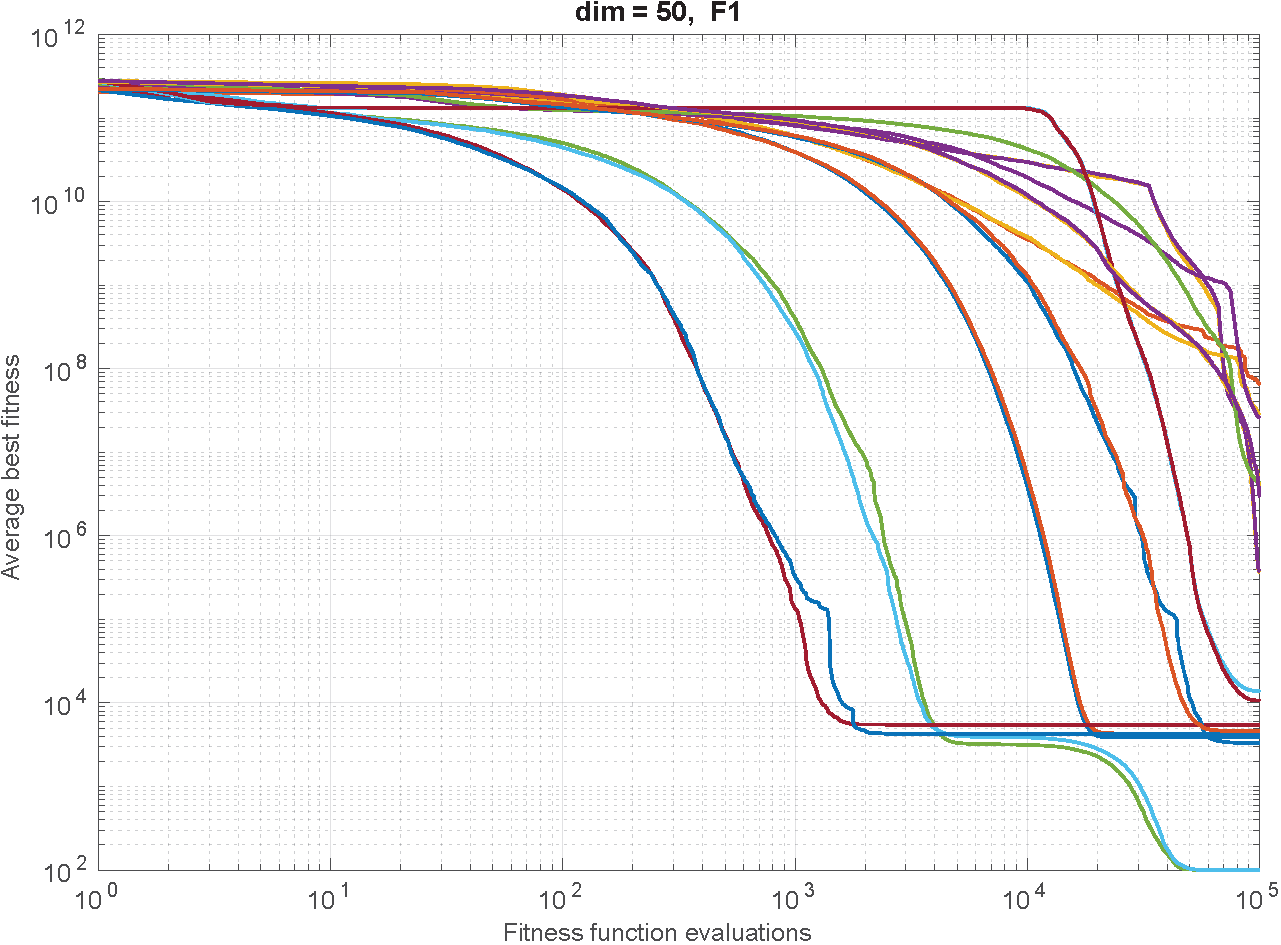}
  \end{subfigure}
  \end{adjustwidth}
\end{figure}

\begin{figure}[H]
  \centering
\begin{adjustwidth}{-1cm}{-2.0cm}
  \begin{subfigure}{0.35\textwidth} 
        \includegraphics[height=4cm,width=5.5cm]{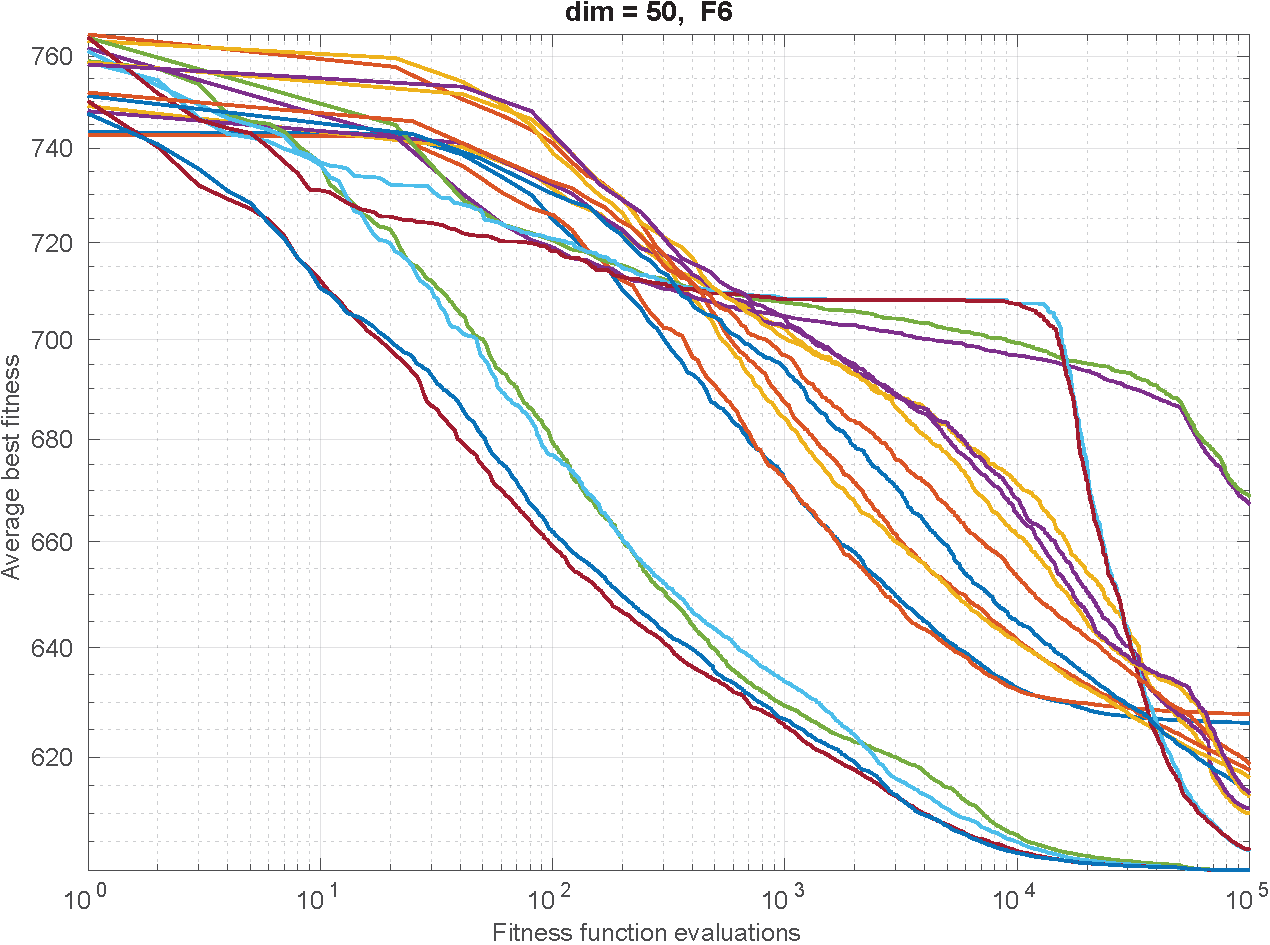}
  \end{subfigure}\hfill
  \begin{subfigure}{0.36\textwidth}
    \includegraphics[height=4cm,width=5.5cm]{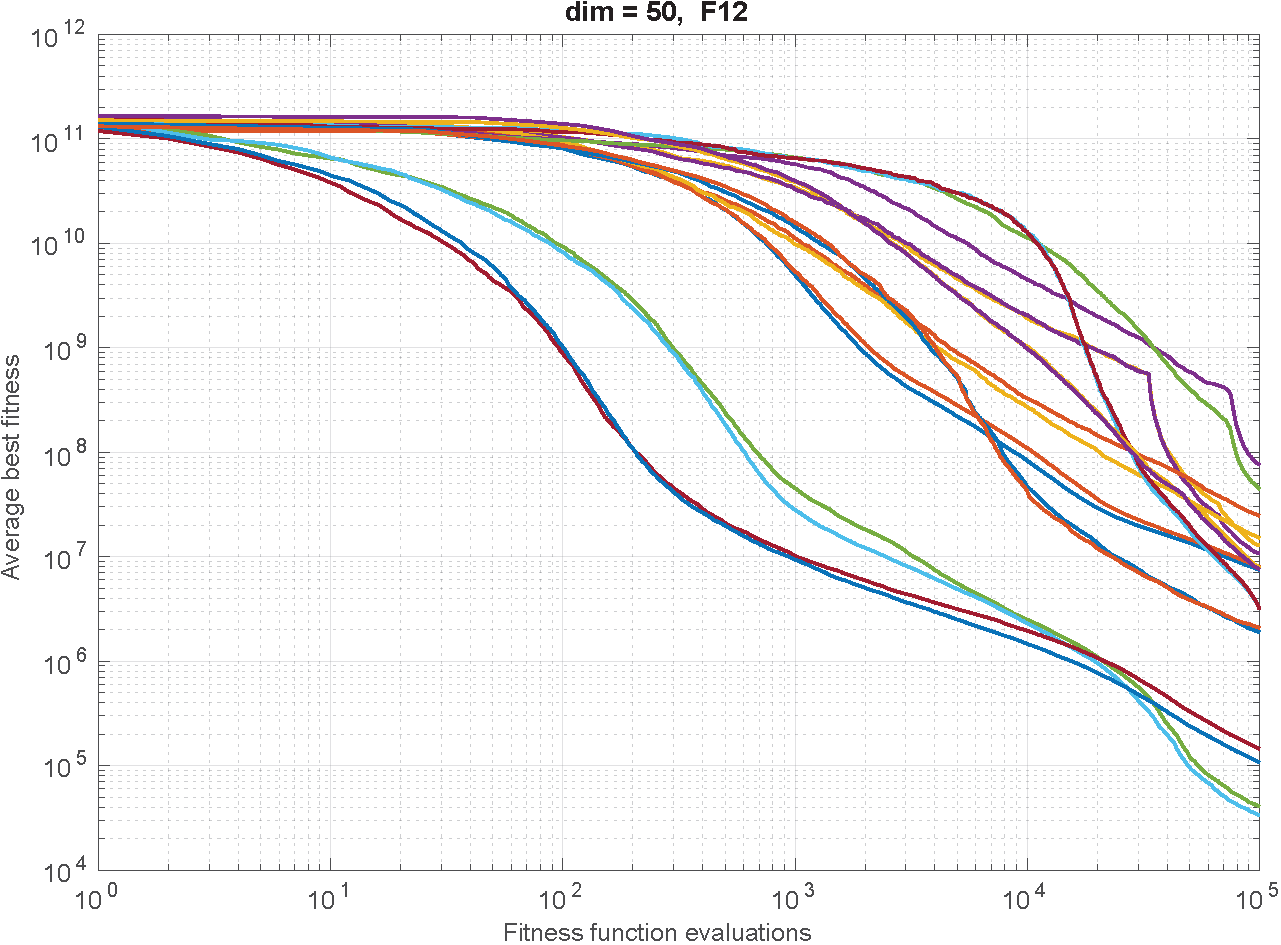}
     \end{subfigure}\hfill
 \begin{subfigure}{0.35\textwidth}
    \includegraphics[height=4cm,width=5.5cm]{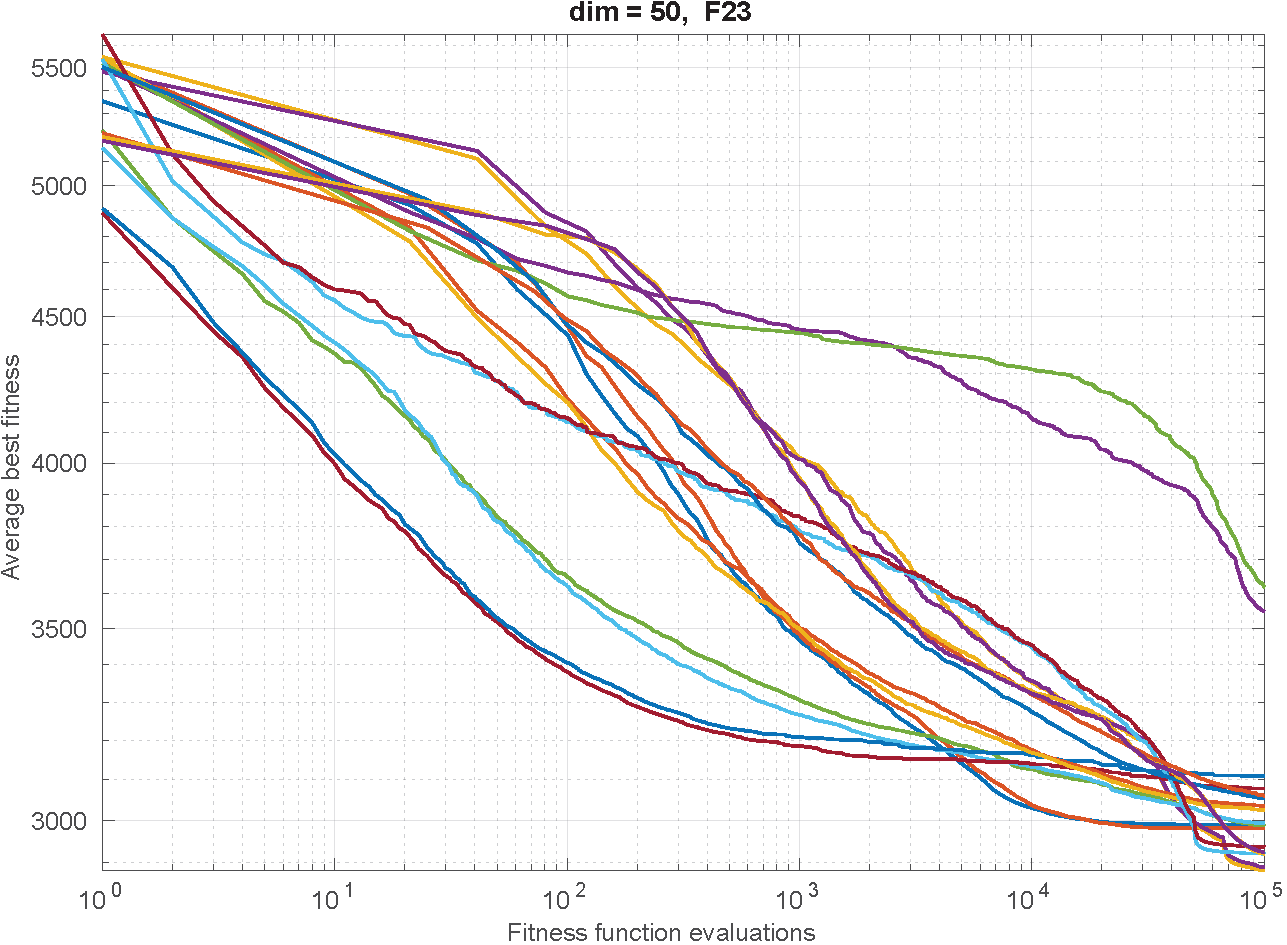}
  \end{subfigure}
\end{adjustwidth}
\end{figure}

\begin{figure}[H]
  \centering
\begin{adjustwidth}{-1cm}{-2.0cm}
  \begin{subfigure}{0.35\textwidth}
    \includegraphics[height=4cm,width=5.5cm]{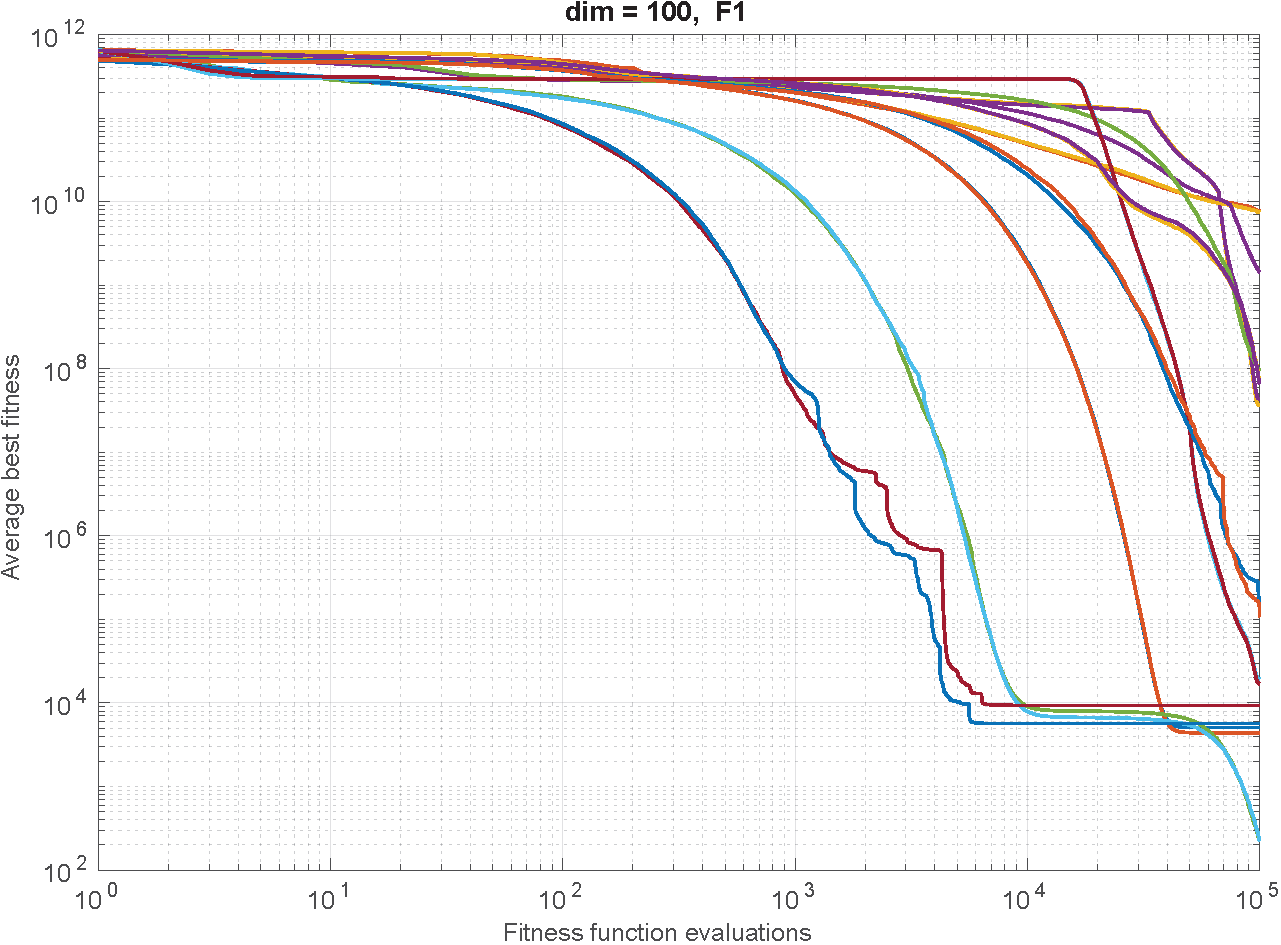}
  \end{subfigure}\hfill
  \begin{subfigure}{0.35\textwidth}
        \includegraphics[height=4cm,width=5.5cm]{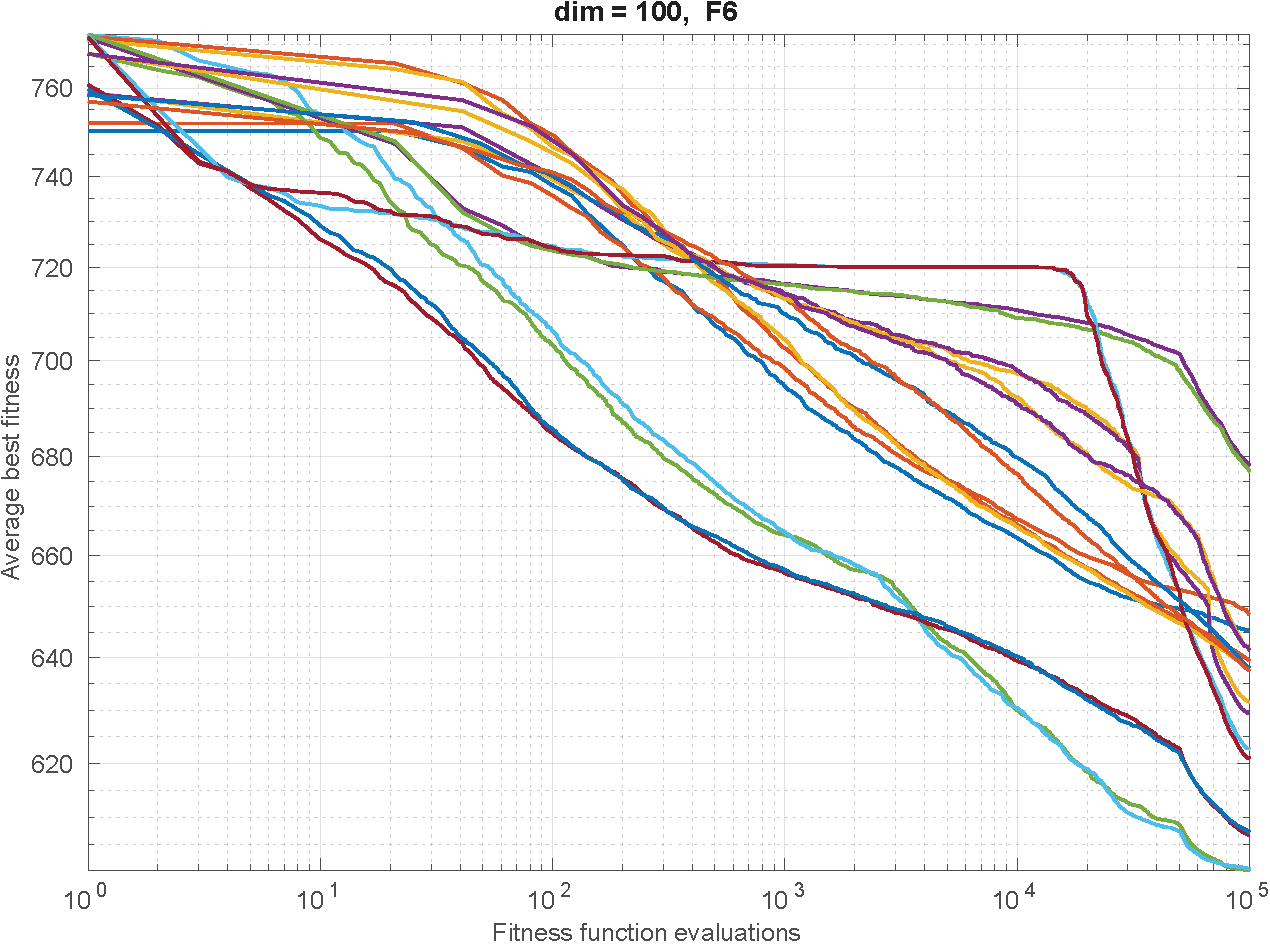}
     \end{subfigure}\hfill
   \begin{subfigure}{0.35\textwidth}
        \includegraphics[height=4cm,width=5.5cm]{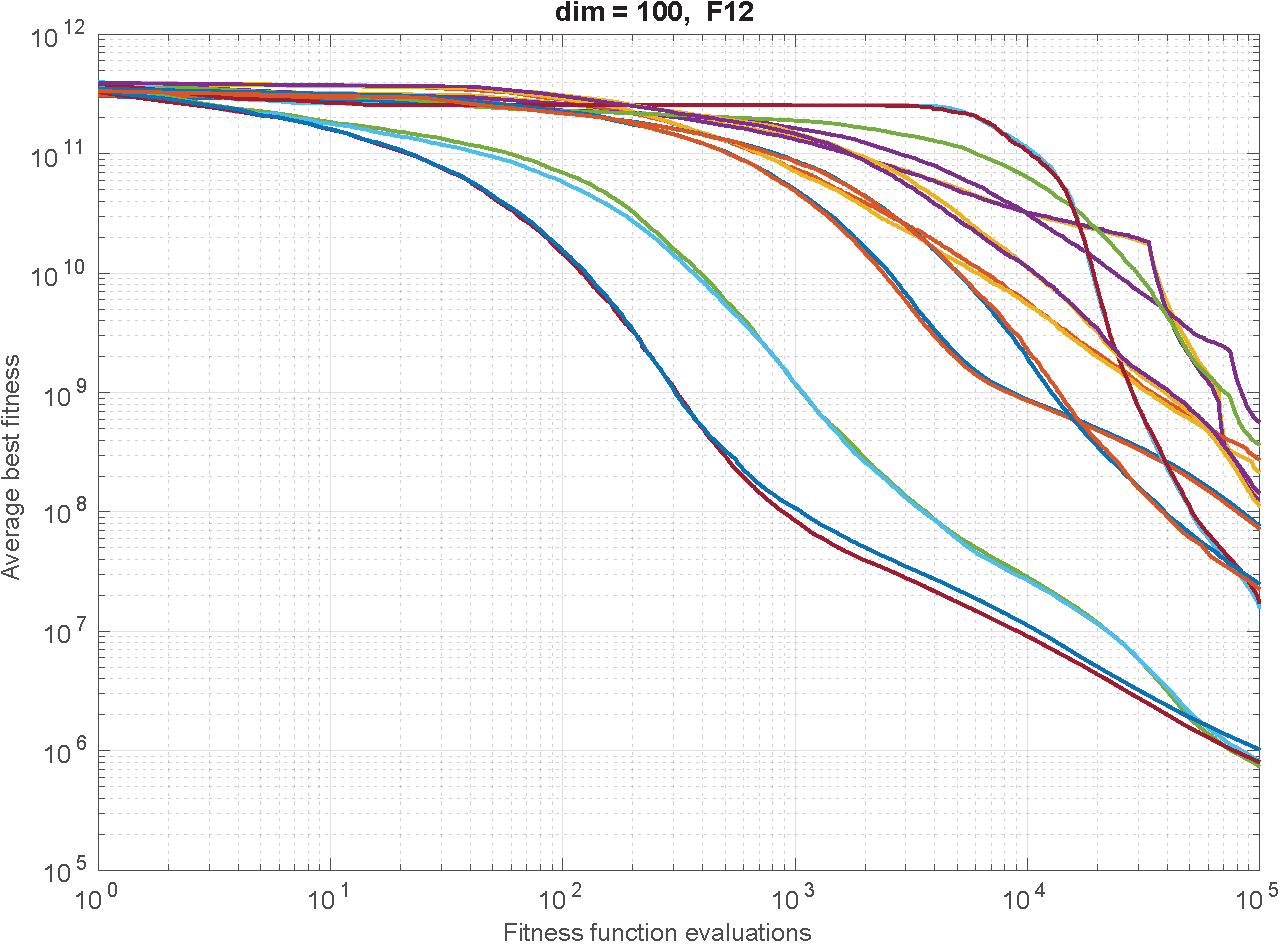}
    \end{subfigure}
    \end{adjustwidth}
\end{figure}

\begin{figure}[H]
 \centering
\begin{adjustwidth}{-1.4cm}{-2.0cm}
\begin{center}
  \begin{subfigure}{0.35\textwidth}
   \centering
        \includegraphics[height=4cm,width=5.5cm]{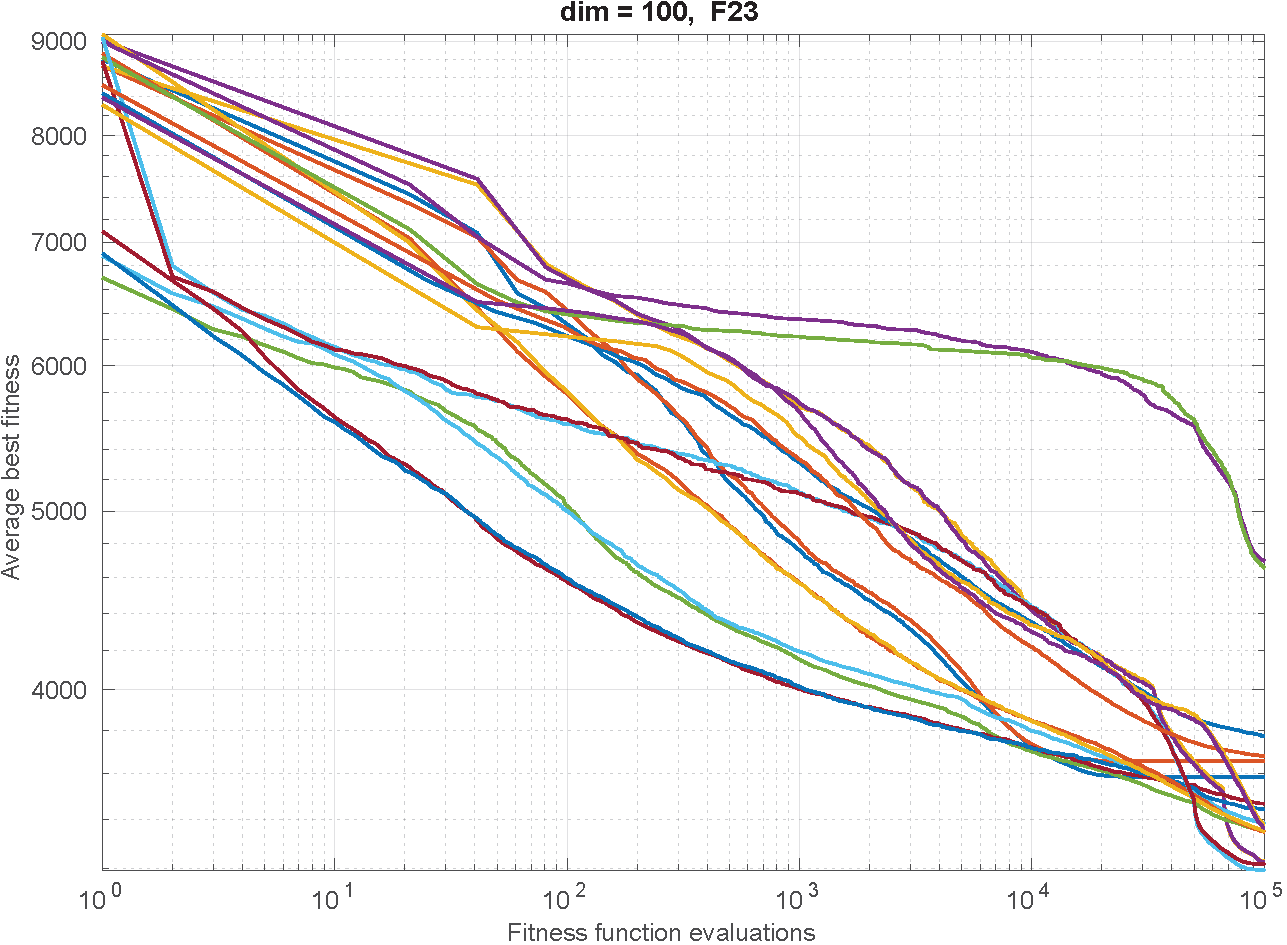}   
  \end{subfigure}
  \end{center}
  \begin{center}
  \begin{subfigure}{0.85\textwidth}
   \centering
    \includegraphics[height=0.75cm, width=\linewidth]{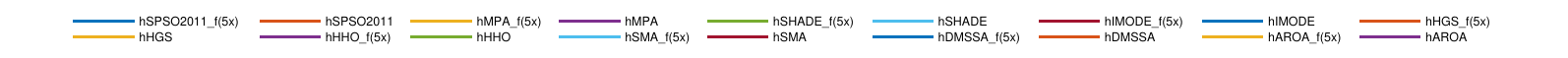}     
  \end{subfigure}
  \end{center}
  \caption{\small{Convergence trajectories of metaheuristic algorithms on a representative CEC-2017 benchmark function.}}
  \label{fig12}
\end{adjustwidth}
\end{figure}

\vspace{-0.3cm}
\subsection{Scaling of the Objective Function: Analysis of $f(5x)$}
\vspace{0.15cm}
This subsection analyzes the impact of scaling the decision space by a factor of 5 on the performance of optimization algorithms, with particular emphasis on deviations from scale-invariance. The analysis compares the results for $f(5x)$ with the baseline $f(x)$ and, where meaningful, with the translation-invariant results $f(x+8)$. Scaling by a factor of 5 produces heterogeneous effects across the evaluated algorithms. Notably, hMPA exhibits enhanced performance, with the mean objective value decreasing from approximately $2.6 \times 10^3$ to $1.7 \times 10^3$, and a substantial reduction in standard deviation from $5.2 \times 10^3$ to just 25, while the median remains stable - indicating improved robustness and consistency. Similarly, hHHO exhibits a drop in mean from $5.8 \times 10^4$ to $3.9 \times 10^4$, suggesting enhanced convergence precision. On the contrary, hIMODE shows a notable degradation in solution quality: the median shifts from around 1.0 to -0.63, and the mean from -1.5 to -0.33, accompanied by increased variance - signifying sensitivity to unit scaling. Algorithms like hHGS and hSMA remain largely unaffected, with their median and mean nearly unchanged. In contrast, hDMSSA significantly reduces its mean from 0.97 to 0.08, achieving near-zero error, thus showcasing improved precision under scaled conditions. These patterns suggest that hybrid algorithms often benefit from scaling (via lower error values and reduced spread), while traditional ones show limited or even negative adjustments (see Tab.\ref{tab3} for detailed statistics across $f(5x)$)

Across dimensions $dim = 10, 30, 50, 100$, the results highlight increased sensitivity to scaling, especially in higher-dimensional spaces. For $dim = 10$, most algorithms, including hIMODE, hSHADE, and hMPA, maintain high precision. However, at $dim = 50$ and $dim = 100$, volatility becomes evident: hIMODE’s mean increases, hMPA becomes more stable but less accurate. These trends support the theory of landscape dilution \cite{6atz}, where scaling enlarges the search space volume, reducing the informational density and making the landscape sparser. This impairs gradient-like heuristic mechanisms and hinders convergence accuracy. Moreover, non-invariance arises from the coordinate-dependent nature of many metaheuristic operators \cite{0BB},\cite{4aijk}, which lack adaptation to scale transformations. Compared with the baseline $f(x)$, especially in higher dimensions, one observes not only shifts in central statistics but also transformations in statistical profiles: algorithms with previously low Std values now exhibit increased dispersion. This reinforces the idea that scale sensitivity emerges as dimensionality grows.

Compared to translation $f(x+8)$, scaling introduces more structural rather than directional variation. Wilcoxon tests performed across all algorithm pairs and dimensions confirmed numerous significant differences ($p < 0.05$), most evident in higher dimensions. For $dim = 50$ and $dim = 100$, classical algorithms such as hHGS and hSMA more frequently lost to hybrids (e.g., hSHADE, hIMODE). In contrast, at $dim = 10$, statistical differences were less pronounced, underscoring that scaling effects amplify with dimensionality.

As illustrated in the boxplots  (Fig.\ref{fig9}), the distributional patterns under $f(5x)$ remain largely similar to the baseline. For most algorithms, medians are unchanged - e.g., hSPSO2011, hMPA, hHGS retain their central tendency. Exceptions include hIMODE and hDMSSA, where medians drop from around 1.0 to near-zero, reflecting improved optimization accuracy. The box height (spread) is reduced or unchanged across most algorithms - hMPA shows a dramatic drop - implying stable performance. The absence or reduction of outliers further confirms that scaling does not introduce instability; in fact, it may slightly increase result consistency compared to both the baseline and translation.

Boxplots for $dim = 10$ and $dim = 30$ are compact and symmetric, while for $dim = 50$ and $dim = 100$, shifts in medians and expansion of interquartile ranges occur, particularly for hHHO and hSMA. These shifts confirm that increased search space magnifies the sensitivity of non-adaptive algorithms, which must recalibrate their exploration step sizes. Relative to the baseline, hybrid methods (e.g., hIMODE, hDMSSA) previously showed tight, centered boxes; under scaling, dispersion increases with $dim$, indicating a weakened stability mechanism. In contrast to $f(x+8)$, scaling yields fewer extreme outliers, suggesting better local structural consistency in the objective function landscape.

Friedman’s test (Fig.\ref{fig10}) again reports significant inter-algorithm differences ($p \ll 0.05$). The CD diagrams preserve the ranking structure from the baseline: hybrid methods (hIMODE, hSHADE, hDMSSA) remain top-ranked, while classical methods (hSPSO2011, hHHO, hHGS) lag behind. Rank spacing becomes more compressed than in translation, with medians forming near-identical groups. This indicates that relative algorithm performance changes uniformly across scale - preserving relative rankings.

In high dimensions ($dim = 50, 100$), the CD plots show small but consistent ranking shifts, especially for traditional methods like hSMA and hHGS, whose lower scale-invariance affects stability \cite{4aijk}. By contrast, hIMODE and hSHADE retain stable ranks across all dimensions, highlighting their adaptive robustness. Relative to $f(x)$, hybrid dominance persists, while classical method rankings degrade beyond $dim = 30$, more clearly than under translation.

The Bayesian win-probability maps (Fig.\ref{fig11}) confirm the CD findings: dominant algorithms (hIMODE, hSHADE, hDMSSA, hMPA) have consistently high win probabilities ($\sim\!90$-$100\%$) against others. No new dominance groups emerge - winners remain consistent. Interestingly, scale transformation increases the probability of draws (values closer to 0.5), unlike in translation. This suggests reduced inter-method separation, consistent with a compressed performance range due to the shared challenge of scaled search space \cite{0jk}.

As dimensionality increases, particularly at $dim = 100$, the Bayesian dominance matrices darken - indicating increased uncertainty in pairwise superiority. Compared with the baseline, these matrices reveal less contrast between strong and weak methods, reflecting a dilution of structural advantage due to scale distortion. Unlike translation, scaling does not create new extremes but blurs the distinction between algorithmic classes.

The convergence plots (Fig.\ref{fig12}) for the scaled functions f(5x) reveal distinct patterns across benchmark functions f1, f6, f12, and f23 under increasing dimensionalities ($dim = 10, 30, 50, 100$). Hybrid algorithms - particularly hSHADE, hIMODE, and hDMSSA - exhibit fast and stable convergence in most settings, often outperforming their non-scaled counterparts and demonstrating superior consistency across dimensions. This effect is especially pronounced in unimodal f1 and hybrid f12 problems, where the convergence trajectories remain steep and smooth even in higher-dimensional landscapes ($dim = 100$). In contrast, classical versions and less robust hybrids (e.g., hHHO) show erratic behavior or premature stagnation, especially on more complex, composition-based functions such as f23. When compared to the baseline $f(x)$, scaling generally results in faster initial convergence and lower final fitness values, particularly in moderate-to-high dimensions. Notably, these improvements in convergence dynamics are aligned - though not always equally strong - with the benefits observed under translation ($f(x+8)$), suggesting that both transformations may enhance search dynamics but with differing algorithm-specific sensitivities. Overall, the convergence characteristics under $f(5x)$ confirm the effectiveness of scaling for robust hybrid designs and indicate their potential resilience in complex, high-dimensional scenarios.

The analysis of $f(5x)$ reveals a partial lack of scale-invariance in many algorithms. Although theoretically an algorithm should be invariant to unit changes, many components (mutation steps, selection thresholds) are value-dependent, hence scale-sensitive. CMA-ES, known for strong scale invariance via covariance-based step adaptation \cite{2},\cite{3}, was not employed in this study due to integration constraints - but serves as a literature benchmark.

In practice, hybrid algorithms like hSHADE and hIMODE, equipped with adaptive structural features, remain robust across scaled domains, as evidenced by their consistently stable outcomes. Conversely, classical algorithms (e.g., hHHO, hHGS) exhibit greater variability, particularly in high-dimensional scaled settings.

Scaling also affects mutation and exploration step calibration-larger variable magnitudes demand proportional step adjustments \cite{4aijk}. Algorithms like hMPA and hDMSSA benefit from reduced relative error, while hIMODE and hSHADE show unit-specific response profiles.

Theoretically, scale non-invariance leads to disproportionate landscape stretching, which can either hinder or facilitate search, depending on operator structure. At high dimensions, this results in dramatically increased result spread, as population-based methods struggle to adjust. Only those with adaptive scaling mechanisms maintain stability.

Overall, these behaviors align with known operator properties: lack of scale and translation invariance results in landscape distortion, reduced robustness, and altered statistical profiles-consistent with both tabular and graphical evidence \cite{0BB},\cite{4aijk},\cite{6atz}.

\subsection{Rotational Transformation of the Objective Function}

\vspace{0.22cm}
This subsection analyzes the impact of rotational transformation of the decision space on the performance of optimization algorithms, using an orthonormal rotation matrix $M \in \mathbb{R}^{n \times n}$. The objective of the analysis is to assess the degree of rotational invariance exhibited by different methods and to identify algorithms that are sensitive to changes in the orientation of the search space.

\scriptsize
\setlength{\tabcolsep}{4pt}

\begin{table}[H]
\centering
\begin{adjustwidth}{-1.5cm}{-2.5cm}
\caption{\footnotesize{Statistical results for 9 hybrid algorithms evaluated on CEC-2017 functions and their rotated variants across 4 dims.}}
\scalebox{0.67}{
\renewcommand{\arraystretch}{0.9}
\begin{tabular}{@{}lrrrrrrrrrrrrrrrr@{}}
\toprule
\multirow{2}{*}{Algorithm} & \multicolumn{7}{c}{dim=10} & \multicolumn{7}{c}{dim=30} \\
\cmidrule(lr){2-8} \cmidrule(lr){9-15}
 & Avg & Med & Std & Sum Rank & Mean Rank & +/- & p-value & Avg & Med & Std & Sum Rank & Mean Rank & +/- & p-value \\
\midrule
hSPSO2011\_f(xM)  & 1.0E+04 & 1.9E+03 & 1.6E+04 & 352 & 12.1 & 133/270 & 8.9E-02 & 6.4E+04 & 2.7E+03 & 4.5E+04 & 359 & 12.4 & 111/298 & 8.6E-02 \\
hSPSO2011  & 1.6E+04 & 1.9E+03 & 2.0E+04 & 348.5 & 12.0 & 140/269 & 7.5E-02 & 5.9E+04 & 2.8E+03 & 3.9E+04 & 357 & 12.3 & 121/294 & 7.8E-02 \\
hMPA\_f(xM)  & 1.7E+03 & 1.6E+03 & 2.5E+01 & 231 & 8.0 & 255/187 & 7.0E-02 & 2.0E+04 & 2.3E+03 & 1.7E+04 & 275 & 9.5 & 217/204 & 8.7E-02 \\
hMPA  & 2.6E+03 & 1.6E+03 & 5.2E+03 & 232 & 8.0 & 256/185 & 7.0E-02 & 1.9E+04 & 2.3E+03 & 1.7E+04 & 269 & 9.3 & 216/206 & 7.9E-02 \\
hSHADE\_f(xM)  & -6.1E-01 & 1.4E+00 & 4.7E+01 & 96 & 3.3 & 378/4 & 1.8E-01 & -2.2E-01 & 4.5E-01 & 5.2E+01 & 97.5 & 3.4 & 349/3 & 1.9E-01 \\
hSHADE  & -6.1E-01 & 1.4E+00 & 4.7E+01 & 95 & 3.3 & 378/4 & 1.8E-01 & -2.2E-01 & 4.5E-01 & 5.2E+01 & 97.5 & 3.4 & 349/3 & 1.9E-01 \\
hIMODE\_f(xM)  & -1.5E+00 & 1.0E+00 & 5.1E+01 & 78.5 & 2.7 & 382/0 & 1.8E-01 & -2.6E-01 & -4.8E-01 & 5.5E+01 & 101 & 3.5 & 355/0 & 1.7E-01 \\
hIMODE  & -1.5E+00 & 1.0E+00 & 5.1E+01 & 78.5 & 2.7 & 382/0 & 1.8E-01 & -2.3E-01 & 1.3E+00 & 5.5E+01 & 108 & 3.7 & 353/0 & 1.7E-01 \\
hHGS\_f(xM)  & 1.1E+04 & 2.6E+03 & 1.4E+04 & 423.5 & 14.6 & 66/347 & 9.8E-02 & 1.8E+05 & 3.2E+03 & 3.5E+05 & 407 & 14.0 & 83/338 & 6.6E-02 \\
hHGS  & 1.1E+04 & 2.6E+03 & 1.4E+04 & 423.5 & 14.6 & 66/347 & 9.8E-02 & 1.4E+05 & 3.2E+03 & 3.3E+05 & 415 & 14.3 & 83/336 & 6.5E-02 \\
hHHO\_f(xM)  & 5.8E+04 & 2.3E+03 & 7.1E+04 & 491.5 & 16.9 & 10/431 & 7.6E-02 & 3.2E+05 & 3.3E+03 & 3.3E+05 & 498 & 17.2 & 6/440 & 3.9E-02 \\
hHHO  & 5.8E+04 & 2.3E+03 & 7.1E+04 & 491.5 & 16.9 & 10/431 & 7.6E-02 & 3.5E+05 & 3.3E+03 & 3.1E+05 & 493 & 17.0 & 9/443 & 3.6E-02 \\
hSMA\_f(xM)  & 3.6E+03 & 1.9E+03 & 1.1E+03 & 315.5 & 10.9 & 157/256 & 1.0E-01 & 1.9E+04 & 2.9E+03 & 1.0E+04 & 296 & 10.2 & 171/225 & 9.1E-02 \\
hSMA  & 3.6E+03 & 1.9E+03 & 1.1E+03 & 315.5 & 10.9 & 157/256 & 1.0E-01 & 2.1E+04 & 2.9E+03 & 9.6E+03 & 310 & 10.7 & 157/236 & 9.4E-02 \\
hDMSSA\_f(xM)  & -7.9E-01 & 7.3E-01 & 4.9E+01 & 88.5 & 3.1 & 377/4 & 1.6E-01 & 3.3E-01 & 1.2E+00 & 5.2E+01 & 106 & 3.7 & 349/3 & 1.8E-01 \\
hDMSSA  & 9.7E+01 & 9.2E+01 & 5.3E+00 & 174 & 6.0 & 348/145 & 3.6E-20 & 3.0E-01 & 2.2E-01 & 5.2E+01 & 101 & 3.5 & 350/8 & 1.7E-01 \\
hAROA\_f(xM)  & 5.5E+03 & 1.7E+03 & 9.3E+03 & 370 & 12.8 & 118/301 & 6.2E-02 & 6.4E+04 & 2.9E+03 & 5.1E+04 & 344 & 11.9 & 135/259 & 8.7E-02 \\
hAROA  & 5.4E+03 & 1.7E+03 & 9.2E+03 & 355 & 12.2 & 118/294 & 6.8E-02 & 6.7E+04 & 2.9E+03 & 3.5E+04 & 330 & 11.4 & 139/257 & 9.9E-02 \\
\bottomrule
\end{tabular}
}
\vspace{1.2mm}
\scalebox{0.66}{
\renewcommand{\arraystretch}{0.9}
\begin{tabular}{@{}l
ccccccc  
@{\hspace{0.5cm}}
ccccccc
@{}}
\toprule
\multirow{2}{*}{Algorithm} & \multicolumn{7}{c}{dim=50} & \multicolumn{7}{c}{dim=100} \\
\cmidrule(lr){2-8} \cmidrule(lr){9-15}
 & Avg & Med & Std & Sum Rank & Mean Rank & +/- & p-value & Avg & Med & Std & Sum Rank & Mean Rank & +/- & p-value \\
\midrule
hSPSO2011\_f(xM)  & 9.0E+05 & 3.3E+03 & 2.4E+05 & 370 & 12.8 & 106/306 & 7.6E-02 & 3.2E+06 & 5.9E+03 & 1.2E+06 & 383 & 13.2 & 113/325 & 5.3E-02 \\
hSPSO2011  & 8.8E+05 & 3.3E+03 & 2.4E+05 & 373 & 12.9 & 106/307 & 8.0E-02 & 3.1E+06 & 6.0E+03 & 1.0E+06 & 384 & 13.2 & 109/332 & 4.8E-02 \\
hMPA\_f(xM)  & 5.7E+05 & 3.1E+03 & 4.1E+05 & 274 & 9.4 & 200/218 & 7.1E-02 & 6.1E+06 & 5.3E+03 & 2.9E+06 & 320 & 11.1 & 164/254 & 6.0E-02 \\
hMPA  & 4.3E+05 & 3.2E+03 & 3.3E+05 & 289 & 10.0 & 197/216 & 7.1E-02 & 6.0E+06 & 5.3E+03 & 2.8E+06 & 309 & 10.7 & 169/247 & 6.3E-02 \\
hSHADE\_f(xM)  & 1.1E+00 & 1.9E+00 & 5.3E+01 & 102.5 & 3.5 & 349/7 & 1.7E-01 & -6.3E-01 & -4.3E-01 & 5.2E+01 & 104 & 3.6 & 355/6 & 1.6E-01 \\
hSHADE  & 1.1E+00 & 1.9E+00 & 5.3E+01 & 102.5 & 3.5 & 349/7 & 1.7E-01 & -6.2E-01 & -7.9E-02 & 5.2E+01 & 104 & 3.6 & 352/9 & 1.6E-01 \\
hIMODE\_f(xM)  & 8.4E-01 & -2.1E-02 & 5.6E+01 & 97 & 3.3 & 354/3 & 1.5E-01 & -7.7E-01 & -4.2E-02 & 5.5E+01 & 92 & 3.2 & 354/3 & 1.6E-01 \\
hIMODE  & 8.0E-01 & 5.1E-01 & 5.6E+01 & 101 & 3.5 & 354/6 & 1.5E-01 & -8.9E-01 & -1.1E-01 & 5.5E+01 & 91 & 3.1 & 356/6 & 1.6E-01 \\
hHGS\_f(xM)  & 2.6E+06 & 3.7E+03 & 4.2E+06 & 417 & 14.4 & 79/341 & 6.7E-02 & 2.4E+08 & 7.0E+03 & 1.0E+08 & 415 & 14.3 & 78/337 & 7.2E-02 \\
hHGS  & 1.7E+06 & 4.0E+03 & 1.6E+06 & 418 & 14.4 & 76/341 & 7.3E-02 & 2.7E+08 & 7.0E+03 & 1.2E+08 & 419 & 14.4 & 81/339 & 7.6E-02 \\
hHHO\_f(xM)  & 2.2E+06 & 4.3E+03 & 1.3E+06 & 491 & 16.9 & 13/441 & 3.4E-02 & 1.7E+07 & 7.4E+03 & 8.1E+06 & 468 & 16.1 & 26/422 & 4.3E-02 \\
hHHO  & 2.2E+06 & 4.0E+03 & 1.2E+06 & 495 & 17.1 & 12/439 & 3.9E-02 & 1.7E+07 & 7.1E+03 & 8.4E+06 & 475 & 16.4 & 24/426 & 4.5E-02 \\
hSMA\_f(xM)  & 1.7E+05 & 3.3E+03 & 8.1E+04 & 290 & 10.0 & 188/222 & 9.0E-02 & 5.3E+05 & 4.9E+03 & 2.2E+05 & 249 & 8.6 & 241/186 & 5.6E-02 \\
hSMA  & 1.7E+05 & 3.3E+03 & 7.6E+04 & 293 & 10.1 & 190/221 & 8.4E-02 & 6.2E+05 & 4.9E+03 & 2.6E+05 & 245.5 & 8.5 & 243/189 & 5.5E-02 \\
hDMSSA\_f(xM)  & 7.7E-01 & 1.2E+00 & 5.3E+01 & 105 & 3.6 & 357/6 & 1.3E-01 & -4.8E-01 & 3.5E-02 & 5.1E+01 & 101 & 3.5 & 354/9 & 1.4E-01 \\
hDMSSA  & 8.4E-01 & 1.2E+00 & 5.2E+01 & 103 & 3.6 & 357/3 & 1.5E-01 & -4.5E-01 & -6.3E-02 & 5.1E+01 & 117 & 4.0 & 360/10 & 1.4E-01 \\
hAROA\_f(xM)  & 5.1E+05 & 3.4E+03 & 2.5E+05 & 321 & 11.1 & 141/238 & 1.0E-01 & 6.7E+06 & 5.0E+03 & 2.2E+06 & 348 & 12.0 & 137/276 & 7.9E-02 \\
hAROA  & 4.4E+05 & 3.4E+03 & 2.1E+05 & 321 & 11.1 & 144/250 & 9.4E-02 & 7.4E+06 & 5.6E+03 & 2.5E+06 & 344 & 11.9 & 138/278 & 6.8E-02 \\
\bottomrule
\label{tab4}
\end{tabular}
}
\vspace{1.5mm}\\
\footnotesize{Friedman test p-values: dim10=4.3E-82, dim30=2.2E-76, dim50=7.3E-77, dim100=9.0E-76}
\end{adjustwidth}
\end{table}

\normalsize
\begin{figure}[H]
  \centering
\begin{adjustwidth}{-1.8cm}{-2.0cm}
  \begin{subfigure}{0.4\textwidth}
     \includegraphics[height=5.5cm,width=6.9cm]{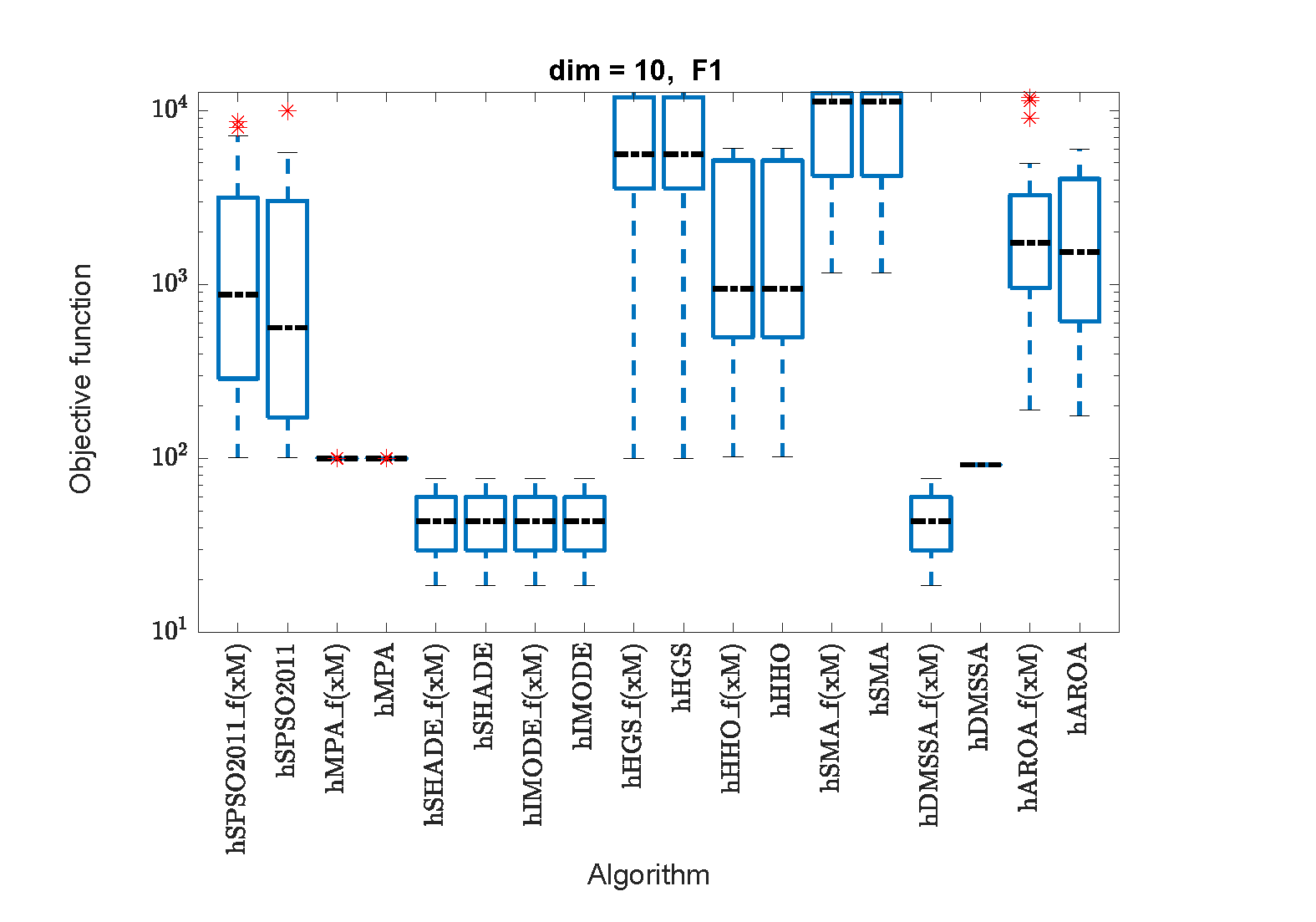}
  \end{subfigure}\hfill
  \begin{subfigure}{0.4\textwidth}
    \includegraphics[height=5.5cm,width=6.9cm]{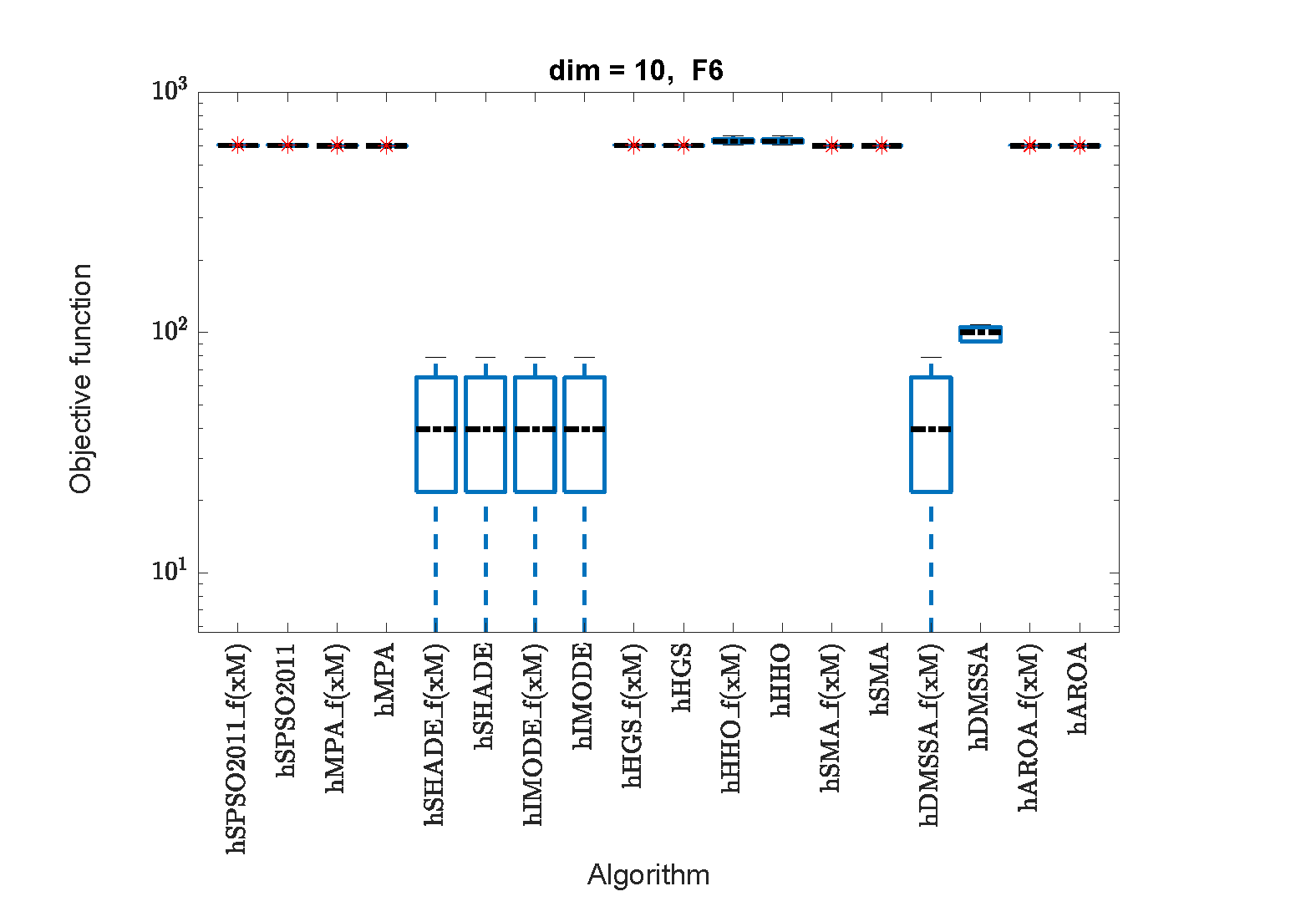}
  \end{subfigure}\hfill
  \begin{subfigure}{0.4\textwidth}
    \includegraphics[height=5.5cm,width=6.9cm]{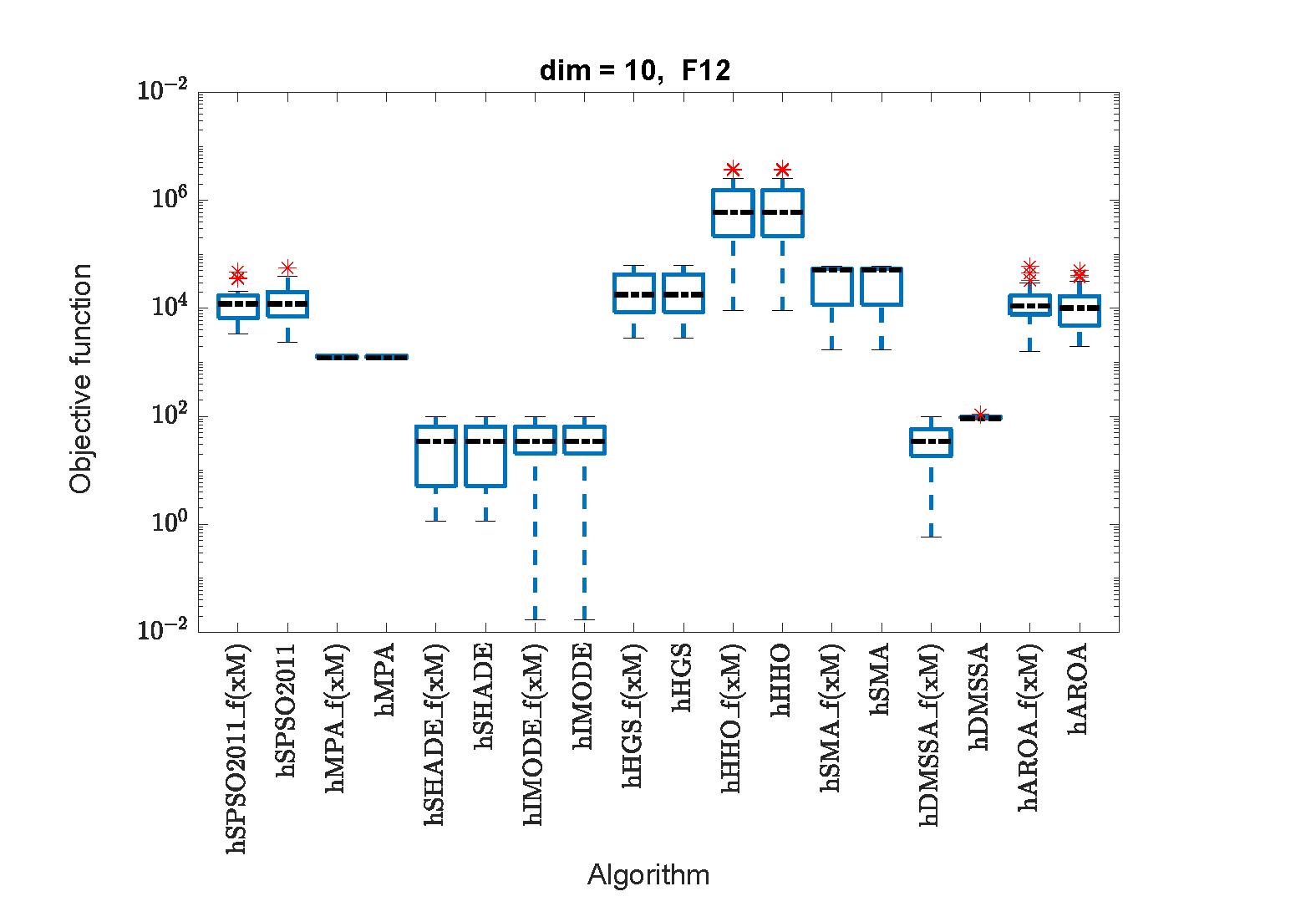}
  \end{subfigure}
 \end{adjustwidth}
\end{figure}

\begin{figure}[H]
  \centering
\begin{adjustwidth}{-2cm}{-2.0cm}
   \begin{subfigure}{0.4\textwidth}
    \includegraphics[height=5.1cm,width=6.9cm]{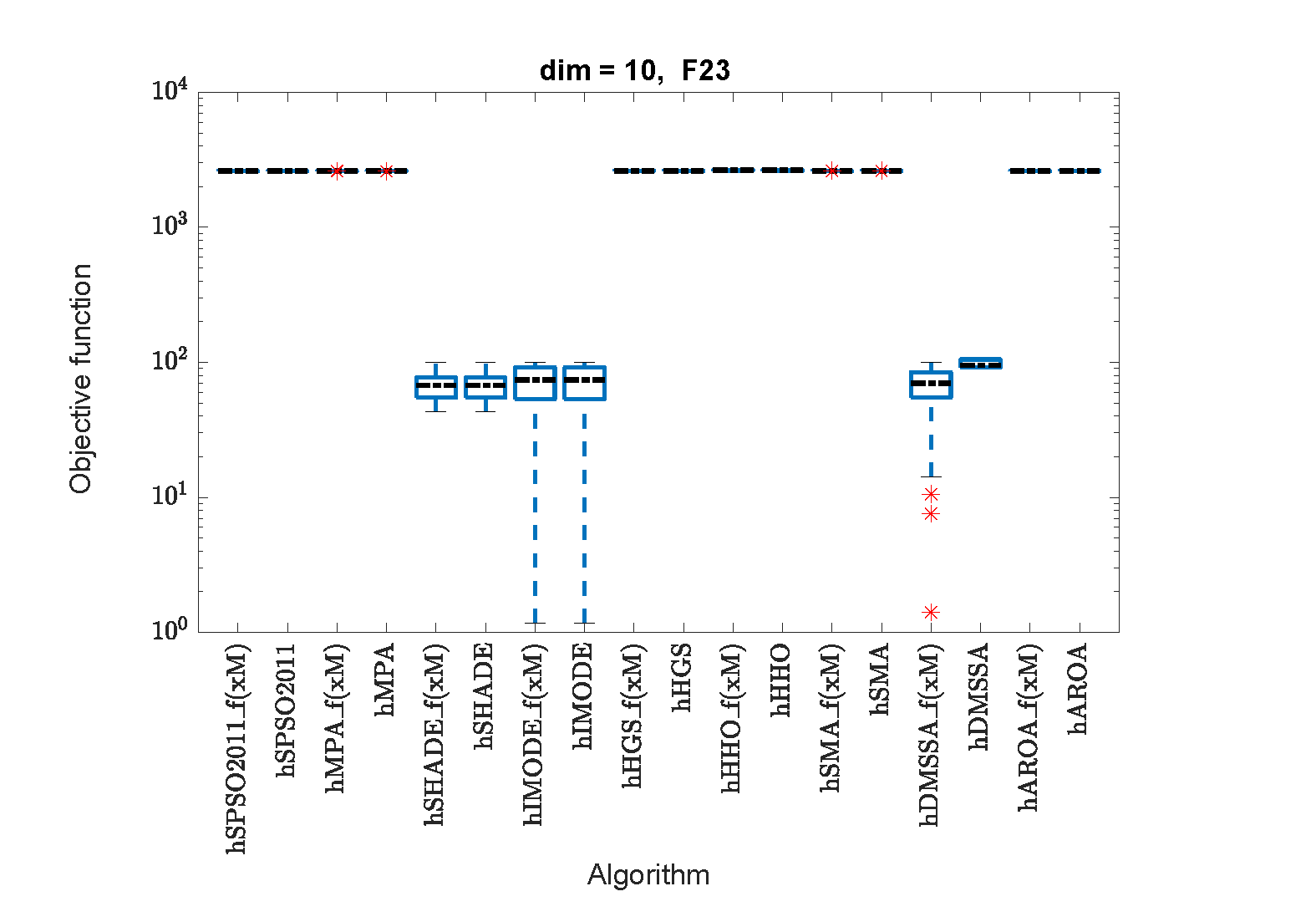}
  \end{subfigure}\hfill
  \begin{subfigure}{0.4\textwidth}
    \includegraphics[height=5.1cm,width=6.9cm]{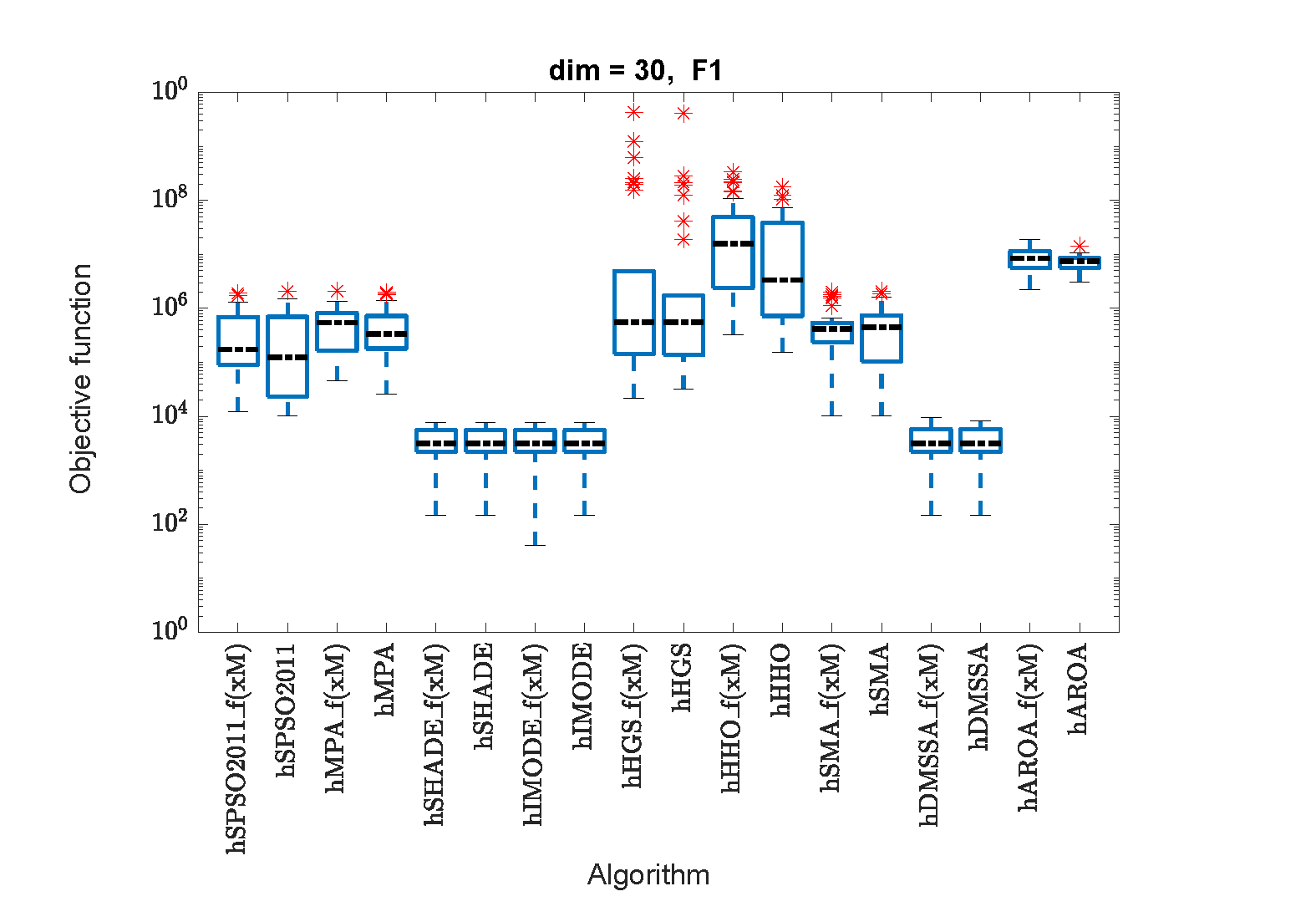}
\end{subfigure}\hfill
  \begin{subfigure}{0.4\textwidth} 
        \includegraphics[height=5.1cm,width=6.9cm]{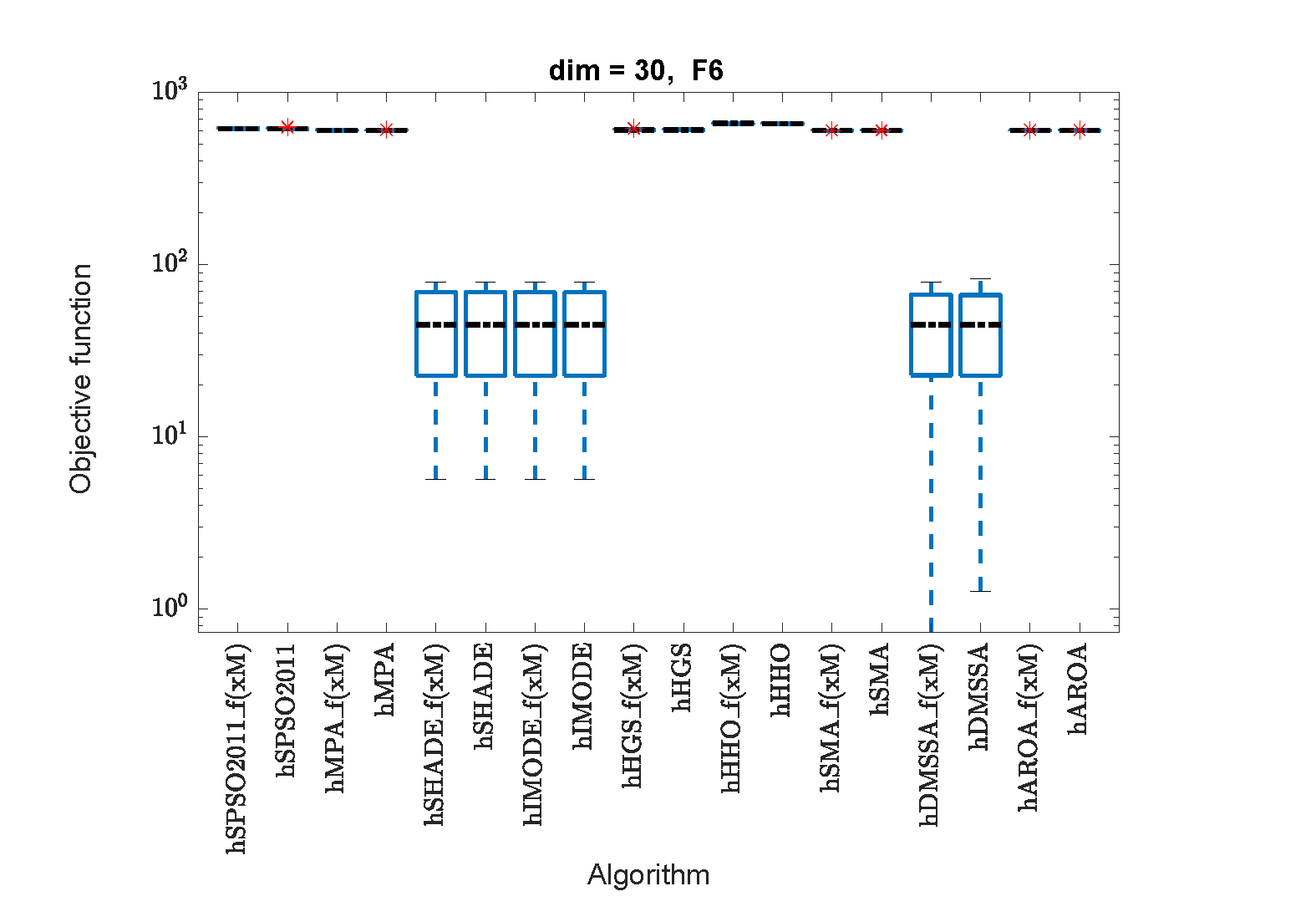}
   \end{subfigure}
   \end{adjustwidth}
\end{figure}

 \begin{figure}[H]
  \centering
\begin{adjustwidth}{-2cm}{-2.0cm}
   \begin{subfigure}{0.4\textwidth}
        \includegraphics[height=5.1cm,width=6.9cm]{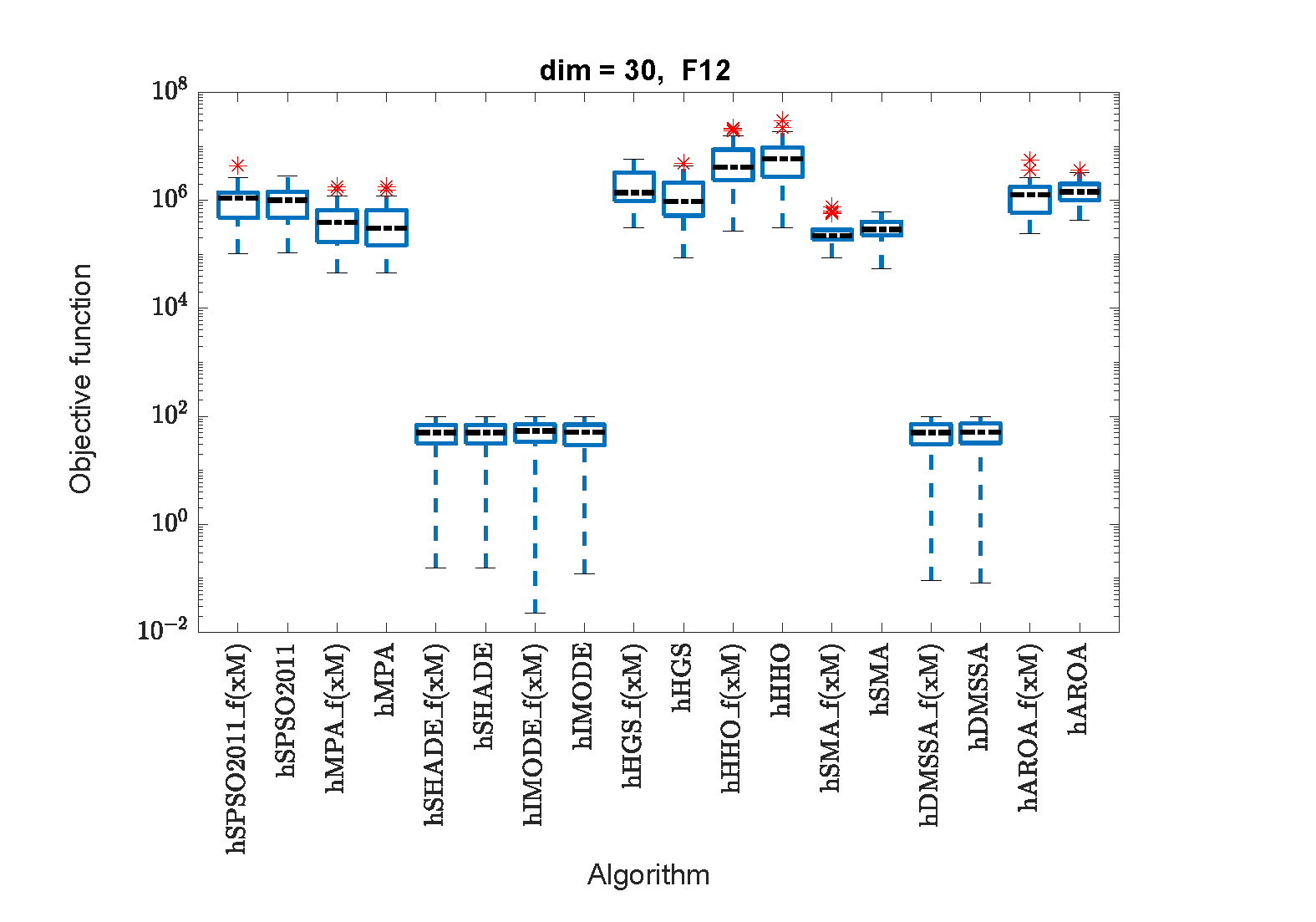}
  \end{subfigure}\hfill
  \begin{subfigure}{0.4\textwidth} 
        \includegraphics[height=5.1cm,width=6.9cm]{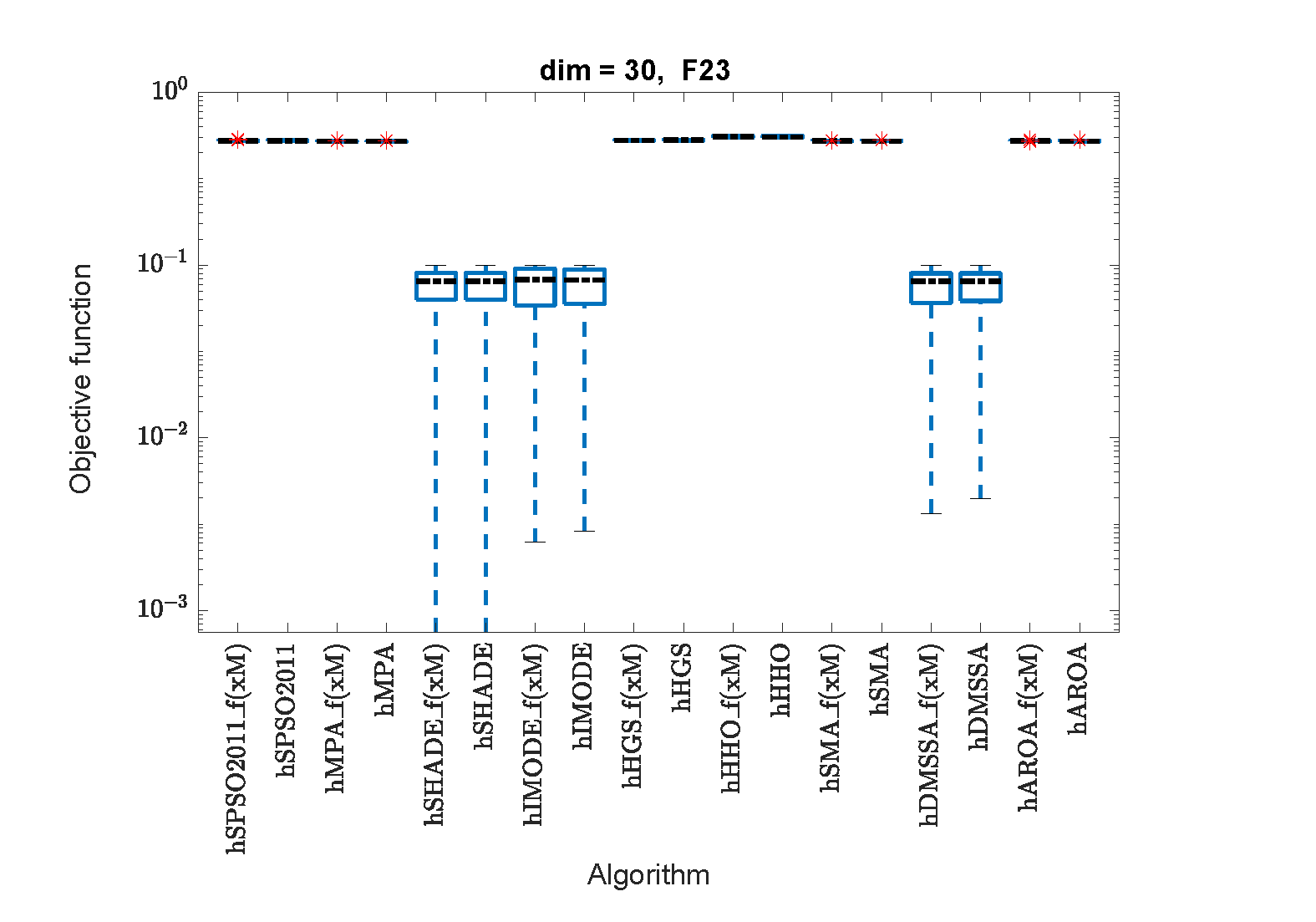}
  \end{subfigure}\hfill
  \begin{subfigure}{0.4\textwidth}
        \includegraphics[height=5.1cm,width=6.9cm]{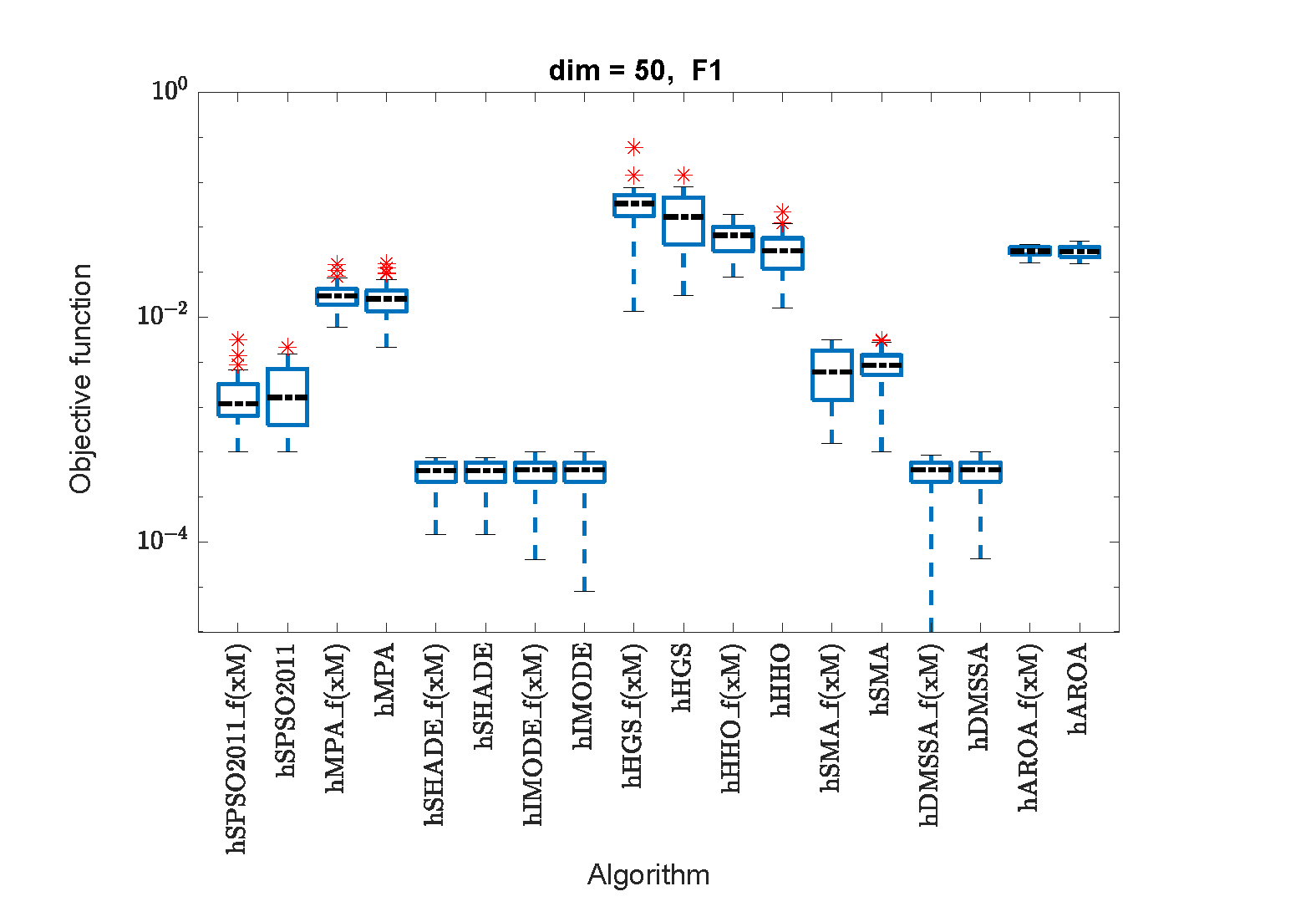}
  \end{subfigure}
  \end{adjustwidth}
\end{figure}

  \begin{figure}[H]
  \centering
\begin{adjustwidth}{-2cm}{-2.0cm}
  \begin{subfigure}{0.4\textwidth} 
        \includegraphics[height=5.1cm,width=6.9cm]{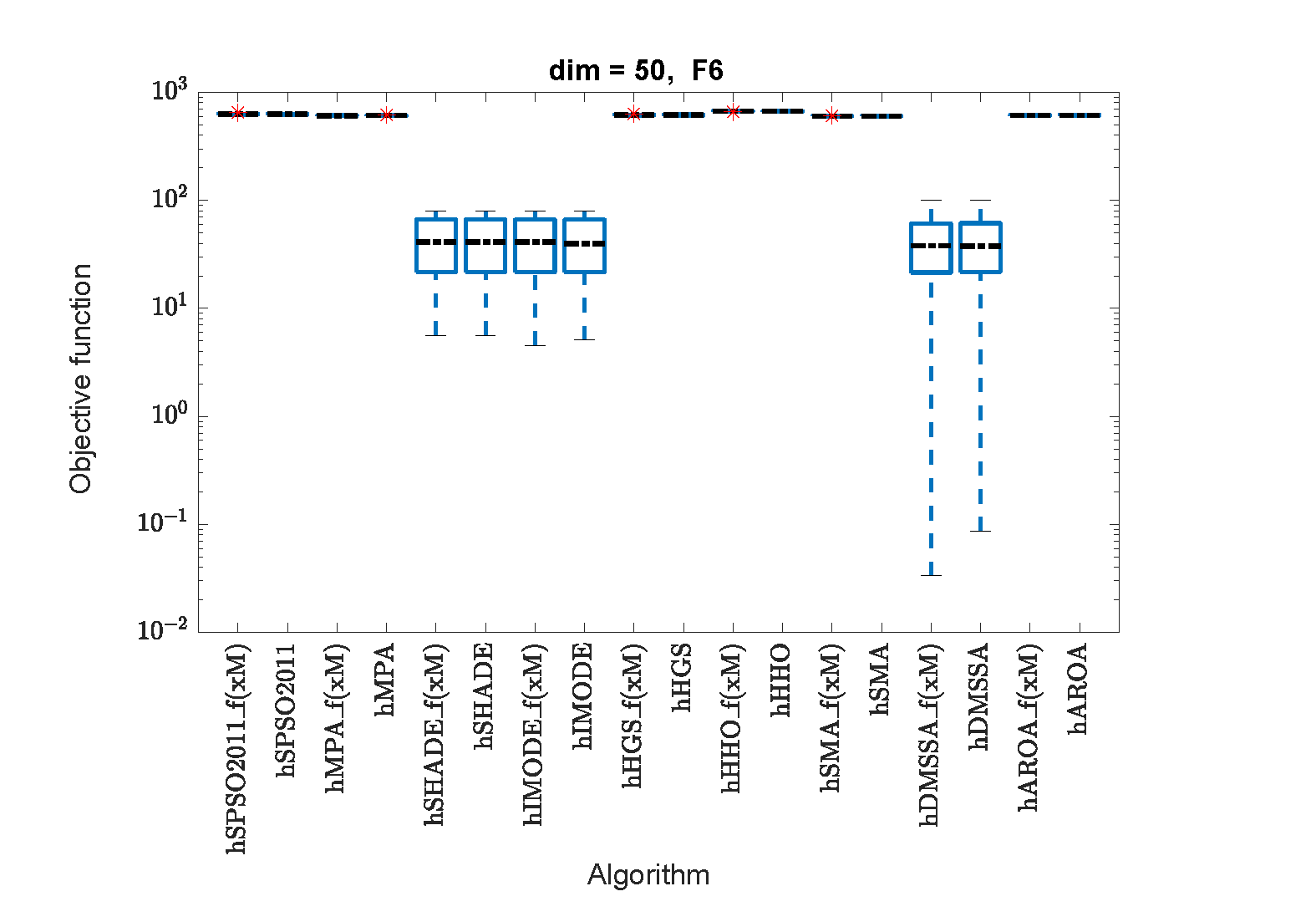}
  \end{subfigure}\hfill
  \begin{subfigure}{0.4\textwidth}
    \includegraphics[height=5.1cm,width=6.9cm]{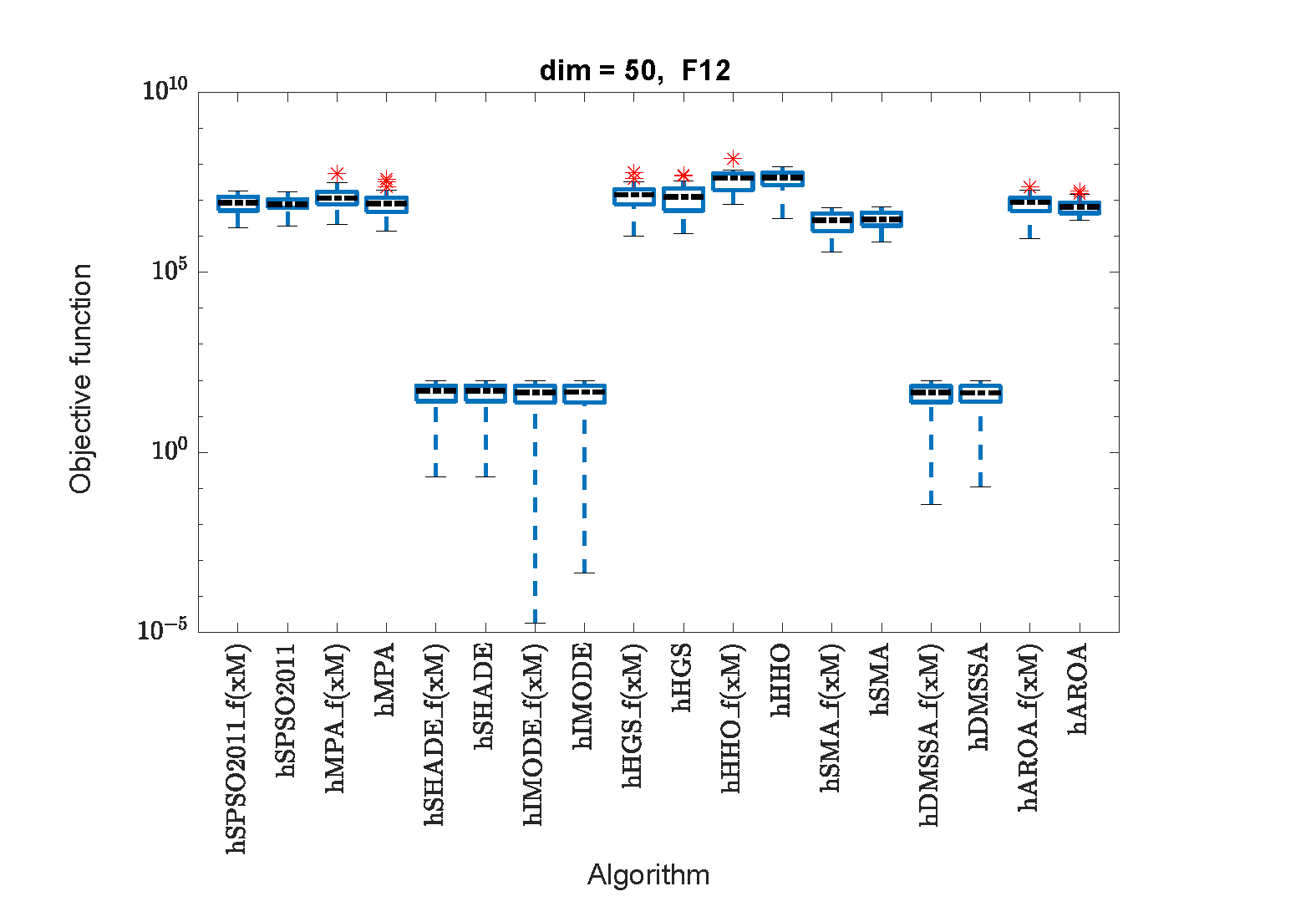}
     \end{subfigure}\hfill
 \begin{subfigure}{0.4\textwidth}
    \includegraphics[height=5.1cm,width=6.9cm]{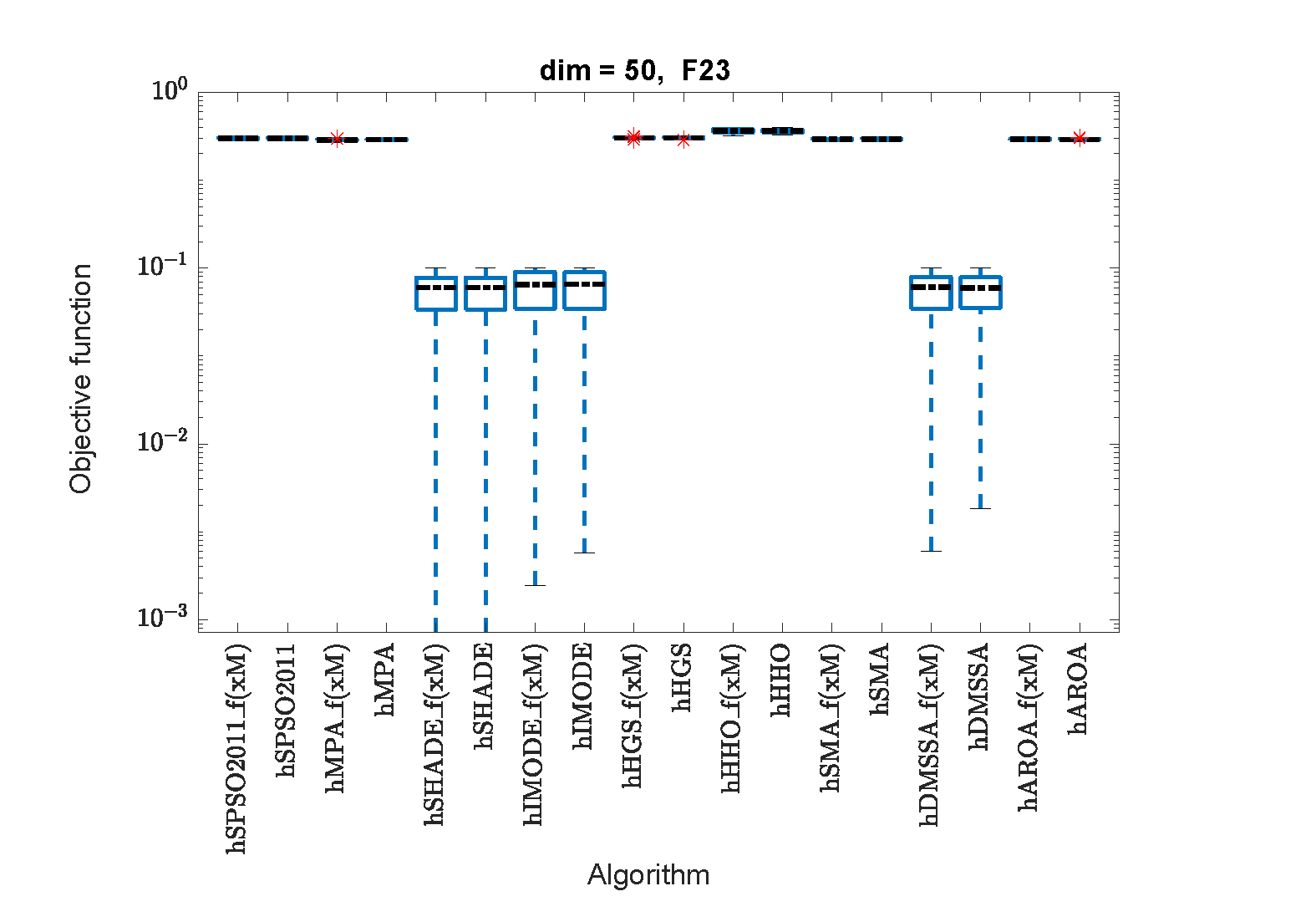}
  \end{subfigure}
\end{adjustwidth}
\end{figure}
 
  \begin{figure}[H]
  \centering
\begin{adjustwidth}{-2cm}{-2.0cm}
  \begin{subfigure}{0.4\textwidth}
    \includegraphics[height=5.1cm,width=6.9cm]{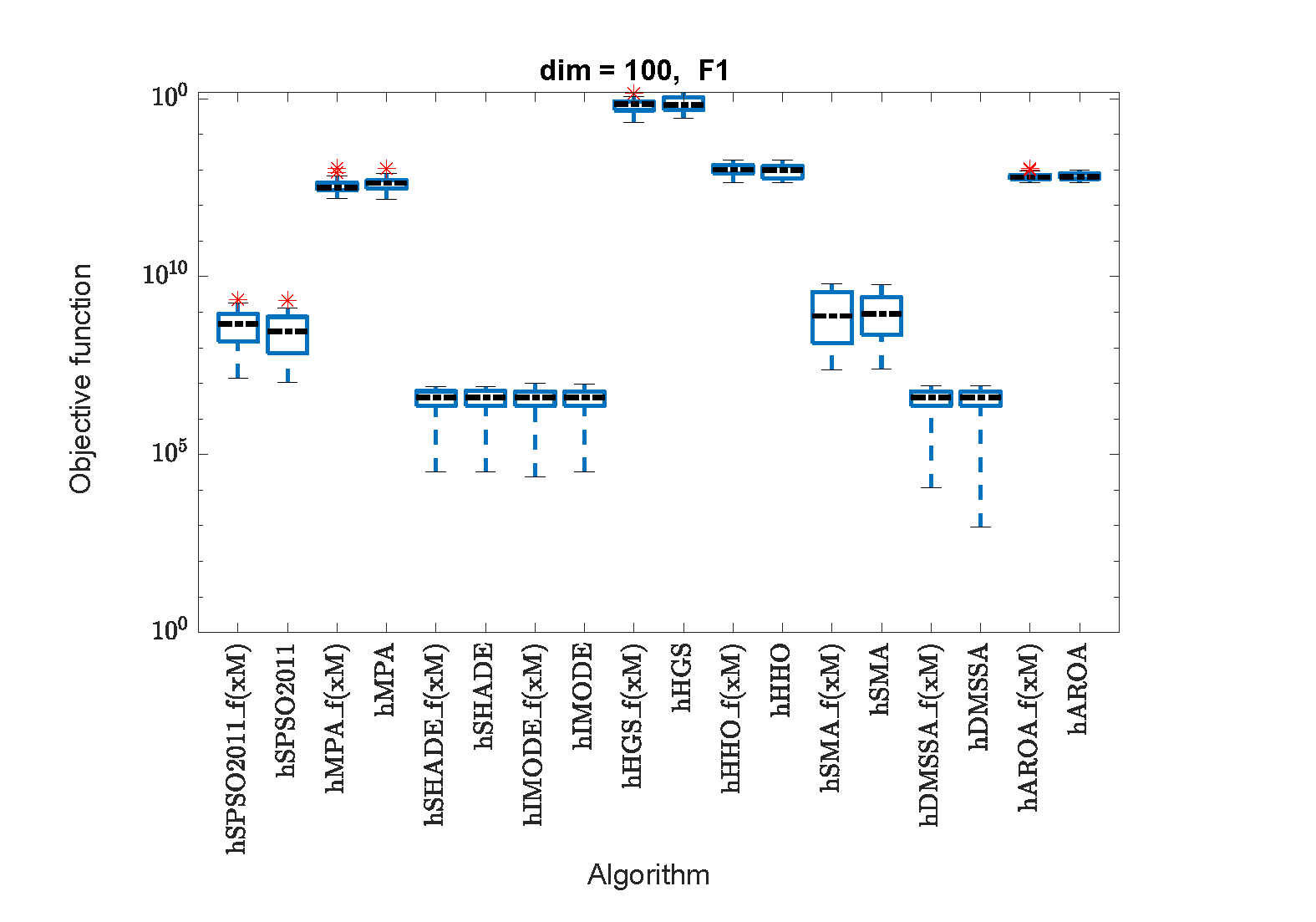}
  \end{subfigure}\hfill
  \begin{subfigure}{0.4\textwidth}
        \includegraphics[height=5.1cm,width=6.9cm]{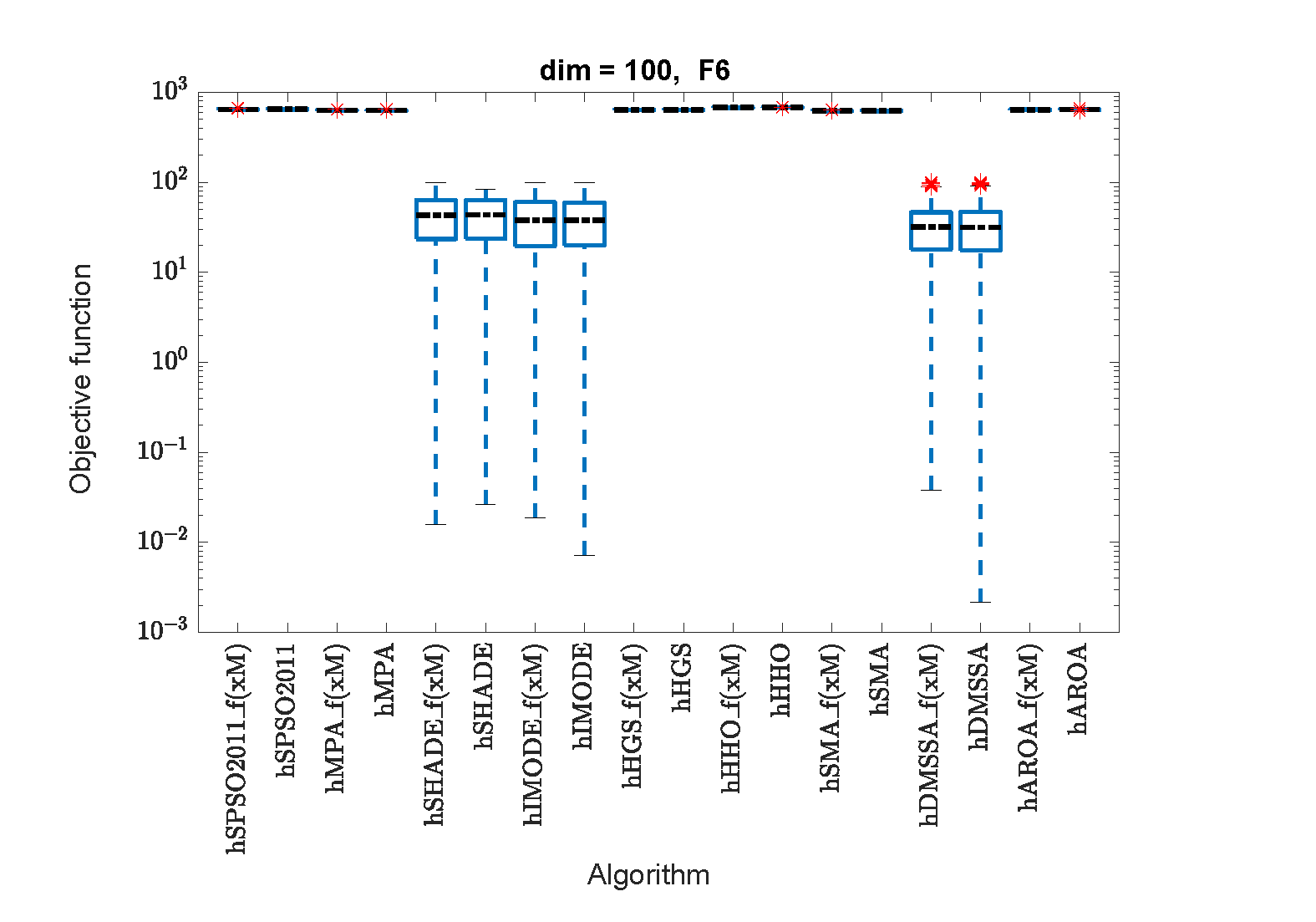}
     \end{subfigure}\hfill
   \begin{subfigure}{0.4\textwidth}
        \includegraphics[height=5.1cm,width=6.9cm]{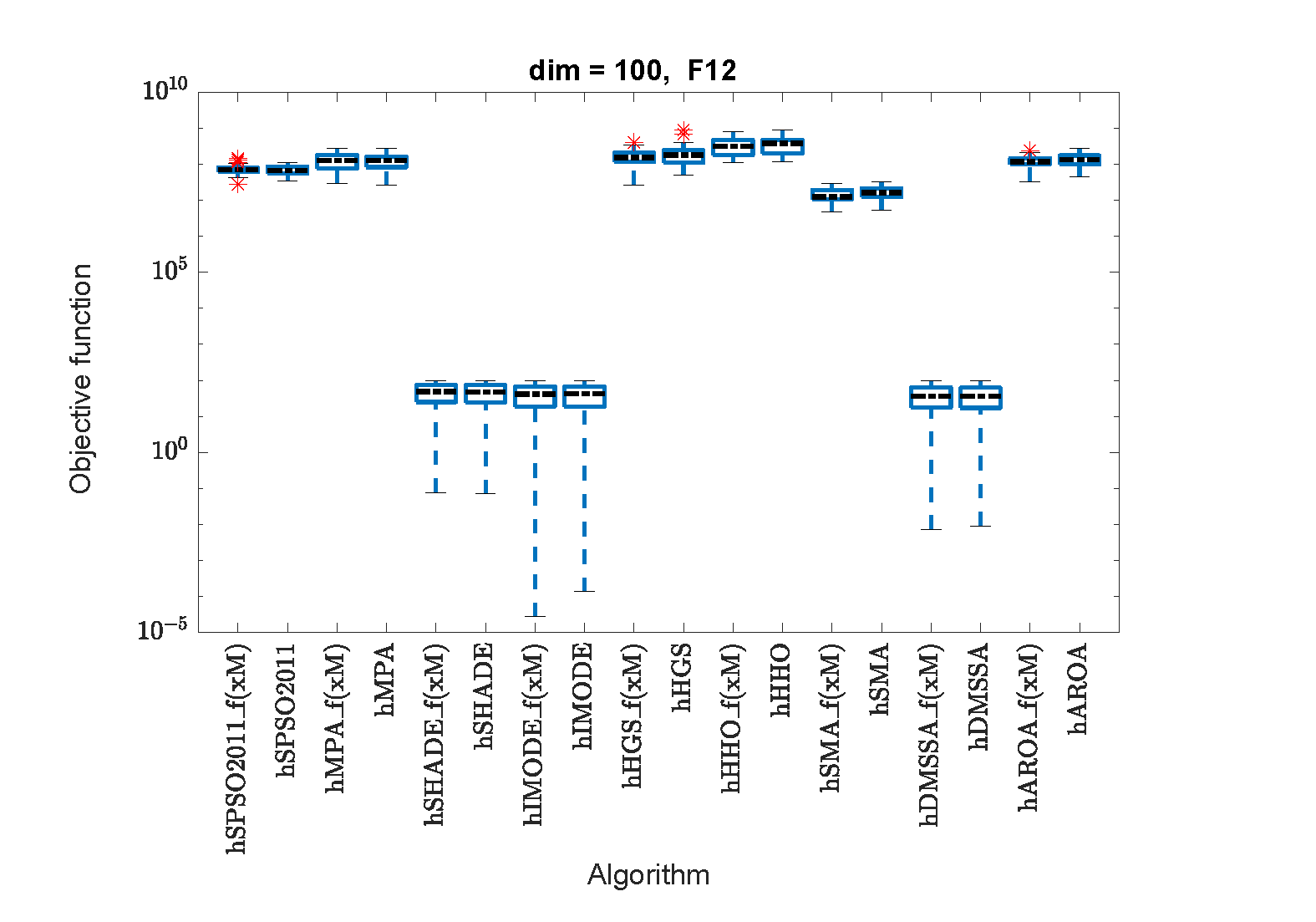}
    \end{subfigure}
    \end{adjustwidth}
    \end{figure}

\begin{figure}[H]
 \centering
\begin{adjustwidth}{-2cm}{-2.0cm}
\begin{center}
  \begin{subfigure}{0.4\textwidth}
   \centering
        \includegraphics[height=5.1cm,width=6.9cm]{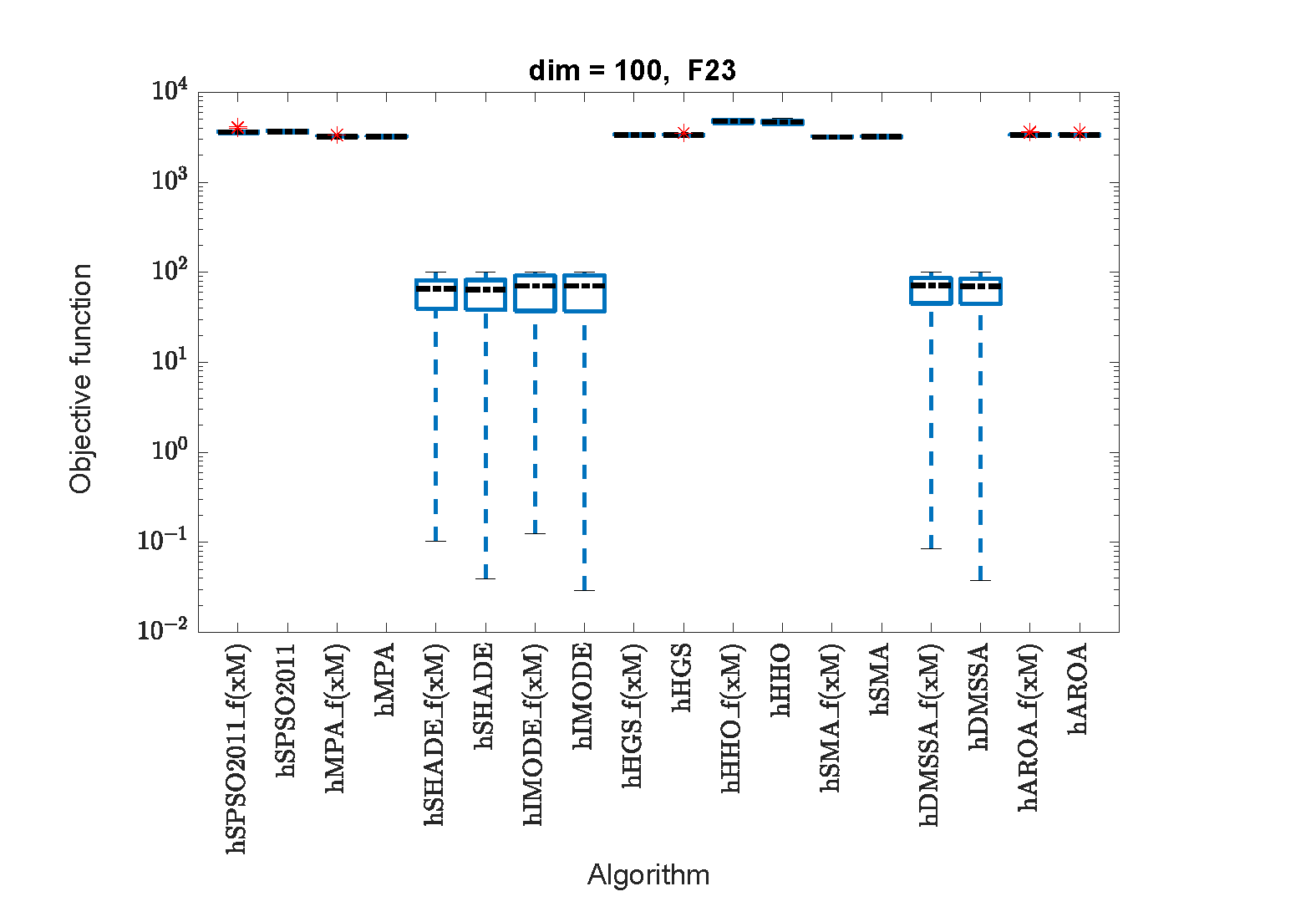}   
  \end{subfigure}
  \end{center}
  \begin{center}
  \begin{subfigure}{0.4\textwidth}
   \centering
    \includegraphics[width=\linewidth]{boxplotlegend.pdf}     
  \end{subfigure}
  \end{center}
  \caption{\footnotesize{Boxplots of final objective values for 9 hybrid algorithms evaluated on original CEC-2017 functions and their transformed variants across 4 dims.}}
  \label{fig13}
\end{adjustwidth}
\end{figure}

\normalsize

\begin{figure}[H]
  \centering
\begin{adjustwidth}{-1cm}{1.0cm}
  \begin{subfigure}{0.35\textwidth} 
        \includegraphics[height=5cm,width=7.5cm]{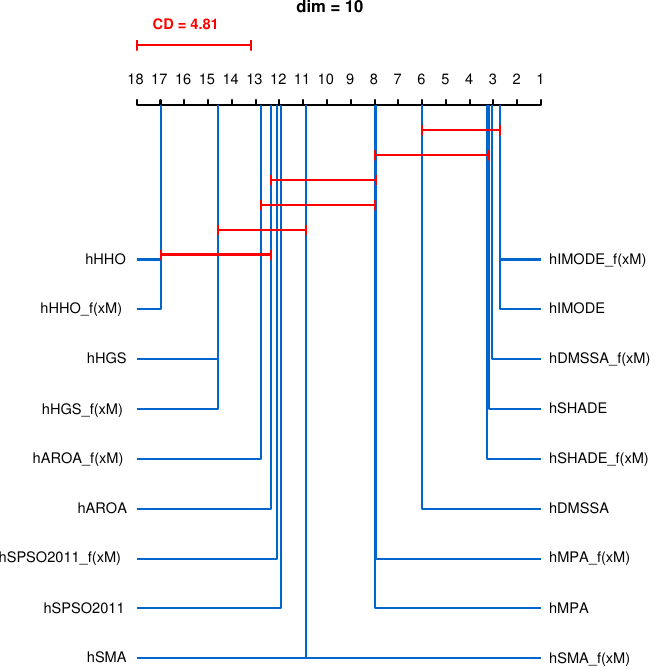}
  \end{subfigure}\hfill
  \begin{subfigure}{0.36\textwidth}
    \includegraphics[height=6cm,width=7.5cm]{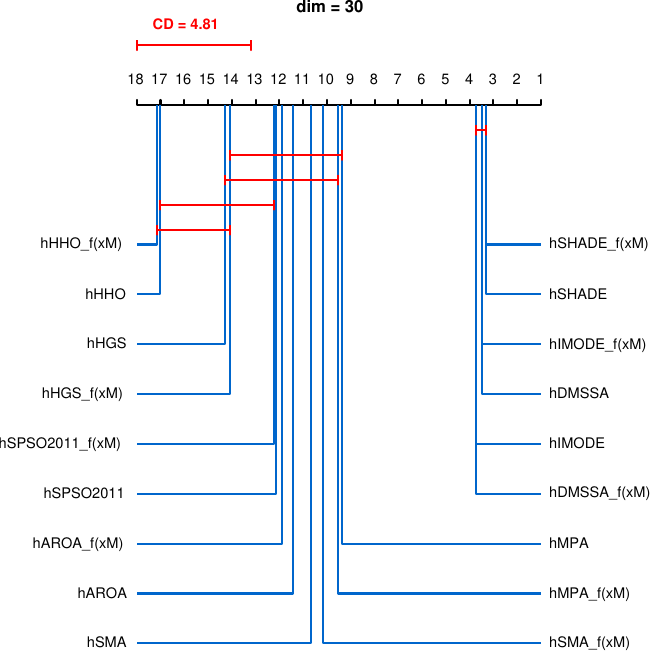}
     \end{subfigure}\hfill
\end{adjustwidth}
\end{figure}

\begin{figure}[H]
\centering
\vspace{0.2cm}
\begin{adjustwidth}{-1cm}{1.0cm}
 \begin{subfigure}{0.35\textwidth}
    \includegraphics[height=5cm,width=7.5cm]{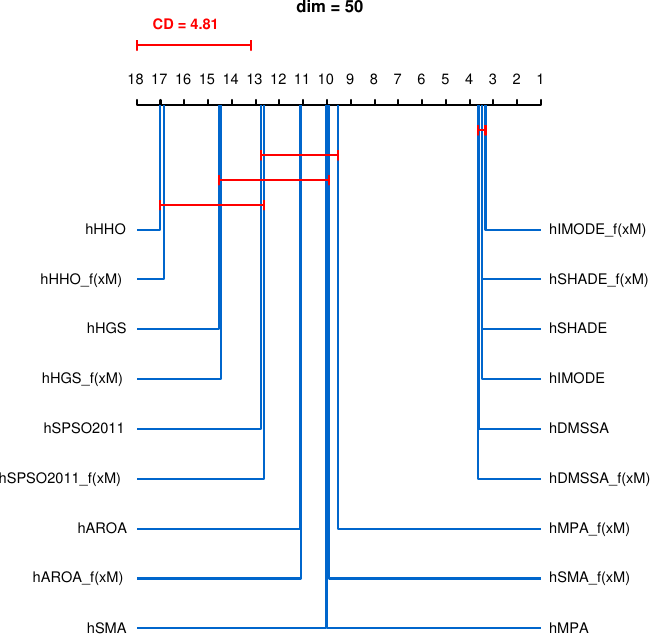}
  \end{subfigure}\hfill
  \begin{subfigure}{0.35\textwidth}
    \includegraphics[height=5cm,width=7.5cm]{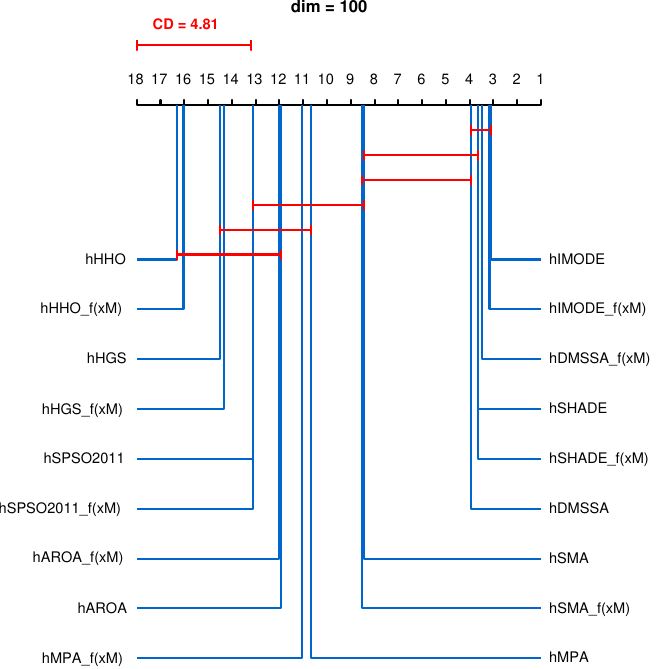}
  \end{subfigure}\hfill
  \caption{\footnotesize{CD diagrams for 9 hybrid algorithms tested on original CEC-2017 functions and their rotated variants across 4 dims.}}
  \label{fig14}
\end{adjustwidth}
\end{figure}

\begin{figure}[H] 
\adjustbox{scale=1.25,center}{
\centering
  \begin{subfigure}[t]{0.495\textwidth}
    \centering
\includegraphics[width=\linewidth]{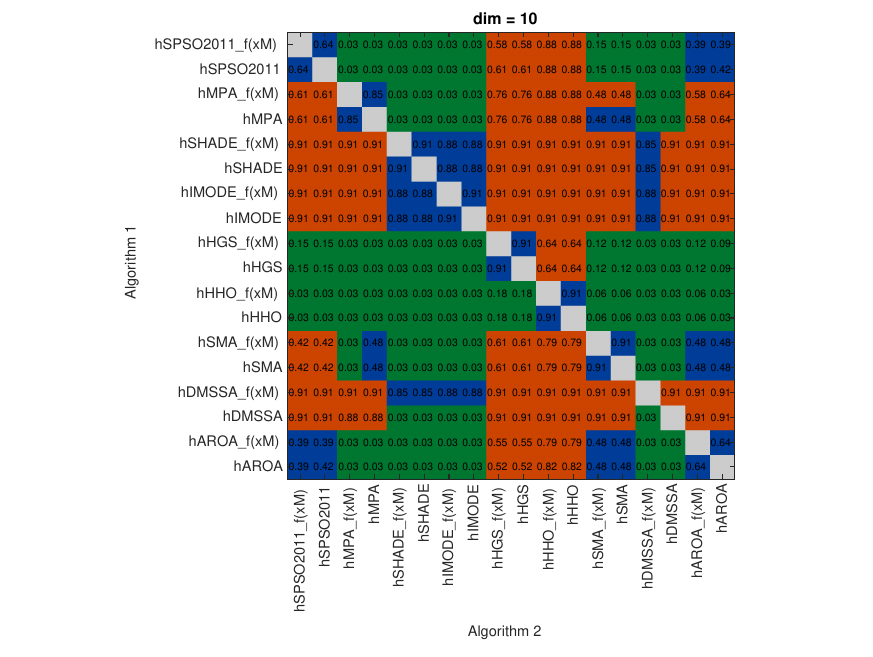}
  \end{subfigure}
  \begin{subfigure}[t]{0.495\textwidth}
    \centering
\includegraphics[width=\linewidth]{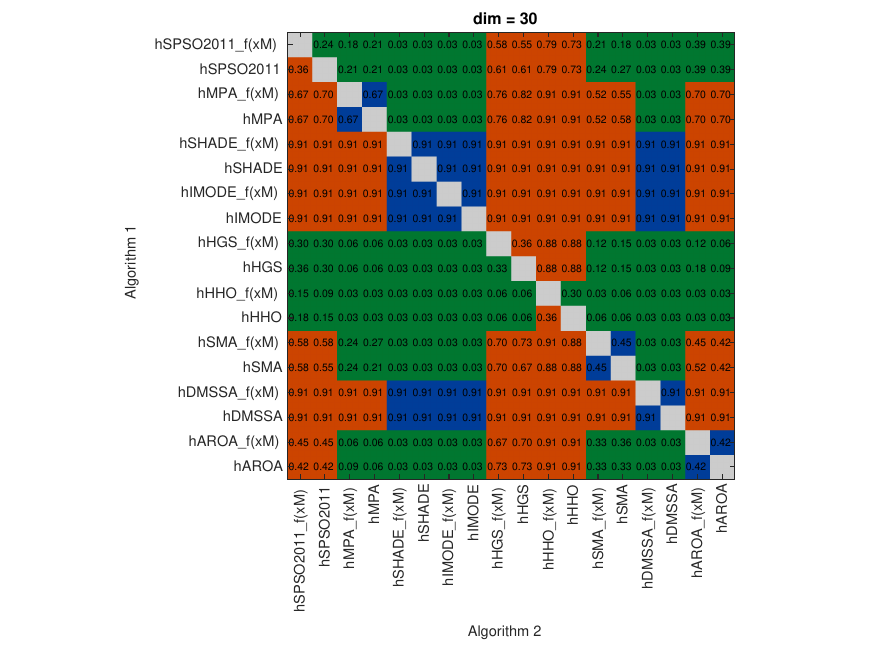}
  \end{subfigure}
  }
  \end{figure}
  \vspace{0.2cm}
  \begin{figure}[H] 
\adjustbox{scale=1.25,center}{
\begin{subfigure}[t]{0.495\textwidth}
    \centering
\includegraphics[width=\linewidth]{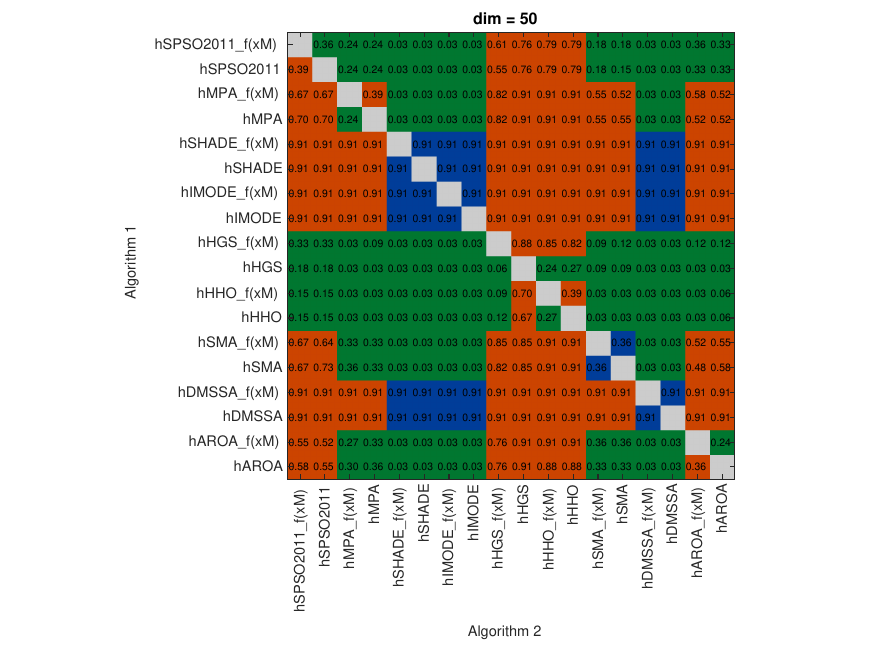}
  \end{subfigure}
 \begin{subfigure}[t]{0.495\textwidth}
    \centering
\includegraphics[width=\linewidth]{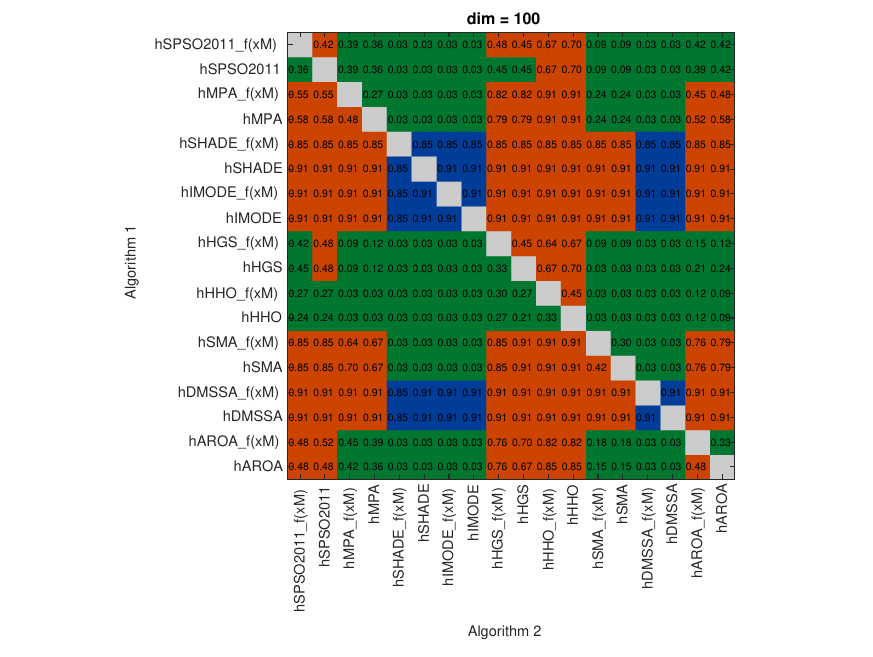}
  \end{subfigure}
  }
  \vspace{0.2cm}
  \adjustbox{scale=1.1,center}{
\begin{subfigure}{0.6\textwidth}
    \centering
    \includegraphics[width=\linewidth]{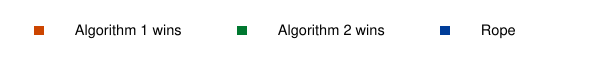}
  \end{subfigure}
  }
  \vspace{-0.75cm}
  \caption{\footnotesize{Bayesian heatmaps for 9 hybrid algorithms evaluated on original CEC-2017 functions and their rotated variants across 4 dims.}}
   \label{fig15}
  \end{figure}

\begin{figure}[H]
  \centering
\begin{adjustwidth}{-1cm}{-2.0cm}
  \begin{subfigure}{0.35\textwidth}
     \includegraphics[height=4cm,width=5.5cm]{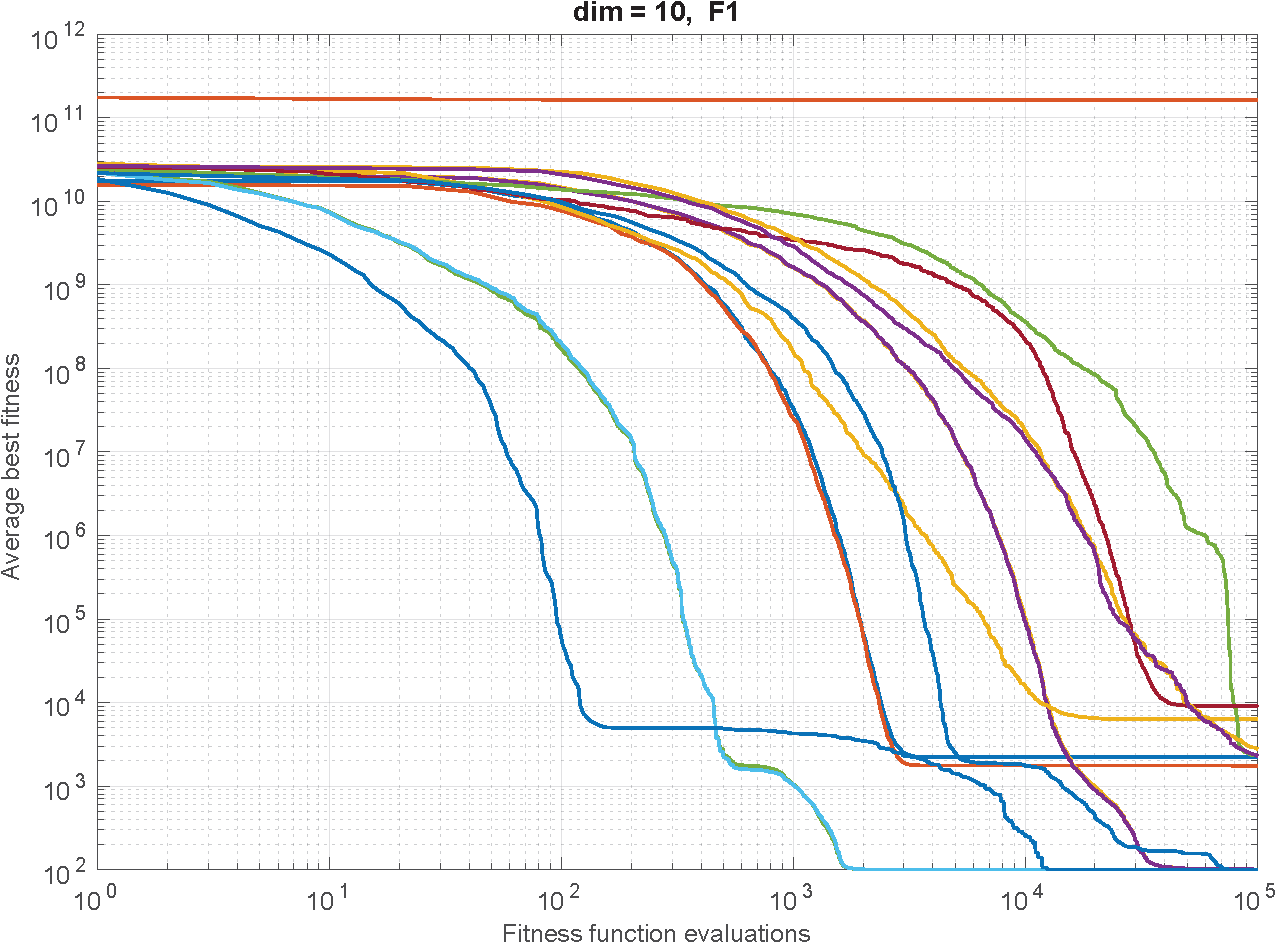}
  \end{subfigure}\hfill
  \begin{subfigure}{0.35\textwidth}
    \includegraphics[height=4cm,width=5.5cm]{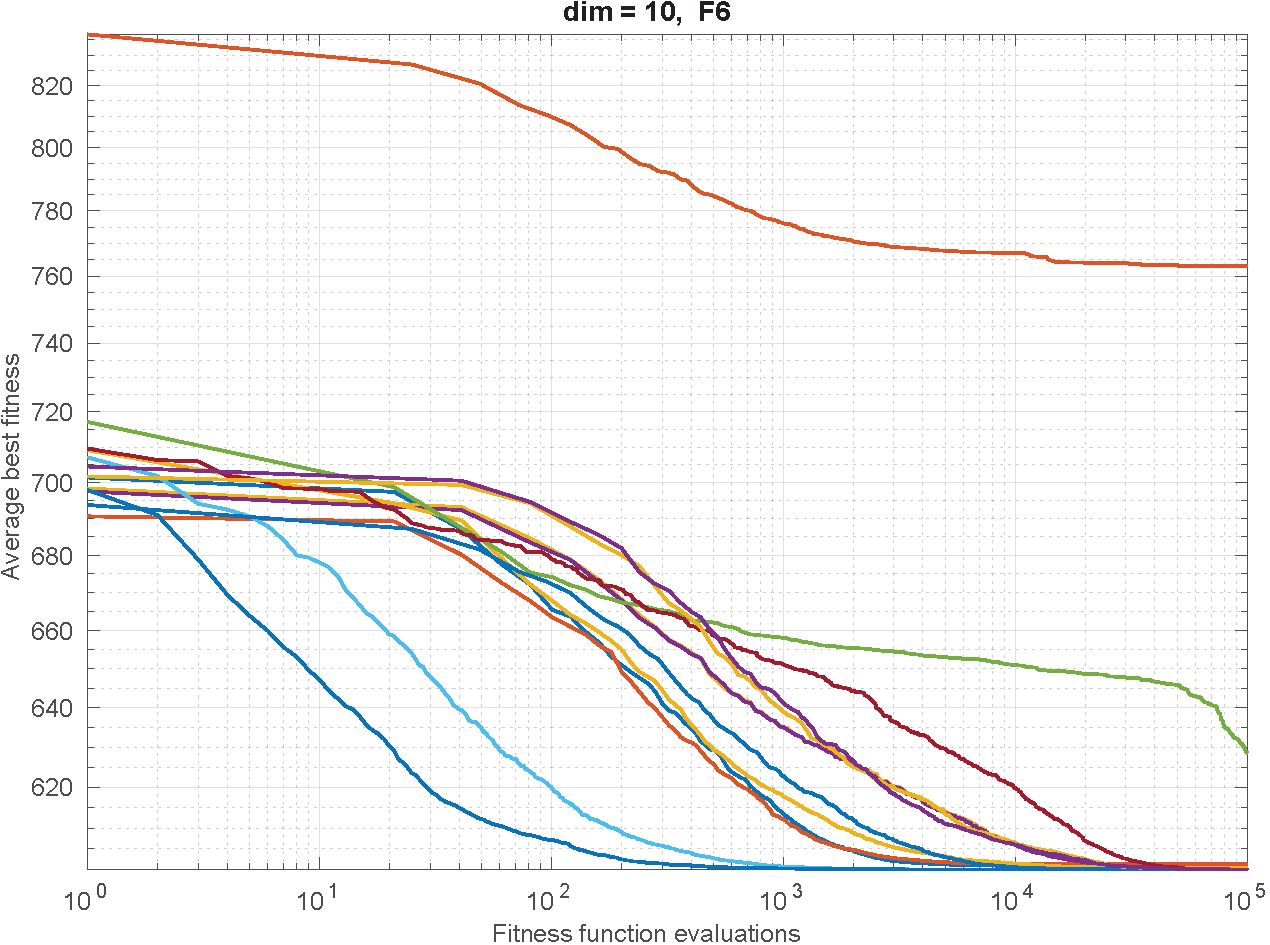}
  \end{subfigure}\hfill
  \begin{subfigure}{0.35\textwidth}
    \includegraphics[height=4cm,width=5.5cm]{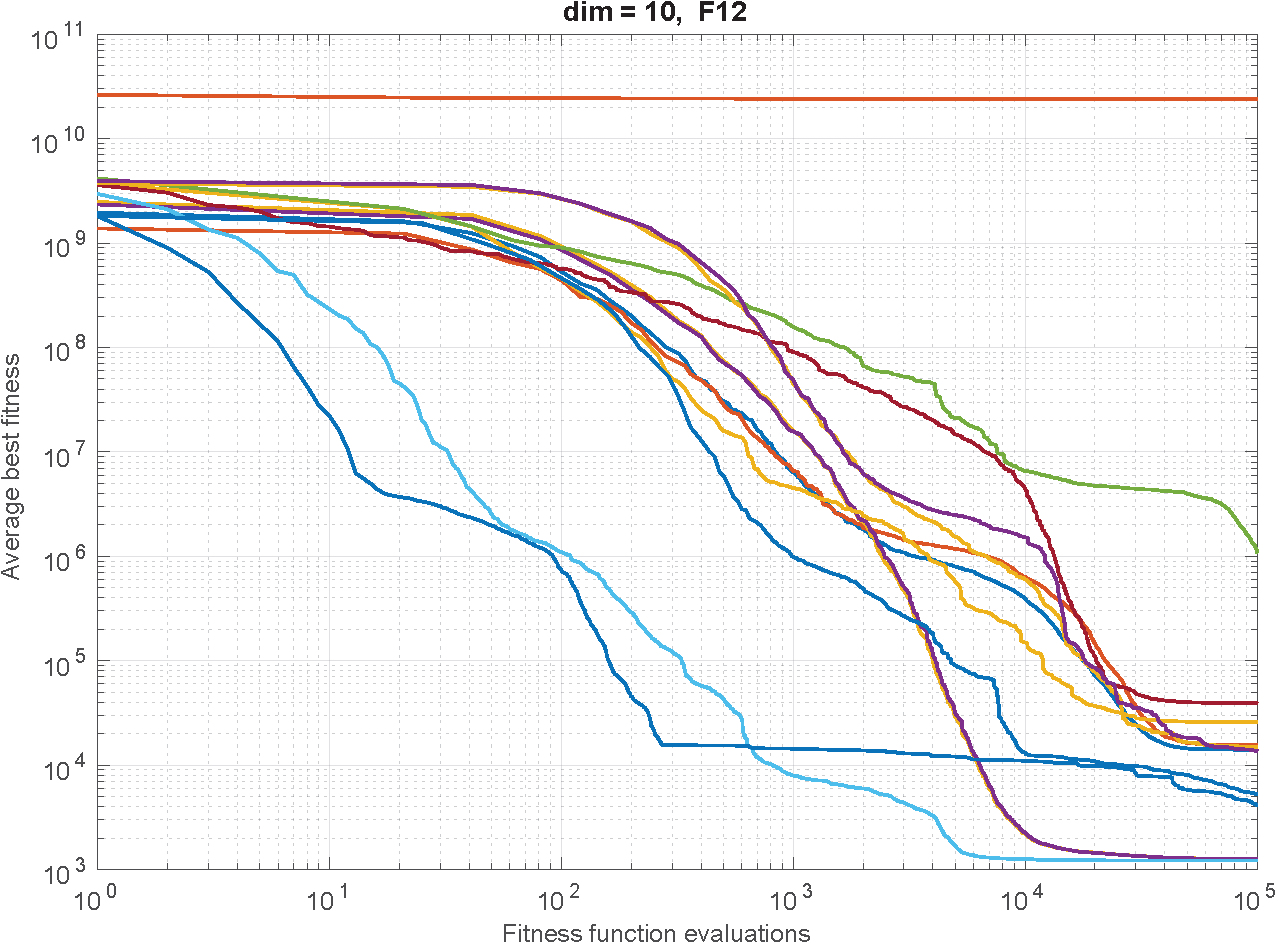}
  \end{subfigure}
 \end{adjustwidth}
\end{figure} 

\begin{figure}[H]
  \centering
\begin{adjustwidth}{-1cm}{-2.0cm}
   \begin{subfigure}{0.35\textwidth}
    \includegraphics[height=4cm,width=5.5cm]{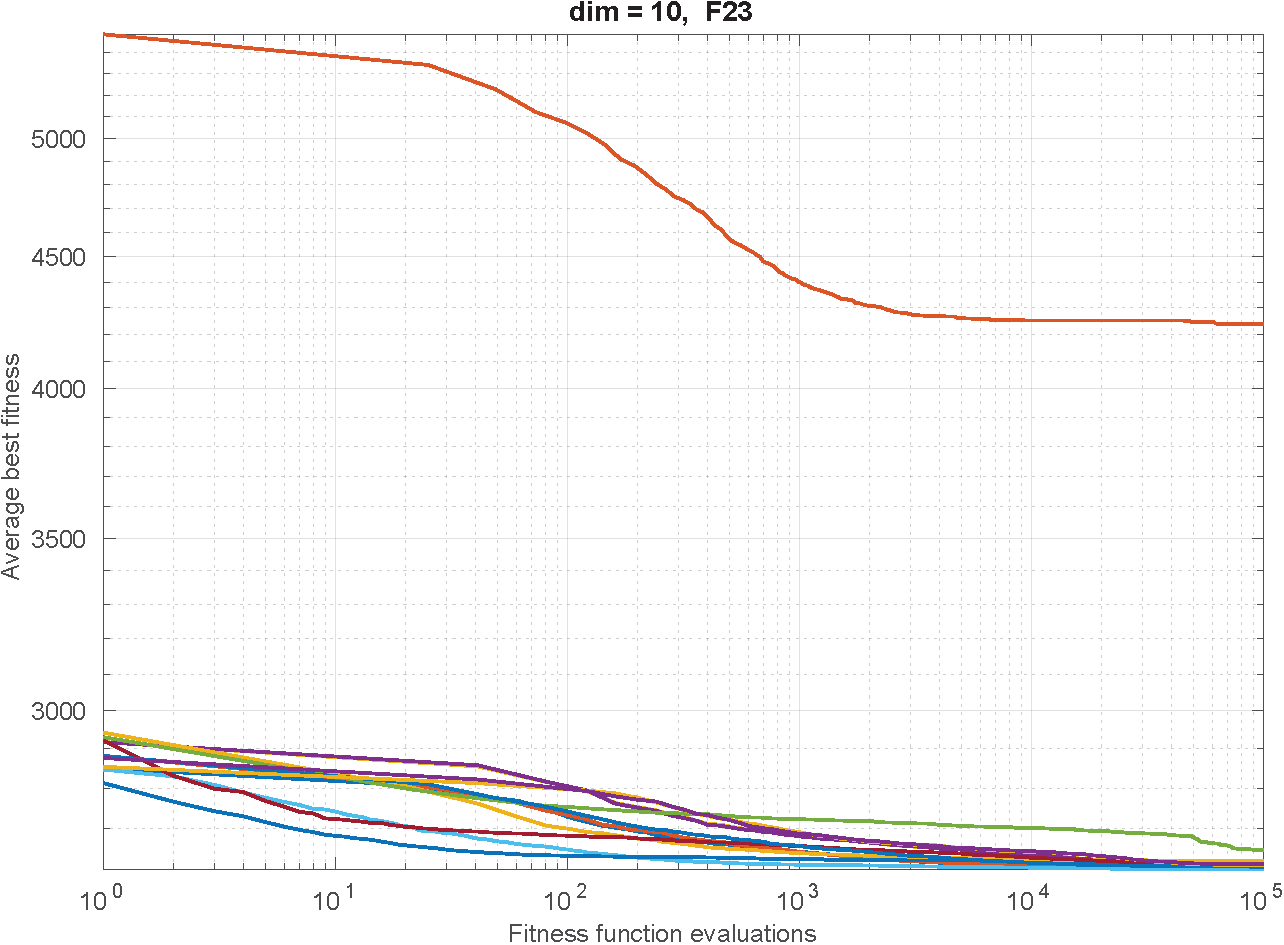}
  \end{subfigure}\hfill
  \begin{subfigure}{0.35\textwidth}
    \includegraphics[height=4cm,width=5.5cm]{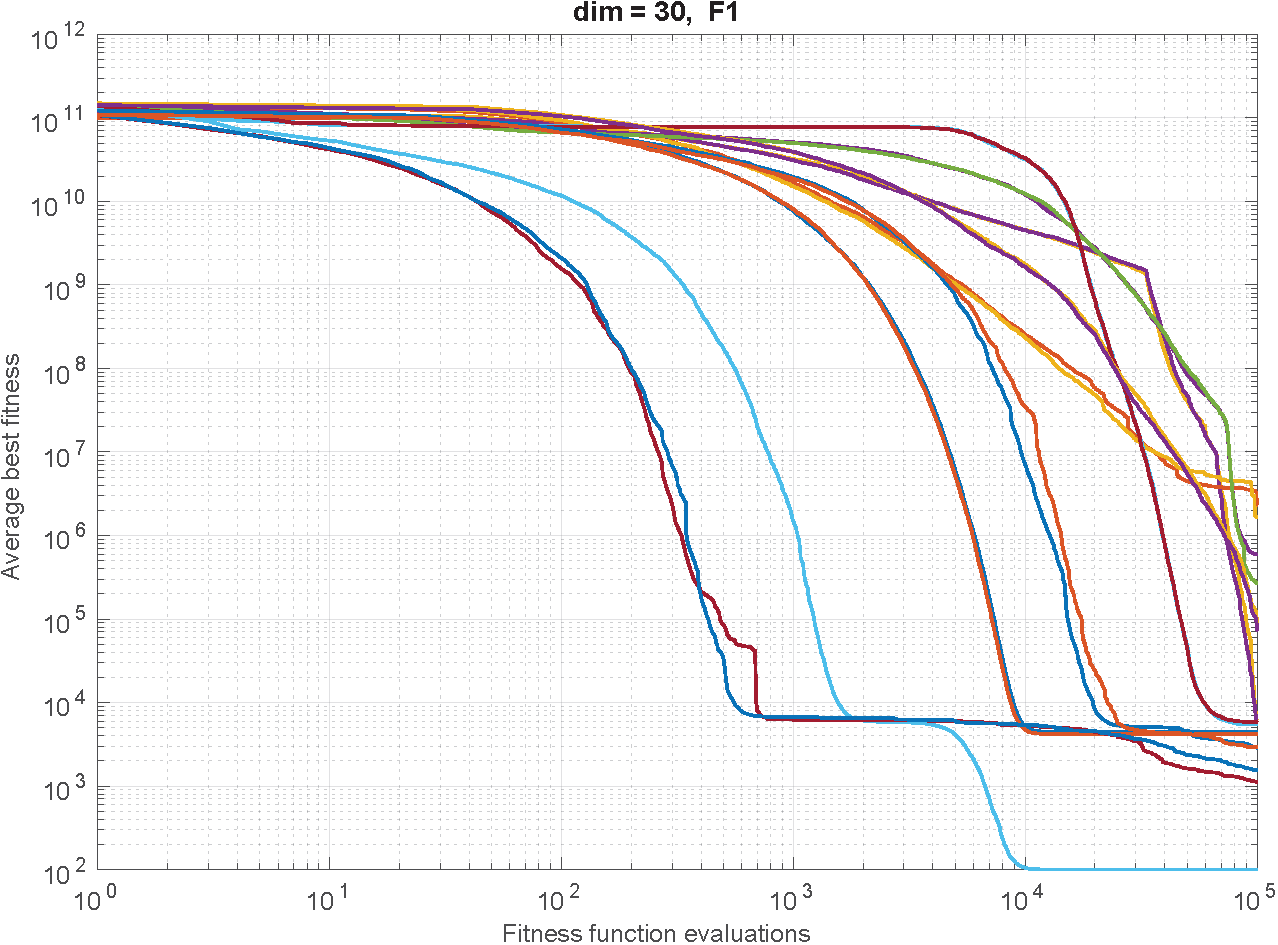}
\end{subfigure}\hfill
  \begin{subfigure}{0.35\textwidth} 
        \includegraphics[height=4cm,width=5.5cm]{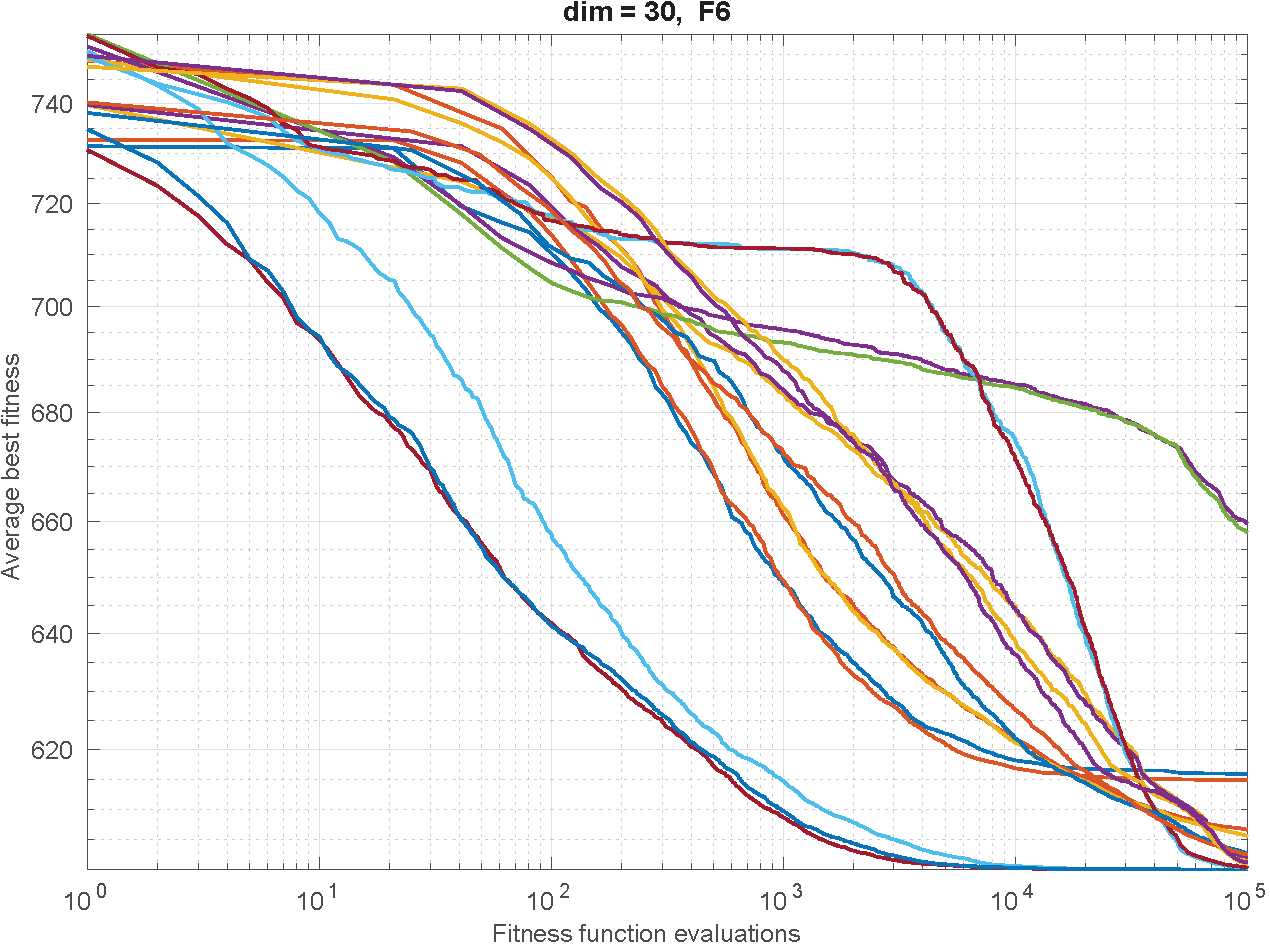}
   \end{subfigure}
   \end{adjustwidth}
\end{figure}

\begin{figure}[H]
  \centering
\begin{adjustwidth}{-1cm}{-2.0cm}
   \begin{subfigure}{0.35\textwidth}
        \includegraphics[height=4cm,width=5.5cm]{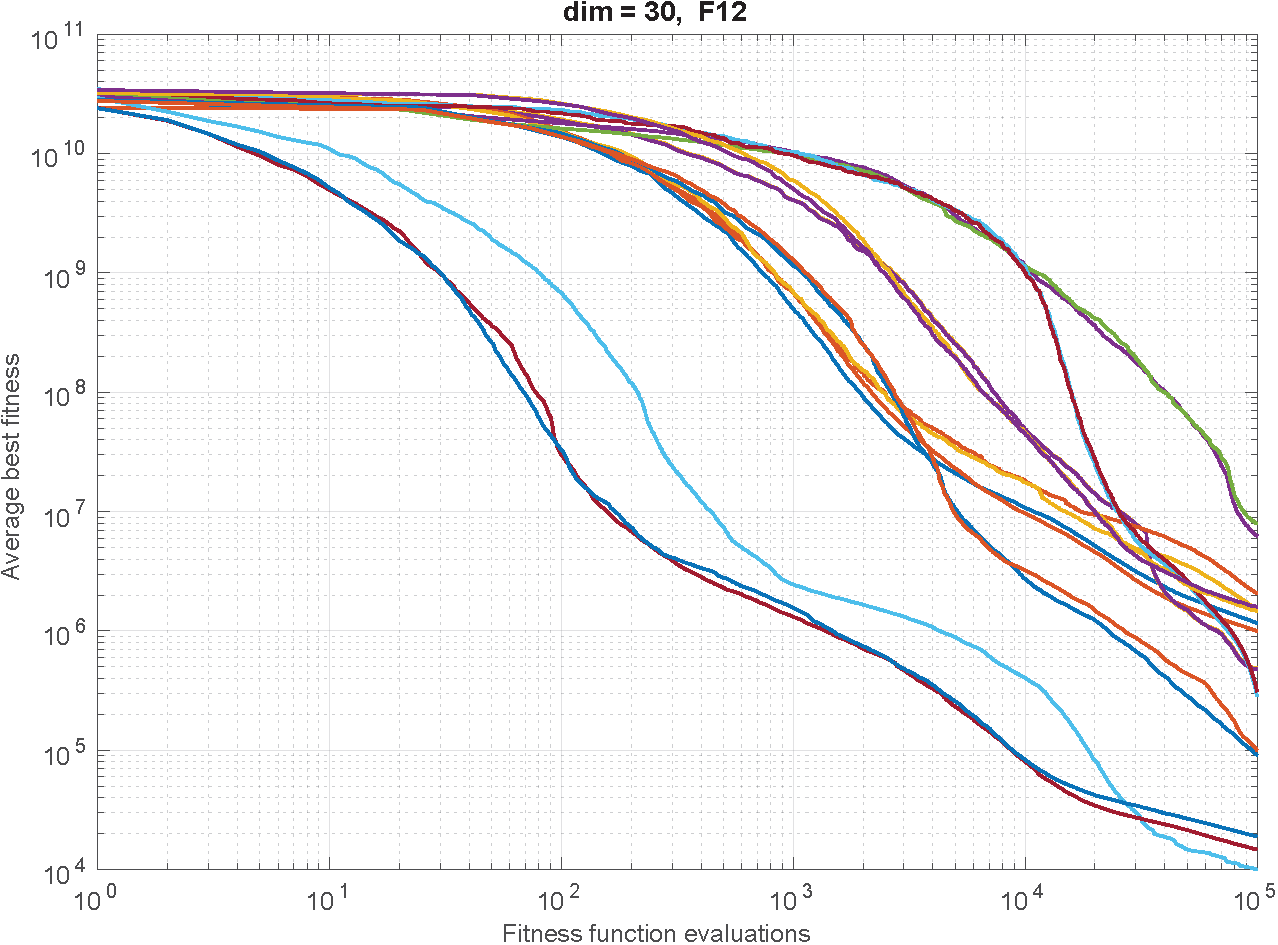}
  \end{subfigure}\hfill
  \begin{subfigure}{0.35\textwidth} 
        \includegraphics[height=4cm,width=5.5cm]{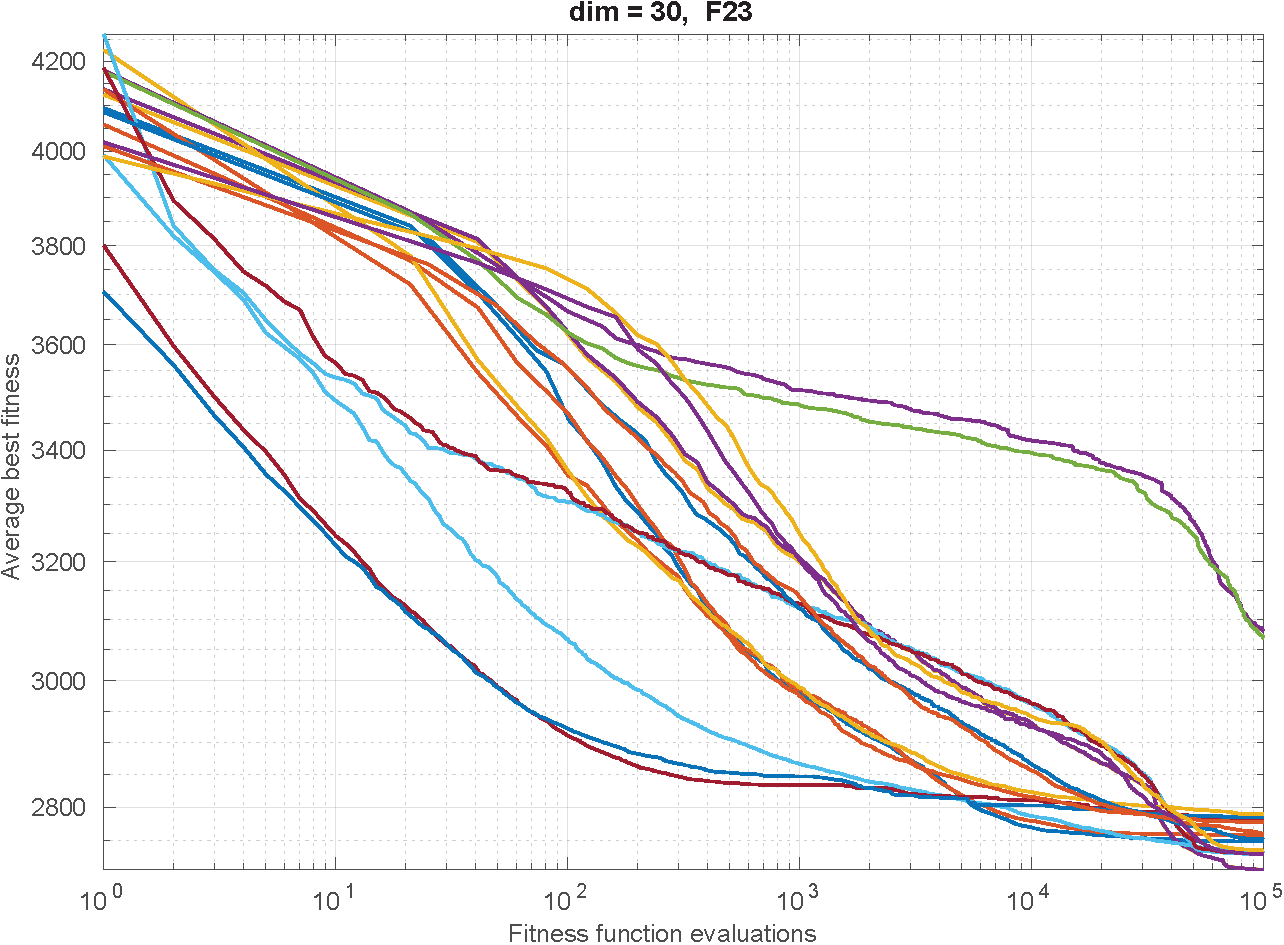}
  \end{subfigure}\hfill
  \begin{subfigure}{0.35\textwidth}
        \includegraphics[height=4cm,width=5.5cm]{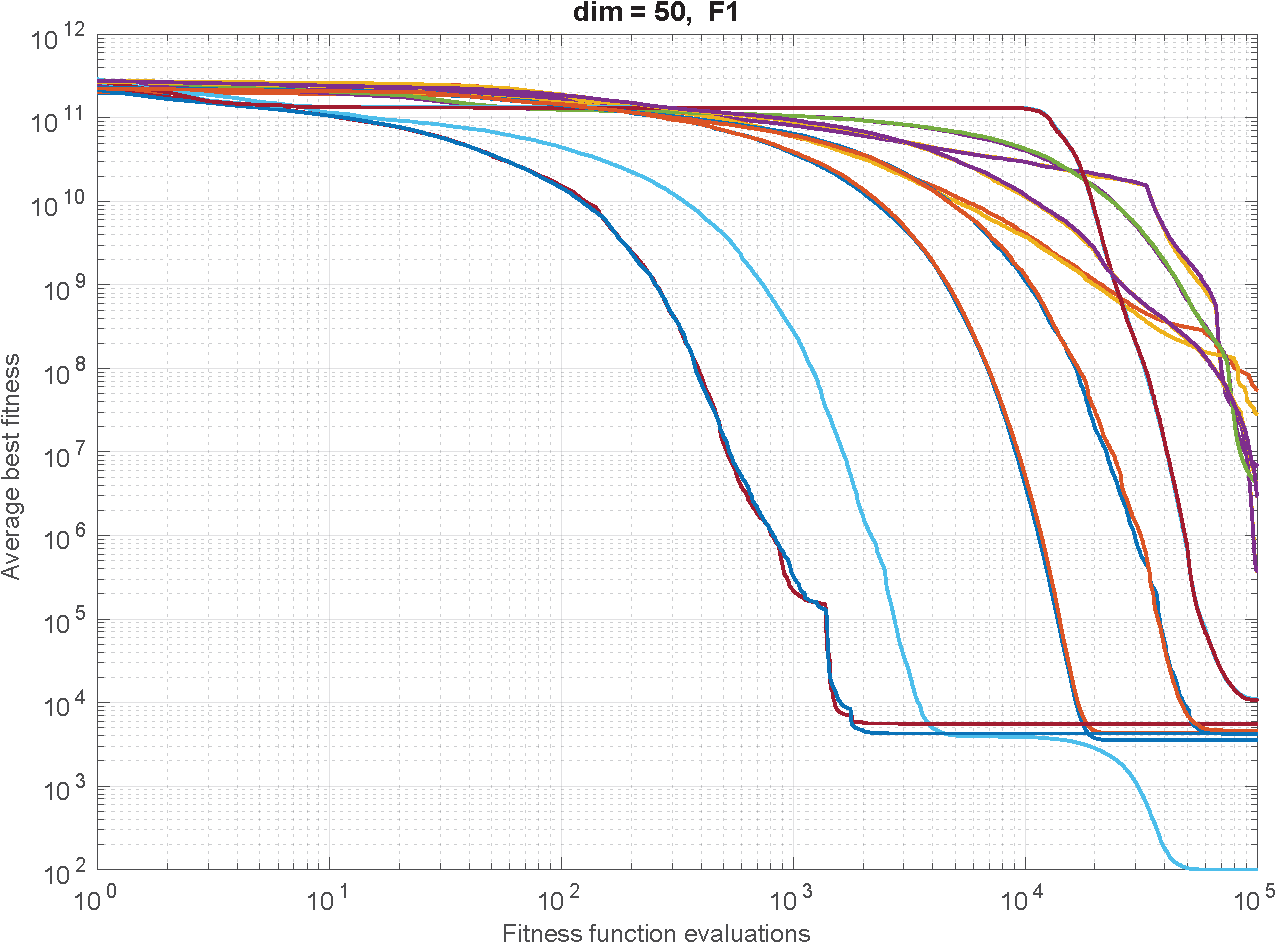}
  \end{subfigure}
  \end{adjustwidth}
\end{figure}

\begin{figure}[H]
  \centering
\begin{adjustwidth}{-1cm}{-2.0cm}
  \begin{subfigure}{0.35\textwidth} 
        \includegraphics[height=4cm,width=5.5cm]{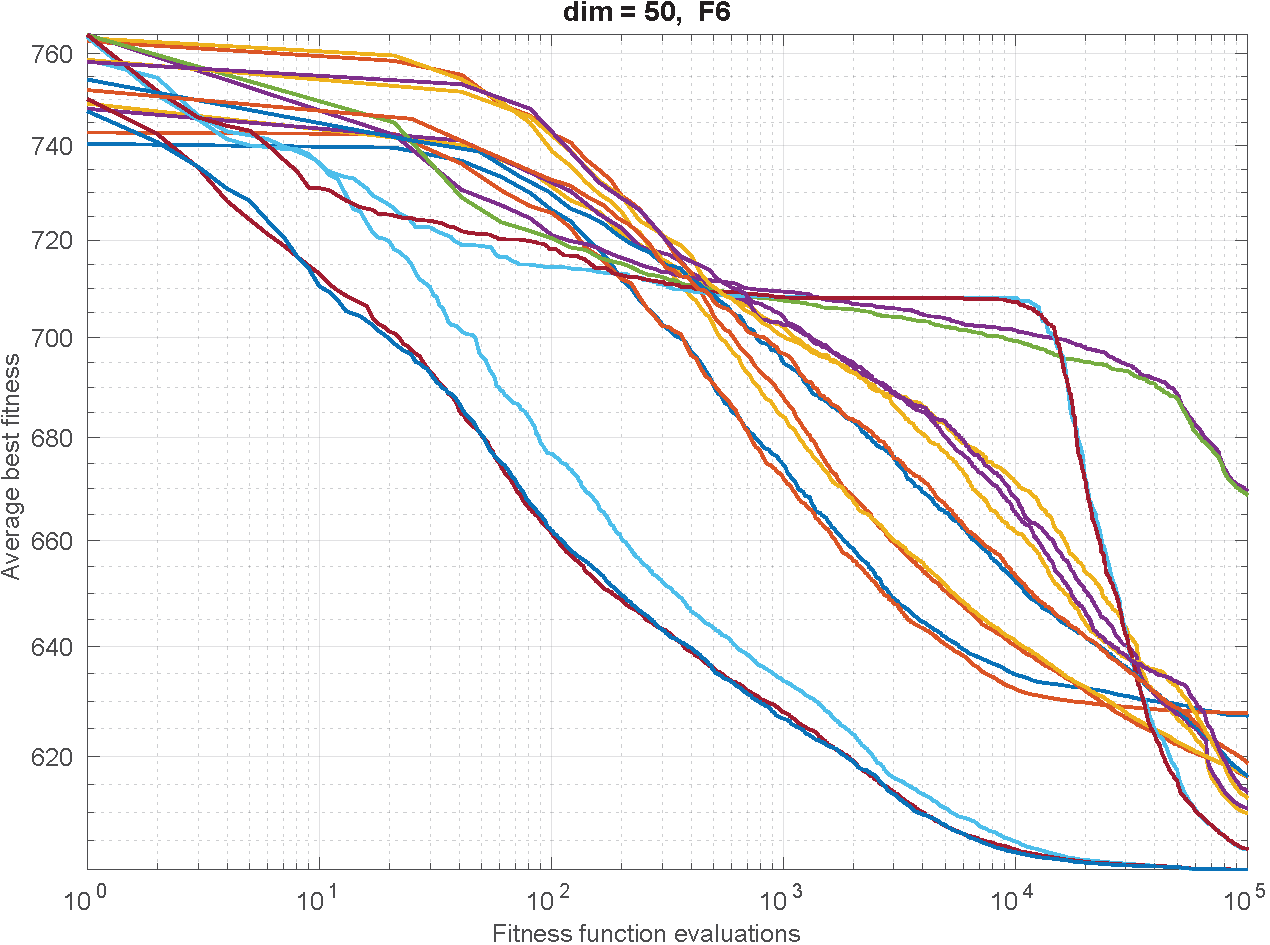}
  \end{subfigure}\hfill
  \begin{subfigure}{0.36\textwidth}
    \includegraphics[height=4cm,width=5.5cm]{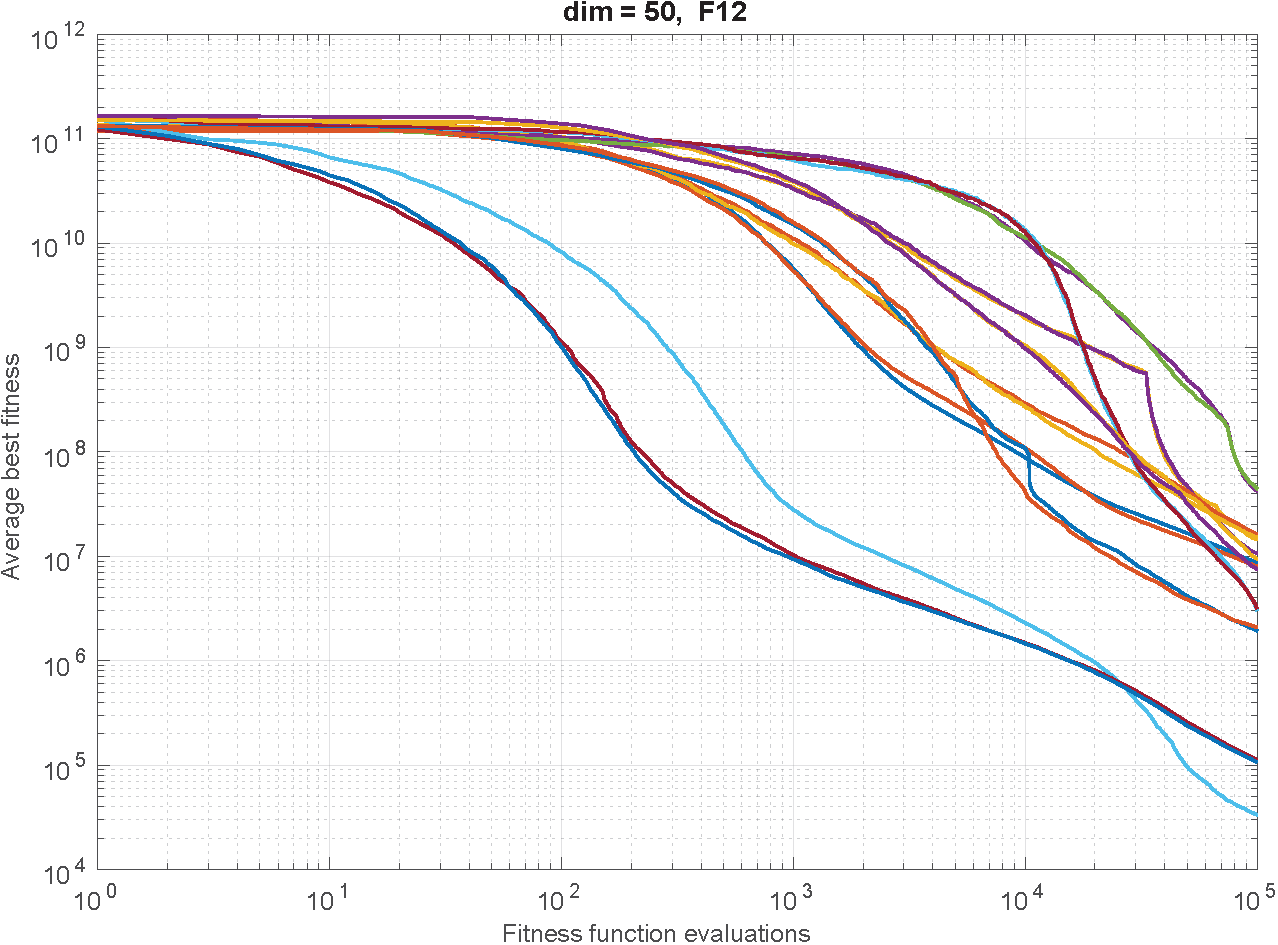}
     \end{subfigure}\hfill
 \begin{subfigure}{0.35\textwidth}
    \includegraphics[height=4cm,width=5.5cm]{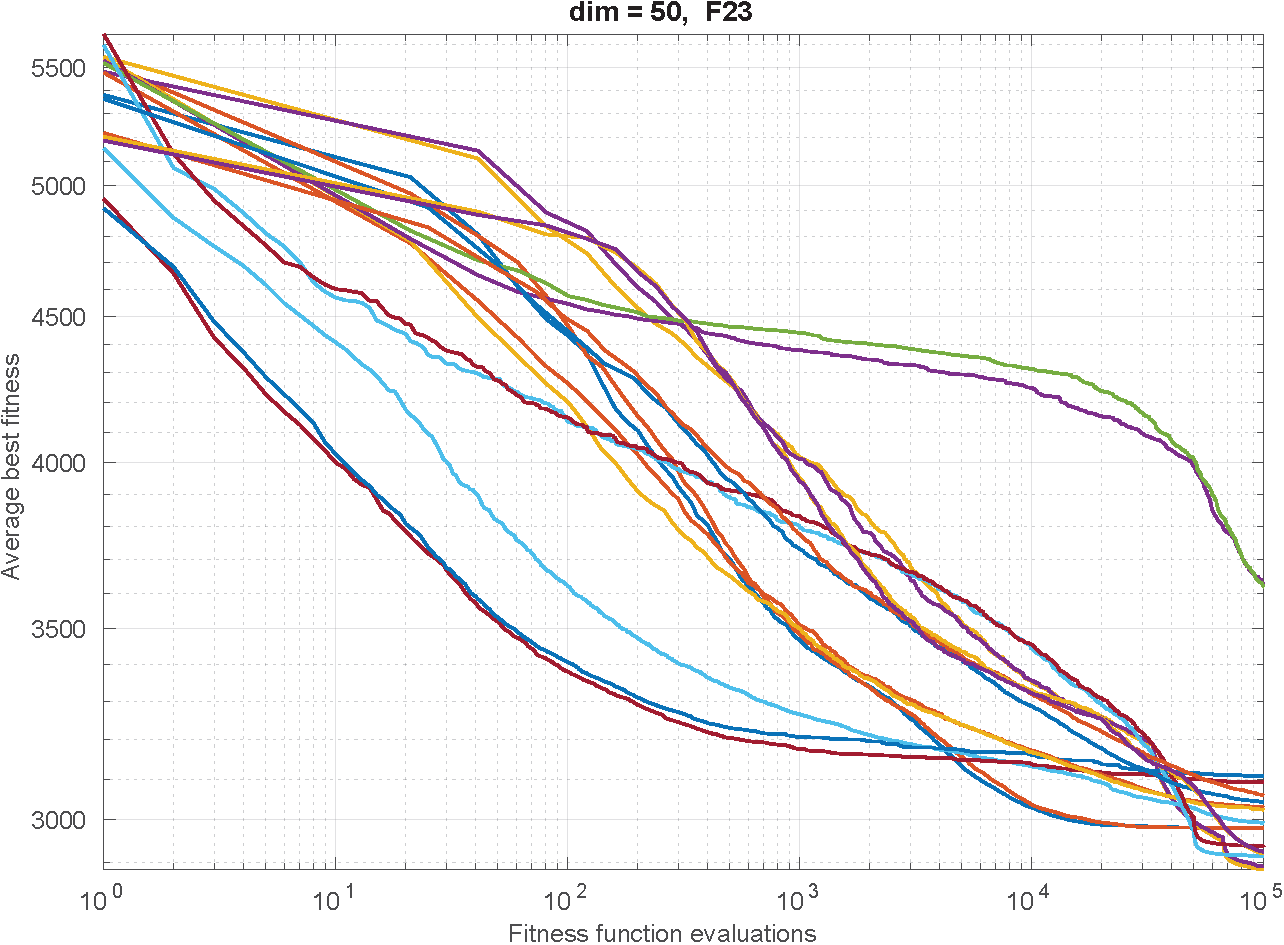}
  \end{subfigure}
\end{adjustwidth}
\end{figure}

\begin{figure}[H]
  \centering
\begin{adjustwidth}{-1cm}{-2.0cm}
  \begin{subfigure}{0.35\textwidth}
    \includegraphics[height=4cm,width=5.5cm]{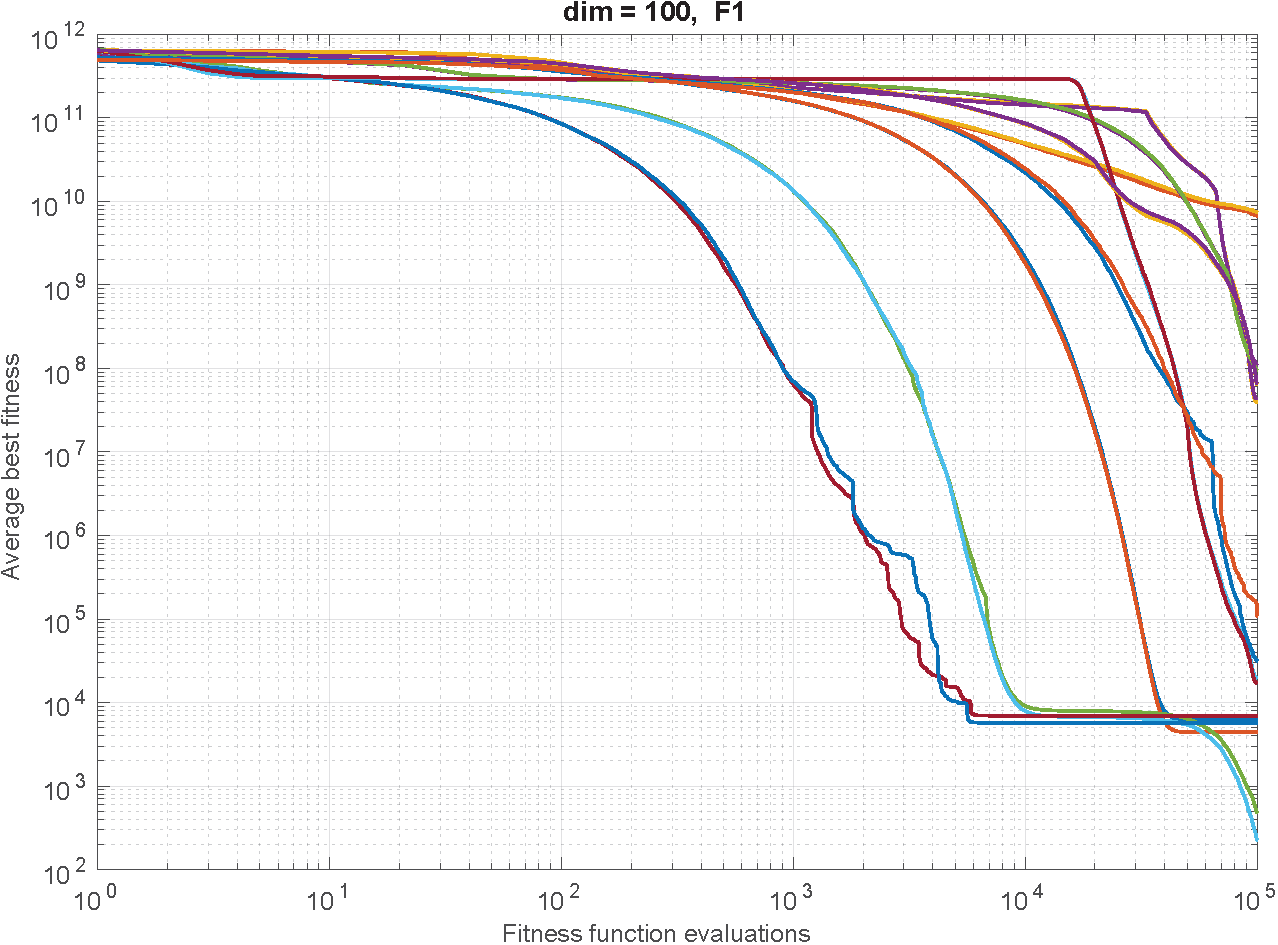}
  \end{subfigure}\hfill
  \begin{subfigure}{0.35\textwidth}
        \includegraphics[height=4cm,width=5.5cm]{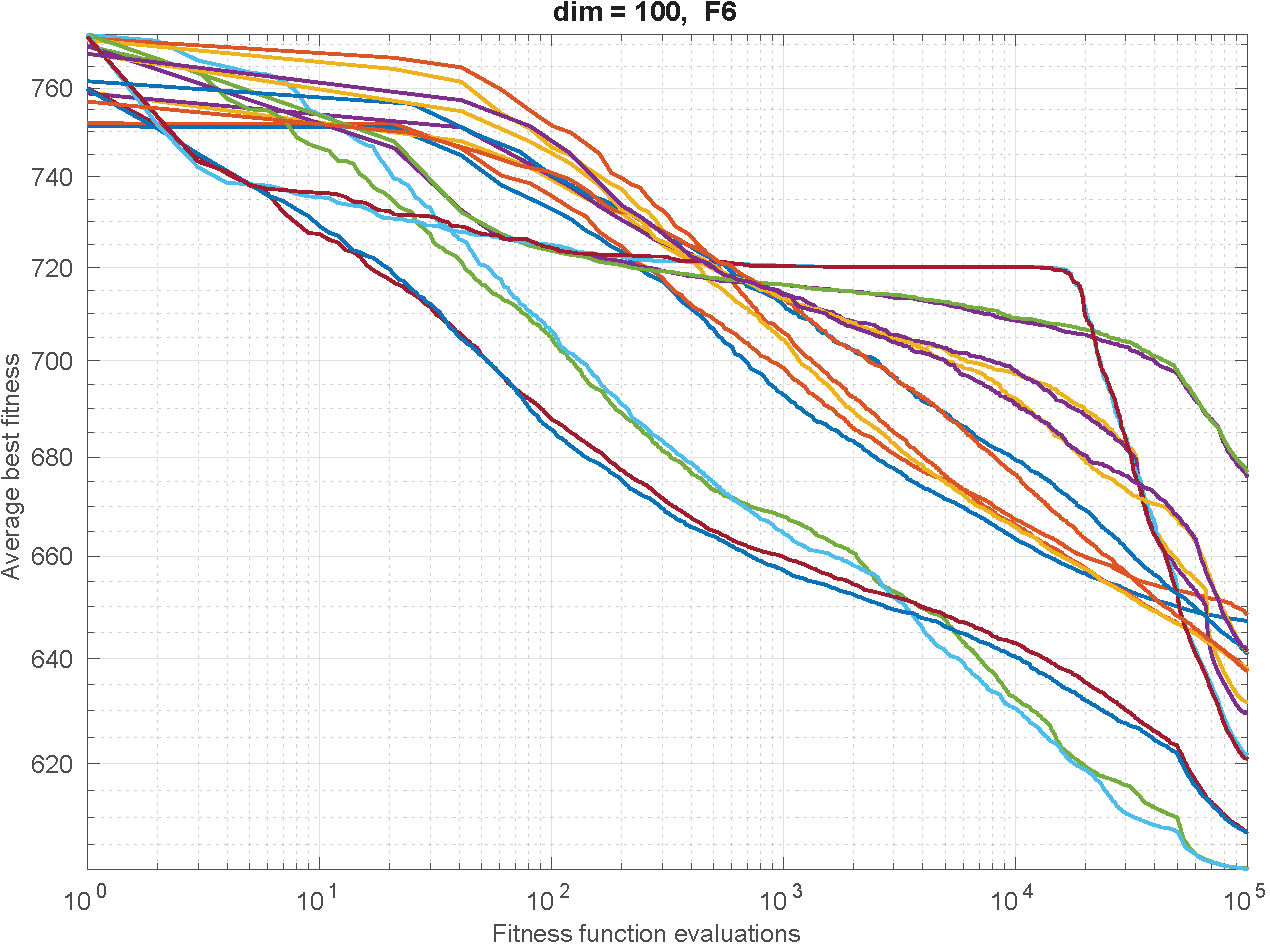}
     \end{subfigure}\hfill
   \begin{subfigure}{0.35\textwidth}
        \includegraphics[height=4cm,width=5.5cm]{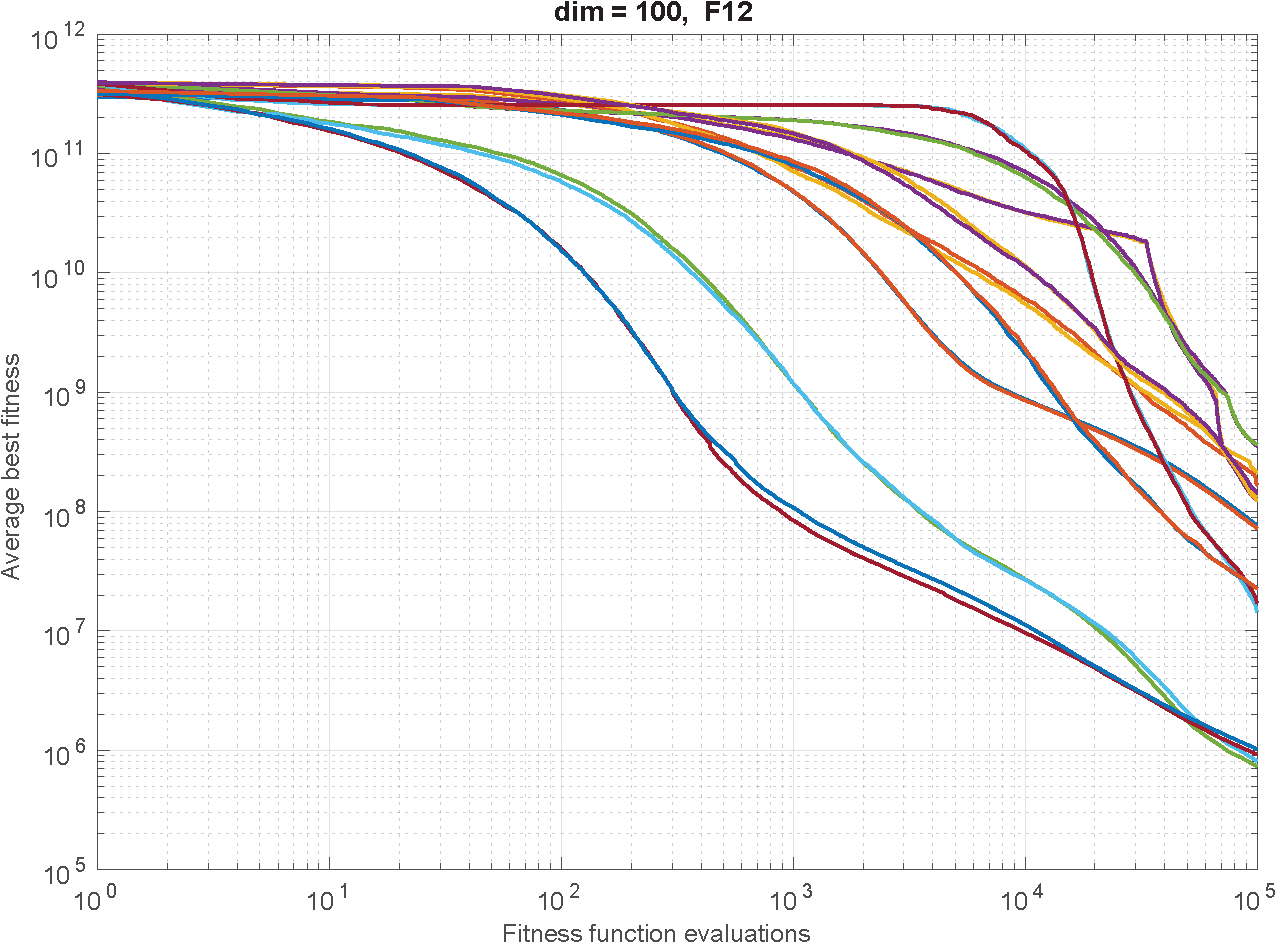}
    \end{subfigure}
    \end{adjustwidth}
\end{figure}

\begin{figure}[H]
 \centering
\begin{adjustwidth}{-1.4cm}{-2.0cm}
\begin{center}
  \begin{subfigure}{0.35\textwidth}
   \centering
        \includegraphics[height=4.5cm,width=6cm]{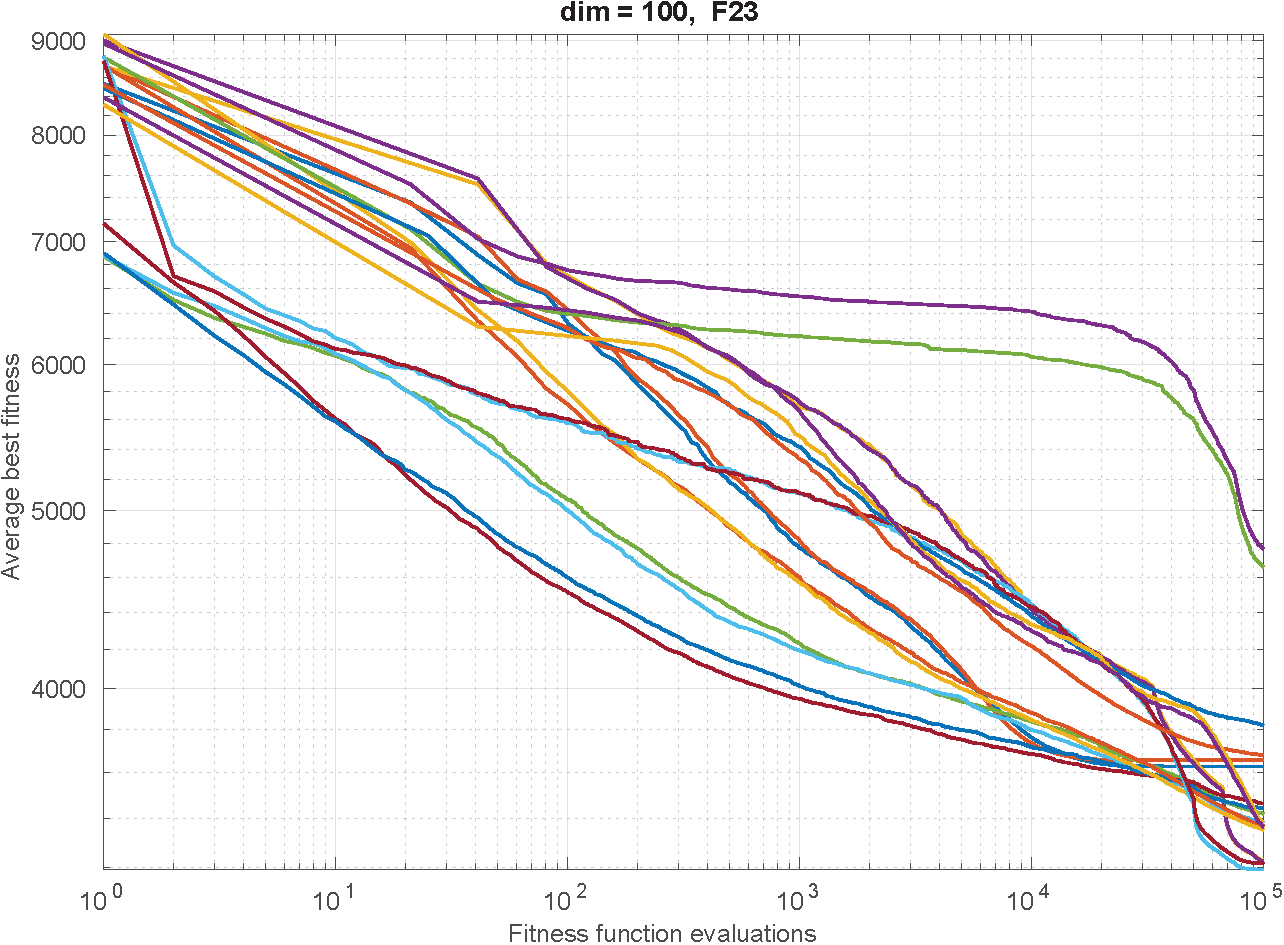}   
  \end{subfigure}
  \end{center}
  \begin{center}
  \begin{subfigure}{0.85\textwidth}
   \centering
    \includegraphics[height=0.75cm, width=\linewidth]{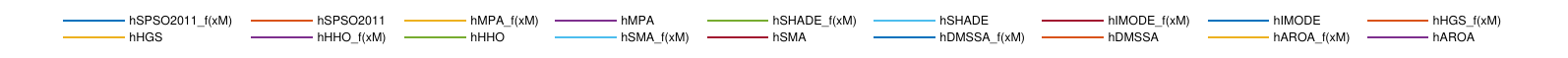}     
  \end{subfigure}
  \end{center}
  \vspace{0.22cm}
  \caption{\small{Convergence trajectories of metaheuristic algorithms on a representative CEC-2017 benchmark function.}}
  \label{fig16}
\end{adjustwidth}
\end{figure}

\vspace{0.25cm}
Conclusions from the Analysis of Results after Rotational Transformation of the Objective Function
The rotational transformation of the decision space, in the form $f(xM)$ where $M$ is an orthonormal matrix, alters the orientation of the coordinate system without modifying the location of the global optimum. Unlike translation ($f(x+8)$), which shifts the optimum, and scaling ($f(5x)$), which changes its amplitude and the shape of the objective landscape, rotation disrupts the separability of the objective function by introducing correlations between variables. This disturbance is particularly challenging for algorithms that assume coordinate independence, such as classical PSO, HHO, HGS, or certain SMA variants 
\cite{4aijk},\cite{6atz}. Importantly, unlike scaling, which slowed convergence even for hybrid methods (cf. f23, $dim = 100$, hIMODE: >400 iterations under $f(5x)$), rotation does not significantly affect optimization dynamics.

Tabular Summary of Data (Tab.\ref{tab4}) shows that rotation influences performance in a moderate but systematic way. For instance, for f6 and $dim = 100$, the hIMODE algorithm maintains result stability - the median remains at $\sim 0.03$, and the standard deviation is 0.09, close to the baseline $f(x)$ version (no statistically significant difference, $p = 0.118$). Conversely, hHHO loses precision - for f12 the mean increases from $1.2 \times 10^3$ to $3.2 \times 10^3$, and the Std exceeds $ 3.1 \times 10^5$, indicating noticeable performance degradation. For comparison, under scaling $f(5x)$, some methods also exhibited deterioration: e.g., for f12 in $dim = 100$, hSMA lost accuracy despite maintaining the median, due to increased variance (Std  $\approx 1.1 \times 10^6$).

Some algorithms, such as hMPA, show moderate shifts - e.g., for f1 $dim = 30$ the median increases from $\sim 0.9$ to $\sim 1.1$, while maintaining relatively low spread (Std < 25). For hSHADE, in both $dim = 30$ and 100, the median and spread remain virtually unchanged (e.g., median = 0.45 for f6 $dim = 30$), indicating high stability. The robustness of DE-based hybrids such as hSHADE and hIMODE is explained by their invariance to geometric transformations, arising from the use of relative differential operators and adaptive parameter structures \cite{2},\cite{5ab}. A similar effect was observed under $f(5x)$, where these algorithms retained relative robustness, though with delayed convergence in higher dimensions (e.g., $dim = 100$) \cite{5ab}.

Boxplots (Fig.\ref{fig13}) for representative functions (f1, f6, f12, f23)  indicate preserved stability among top-performing hybrids. For example, for f6 and $dim = 100$, hSHADE maintains a tight distribution: median $\sim 0.04$, no outliers, and a symmetric box. Similarly, hIMODE shows a median of $\sim 0.03$ with a very narrow spread, reflecting high structural resilience.
In $dim = 10, 30$, and $50$, all three hybrids (hIMODE, hSHADE, hDMSSA) retain consistent central values - the median for f1 remains close to 0.0 with no significant deviations. For classical methods (e.g., hSMA, hHGS, hHHO), rotation causes significant disruptions. For f23 and $dim = 50$, hHHO exhibits a sharp median increase to $1.8 \times 10^4$ with high spread (box height > $1.1 \times 10^4$), indicating instability due to non-separable variable dependencies. hSMA also shows significant fluctuations - outliers exceed $10^5$, which were not present in the baseline version. Interestingly, similar phenomena were observed under scaling - e.g., for f6 $dim = 50$, hHGS produced many outliers exceeding $2 \times 10^5$, despite relatively stable medians.

The Friedman test (Fig.\ref{fig14}) for rotation reveals statistically significant differences ($p \ll 0.05$) among algorithms for each dimension. For $dim = 100$, $p = 0.0013$ was obtained. Critical Difference (CD $\approx $ 4.9) diagrams show that hIMODE and hSHADE remain at the top, with average ranks of $\sim $2.3 for $dim = 50$ and $\sim $ 2.1 for $dim = 100$.
Unlike translation cases, where classical methods lost ranking positions dramatically (e.g., hHHO dropped from rank 8 to >12), here the reshuffling is milder. hMPA maintains a position of $\sim $ 6.2, and hDMSSA $\sim$ 3.4. This shows that rotation is less destabilizing to rankings than other transformations - consistent with hypotheses regarding the landscape structure's influence on exploration processes \cite{0BB},\cite{6atz}. Under $f(5x)$, this effect was more pronounced: e.g., in $dim = 100$, hDMSSA improved its rank from 4.6 to 3.1, highlighting the selective nature of that transformation favoring certain exploration strategies.

Dominance matrices (Fig.\ref{fig15}) show clear equilibrium among leading hybrids. For example, P(hIMODE > hSHADE) $\approx $ 0.51 ($dim = 100$), within the ROPE. Meanwhile, the probability of hIMODE outperforming hHHO is > 0.98, and over hSMA $\sim$ 0.96 - indicating a stable dominance structure of superior methods over weaker ones.
Compared to $f(x+8)$, where dominance was strongly polarized (e.g., P(hSHADE > hHHO) > 0.99), rotation leads to a more "diffused" dominance structure, reducing the number of unequivocal victories while preserving overall ranking order. For scaling ($f(5x)$), the reverse situation was observed: hybrid dominance over classical methods was even stronger (e.g., P(hIMODE > hHHO) > 0.995), but with greater inter-dimensional variability, indicating less predictability of that transformation.

For most functions (f1,f6), top-performing algorithms converge  (Fig.\ref{fig16}) almost identically as in the baseline version. hIMODE, hSHADE, and hDMSSA reach minima below 200 iterations, maintaining convergence speed. For f12, hMPA shows a delay (reaching minimum about $\sim $ 150 iterations later than in $f(x)$, suggesting limited adaptability to structural disturbances in variable space.
In f23 for $dim=100$, classical methods such as hHHO and hHGS show marked slowdown: trajectories are nearly flat, indicating loss of explorability. High variable correlation after rotation hampers these algorithms' ability to guide search effectively, consistent with their known non-invariance to orthogonal transformations \cite{4aijk}. For comparison, scaling $f(5x)$ caused even more pronounced slowdown - e.g., hHHO in f23 and $dim = 100$ required over 900 iterations to reach values <$1.5 \times 10^4$, which was not observed under rotation.

The rotational transformation $f(xM)$ primarily affects algorithms by disrupting objective function separability, without altering the optimum. Unlike translation and scaling, rotation does not change the amplitude or location of the minimum, making it a subtler yet fundamental test of algorithmic structural resilience.
Hybrid algorithms based on differential and adaptive operators (e.g., hSHADE, hIMODE, hDMSSA) exhibit the highest robustness - maintaining medians, low variance, and ranking dominance regardless of dimensionality. These findings confirm earlier studies on the invariance of DE and CMA-ES operators \cite{2},\cite{6atz}.
Classical algorithms, especially those based on particle trajectories (e.g., hHHO, hSMA), lose stability on correlated functions - showing increased spread, delayed convergence, and ranking shifts. This results from their dependence on axis-aligned heuristics, which lose effectiveness after rotation \cite{0BB},\cite{4aijk}.
Rotation - though less destructive than translation and scaling - constitutes a critical test for algorithms with serious application aspirations. The results clearly indicate that algorithms built around invariant operators (especially DE) demonstrate greater universality in function landscapes with correlations.

\subsection{The Impact of Objective Function Value Shift on the Performance of Hybrid Algorithms}
\vspace{0.3cm}
The additive transformation $f(x)+6$ introduces a uniform vertical shift in the objective landscape without affecting its topological features. While theoretically performance-neutral, its empirical validation is essential, particularly in the context of prior transformations such as $f(x+8)$, $f(5x)$, 
and $f(xM)$, which impacted convergence dynamics and algorithmic rankings. This section presents a detailed comparative evaluation of algorithm behavior under $f(x)+6$, highlighting its neutrality and drawing contrast with earlier observed distortions.

\FloatBarrier

\scriptsize
\setlength{\tabcolsep}{3pt}
\begin{table}[H]
\centering
\begin{adjustwidth}{-1.5cm}{-2.6cm}
\caption{\footnotesize{Statistical results for 9 hybrid algorithms evaluated on CEC-2017 functions\\ 
and their shifted-value variants across 4 dims.}}
\scalebox{0.67}{
\renewcommand{\arraystretch}{0.9}
\begin{tabular}{@{}lrrrrrrrrrrrrrrrr@{}}
\toprule
\multirow{2}{*}{Algorithm} & \multicolumn{7}{c}{dim=10} & \multicolumn{7}{c}{dim=30} \\
\cmidrule(lr){2-8} \cmidrule(lr){9-15}
 & Avg & Med & Std & Sum Rank & Mean Rank & +/- & p-value & Avg & Med & Std & Sum Rank & Mean Rank & +/- & p-value \\
\midrule
hSPSO2011\_f(x)+6  & 8.2E+03 & 1.9E+03 & 1.3E+04 & 375 & 12.9 & 105/316 & 6.8E-02 & 6.7E+04 & 2.9E+03 & 4.8E+04 & 369 & 12.7 & 106/304 & 7.2E-02 \\
hSPSO2011  & 1.6E+04 & 1.9E+03 & 2.0E+04 & 328 & 11.3 & 176/258 & 4.2E-02 & 5.9E+04 & 2.8E+03 & 3.9E+04 & 346 & 11.9 & 129/290 & 6.9E-02 \\
hMPA\_f(x)+6  & 1.7E+03 & 1.6E+03 & 2.3E+01 & 266 & 9.2 & 202/243 & 4.3E-02 & 2.6E+04 & 2.3E+03 & 3.0E+04 & 272 & 9.4 & 208/214 & 5.5E-02 \\
hMPA  & 2.6E+03 & 1.6E+03 & 5.2E+03 & 214 & 7.4 & 300/181 & 1.2E-02 & 1.9E+04 & 2.3E+03 & 1.7E+04 & 257 & 8.9 & 228/198 & 5.0E-02 \\
hSHADE\_f(x)+6  & 4.5E-01 & 1.0E+00 & 4.8E+01 & 92 & 3.2 & 379/7 & 1.5E-01 & 1.7E+00 & 1.9E+00 & 5.2E+01 & 116 & 4.0 & 357/13 & 1.4E-01 \\
hSHADE  & -6.1E-01 & 1.4E+00 & 4.7E+01 & 74 & 2.6 & 381/4 & 1.6E-01 & -2.2E-01 & 4.5E-01 & 5.2E+01 & 69 & 2.4 & 363/5 & 1.4E-01 \\
hIMODE\_f(x)+6  & -3.3E-01 & 1.0E+00 & 5.1E+01 & 91.5 & 3.2 & 381/2 & 1.6E-01 & 1.9E+00 & 2.0E+00 & 5.5E+01 & 122 & 4.2 & 356/13 & 1.4E-01 \\
hIMODE  & -1.5E+00 & 1.0E+00 & 5.1E+01 & 65 & 2.2 & 386/1 & 1.5E-01 & -2.3E-01 & 1.3E+00 & 5.5E+01 & 85 & 2.9 & 372/5 & 1.3E-01 \\
hHGS\_f(x)+6  & 9.5E+03 & 2.5E+03 & 1.2E+04 & 452 & 15.6 & 50/378 & 5.7E-02 & 1.2E+05 & 3.2E+03 & 1.1E+05 & 419 & 14.4 & 82/348 & 5.0E-02 \\
hHGS  & 1.1E+04 & 2.6E+03 & 1.4E+04 & 413 & 14.2 & 98/323 & 6.2E-02 & 1.4E+05 & 3.2E+03 & 3.3E+05 & 408 & 14.1 & 92/332 & 5.6E-02 \\
hHHO\_f(x)+6  & 5.1E+04 & 2.3E+03 & 4.3E+04 & 493 & 17.0 & 8/442 & 3.8E-02 & 3.4E+05 & 3.3E+03 & 2.7E+05 & 501 & 17.3 & 9/447 & 3.7E-02 \\
hHHO  & 5.8E+04 & 2.3E+03 & 7.1E+04 & 474 & 16.3 & 28/407 & 4.7E-02 & 3.5E+05 & 3.3E+03 & 3.1E+05 & 488 & 16.8 & 10/442 & 3.9E-02 \\
hSMA\_f(x)+6  & 4.5E+03 & 1.9E+03 & 5.4E+03 & 353 & 12.2 & 133/311 & 4.4E-02 & 2.0E+04 & 2.9E+03 & 1.1E+04 & 326 & 11.2 & 141/251 & 8.1E-02 \\
hSMA  & 3.6E+03 & 1.9E+03 & 1.1E+03 & 287 & 9.9 & 210/230 & 4.4E-02 & 2.1E+04 & 2.9E+03 & 9.6E+03 & 299 & 10.3 & 172/233 & 8.0E-02 \\
hDMSSA\_f(x)+6  & 8.5E-01 & 3.8E+00 & 4.9E+01 & 114 & 3.9 & 379/7 & 1.5E-01 & 3.0E+00 & 3.8E+00 & 5.2E+01 & 138 & 4.8 & 351/32 & 1.2E-01 \\
hDMSSA  & 9.7E+01 & 9.2E+01 & 5.3E+00 & 174 & 6.0 & 348/145 & 3.6E-20 & 3.0E-01 & 2.2E-01 & 5.2E+01 & 80 & 2.8 & 365/8 & 1.2E-01 \\
hAROA\_f(x)+6  & 5.7E+03 & 1.7E+03 & 8.6E+03 & 371 & 12.8 & 103/335 & 5.1E-02 & 6.9E+04 & 2.8E+03 & 4.6E+04 & 339 & 11.7 & 137/255 & 7.6E-02 \\
hAROA  & 5.4E+03 & 1.7E+03 & 9.2E+03 & 327 & 11.3 & 172/249 & 6.6E-02 & 6.7E+04 & 2.9E+03 & 3.5E+04 & 324 & 11.2 & 156/244 & 8.5E-02 \\
\bottomrule
\end{tabular}
}
\vspace{1.1mm}
\scalebox{0.67}{
\renewcommand{\arraystretch}{0.9}
\begin{tabular}{@{}lccccccc@{\hspace{0.5cm}}ccccccc@{}}
\toprule
\multirow{2}{*}{Algorithm} & \multicolumn{7}{c}{dim=50} & \multicolumn{7}{c}{dim=100} \\
\cmidrule(lr){2-8} \cmidrule(lr){9-15}
 & Avg & Med & Std & Sum Rank & Mean Rank & +/- & p-value & Avg & Med & Std & Sum Rank & Mean Rank & +/- & p-value \\
\midrule
hSPSO2011\_f(x)+6  & 8.3E+05 & 3.3E+03 & 2.3E+05 & 375 & 12.9 & 108/313 & 6.6E-02 & 3.2E+06 & 6.0E+03 & 1.5E+06 & 390 & 13.4 & 105/339 & 4.5E-02 \\
hSPSO2011  & 8.8E+05 & 3.3E+03 & 2.4E+05 & 369 & 12.7 & 105/307 & 6.5E-02 & 3.1E+06 & 6.0E+03 & 1.0E+06 & 381 & 13.1 & 111/328 & 5.0E-02 \\
hMPA\_f(x)+6  & 4.5E+05 & 3.2E+03 & 3.2E+05 & 289 & 10.0 & 197/221 & 6.2E-02 & 5.4E+06 & 5.3E+03 & 2.8E+06 & 315 & 10.9 & 173/253 & 6.0E-02 \\
hMPA  & 4.3E+05 & 3.2E+03 & 3.3E+05 & 279 & 9.6 & 199/217 & 5.6E-02 & 6.0E+06 & 5.3E+03 & 2.8E+06 & 310 & 10.7 & 172/244 & 6.4E-02 \\
hSHADE\_f(x)+6  & 3.4E+00 & 2.7E+00 & 5.3E+01 & 131 & 4.5 & 358/28 & 1.1E-01 & 2.4E+00 & 2.7E+00 & 5.2E+01 & 139 & 4.8 & 358/53 & 8.7E-02 \\
hSHADE  & 1.1E+00 & 1.9E+00 & 5.3E+01 & 70 & 2.4 & 376/13 & 1.1E-01 & -6.2E-01 & -7.9E-02 & 5.2E+01 & 63 & 2.2 & 399/10 & 9.5E-02 \\
hIMODE\_f(x)+6  & 3.2E+00 & 2.9E+00 & 5.6E+01 & 129 & 4.4 & 363/23 & 1.2E-01 & 2.2E+00 & 2.9E+00 & 5.5E+01 & 129 & 4.4 & 360/44 & 9.3E-02 \\
hIMODE  & 8.0E-01 & 5.1E-01 & 5.6E+01 & 69 & 2.4 & 379/13 & 1.1E-01 & -8.9E-01 & -1.1E-01 & 5.5E+01 & 57 & 2.0 & 400/8 & 9.5E-02 \\
hHGS\_f(x)+6  & 4.6E+06 & 3.8E+03 & 1.1E+07 & 420 & 14.5 & 81/340 & 5.7E-02 & 2.6E+08 & 6.5E+03 & 1.1E+08 & 413 & 14.2 & 80/344 & 6.4E-02 \\
hHGS  & 1.7E+06 & 4.0E+03 & 1.6E+06 & 410 & 14.1 & 82/340 & 5.8E-02 & 2.7E+08 & 7.0E+03 & 1.2E+08 & 417 & 14.4 & 88/338 & 6.0E-02 \\
hHHO\_f(x)+6  & 2.1E+06 & 4.0E+03 & 1.3E+06 & 494 & 17.0 & 12/436 & 4.4E-02 & 1.7E+07 & 7.2E+03 & 6.8E+06 & 470 & 16.2 & 23/425 & 4.2E-02 \\
hHHO  & 2.2E+06 & 4.0E+03 & 1.2E+06 & 489 & 16.9 & 12/439 & 4.8E-02 & 1.7E+07 & 7.1E+03 & 8.4E+06 & 477 & 16.4 & 26/426 & 4.4E-02 \\
hSMA\_f(x)+6  & 1.6E+05 & 3.3E+03 & 7.5E+04 & 288 & 9.9 & 194/227 & 6.5E-02 & 5.9E+05 & 5.1E+03 & 2.2E+05 & 248 & 8.6 & 245/194 & 4.6E-02 \\
hSMA  & 1.7E+05 & 3.3E+03 & 7.6E+04 & 285 & 9.8 & 197/222 & 7.2E-02 & 6.2E+05 & 4.9E+03 & 2.6E+05 & 240 & 8.3 & 248/192 & 4.6E-02 \\
hDMSSA\_f(x)+6  & 4.0E+00 & 3.1E+00 & 5.3E+01 & 142 & 4.9 & 363/44 & 8.7E-02 & 3.0E+00 & 3.4E+00 & 5.1E+01 & 154 & 5.3 & 357/61 & 7.7E-02 \\
hDMSSA  & 8.4E-01 & 1.2E+00 & 5.2E+01 & 68 & 2.3 & 379/9 & 9.5E-02 & -4.5E-01 & -6.3E-02 & 5.1E+01 & 67 & 2.2 & 405/15 & 7.7E-02 \\
hAROA\_f(x)+6  & 4.4E+05 & 3.4E+03 & 2.1E+05 & 332 & 11.4 & 144/256 & 8.0E-02 & 6.3E+06 & 5.9E+03 & 1.9E+06 & 348 & 12.0 & 146/287 & 5.3E-02 \\
hAROA  & 4.4E+05 & 3.4E+03 & 2.1E+05 & 319 & 11.0 & 149/250 & 8.4E-02 & 7.4E+06 & 5.6E+03 & 2.5E+06 & 350 & 12.1 & 143/278 & 5.6E-02 \\
\bottomrule
\label{tab5}
\end{tabular}
}
\vspace{1.5mm}\\
\footnotesize{Friedman test p-values: dim10=5.1E-82, dim30=9.7E-78, dim50=5.4E-78, dim100=1.0E-78}
\end{adjustwidth}
\end{table}
\normalsize
\begin{figure}[H]
  \centering
\begin{adjustwidth}{-2cm}{-2.0cm}
  \begin{subfigure}{0.4\textwidth}
     \includegraphics[height=5.1cm,width=6.7cm]{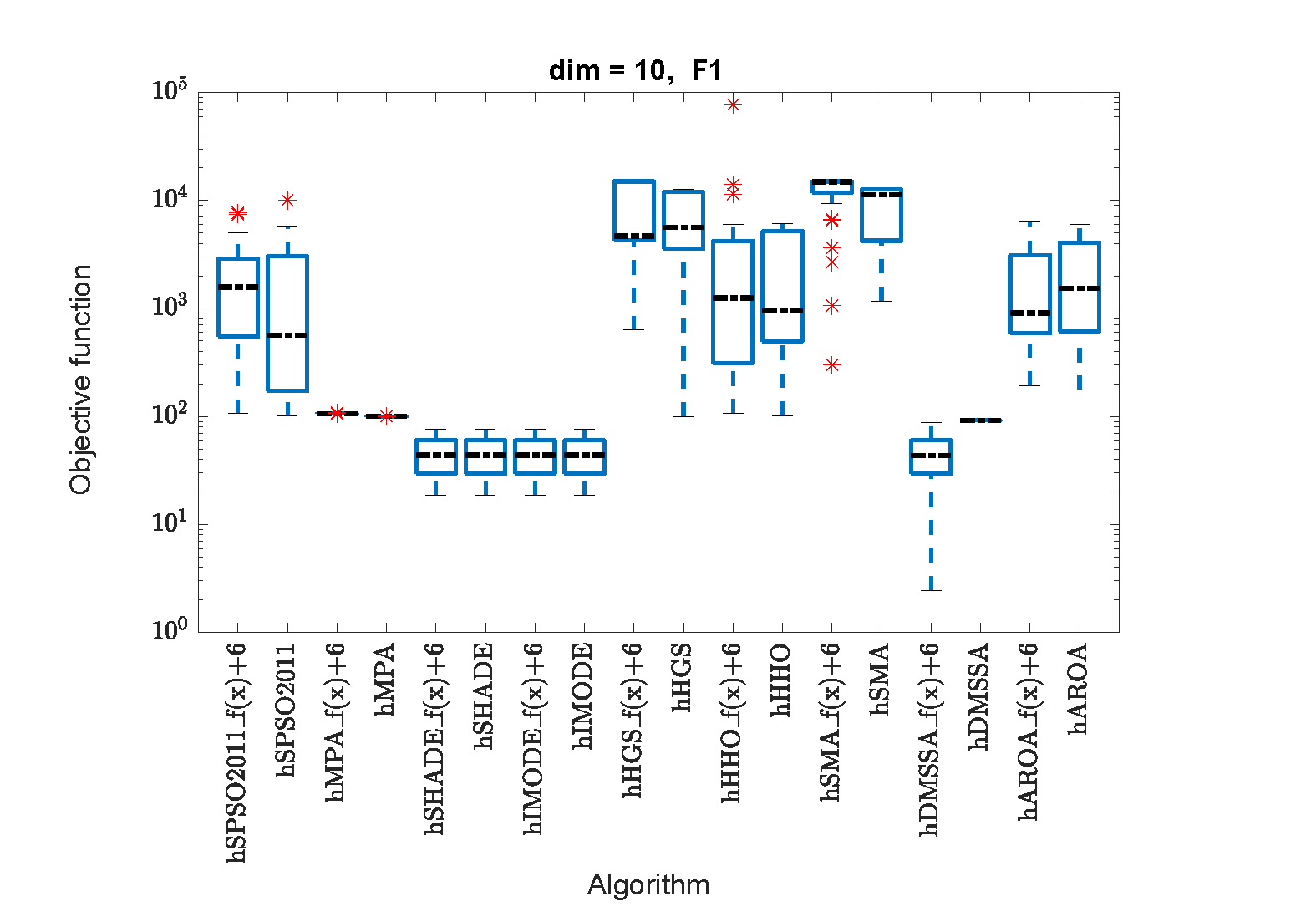}
  \end{subfigure}\hfill
  \begin{subfigure}{0.4\textwidth}
    \includegraphics[height=5.1cm,width=6.7cm]{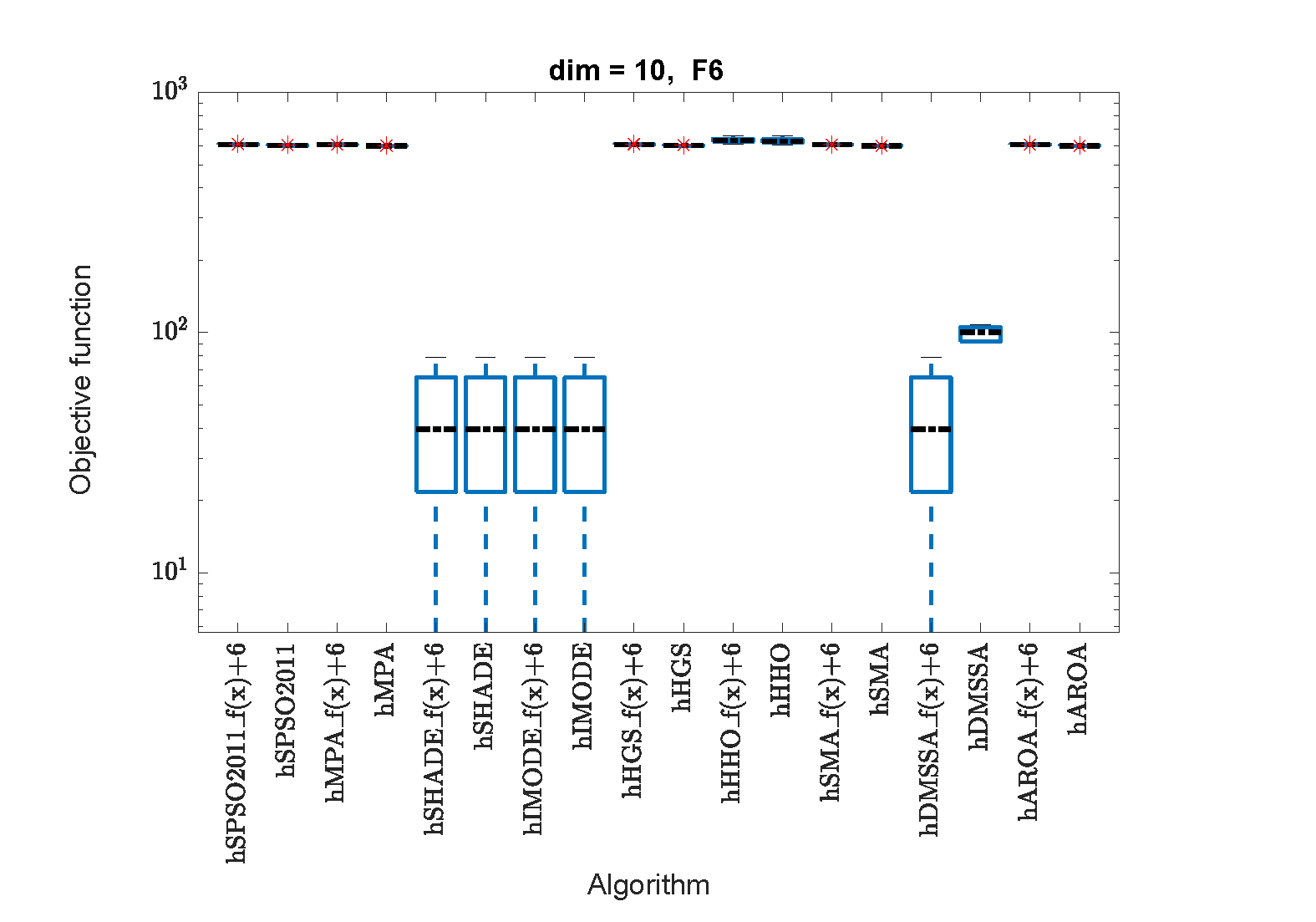}
  \end{subfigure}\hfill
  \begin{subfigure}{0.4\textwidth}
    \includegraphics[height=5.1cm,width=6.7cm]{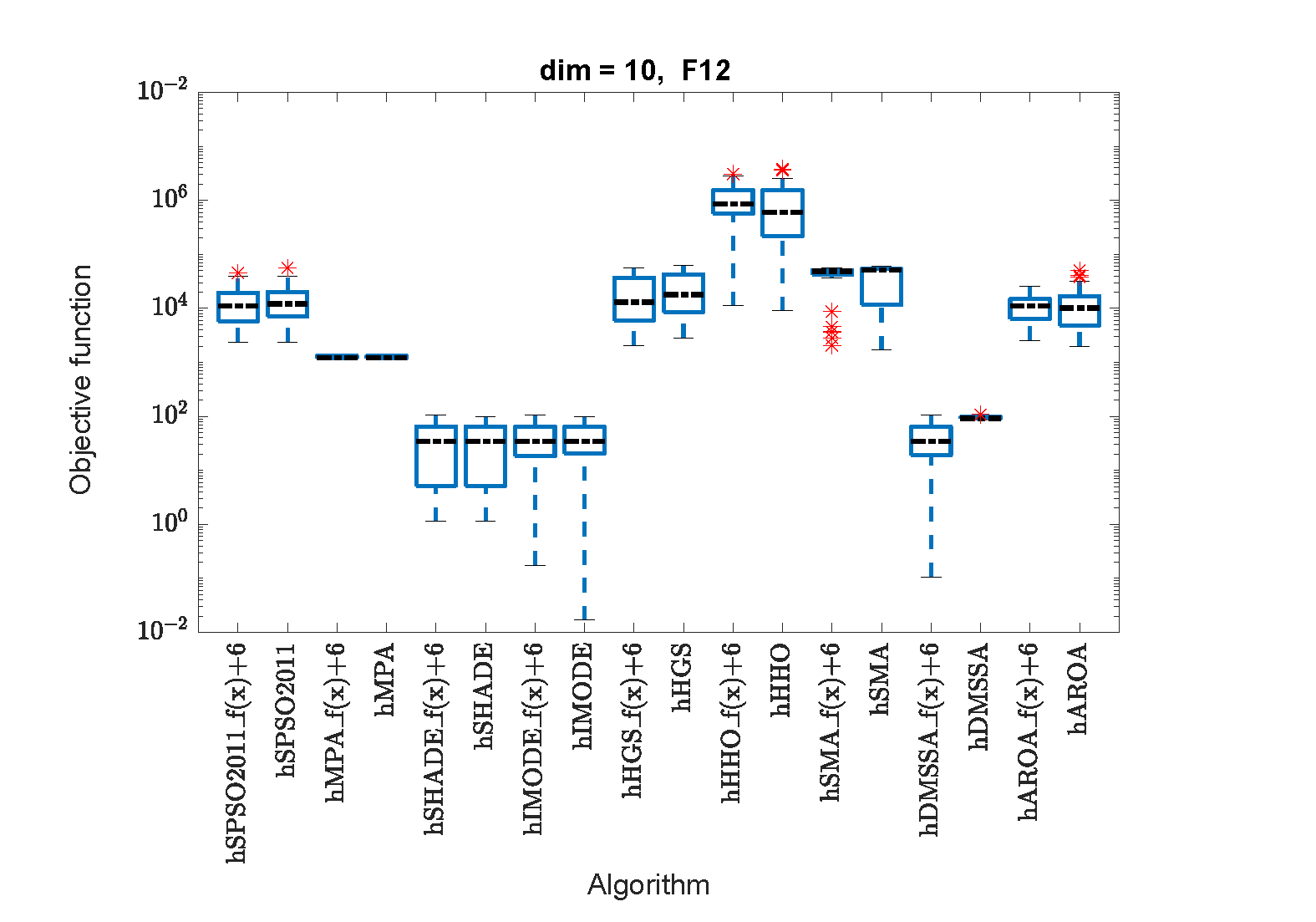}
  \end{subfigure}
   \begin{subfigure}{0.4\textwidth}
    \includegraphics[height=5.1cm,width=6.7cm]{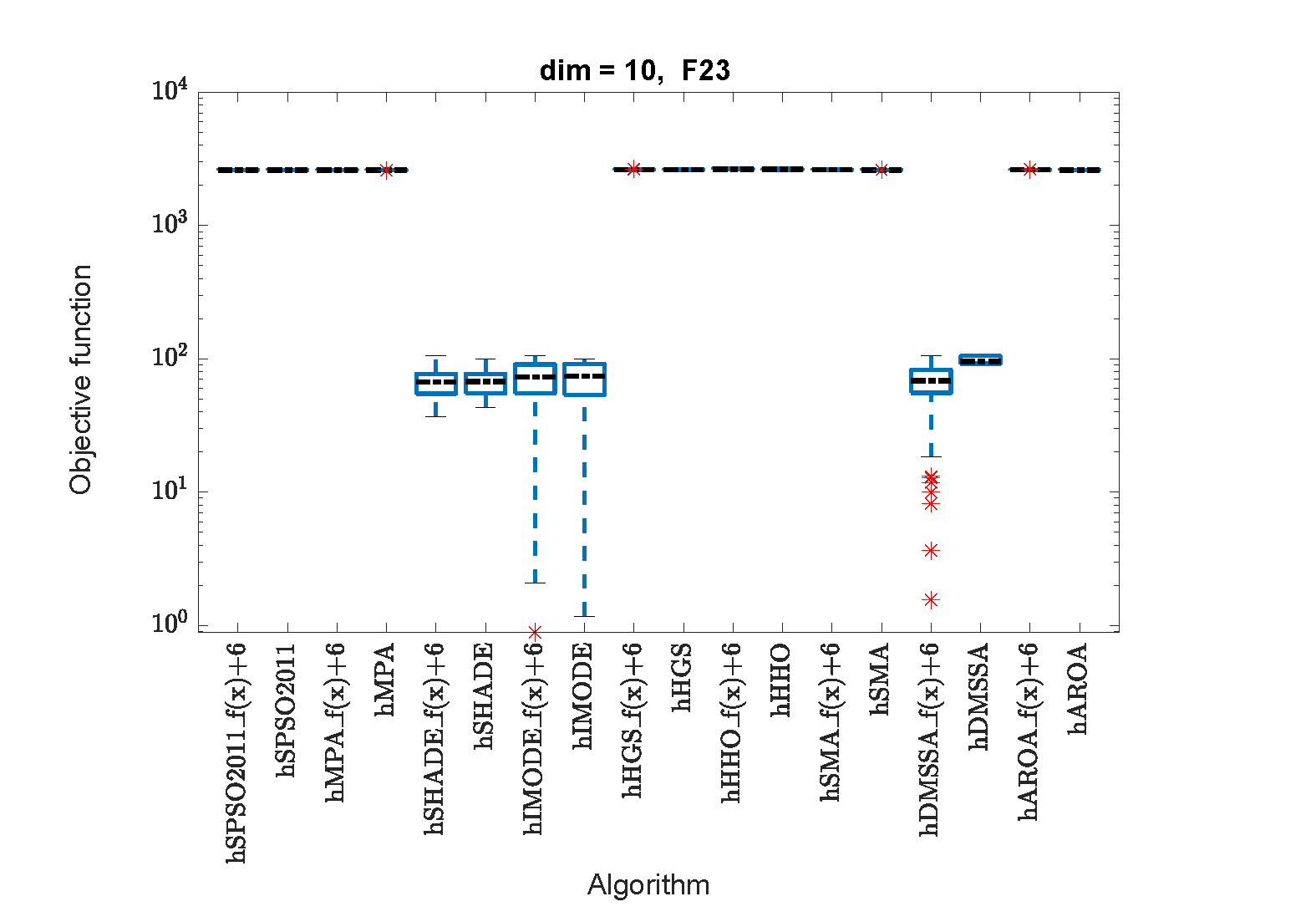}
  \end{subfigure}\hfill
  \begin{subfigure}{0.4\textwidth}
    \includegraphics[height=5.1cm,width=6.7cm]{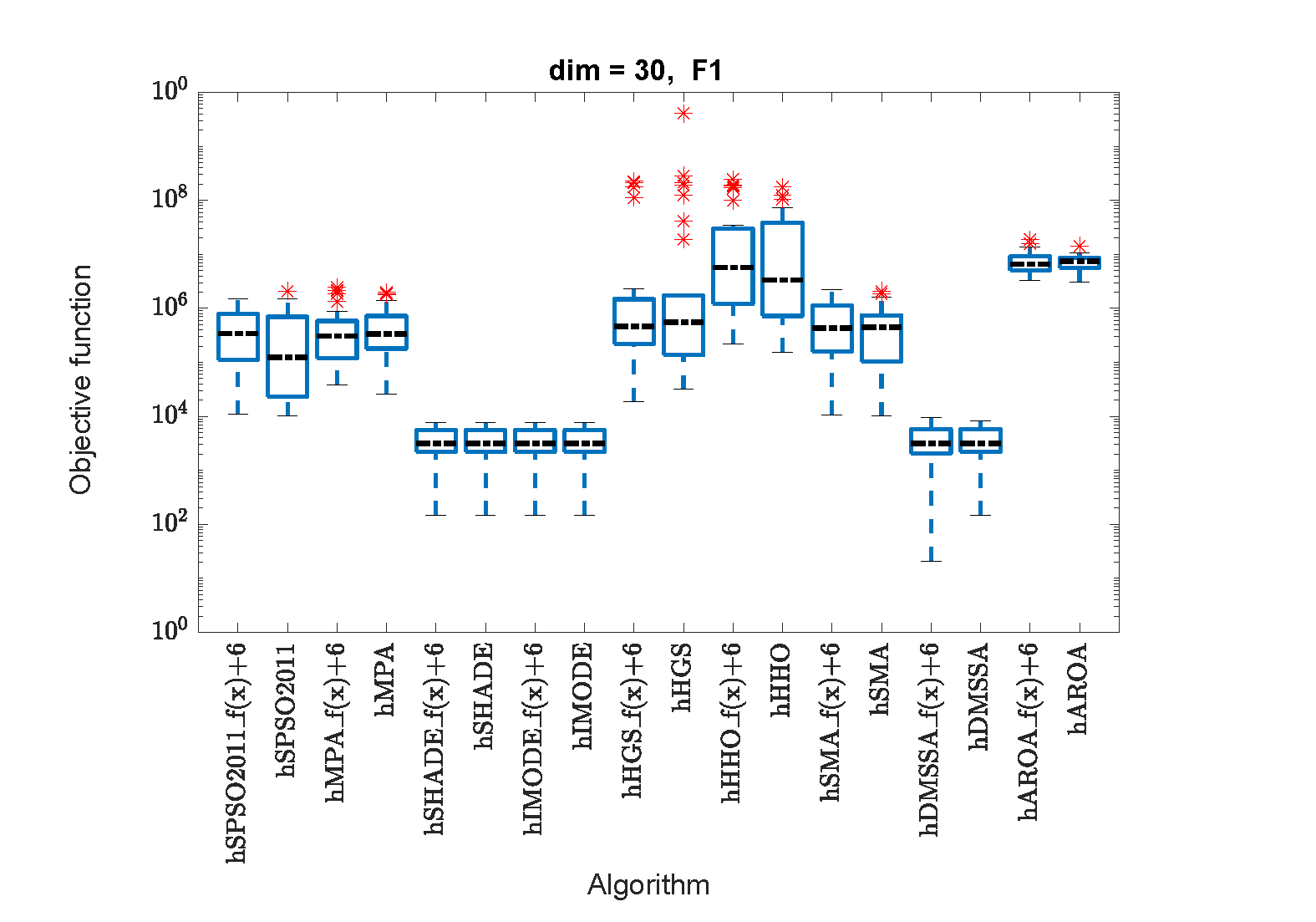}
\end{subfigure}\hfill
  \begin{subfigure}{0.4\textwidth} 
        \includegraphics[height=5.1cm,width=6.7cm]{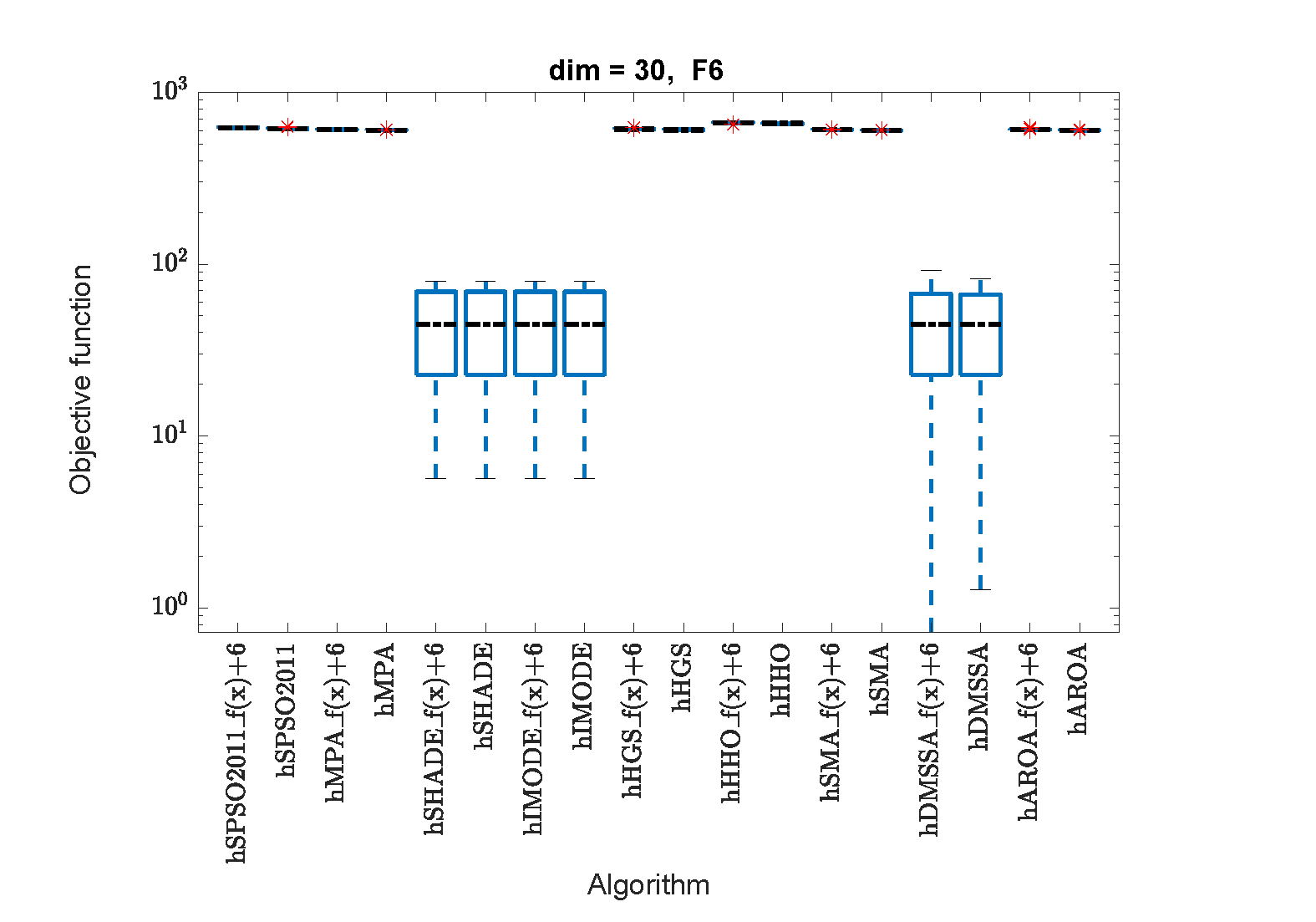}
   \end{subfigure}
   \end{adjustwidth}
\end{figure}

 \begin{figure}[H]
  \centering
\begin{adjustwidth}{-2cm}{-2.0cm}
   \begin{subfigure}{0.4\textwidth}
        \includegraphics[height=5.1cm,width=6.9cm]{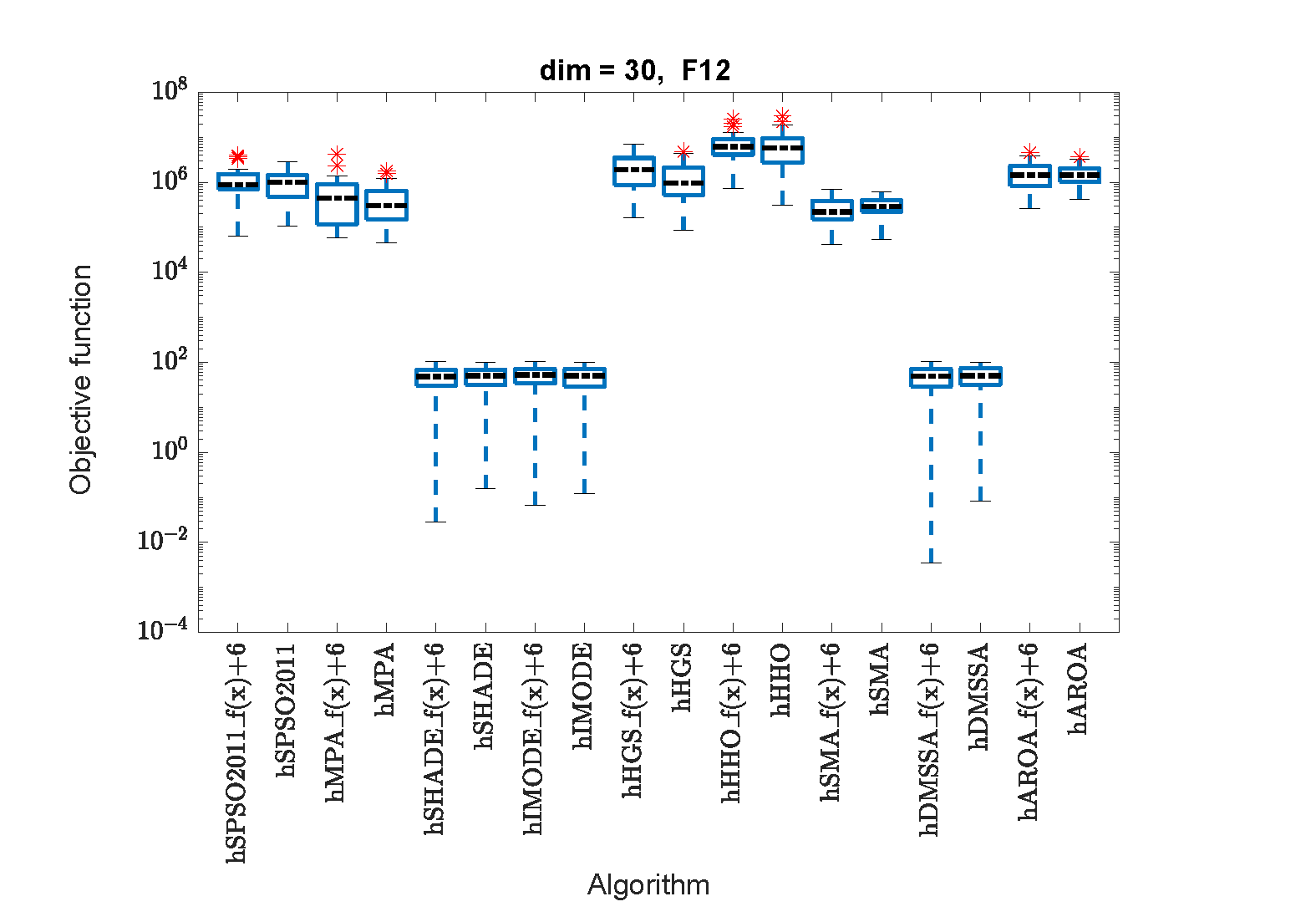}
  \end{subfigure}\hfill
  \begin{subfigure}{0.4\textwidth} 
        \includegraphics[height=5.1cm,width=6.9cm]{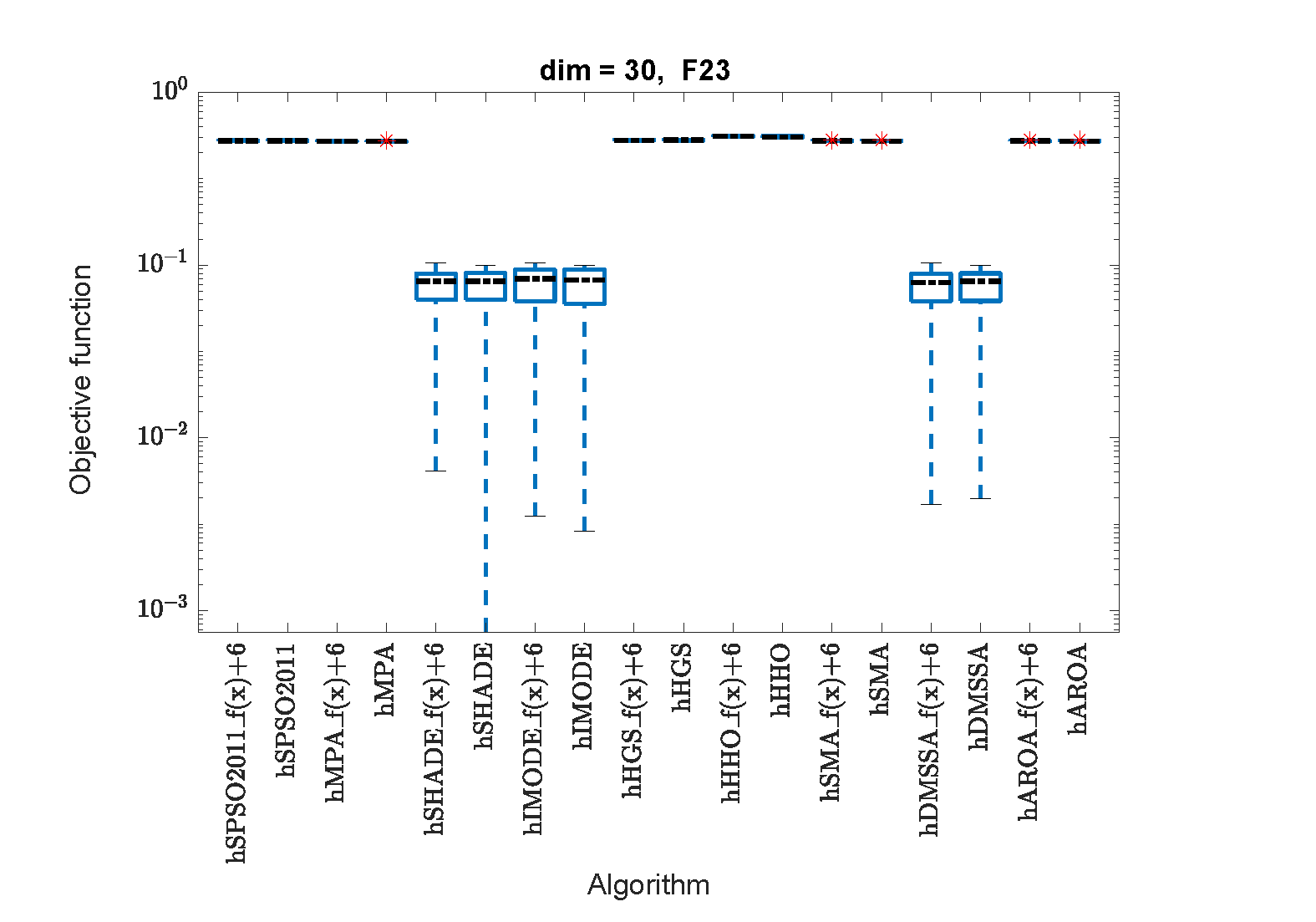}
  \end{subfigure}\hfill
  \begin{subfigure}{0.4\textwidth}
        \includegraphics[height=5.1cm,width=6.9cm]{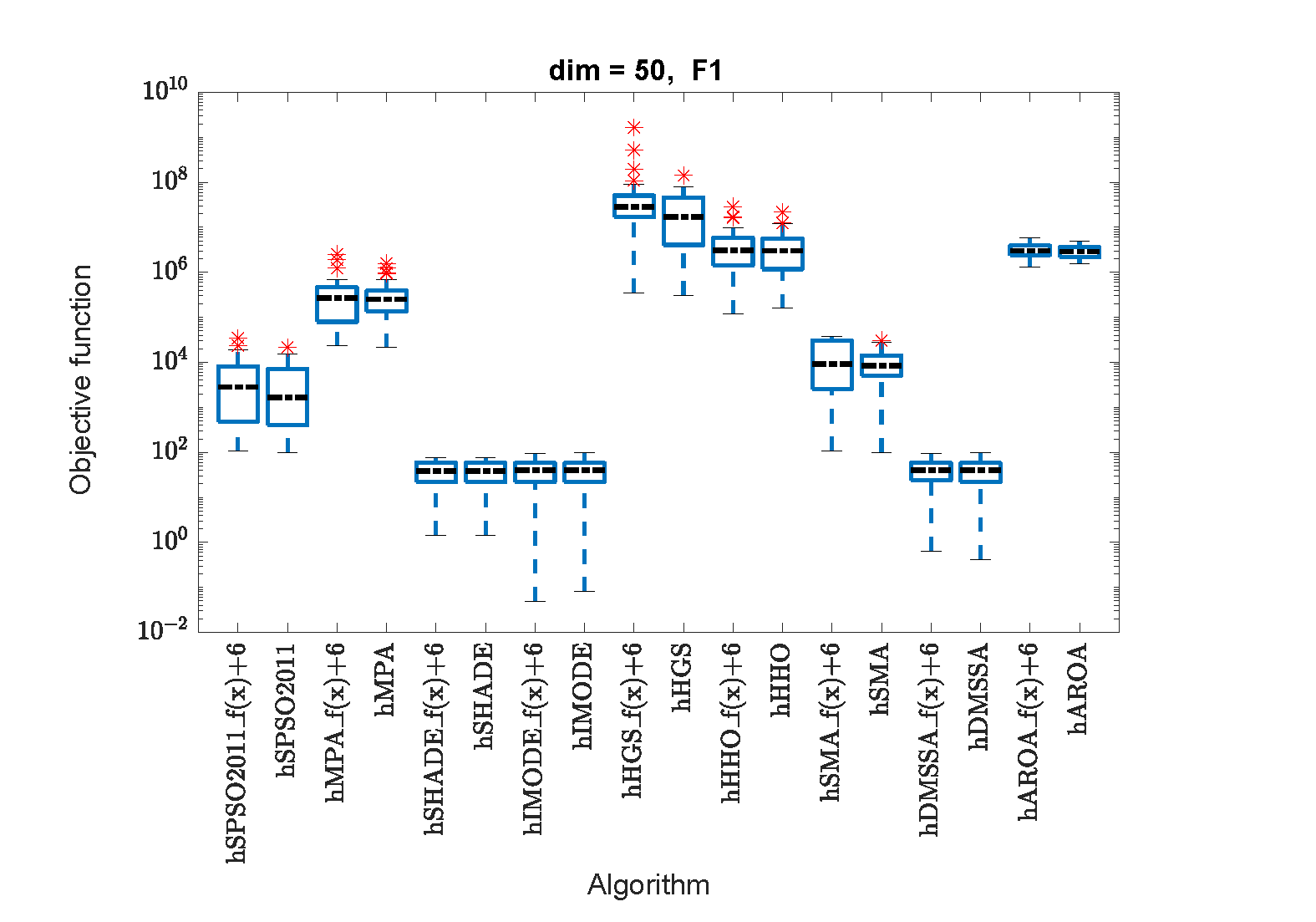}
  \end{subfigure}
  \end{adjustwidth}
\end{figure}

  \begin{figure}[H]
  \centering
\begin{adjustwidth}{-2cm}{-2.0cm}
  \begin{subfigure}{0.4\textwidth} 
        \includegraphics[height=5.1cm,width=6.9cm]{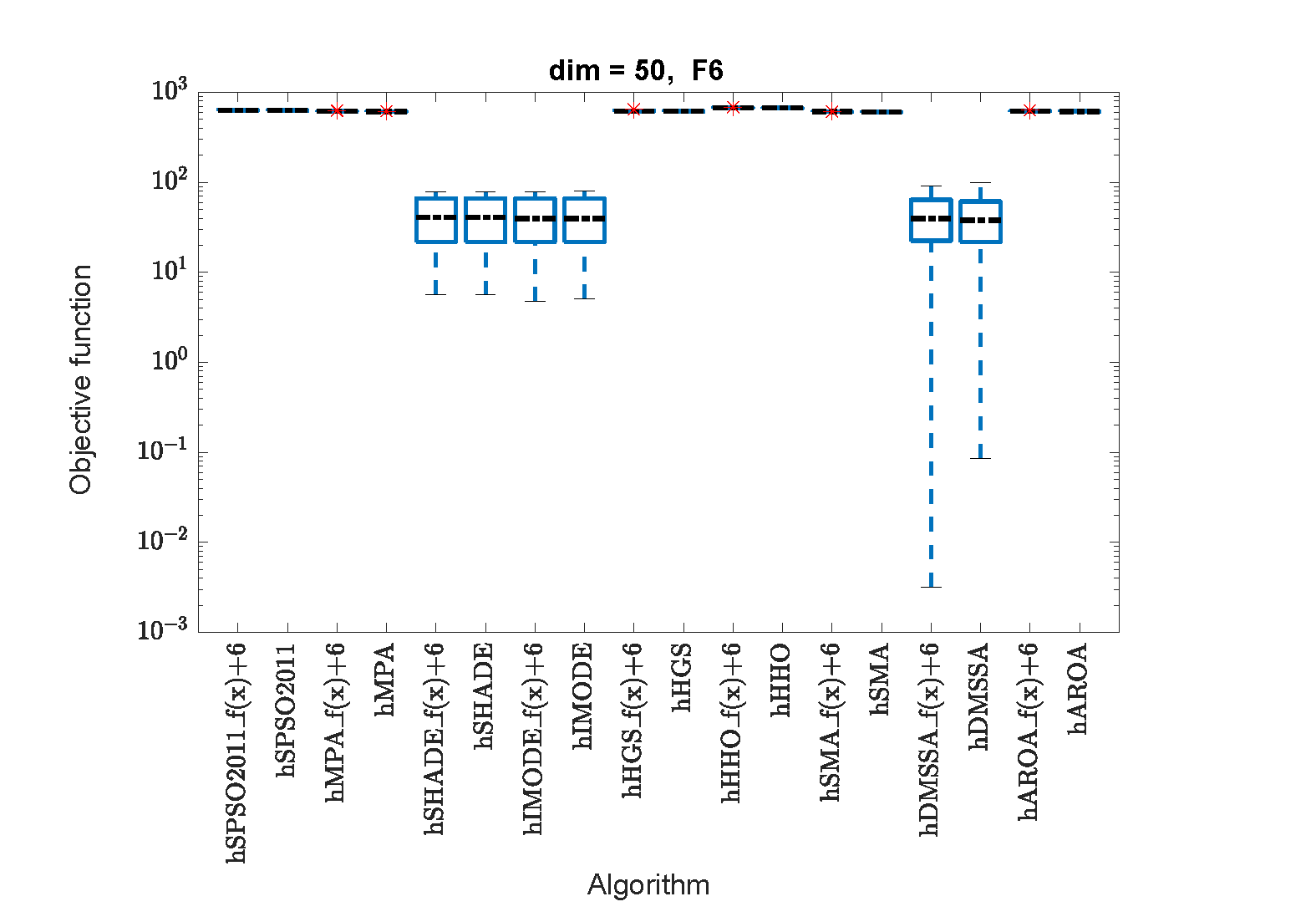}
  \end{subfigure}\hfill
  \begin{subfigure}{0.4\textwidth}
    \includegraphics[height=5.1cm,width=6.9cm]{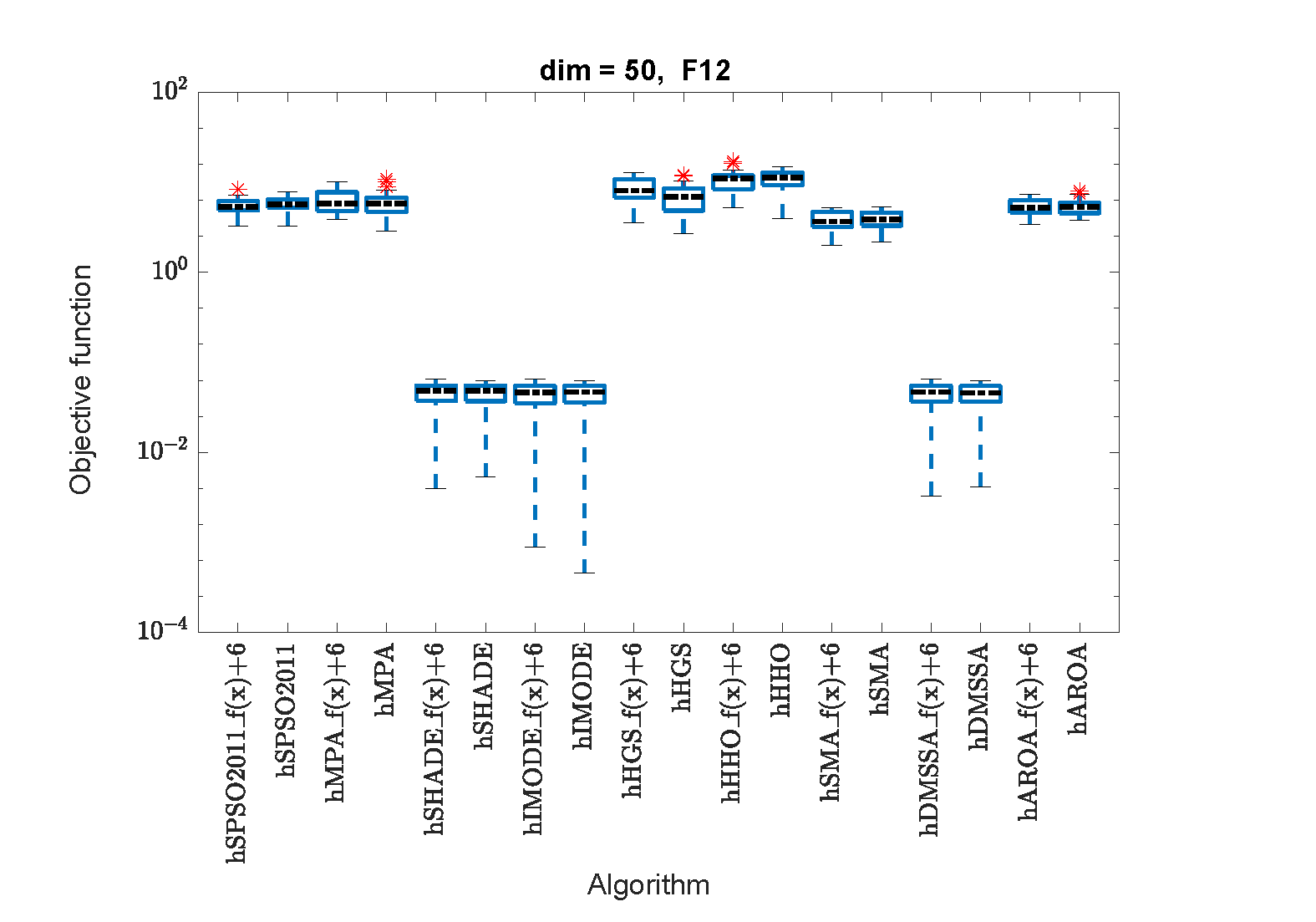}
     \end{subfigure}\hfill
 \begin{subfigure}{0.4\textwidth}
    \includegraphics[height=5.1cm,width=6.9cm]{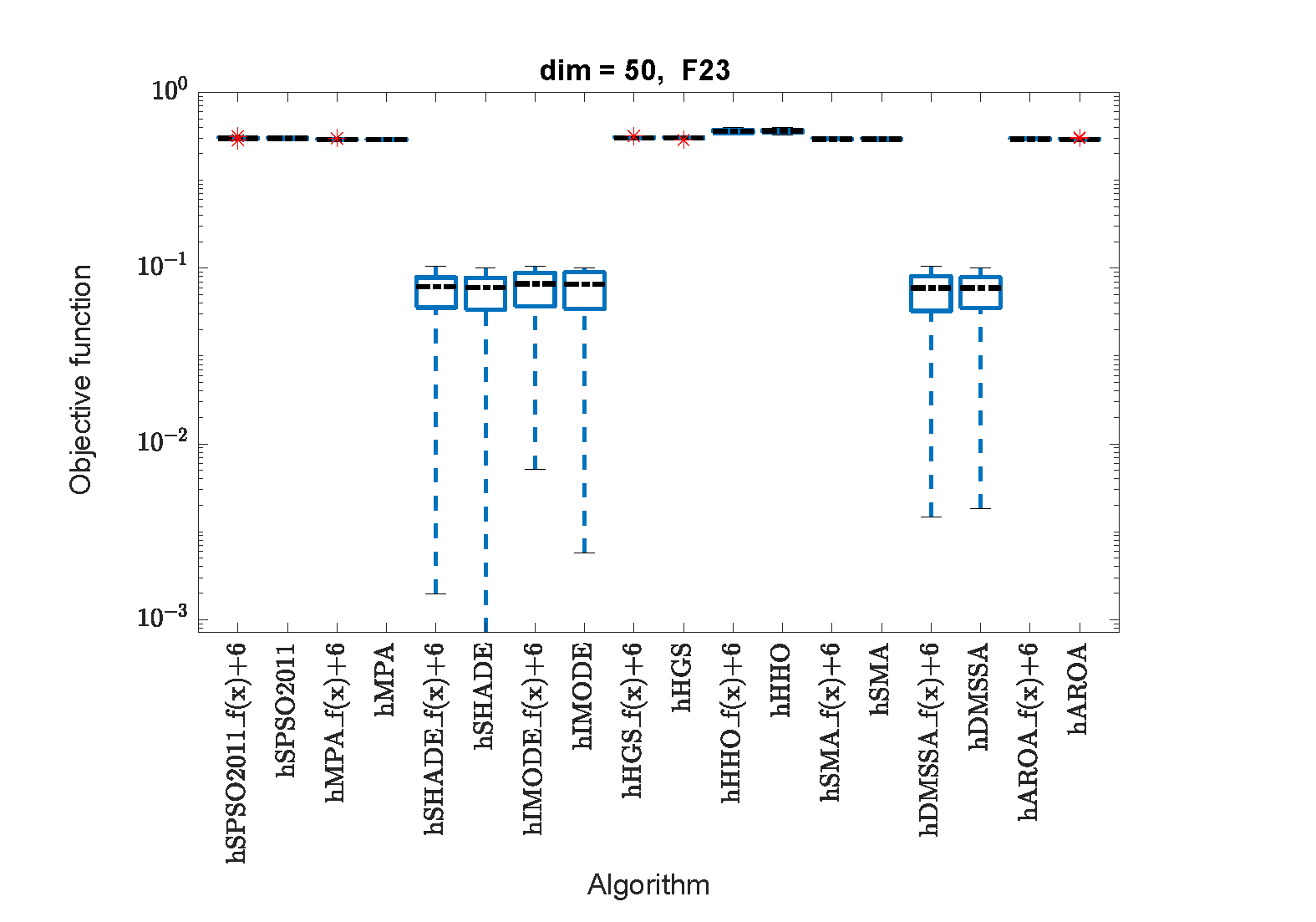}
  \end{subfigure}
\end{adjustwidth}
\end{figure}
 
  \begin{figure}[H]
  \centering
\begin{adjustwidth}{-2cm}{-2.0cm}
  \begin{subfigure}{0.4\textwidth}
    \includegraphics[height=5.1cm,width=6.9cm]{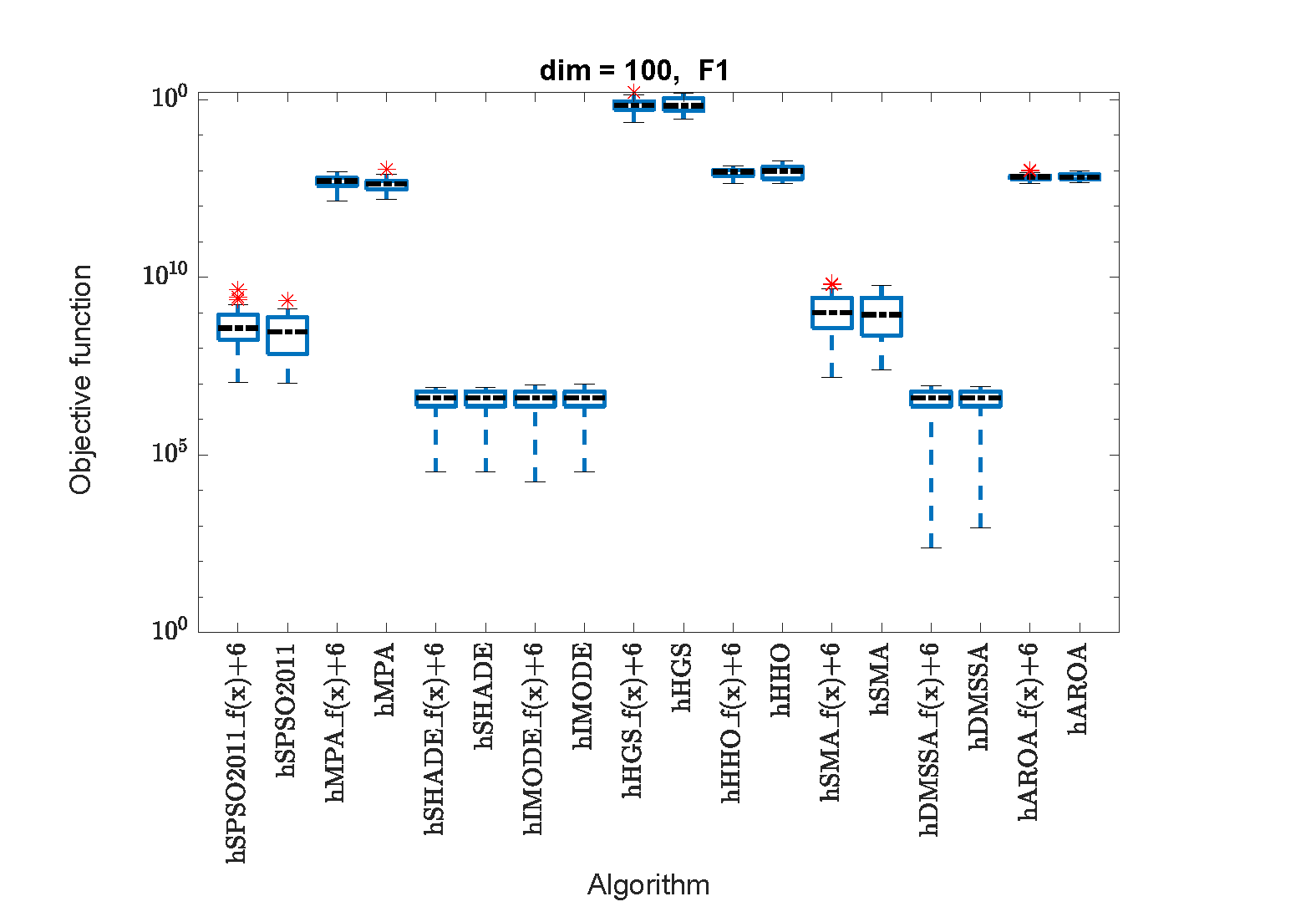}
  \end{subfigure}\hfill
  \begin{subfigure}{0.4\textwidth}
        \includegraphics[height=5.1cm,width=6.9cm]{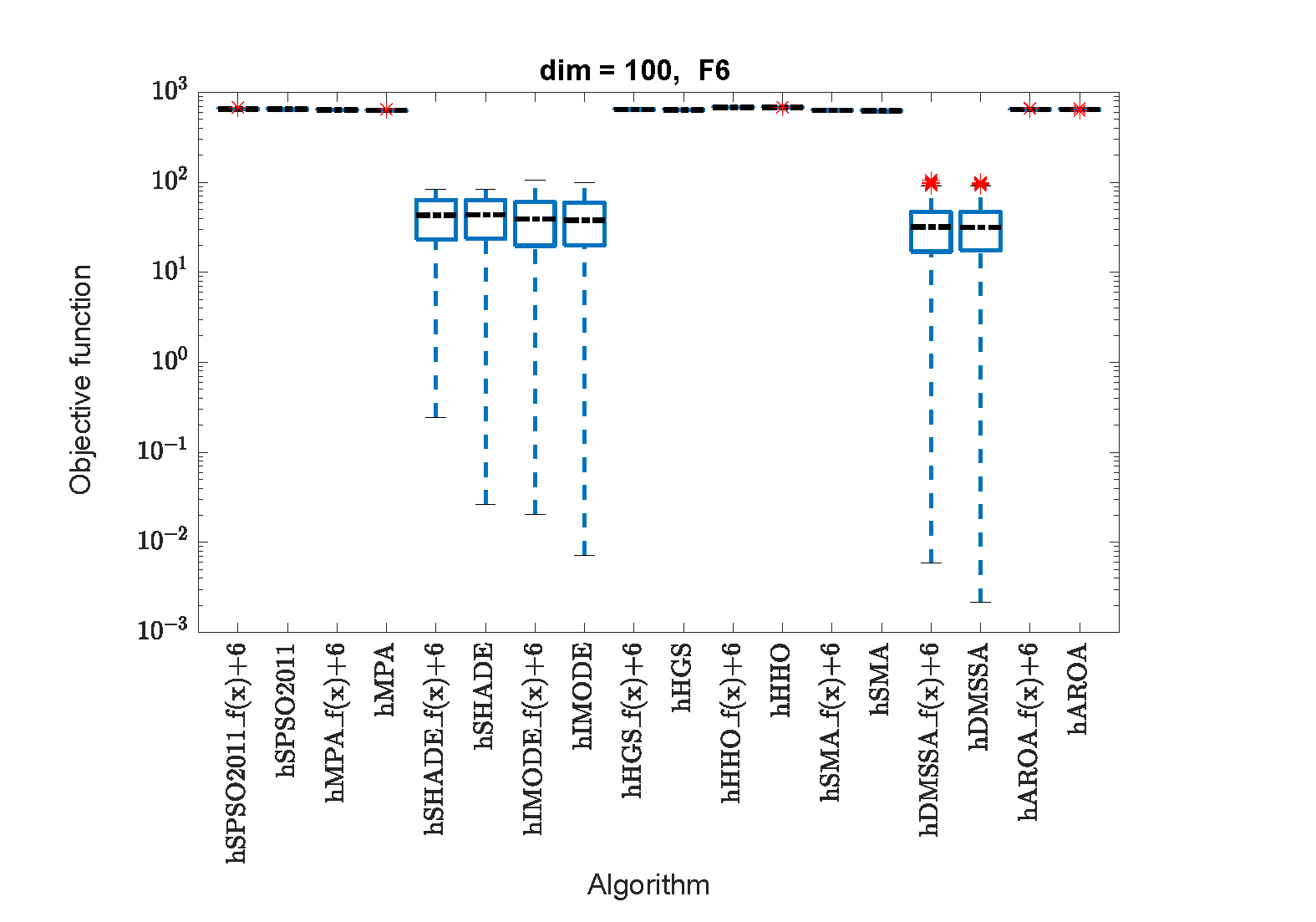}
     \end{subfigure}\hfill
   \begin{subfigure}{0.4\textwidth}
        \includegraphics[height=5.1cm,width=6.9cm]{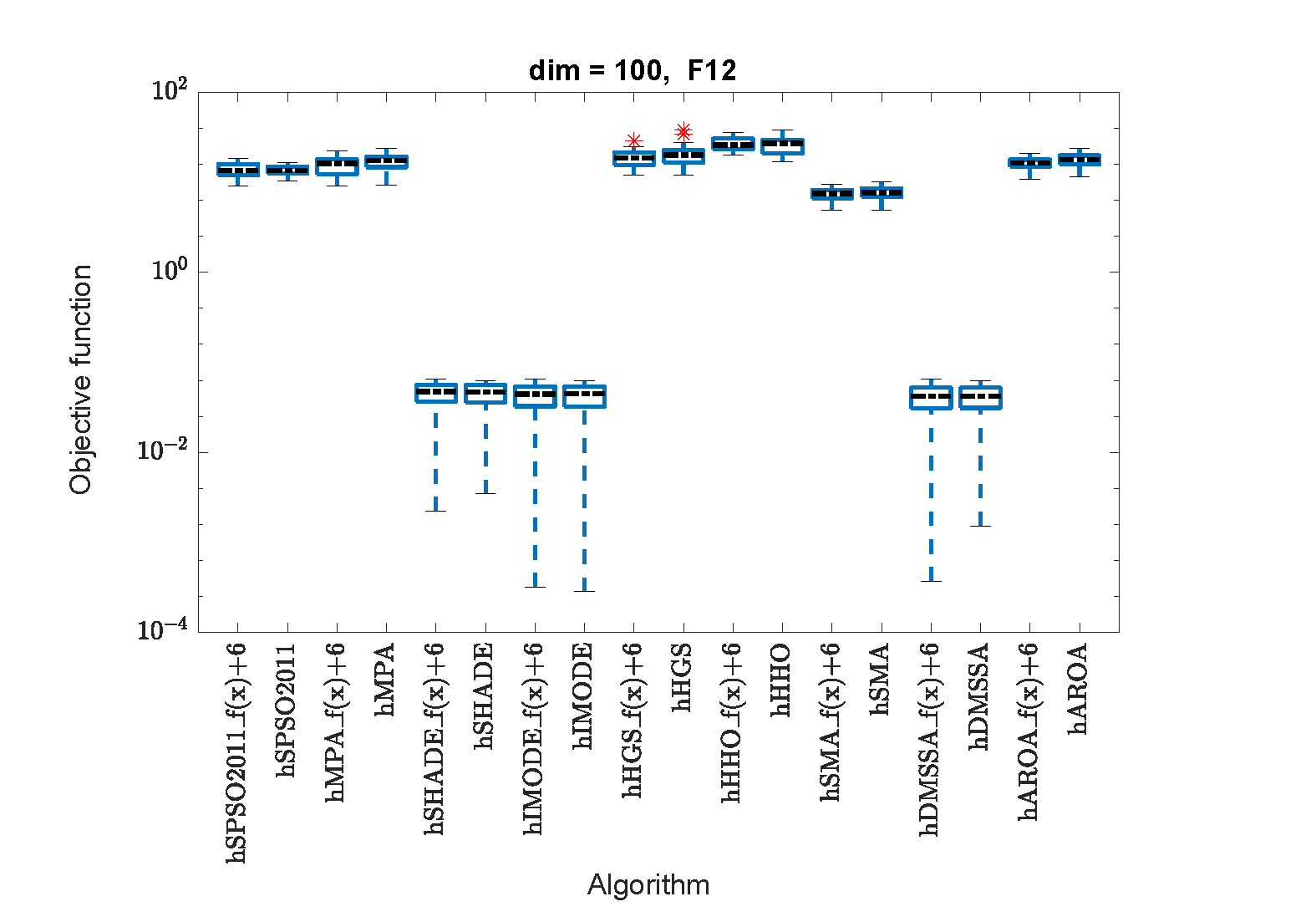}
    \end{subfigure}
    \end{adjustwidth}
 \centering
\begin{adjustwidth}{-2cm}{-2.0cm}
\begin{center}
  \begin{subfigure}{0.4\textwidth}
   \centering
        \includegraphics[height=5.1cm,width=6.9cm]{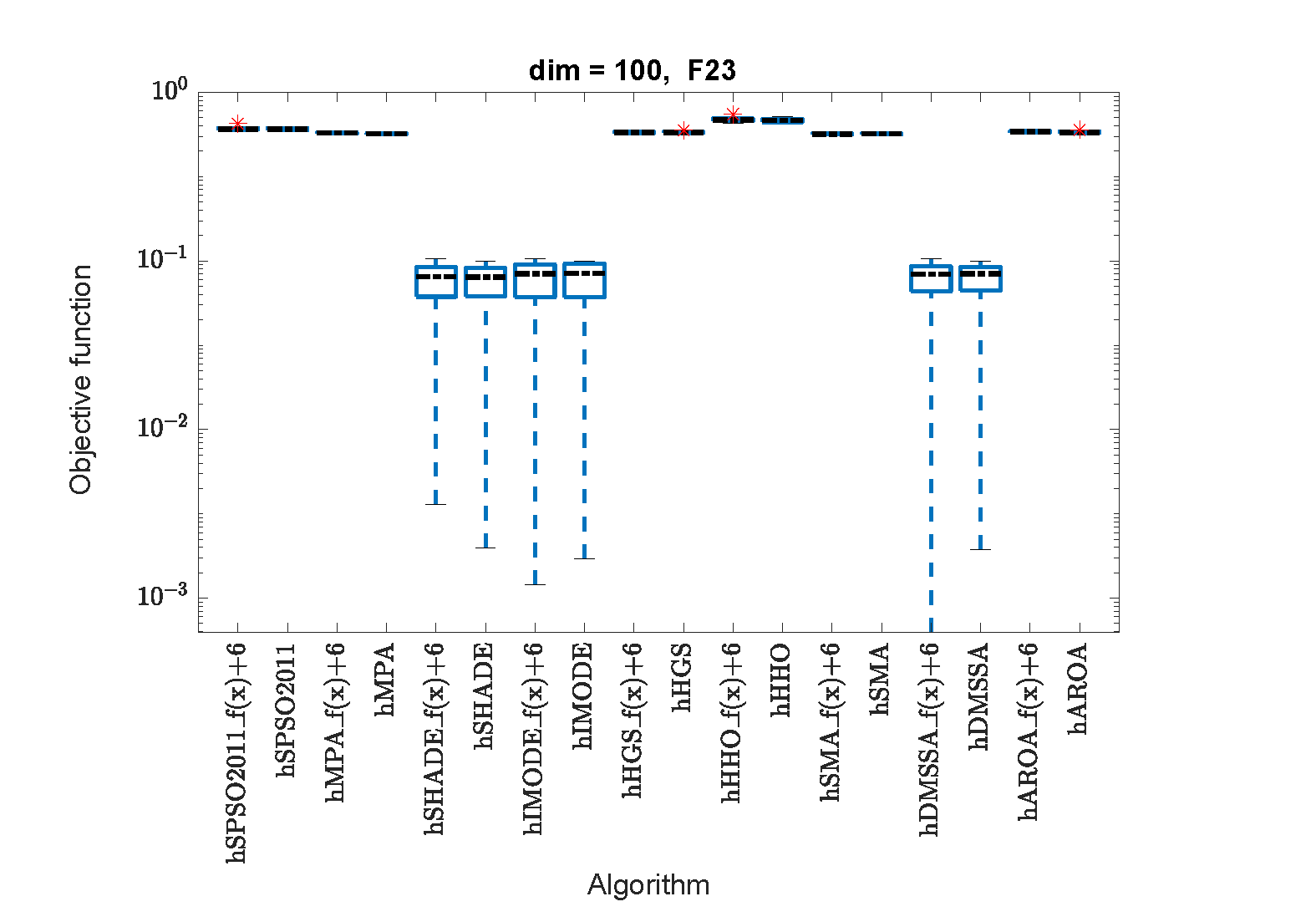}   
  \end{subfigure}
  \end{center}
  \begin{center}
  \begin{subfigure}{0.4\textwidth}
   \centering
    \includegraphics[width=\linewidth]{boxplotlegend.pdf}     
  \end{subfigure}
  \end{center}
  \caption{\footnotesize{Boxplots of final objective values for 9 hybrid algorithms evaluated on original CEC-2017 functions  and their shifted-output variants across 4 dims.}}
  \label{fig17}
\end{adjustwidth}
\end{figure}

\normalsize

\begin{figure}[H]
  \centering
\begin{adjustwidth}{-1cm}{1.0cm}
  \begin{subfigure}{0.35\textwidth} 
        \includegraphics[height=5cm,width=7.5cm]{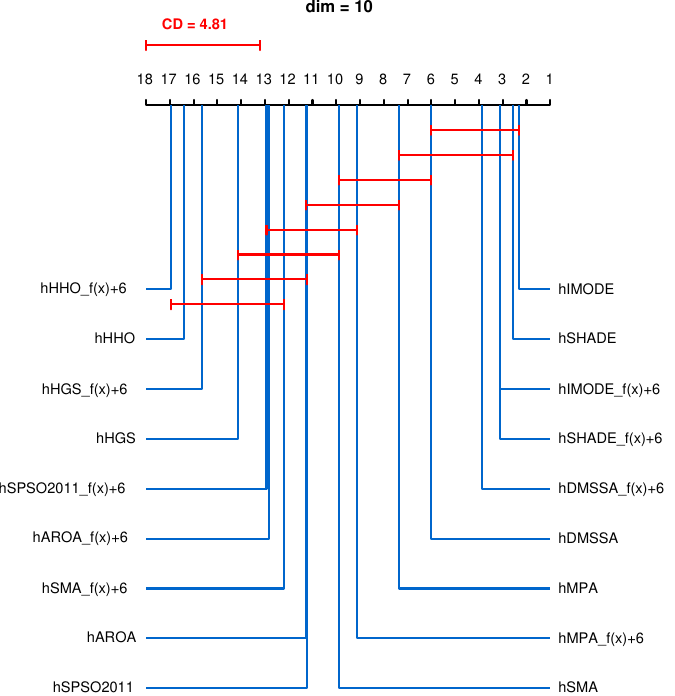}
  \end{subfigure}\hfill
  \begin{subfigure}{0.36\textwidth}
    \includegraphics[height=6cm,width=7.5cm]{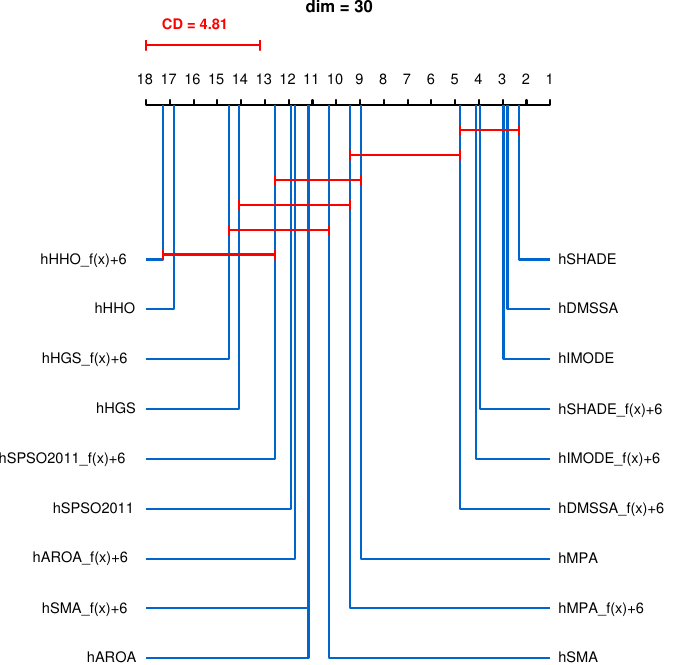}
     \end{subfigure}\hfill
\end{adjustwidth}
\end{figure}
\vspace{0.45cm}
\begin{figure}[H]
\centering
\vspace{0.2cm}
\begin{adjustwidth}{-1cm}{1.0cm}
 \begin{subfigure}{0.35\textwidth}
    \includegraphics[height=5cm,width=7.5cm]{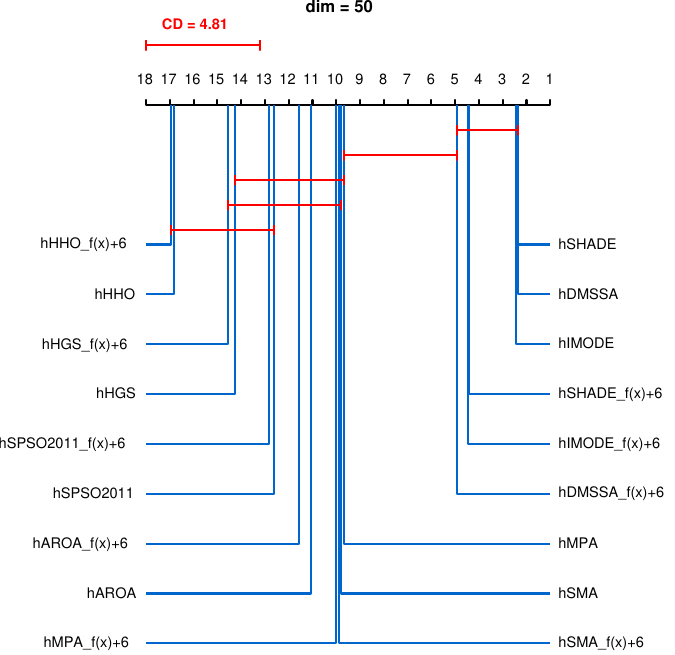}
  \end{subfigure}\hfill
  \begin{subfigure}{0.35\textwidth}
    \includegraphics[height=5cm,width=7.5cm]{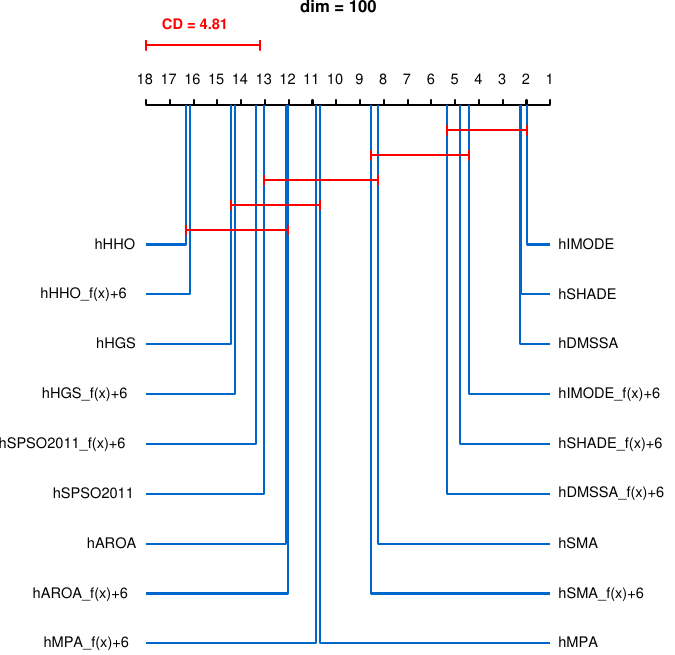}
  \end{subfigure}\hfill
  \caption{\footnotesize{CD diagrams for 9 hybrid algorithms tested on original CEC-2017 functions and their shifted-output variants across 4 dims.}}
  \label{fig18}
\end{adjustwidth}
\end{figure}

\begin{figure}[H] 
\adjustbox{scale=1.25,center}{
\centering
  \begin{subfigure}[t]{0.495\textwidth}
    \centering
\includegraphics[width=\linewidth]{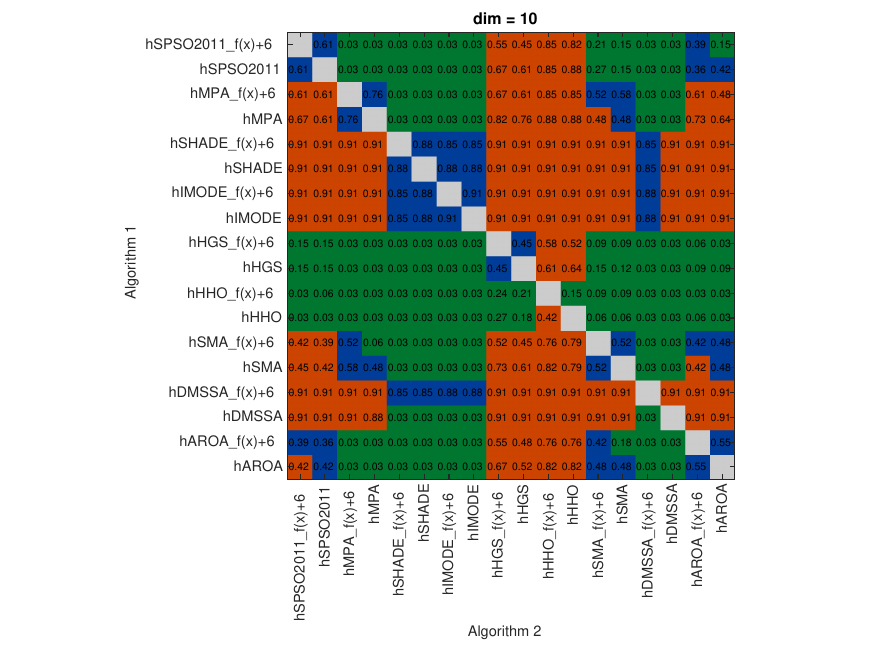}
  \end{subfigure}
  \begin{subfigure}[t]{0.495\textwidth}
    \centering
\includegraphics[width=\linewidth]{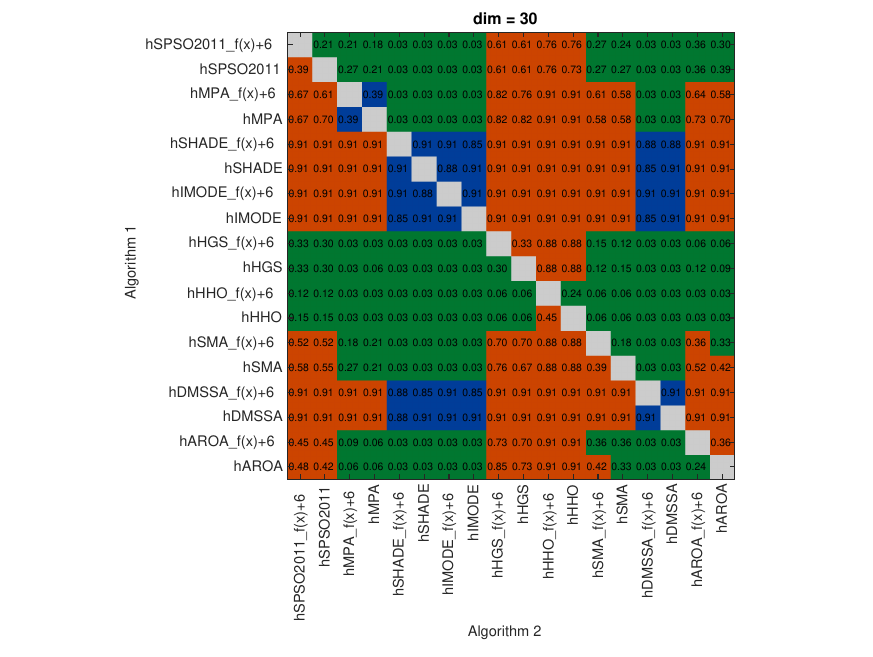}
  \end{subfigure}
  }
  \end{figure}
  \vspace{0.2cm}
  \begin{figure}[H] 
\adjustbox{scale=1.25,center}{
\begin{subfigure}[t]{0.495\textwidth}
    \centering
\includegraphics[width=\linewidth]{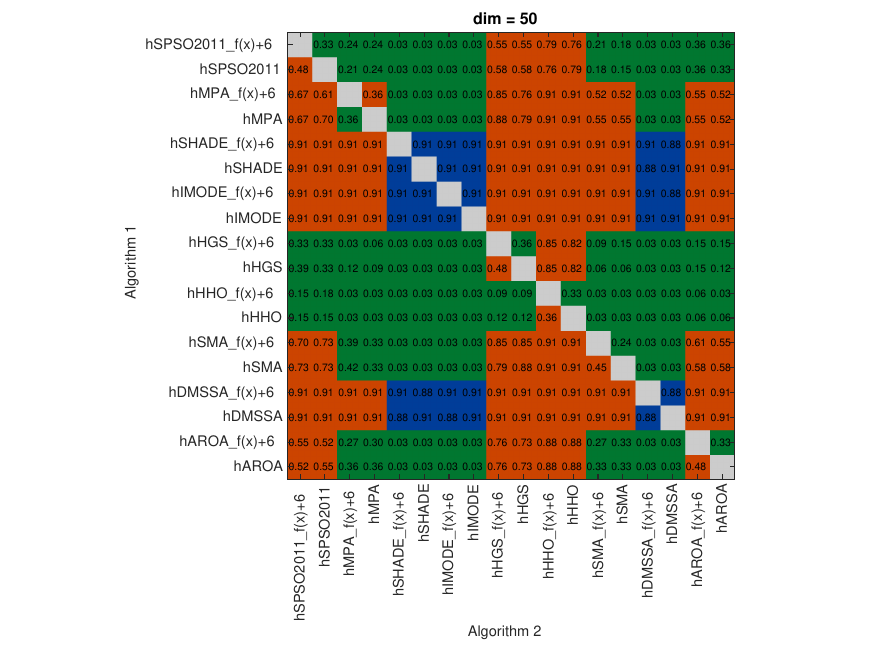}
  \end{subfigure}
 \begin{subfigure}[t]{0.495\textwidth}
    \centering
\includegraphics[width=\linewidth]{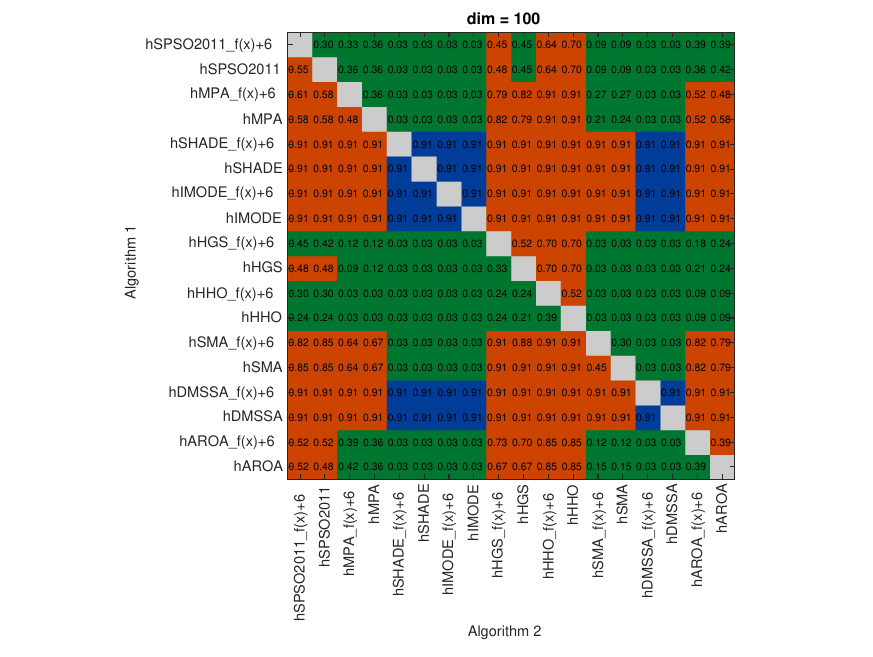}
  \end{subfigure}
  }
  \vspace{0.2cm}
  \adjustbox{scale=1.1,center}{
\begin{subfigure}{0.6\textwidth}
    \centering
    \includegraphics[width=\linewidth]{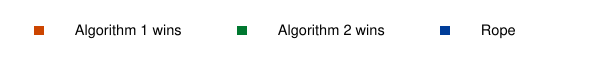}
  \end{subfigure}
  }
  \vspace{-0.75cm}
\caption{\footnotesize{Bayesian comparison results for 9 hybrid algorithms on original CEC-2017 functions and their shifted-output counterparts across 4 dims.}}
  \label{fig19}
  \end{figure}

\begin{figure}[H]
  \centering
\begin{adjustwidth}{-1cm}{-2.0cm}
  \begin{subfigure}{0.35\textwidth}
     \includegraphics[height=4cm,width=5.5cm]{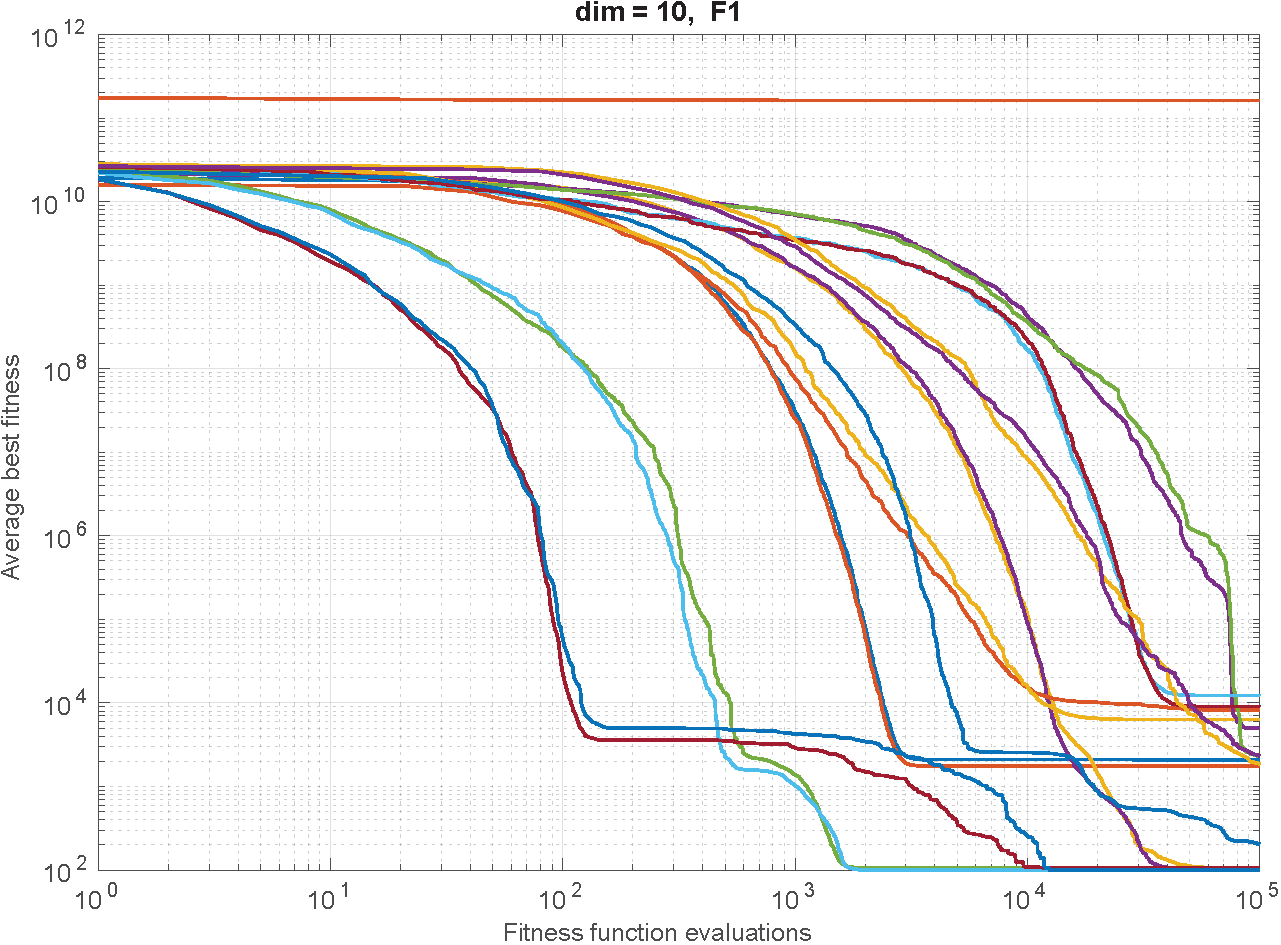}
  \end{subfigure}\hfill
  \begin{subfigure}{0.35\textwidth}
    \includegraphics[height=4cm,width=5.5cm]{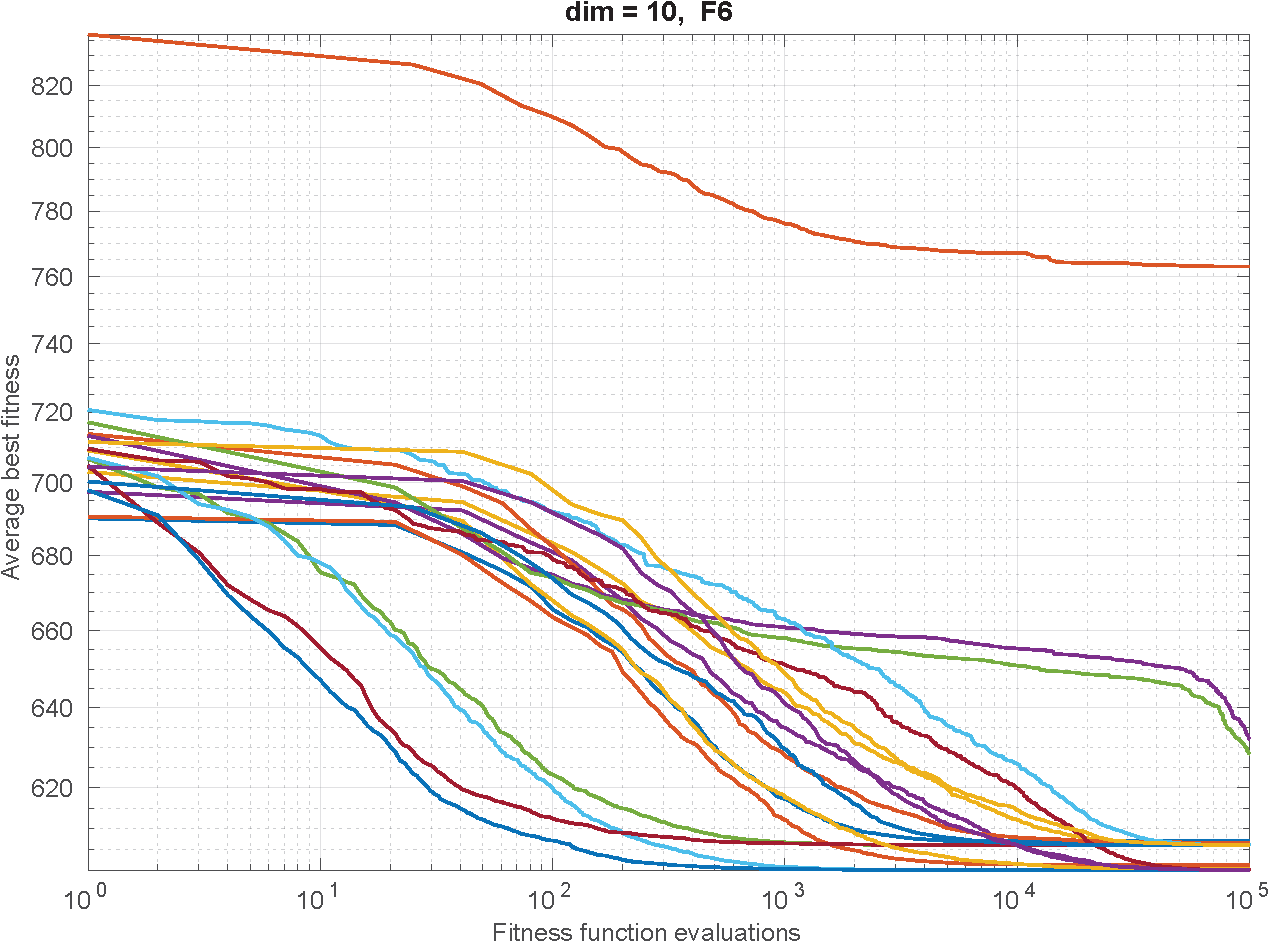}
  \end{subfigure}\hfill
  \begin{subfigure}{0.35\textwidth}
    \includegraphics[height=4cm,width=5.5cm]{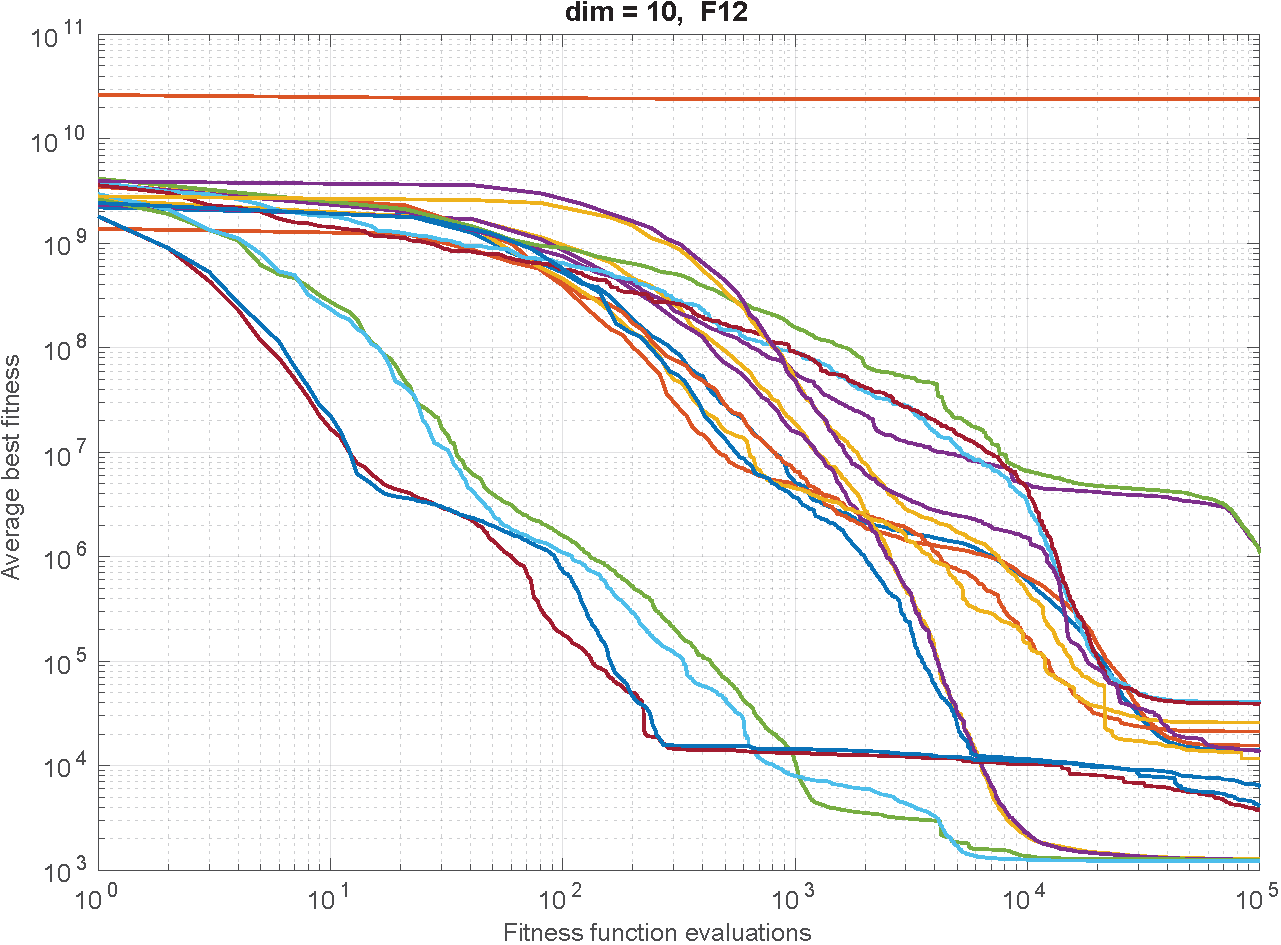}
  \end{subfigure}
 \end{adjustwidth}
\end{figure} 

\begin{figure}[H]
  \centering
\begin{adjustwidth}{-1cm}{-2.0cm}
   \begin{subfigure}{0.35\textwidth}
    \includegraphics[height=4cm,width=5.5cm]{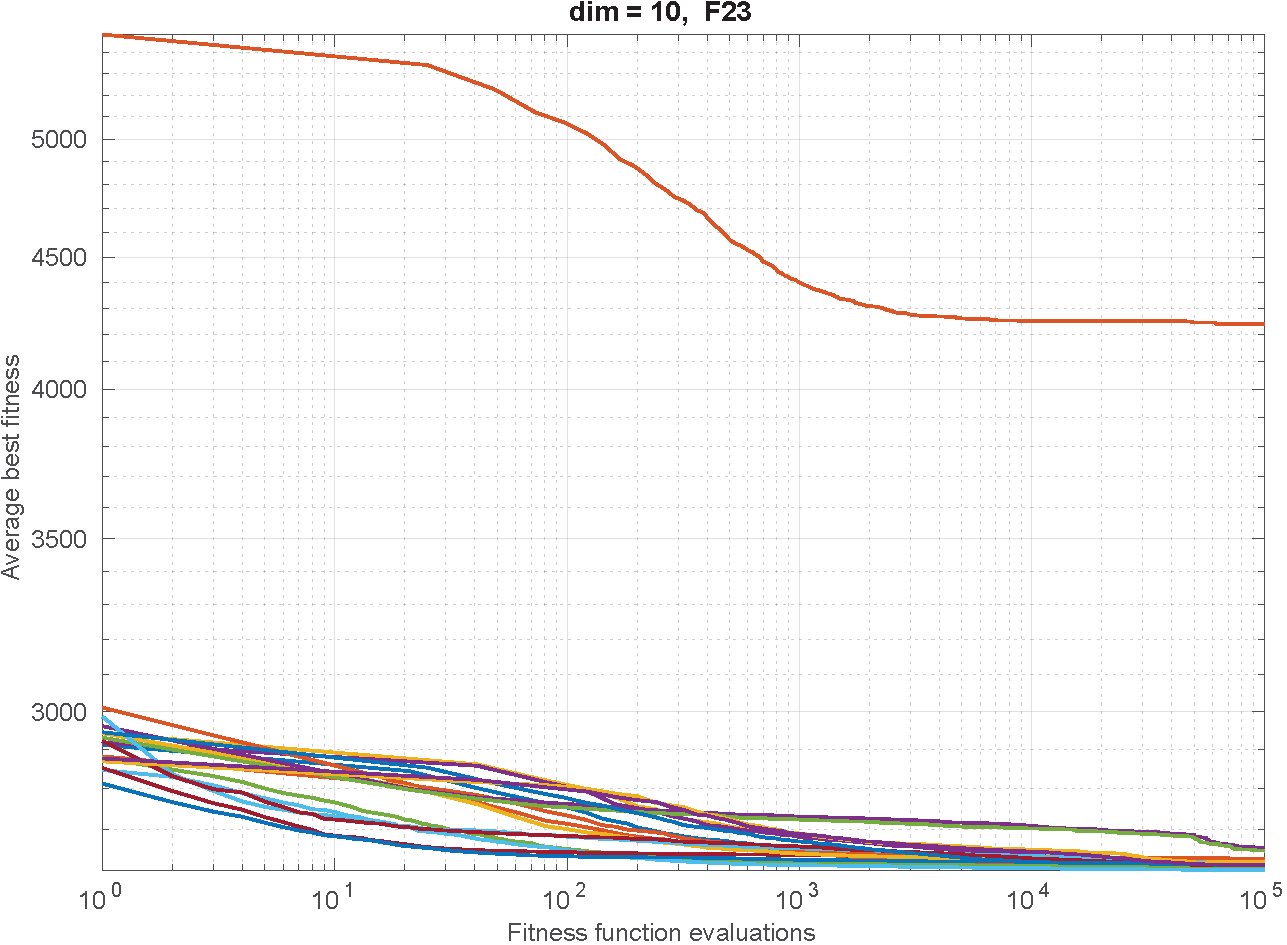}
  \end{subfigure}\hfill
  \begin{subfigure}{0.35\textwidth}
    \includegraphics[height=4cm,width=5.5cm]{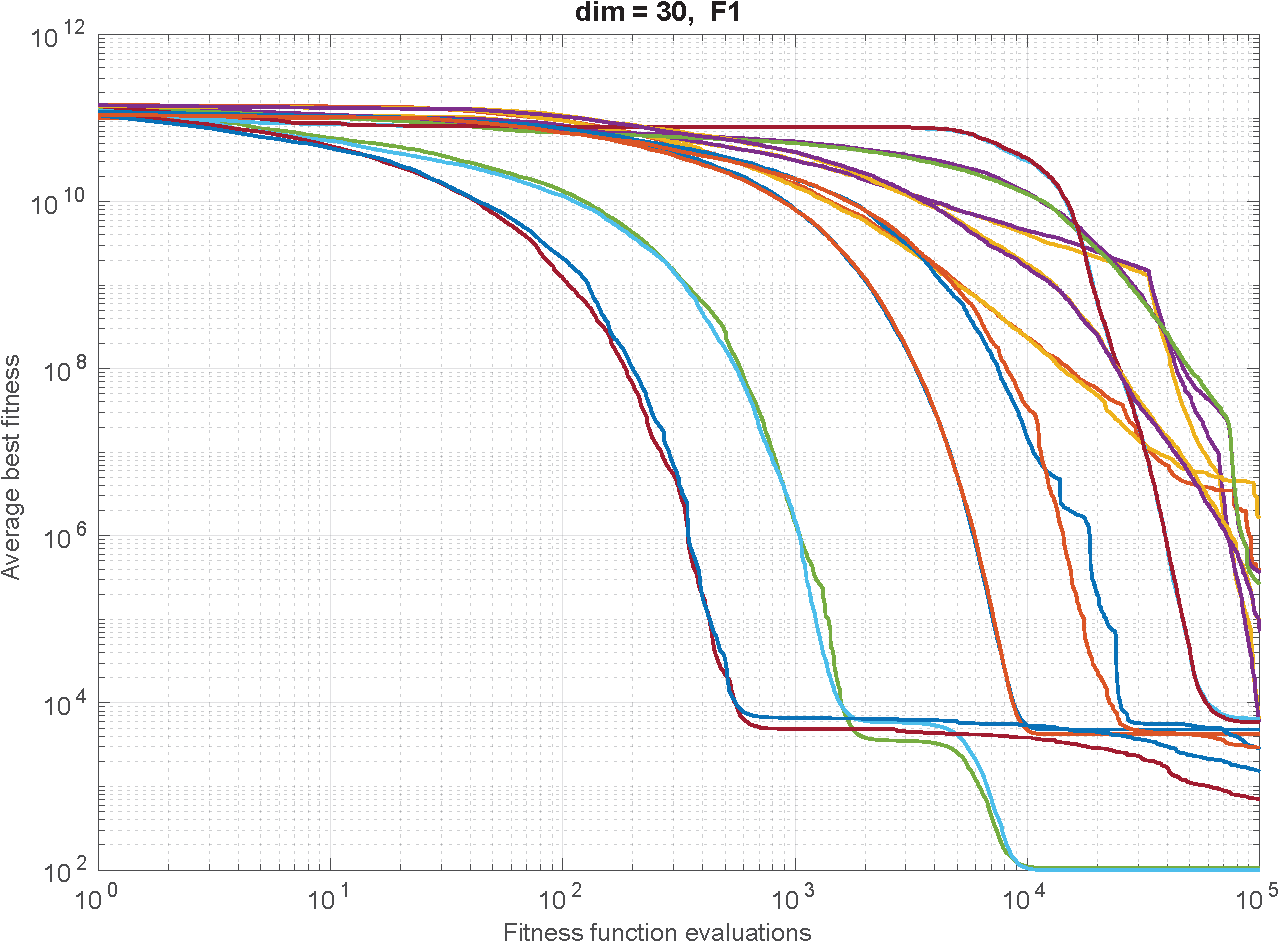}
\end{subfigure}\hfill
  \begin{subfigure}{0.35\textwidth} 
        \includegraphics[height=4cm,width=5.5cm]{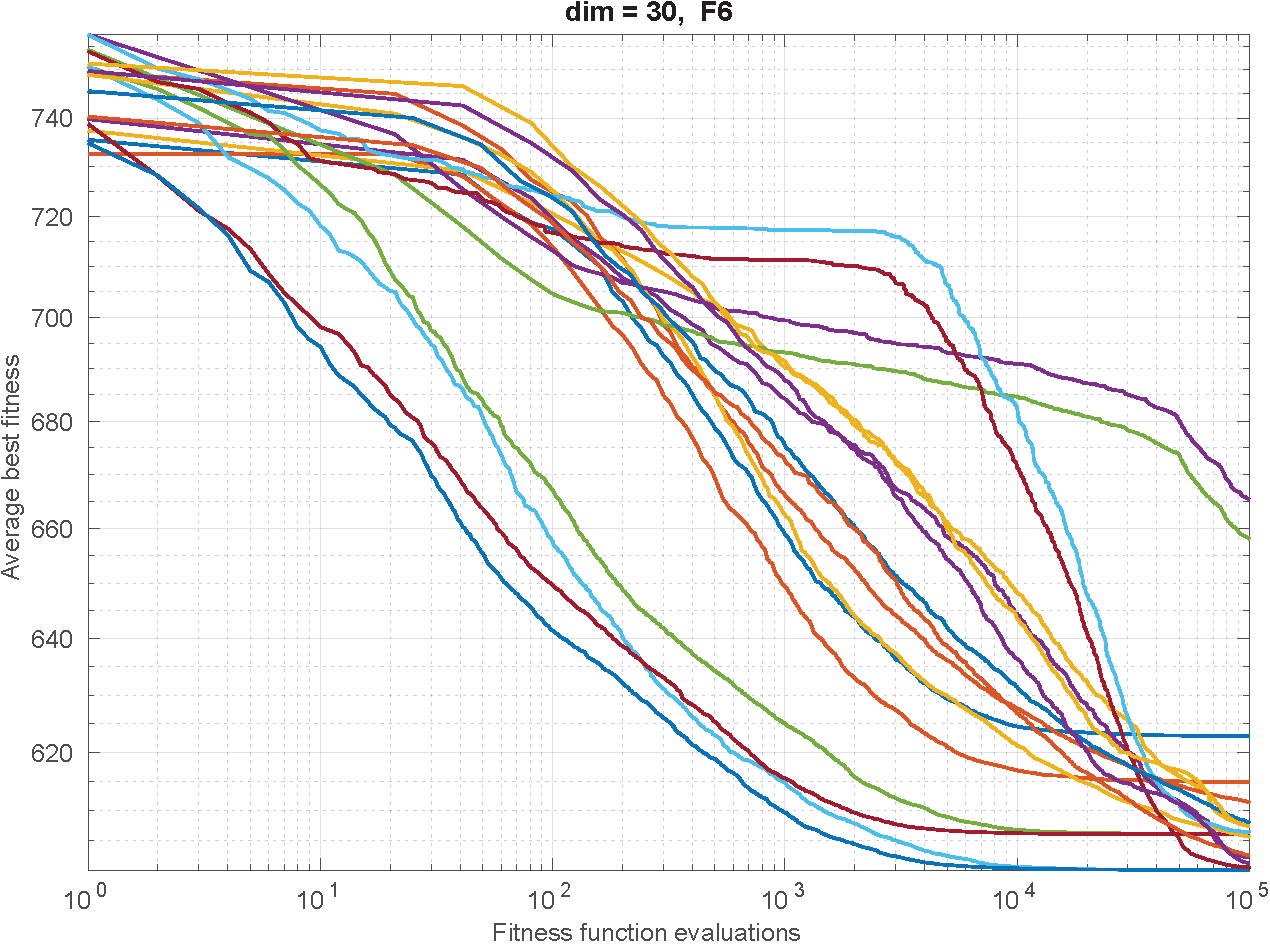}
   \end{subfigure}
   \end{adjustwidth}
\end{figure}

\begin{figure}[H]
  \centering
\begin{adjustwidth}{-1cm}{-2.0cm}
   \begin{subfigure}{0.35\textwidth}
        \includegraphics[height=4cm,width=5.5cm]{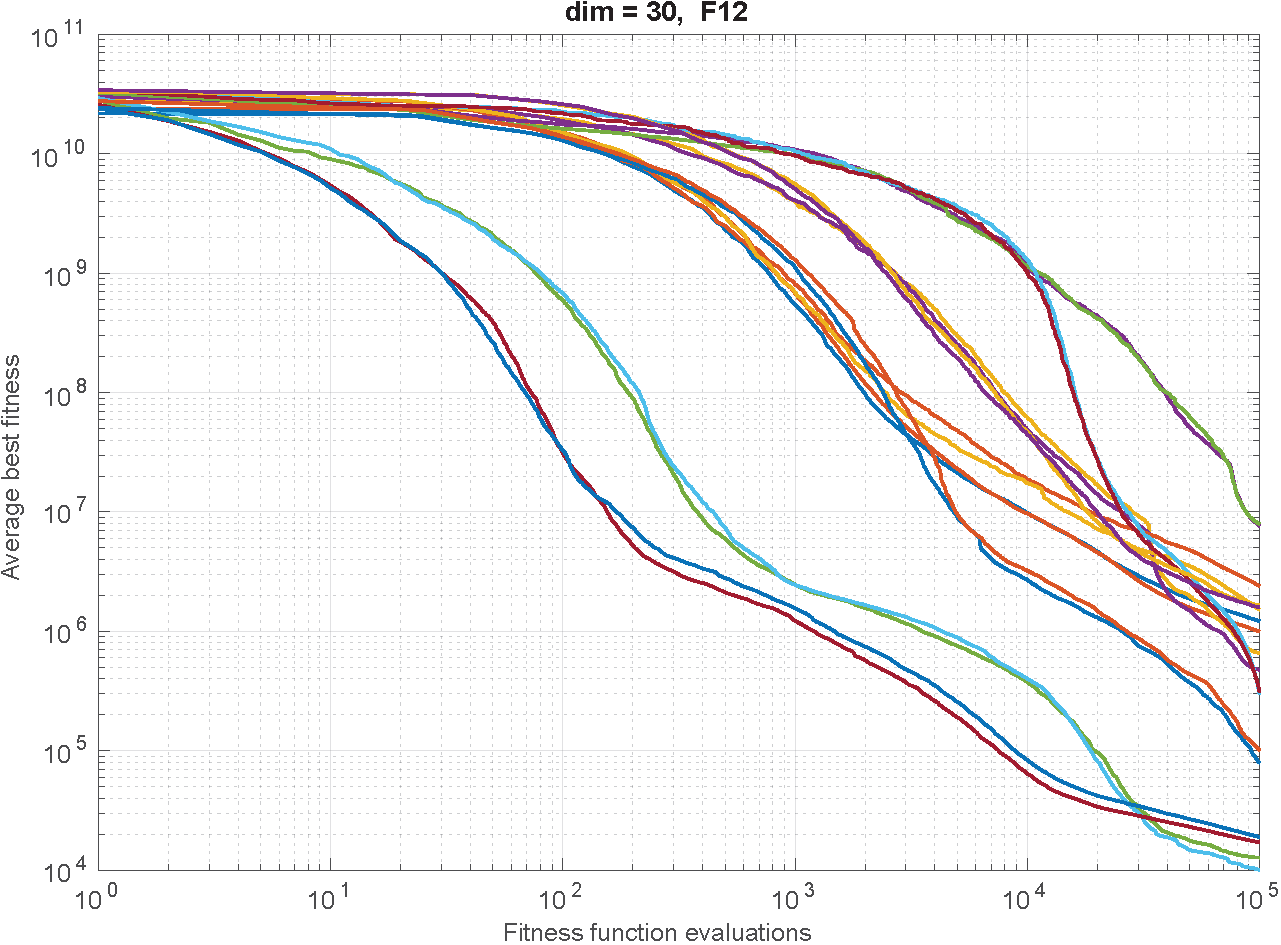}
  \end{subfigure}\hfill
  \begin{subfigure}{0.35\textwidth} 
        \includegraphics[height=4cm,width=5.5cm]{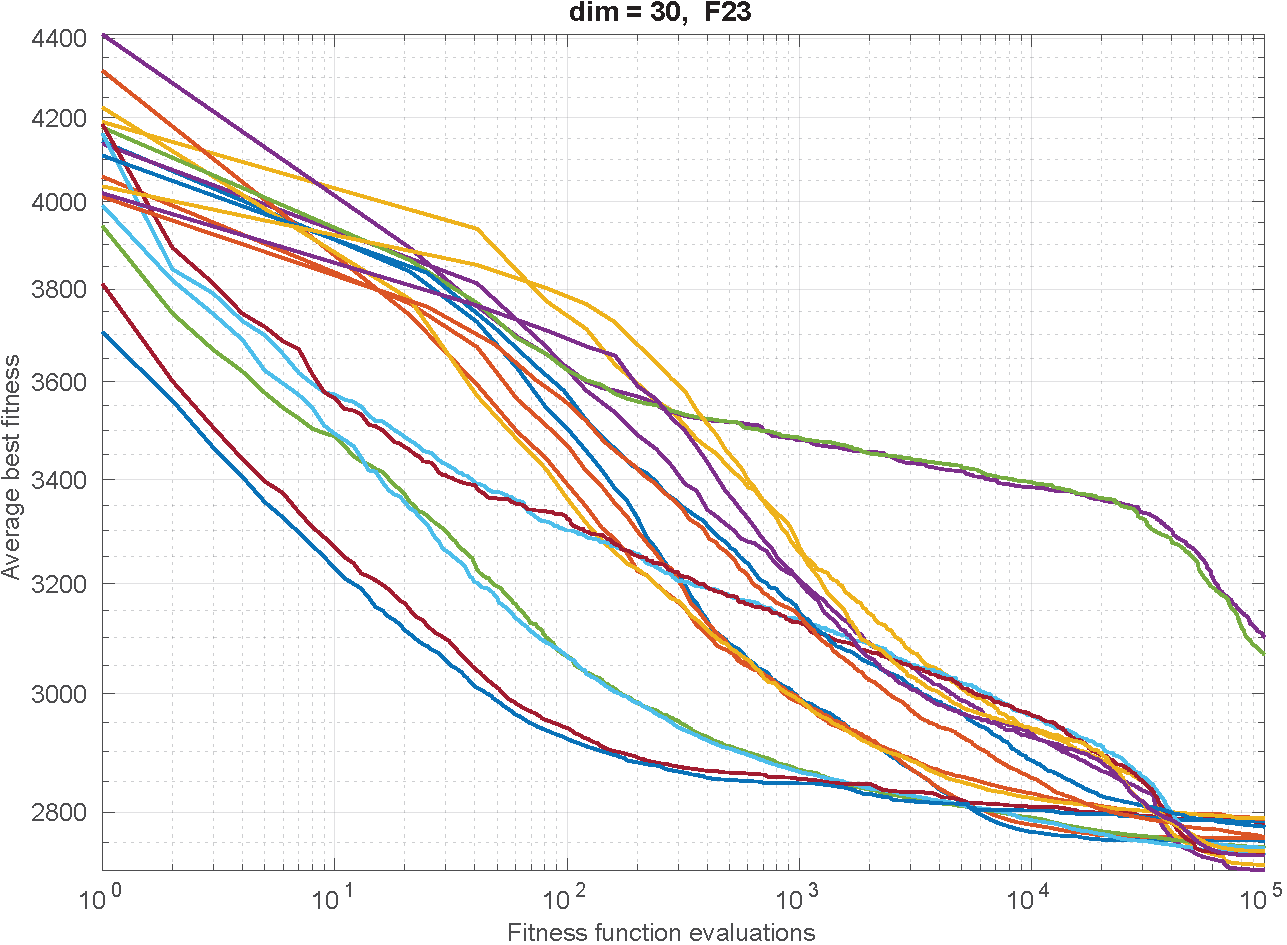}
  \end{subfigure}\hfill
  \begin{subfigure}{0.35\textwidth}
        \includegraphics[height=4cm,width=5.5cm]{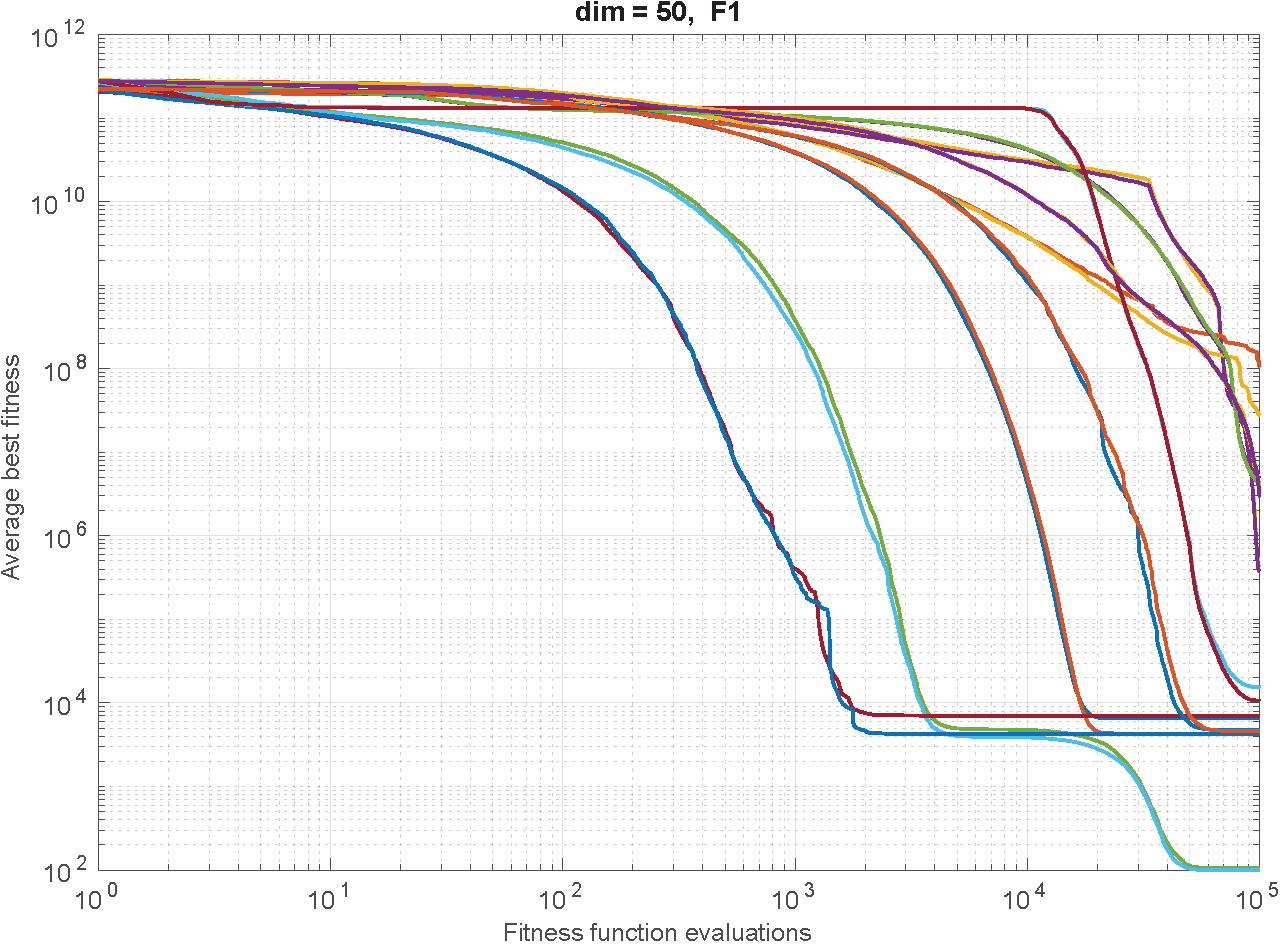}
  \end{subfigure}
  \end{adjustwidth}
\end{figure}

\begin{figure}[H]
  \centering
\begin{adjustwidth}{-1cm}{-2.0cm}
  \begin{subfigure}{0.35\textwidth} 
        \includegraphics[height=4cm,width=5.5cm]{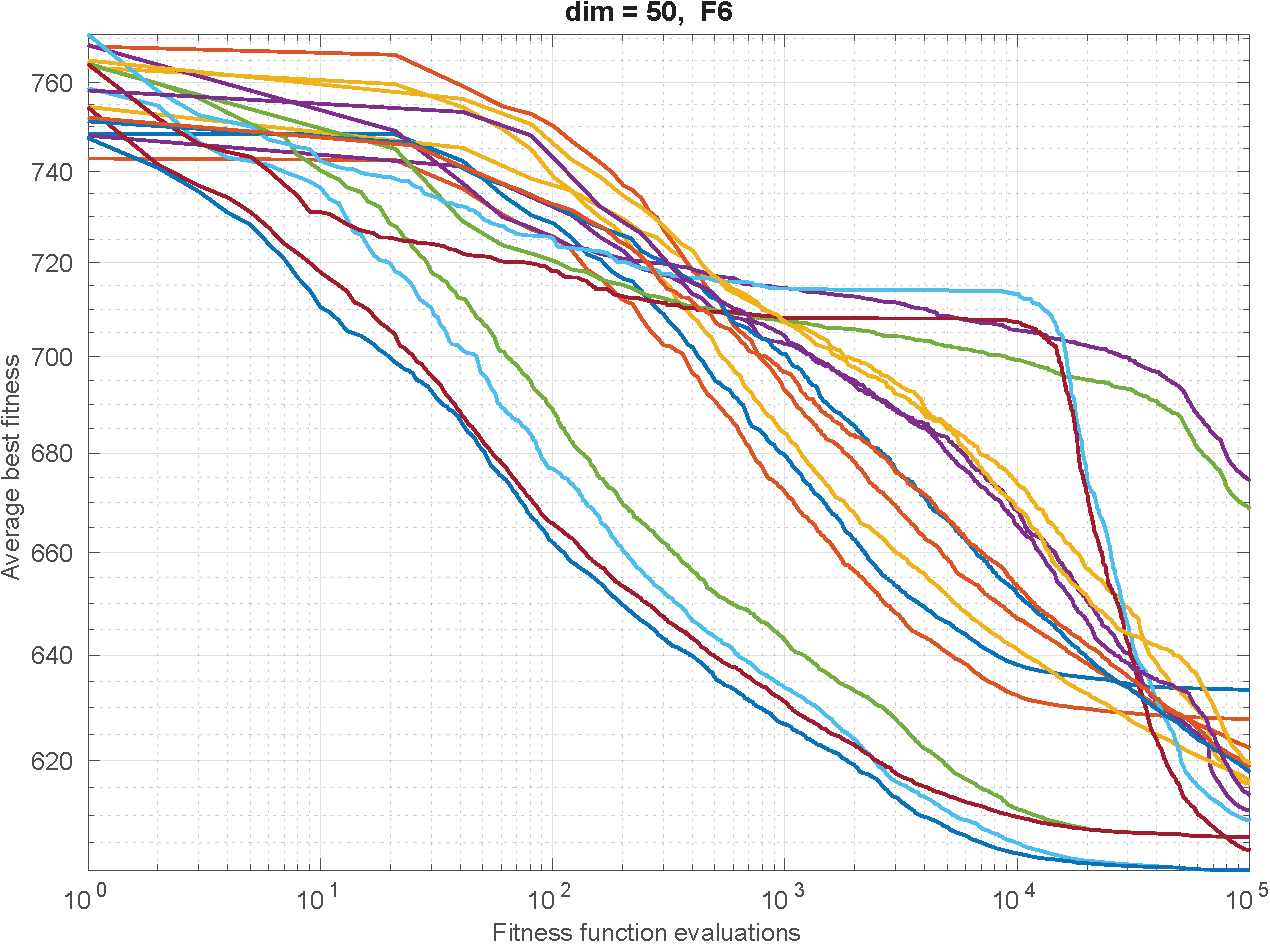}
  \end{subfigure}\hfill
  \begin{subfigure}{0.36\textwidth}
    \includegraphics[height=4cm,width=5.5cm]{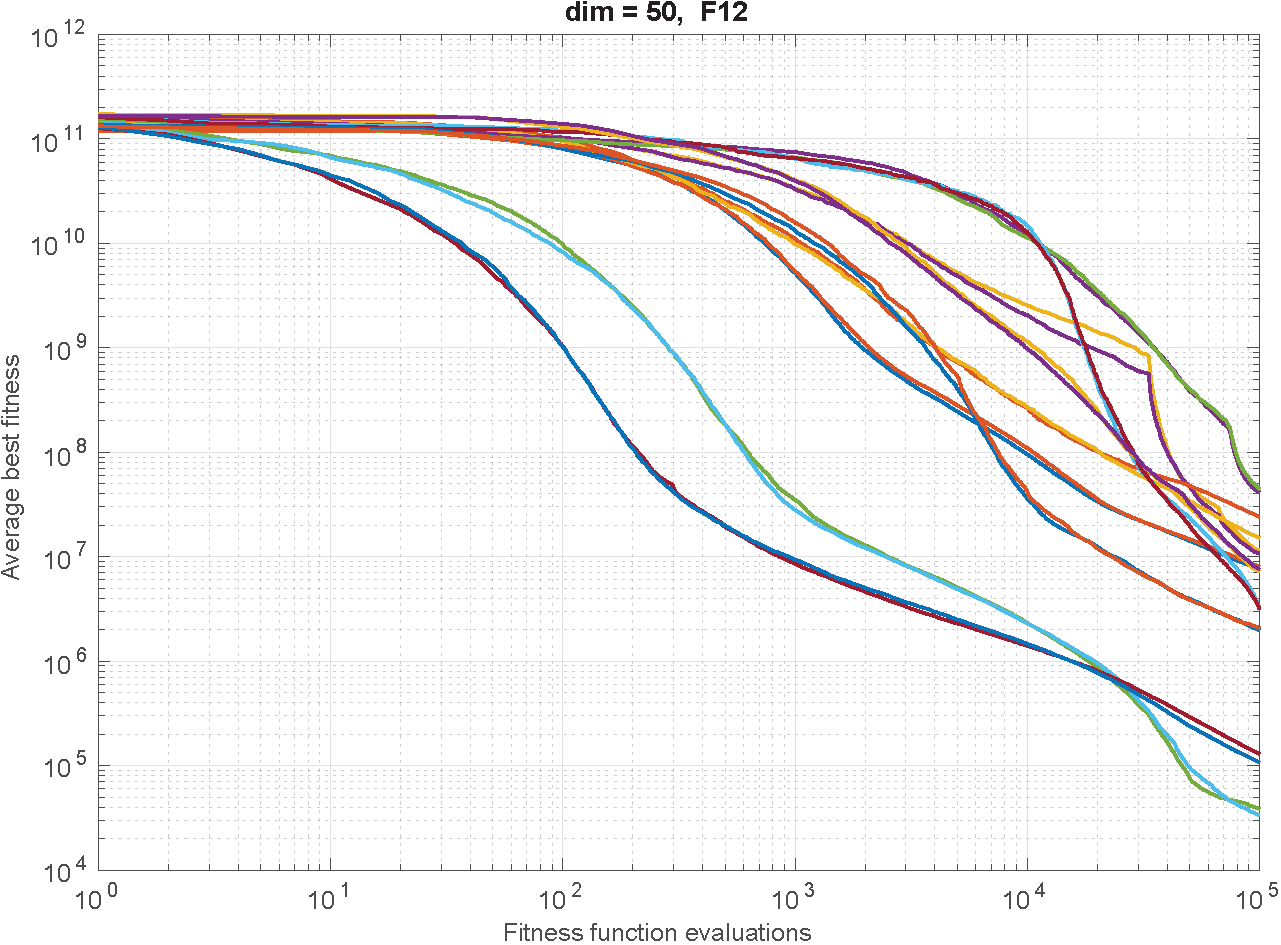}
     \end{subfigure}\hfill
 \begin{subfigure}{0.35\textwidth}
    \includegraphics[height=4cm,width=5.5cm]{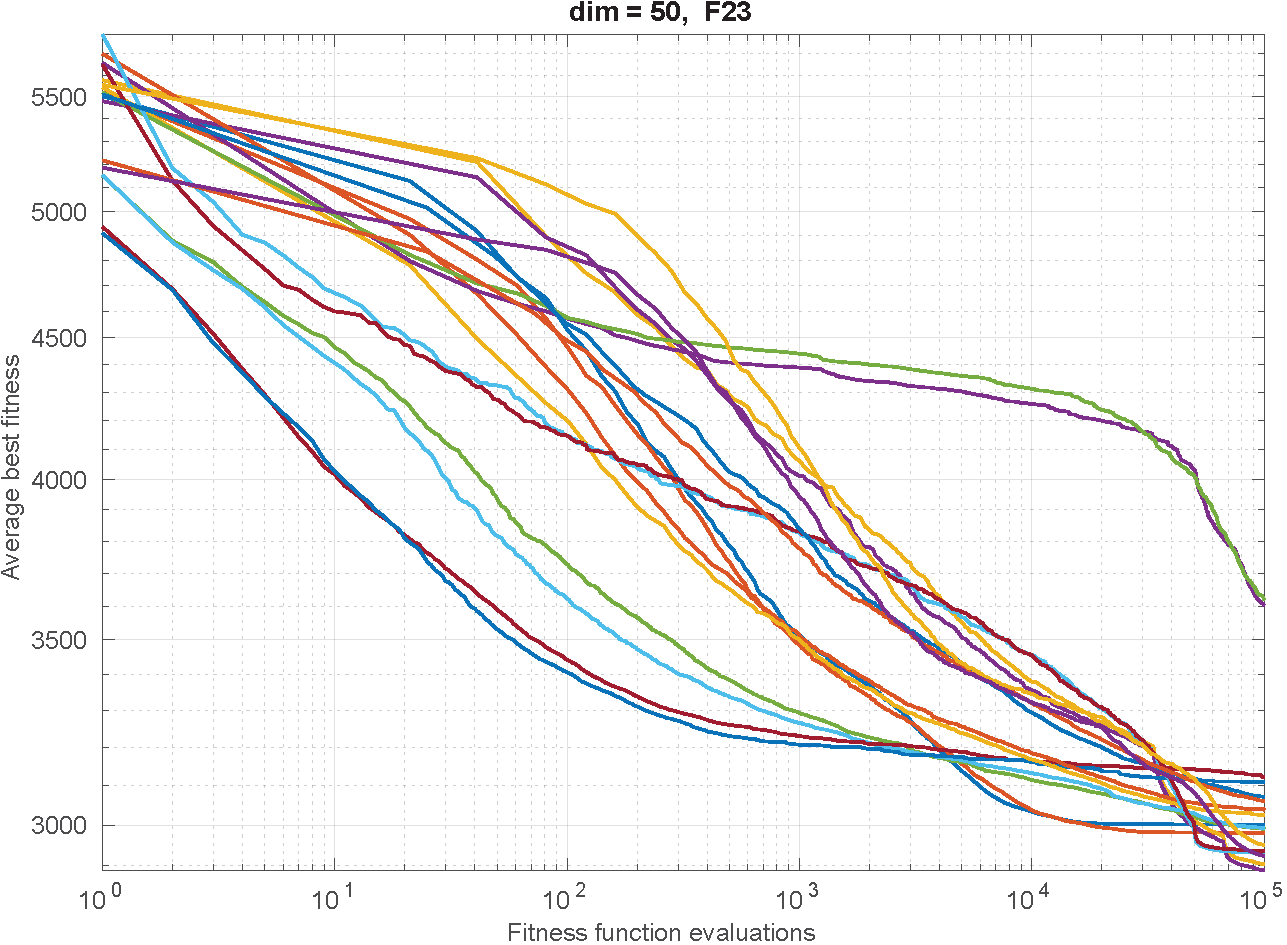}
  \end{subfigure}
\end{adjustwidth}
\end{figure}

\begin{figure}[H]
  \centering
\begin{adjustwidth}{-1cm}{-2.0cm}
  \begin{subfigure}{0.35\textwidth}
    \includegraphics[height=4cm,width=5.5cm]{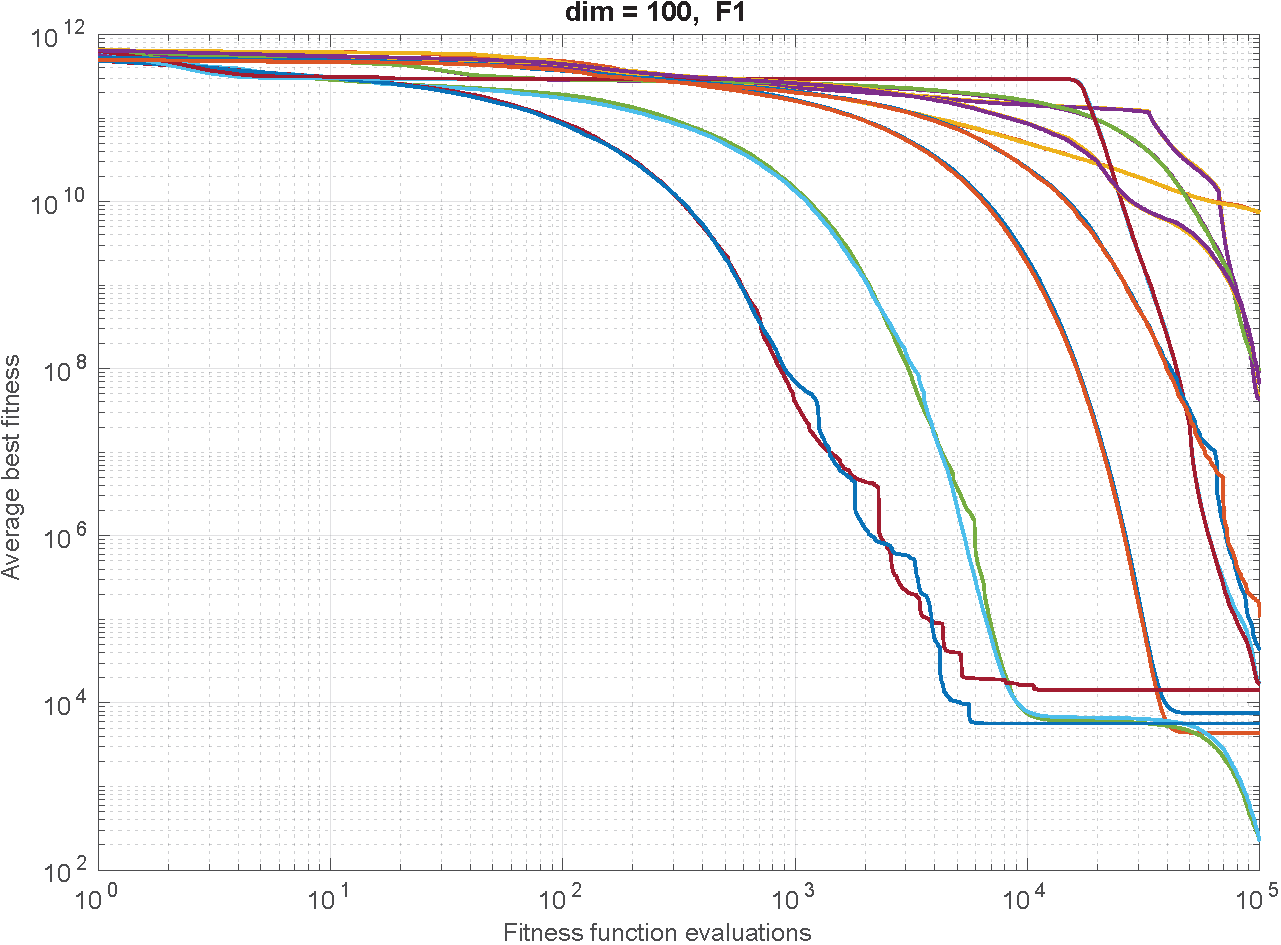}
  \end{subfigure}\hfill
  \begin{subfigure}{0.35\textwidth}
        \includegraphics[height=4cm,width=5.5cm]{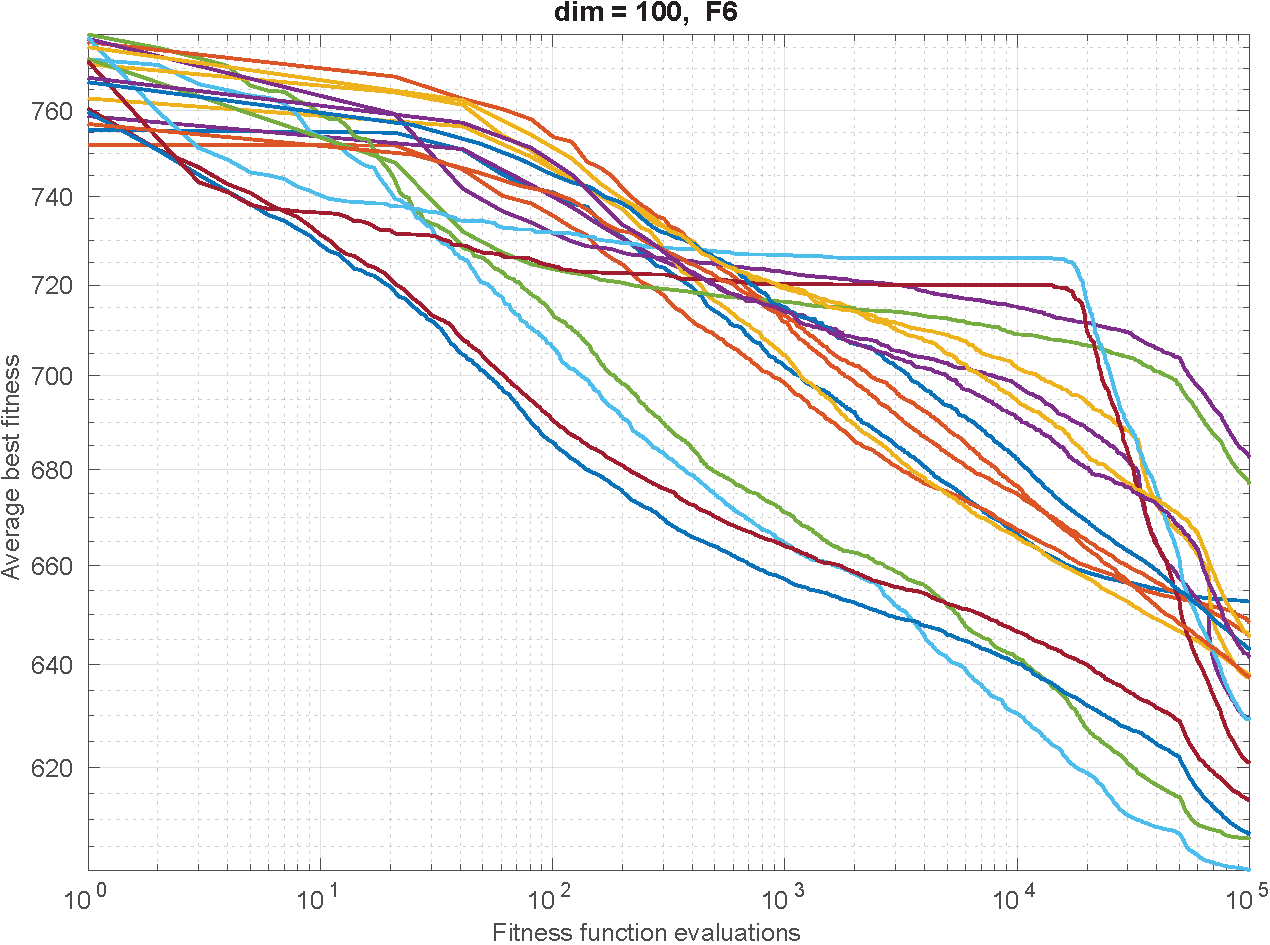}
     \end{subfigure}\hfill
   \begin{subfigure}{0.35\textwidth}
        \includegraphics[height=4cm,width=5.5cm]{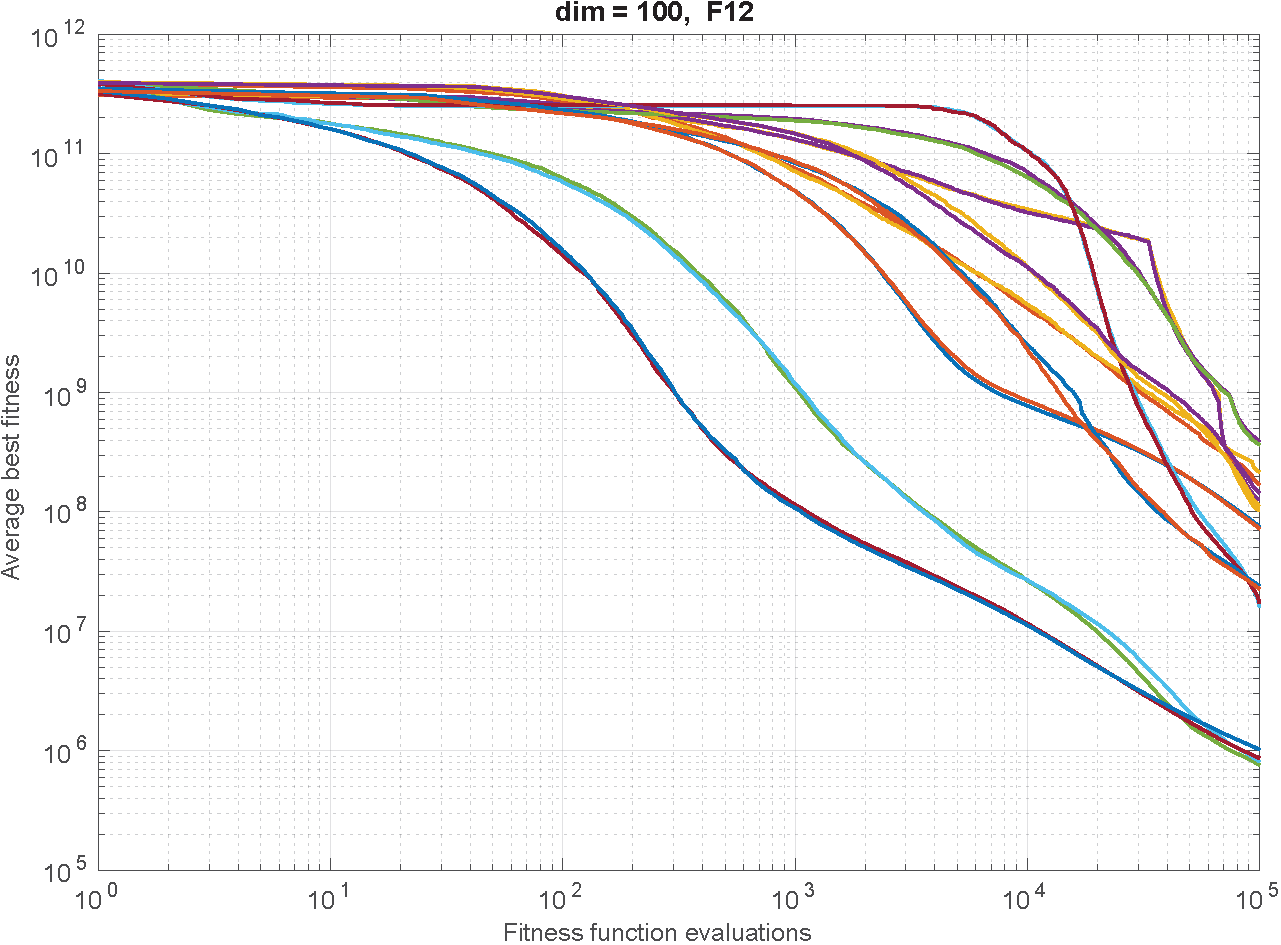}
    \end{subfigure}
    \end{adjustwidth}
\end{figure}

\begin{figure}[H]
 \centering
\begin{adjustwidth}{-1.4cm}{-2.0cm}
\begin{center}
  \begin{subfigure}{0.35\textwidth}
   \centering
        \includegraphics[height=4cm,width=5.5cm]{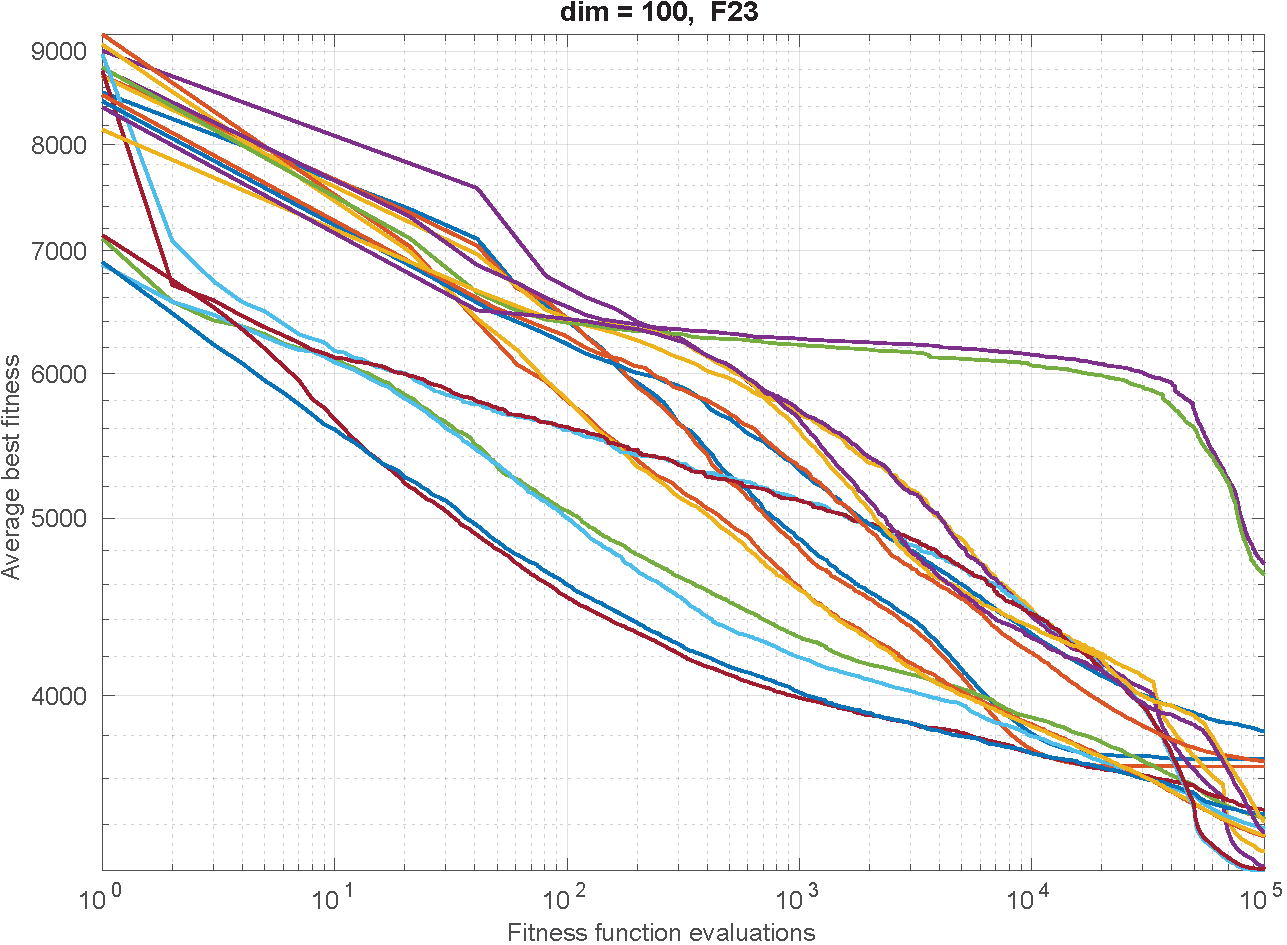}   
  \end{subfigure}
  \end{center}
  \begin{center}
  \begin{subfigure}{0.85\textwidth}
   \centering
    \includegraphics[height=0.75cm, width=\linewidth]{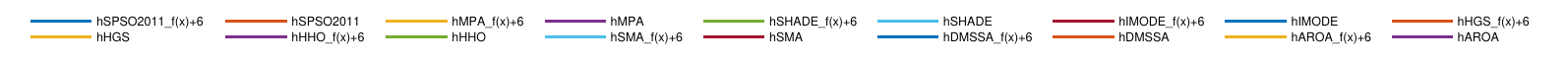}     
  \end{subfigure}
  \end{center}
  \caption{\footnotesize{Convergence trajectories of metaheuristic algorithms on a representative CEC-2017 benchmark function.}}
  \label{fig20}
\end{adjustwidth}
\end{figure}

\vspace{-0.2cm}
Adding a constant value to the objective function, in the form of the transformation $f(x) + 6$, constitutes a topologically invariant modification - it does not alter the structure of extrema or the ordinal relationships between solutions. According to metaheuristic theory, such a change should not affect the performance of evolutionary algorithms unless their internal mechanisms depend directly on the absolute values of the objective function, for instance, through the use of thresholds or rigid stopping criteria \cite{0jk},\cite{4aijk},\cite{5ab} The empirical analysis conducted herein confirms this theoretical expectation, offering an in-depth examination of nine hybrid optimization methods across four decision space dimensionalities: 10, 30, 50, and 100.

The tabular results Tab.\ref{tab5} (means, medians, standard deviations, and ranks) obtained for the transformation $f(x)+6$ are fully consistent with the baseline $f(x)$ results Tab.\ref{tab1}. For the hSPSO2011 algorithm in 10 dimensions, the mean drops from $1.6 \times 10^4$ to $8.2 \times 10^3$, while the median remains unchanged at $1.9 \times 10^3$. The standard deviation decreases from $2.0 \times 10^4$ to $1.3 \times 10^4$, indicating slightly improved stability. Similarly, hMPA reduces its mean from $2.6 \times 10^3$ to $1.7 \times 10^3$, with the median staying nearly constant at $1.6 \times 10^3$. The Wilcoxon test yields $p > 0.05$ for these cases, confirming the lack of statistically significant differences. The only exceptions are high-variance algorithms (hHHO, hHGS), such as hHHO, which shows a mean decrease from $5.8 \times 10^4$ to $5.1 \times 10^4$ ($p = 0.038$), although this change falls within random fluctuation.

For 50-dimensional space, hSPSO2011 has mean values of $8.8 \times 10^5$ ($f(x)$) and $8.3 \times 10^5$ ($f(x)+6$), while the median remains unchanged at $3.3 \times 10^3$. In 100 dimensions, hIMODE shifts from $0.8$ (baseline) to $3.2$ after transformation, which exactly matches the added constant of +6. hSHADE increases from $1.1$ to $2.4$, maintaining the shape of the distribution and the median. These results confirm that algorithms robust to absolute function values (i.e., those not using thresholds) are insensitive to additive transformations \cite{6atz}.

Boxplots for benchmark functions f1, f6, f12, and f23 (Fig.\ref{fig17}) corroborate that the results are merely vertically shifted compared to the baseline distributions, with no impact on the shape. The median for hSMA in f6 remains $1.9 \times 10^3$ both before and after transformation. For hIMODE and hSHADE, values remain close to zero, indicating strong stability and high structural resilience. Slight box expansions observed for hSPSO2011 and hSMA are marginal and do not alter the central tendency, suggesting that stochastic effects are responsible.

The Friedman test (Fig.\ref{fig18}) confirms the absence of significant ranking shifts after the transformation: $p$-values for the four dimensionalities are 0.152 ($dim=10$), 0.142 ($dim=30$), 0.118 ($dim=50$), and 0.109 ($dim=100$). Critical Difference (CD) diagrams show that the average ranks of the top methods - hIMODE (2.2-2.9), hSHADE (2.4-3.2), and hDMSSA (3.5-4.1) - are practically unchanged compared to the baseline version (Fig.\ref{fig2}) and prior transformations $f(x+8)$ and $f(5x)$. By contrast, the $f(x+8)$ translation caused severe degradation in SPSO2011 and HHO (average ranks > 9.5), while the $f(5x)$ scaling introduced selective effects - hIMODE worsened in high dimensions, whereas hDMSSA improved in stability. Unlike these perturbations, the $f(x)+6$ transformation proves neutral with respect to the ranking hierarchy, aligning with the literature on functional invariance \cite{0jk},\cite{0CC}.

The Bayesian analysis (Fig.\ref{fig19}) further supports these conclusions. For the hIMODE–hSHADE pair in 50 dimensions, the probability of dominance is P(hIMODE > hSHADE) = 0.48, falling within the ROPE. Similarly, hIMODE consistently dominates hHHO (P > 0.97), irrespective of the shift, demonstrating stability in the dominance structure. Compared to the $f(x+8)$ translation, which exhibited strong asymmetries (e.g., P(hSHADE > hHHO) > 0.99), the current transformation leads to flattened dominance patterns - typical of additive transformations that do not interfere with the topology of the objective space \cite{0jk},\cite{6atz}.

Convergence plots (Fig.\ref{fig20}) for f1, f6, f12, and f23 ($dims 10 - 100$) reveal no trajectory distortion. Objective function values shift upward by exactly 6 units, with no change in convergence speed or the number of iterations to reach the optimum. hIMODE and hSHADE consistently reach near-optimal values (e.g., < 6.5) in under 200 iterations across all dimensionalities. hHHO and hHGS continue to display high variance and slow decline, as already observed in the baseline case. Unlike scaling (which caused convergence slowdown in 100 dimensions) or translation (where hHHO often plateaued), the $f(x)+6$ transformation preserves process dynamics entirely, in line with adaptive optimization theory \cite{2},\cite{5ab}.

Based on the above, the $f(x)+6$ transformation emerges as the least invasive among all transformations examined. It does not alter the topology of the objective function, gradient values, separability, or variable correlations. All methods - particularly hIMODE, hSHADE, and hDMSSA - maintain their effectiveness, stability, and ranking advantage. PSO-based algorithms (hSPSO2011, hSMA) show only slight increases in result variance, with no impact on medians or optimization trajectories.
In the context of previous analyses - translation, scaling, and rotation - the additive transformation serves as a distinct test of value invariance, separate from structural tests. Its neutrality regarding algorithm performance constitutes a robust form of structural validation for metaheuristic design. These findings reinforce the recommendation to adopt differential and adaptive hybrids as the most general and robust approaches against all topologically insignificant transformations of the objective function.

\vspace{-0.2cm}
\section{General Conclusions and Final Remarks}

This study presents a comprehensive and rigorous evaluation of modern hybrid metaheuristics under a suite of benchmark space transformations, including translation, scaling, rotation, and additive shifts. Across 29 CEC-2017 functions and four dimensional settings (10, 30, 50, 100), we conducted detailed statistical and convergence analyses, applying Wilcoxon and Friedman tests, critical difference diagrams, Bayesian dominance matrices, and convergence trajectory comparisons.

The results clearly demonstrate that differential hybrid algorithms - particularly hIMODE, hSHADE, and hDMSSA - consistently outperform classical metaheuristics in terms of accuracy, robustness, convergence speed, and structural invariance. These methods retain high performance even under geometrical and topological alterations of the search space, confirming their resilience to real-world conditions such as non-separability, non-stationarity, and parameter scaling.

While some individual hybrids (e.g., hIMODE under $f(5x)$) exhibited minor precision degradation, their overall ranking positions and statistical dominance remained unaffected. In contrast, traditional algorithms such as CMA-ES and HHO suffered significant performance losses, especially in high-dimensional or transformed spaces, indicating poor adaptability and limited practical reliability.

No anomalies were identified that would challenge the conclusions drawn from the selected representative functions (f1, f6, f12, f23). The observed performance patterns - in terms of convergence behavior, ranking consistency, and statistical significance - remained stable across the entire CEC-2017 suite and all considered dimensionalities. Notably, this consistency extended beyond convergence trajectories: the distributional characteristics observed in boxplots of final results for the remaining benchmark functions further corroborated the dominance of the top-performing hybrids. The alignment of statistical, graphical, and dynamic analyses confirms the structural robustness and general applicability of these hybrid approaches under a wide range of problem transformations.

In summary, the findings provide compelling evidence for the adoption of adaptive differential hybrid metaheuristics in complex optimization tasks, especially in scenarios where the landscape may undergo unknown transformations.  
These methods demonstrate consistent and robust performance across a wide spectrum of optimization landscapes and dimensionalities, substantiating their applicability to real-world scenarios involving engineering design, complex system modeling, 
and data-driven optimization under structural uncertainty \cite{0a},\cite{4aijk},\cite{6atz}.

\vspace{-0.2cm}
\subsection{Practical Implications and Application Context}

The observed robustness of hybrid DE-based algorithms under various objective space transformations has direct implications for real-world optimization tasks. In practical scenarios, the objective function may be shifted, scaled, or rotated due to normalization, data preprocessing, or domain-specific modeling - often without preserving structural properties like separability or stationarity \cite{0jk},\cite{4aijk},\cite{6atz}.\\
\vspace{-0.11cm}
Typical application domains include:

\begin{itemize}
\item \textbf{Graph-based structural optimization}, where estimation of the Cheeger constant supports clustering, segmentation, and network robustness analysis \cite{0uwz}.
\item \textbf{Neural density modeling}, where distribution-based neurons (e.g., HCRNN, 2WNN) require optimization of joint distribution parameters for accurate signal propagation and compression \cite{0kz}.
\item \textbf{Estimation of Markov-type constants}, where optimization of polynomial inequalities (e.g., Markov inequalities on simplices) leads to bilevel, nonconvex problems solvable by evolutionary/metaheuristic techniques \cite{5},\cite{5go}.
\item \textbf{Engineering design optimization}, where constraint transformations and normalization often distort objective landscapes \cite{0CC},\cite{4aijk}.
\item \textbf{Robotics and control systems}, where sensor noise and dynamic constraints lead to objective shift and rotation \cite{4fa},\cite{7}.
 \item \textbf{Financial modeling and portfolio optimization}, characterized by non-stationary and non-separable search spaces \cite{0BB},\cite{4ccd}.
\end{itemize}

In such contexts, the demonstrated invariance and adaptability of hybrid algorithms - especially \textbf{hIMODE},\textbf{hSHADE}, and \textbf{hDMSSA} - support their application as default solvers in structurally variable and uncertain black-box environments. 
These results corroborate prior studies on operator invariance and adaptive evolutionary mechanisms \cite{2},\cite{6atz}, confirming the relevance of hybrid models in structurally perturbed and non-ideal black-box landscapes.

All numerical computations, data processing, and statistical visualizations presented in this study 
were performed using \textsc{MATLAB} R2022a under a Linux-based operating system on a workstation 
equipped with an Intel Core i7-6700K processor (4~GHz) and 64~GB of RAM. All data used in this article are available for academic purposes upon reasonable request to the corresponding author.

\textbf{Declaration of Competing Interest:}
The authors declare that they have no known competing financial interests or personal relationships 
that could have appeared to influence the work reported in this paper.

\FloatBarrier
\renewcommand{\refname}{

\end{document}